\documentclass[letterpaper]{article} % DO NOT CHANGE THIS
\usepackage{aaai25}
\usepackage{times}  % DO NOT CHANGE THIS
\usepackage{helvet}  % DO NOT CHANGE THIS
\usepackage{courier}  % DO NOT CHANGE THIS
\usepackage[hyphens]{url}  % DO NOT CHANGE THIS
\usepackage{graphicx} % DO NOT CHANGE THIS
\usepackage{natbib}  % DO NOT CHANGE THIS AND DO NOT ADD ANY OPTIONS TO IT
\usepackage{caption} % DO NOT CHANGE THIS AND DO NOT ADD ANY OPTIONS TO IT
\usepackage{algorithm}
\usepackage{algorithmic}

\usepackage[inkscapelatex=false]{svg}
\usepackage{mdframed}
\usepackage{tabularx}
\usepackage{adjustbox}

\usepackage{placeins}

\PassOptionsToPackage{table,xcdraw}{xcolor}
\usepackage{xcolor}

\usepackage{newfloat}
\usepackage{listings}
\DeclareCaptionStyle{ruled}{labelfont=normalfont,labelsep=colon,strut=off} % DO NOT CHANGE THIS
\floatstyle{ruled}
\newfloat{listing}{tb}{lst}{}
\floatname{listing}{Listing}
\usepackage{multirow}
\usepackage{colortbl}
\usepackage{array}
\newcolumntype{Q}{>{\footnotesize}p{1.25cm}}
\newcolumntype{R}{>{\footnotesize}p{1.5cm}}
\definecolor{fgreen}{rgb}{0.196, 0.804, 0.196}
\title{
Bridging the Gap: Enhancing LLM Performance for Low-Resource African Languages with New Benchmarks, Fine-Tuning, and Cultural Adjustments
}

\author{
    Tuka Alhanai\textsuperscript{\rm $\dagger$}\equalcontrib,
    Adam Kasumovic\textsuperscript{\rm $\dagger$}\equalcontrib,
    Mohammad Ghassemi\textsuperscript{\rm $\dagger$}\equalcontrib, \\
    Aven Zitzelberger\textsuperscript{\rm $\dagger$},
    Jessica Lundin\textsuperscript{\rm $\ddagger$},
    Guillaume Chabot-Couture\textsuperscript{\rm $\ddagger$}
}

\affiliations{
    \textsuperscript{\rm $\dagger$}Ghamut Corporation, MI,  USA\\
    \textsuperscript{\rm $\ddagger$}Bill and Melinda Gates Foundation, WA, USA\\
}

\usepackage{bibentry}
\definecolor{mygreen}{rgb}{0.855, 0.969, 0.651}  % Define custom green
\definecolor{myblue}{rgb}{0.855,0.902,0.996}
\usepackage{soul}
\renewcommand{\ul}{\underline}

\begin{document}

\maketitle

% Numerous benchmarks designed to evaluate the reasoning capacities of large language models (LLMs) are not publicly available for low-resourced languages; the absence of these benchmarks increases the already steep challenge of developing models for these languages. 

\begin{abstract}
Large Language Models (LLMs) have shown remarkable performance across various tasks, yet significant disparities remain for non-English languages, and especially native African languages. This paper addresses these disparities by creating approximately 1 million human-translated words of new benchmark data in 8 low-resource African languages, covering a population of over 160 million speakers of: Amharic, Bambara, Igbo, Sepedi (Northern Sotho), Shona, Sesotho (Southern Sotho), Setswana, and Tsonga. Our benchmarks are translations of Winogrande and three sections of MMLU: college medicine, clinical knowledge, and virology. Using the translated benchmarks, we report previously unknown performance gaps between state-of-the-art (SOTA) LLMs in English and African languages. Finally, using results from over 400 fine-tuned models, we explore several methods to reduce the LLM performance gap, including high-quality dataset fine-tuning (using an LLM-as-an-Annotator), cross-lingual transfer, and cultural appropriateness adjustments. Key findings include average mono-lingual improvements of 5.6\% with fine-tuning (with 5.4\% average mono-lingual improvements when using high-quality data over low-quality data), 2.9\% average gains from cross-lingual transfer, and a 3.0\% out-of-the-box performance boost on culturally appropriate questions. The publicly available benchmarks, translations, and code from this study support further research and development aimed at creating more inclusive and effective language technologies. 

\end{abstract}

% Change these as needed once they are made
% Remove extended version from arXiv version
\begin{links}
    \link{Code}{https://github.com/InstituteforDiseaseModeling/Bridging-the-Gap-Low-Resource-African-Languages}
    \link{Datasets}{https://huggingface.co/datasets/Institute-Disease-Modeling/mmlu-winogrande-afr}
    %\link{Extended version}{arXiv-link-here}
\end{links}

\section{Introduction}

For many tasks, Large Language Models (LLMs) perform on-par with or approaching human performance. Furthermore, LLM capabilities are improving: the performance gap between state-of-the-art LLMs (e.g. GPT-4) and humans for many benchmarks is much smaller than the gap between previous LLM generations (e.g. GPT 3.5) and humans \cite{achiam2023gpt, bandarkar-etal-2024-belebele, sakaguchi2021winogrande, lin2021few, hendryckstest2021}. Despite impressive advancements, LLMs are significantly less capable when assessed in non-English languages. When assessing LLMs in native African languages, which are predominantly low-resource \cite{joshi2020state}, the gap between human and LLM performance is notable (when known), but generally remains unknown because many standard benchmarks do not exist in native African languages. 

The performance discrepancy between LLMs in English and African languages is not just a technical challenge; it is a significant issue of equity. All 2,123 native African languages are low-resource, including the 31 languages with more than 10 million speakers \cite{hammarstrom2015ethnologue, joshi2020state}\footnote{It was assumed that if a resource level rating in \cite{joshi2020state} for a language was not given, then it was low-resource.}. Naturally, lower language resource levels result in poorer-performing LLMs. This is particularly tragic because LLMs are least reliable for language speakers who have the most to gain. Of the world's poor, ~66\% live in Africa \cite{Galal_2024}, ~66\% of those don’t have access to the Internet \cite{Facts_and_Figures_2023_2023a}, and 80\% do not speak English \cite{Central_Intelligence_Agency_2024}. Thus, even in the unlikely event that the world's poorest could access the Internet and afford the costs of a state-of-the-art LLM, their ability to read and write English would prevent them from making use of the tools. Ultimately, the differences in LLM performance between languages result in a ``rich-get-richer" effect: LLMs are more helpful to (English-speaking) people who are better off, and who may then provide better content to train better LLMs. 

One approach to bridging the LLM performance gap is to translate all non-English language queries into English, query an English LLM, and backtranslate the response. This approach is only viable if the cumulative errors from translation and backtranslation are smaller than the errors from using non-English language LLMs alone. Previous studies report that errors from leading machine translation tools (i.e. Google Translate) can be substantial in African languages \cite{bapna2022building, gt_tarzan}; however, the extent to which translation errors impact the \textit{substantive} reasoning capabilities of multilingual LLMs versus their \textit{style} (``translationese"), remains unclear. Insofar as the \textit{substantive} errors are minimal, machine translation may serve as a workaround when: (i) the goal is to leverage an LLM to answer questions based on facts, and (ii) when the LLM being used has limited non-English experience (e.g. Phi 3 \cite{abdin2024phi}).

%However, even though the gap between machine-translated queries and directly using an LLM in native African languages is not always substantial, it is crucial to address the performance gap for several reasons. First, reliance on translation technologies can still result in loss of nuanced meaning, especially for culturally specific or context-sensitive queries, where even minor errors can lead to significantly different interpretations. Second, not all tasks benefit equally from translation; tasks requiring deep cultural understanding or subtleties of language may still see a performance drop when translated. Finally, while large multilingual models like GPT-4o show promise, their superior performance in low-resource languages is not universal across all models or tasks, indicating that further improvements are necessary. Therefore, there is still a need for more human-made tools, resources, and studies to help the research community understand (1) how large the LLM performance gap in African languages truly is, and (2) what can be done to close the performance gap, once known.

In summary, existing LLMs under-perform in native African languages, and it is unknown if translation technologies introduce too much error to serve as a viable workaround. Thus, more tools, resources, and studies are needed to help the research community understand (1) how large is the LLM performance gap in African languages and (2) what can be done to close the performance gap, once known?

\subsection{Aims}

Our work has three specific aims, listed below:

\begin{itemize}
\item \textbf{Aim 1 - Benchmark Translation}: We translate the popular multiple choice reasoning benchmark Winogrande, as well as three clinical sections of MMLU (college medicine, clinical knowledge, and virology) into 8 low-resourced and under-studied African languages (Amharic, Bambara, Igbo, Sepedi, Shona, Sesotho, Setswana, and Tsonga), allowing assessment of the (unknown) capabilities of LLMs in African languages.
\item \textbf{Aim 2 - Performance Assessment:} We apply several state-of-the-art (SOTA) LLMs to the newly translated benchmarks from Aim 1 and measure the extent of the performance gap between English and each of the African languages on the benchmarks. To assess the viability of machine translation, we compare the performance of SOTA LLMs on machine-translated benchmarks versus human-translated benchmarks. We also assessed the performance gap between culturally appropriate and inappropriate benchmark questions. This quantitative assessment highlights the areas and languages where LLM improvements are most needed.
\item \textbf{Aim 3 - Performance Enhancement}: We explore various fine-tuning strategies to determine their impact on closing the LLM performance gap in African languages. This includes adjusting the fine-tuning data used based on the data domain, language, data quality, and the volume of training samples. Understanding how fine-tuning characteristics impact LLM performance will inform prospective data collection efforts for the community at large.

\end{itemize}

The benchmarks, translations, and all the code needed to recreate the results herein may be found online and are made publicly available under the MIT license.

\section{Related Work}

\subsection{Benchmarks in African Languages}
Recent advances in natural language processing have seen a growing interest in assessing language modeling performance in African languages. This interest has resulted in benchmarks in several tasks, including language identification with AfroLID \cite{adebara2022afrolid}, machine translation with FLORES-200 \cite{nllb2024scaling}, and natural language inference with XTREME \cite{hu2020xtreme}. More recently, benchmarks have emerged for African languages to assess LLM \textit{reasoning} using multiple choice questions. In particular, the manually curated reading comprehension benchmark Belebele provides the most extensive coverage of African languages (25 languages; 115 languages in total) \cite{bandarkar-etal-2024-belebele}. In addition to Belebele, Irokobench provides manual translations of reasoning tasks into 15 African languages \cite{adelani2024irokobench}, while Winogrande-MMLU-Clinical-ZA also provides manual translations of reasoning tasks into 3 African languages \cite{winmmluza2024}. Given the lack of reasoning benchmarks available for African languages, particularly benchmarks that have been manually translated or otherwise sourced from human-written text (i.e. not machine-translated or AI-generated), we aim to translate two established reasoning and domain-knowledge benchmarks, Winogrande \cite{sakaguchi2021winogrande} and MMLU \cite{hendryckstest2021, hendrycks2021ethics}, into 8 African languages. The translations of these two benchmarks provide valuable additions to existing African language datasets, enabling the proper evaluation of LLMs in African languages in popular LLM evaluation tasks. This effort helps pave the way for developing LLMs that perform as well in African languages as they do in English, as the availability of African language benchmarks allows developers to continually refine and enhance their LLMs for use in African languages.

\subsection{Impact of Culture on LLM Performance}

When utilizing data from different languages, cultural factors have been shown to affect the accuracy, relevance, and sensitivity of LLM output. Researchers have focused on the ``toxicity" of model outputs (i.e. the output of racist, violent, or harmful content) and have released several benchmarks to measure toxicity in LLMs \cite{wen2023unveiling}. However, the impact of cultural nuances has been relatively less explored, in which content that in one context (language) is mundane or inoffensive (e.g. \textit{``a child dislikes broccoli"}) may be considered culturally strange, incoherent, or disrespectful in another context (language) \cite{ids_use}. Cultural nuances have been shown to impact both the quality of translations \cite{Yao2023BenchmarkingLM} as well as the model's understanding and generation of responses \cite{putri2024llmgenerateculturallyrelevant}. Strategies to identify culture-specific references generally require ground-truth labels generated by human annotators, such as a list of offensive language-specific words \cite{nllb2024scaling}, a list of universally acceptable words \cite{ids_source}, or predefined dimensions of culture \cite{arora-etal-2023-probing}. Word lists identify explicit expressions in the data and thus have the advantage of scaling to large amounts of data; however, word lists do not capture data that are \textit{implicitly} inappropriate. Therefore, several efforts have been made to annotate implicit expressions of cultural (in)appropriateness \cite{xu2021bot, hartvigsen2022toxigen}. In this study, we evaluate the impact of implicit cultural appropriateness on LLM performance (via human annotators), when translating seemingly inoffensive data (i.e. Winogrande) from English. We also share the cultural annotation data to enrich the translated benchmark.

\subsection{Impact of Fine-tuning on Cross-lingual Transfer}
Previous studies report that cross-lingual instruction tuning improves performance by nearly as much as mono-lingual instruction tuning \cite{shaham2024multilingual}. Given that even the most widely spoken African languages are low-resource, cross-lingual tuning may provide a way to ``boost" the amount of data (and thus the performance) of LLMs in low-resource languages \cite{beukman2023analysing, whitehouse2023llm}. However, the impact of cross-lingual tuning is unknown for most African languages \cite{hu2020xtreme, liang2020xglue, conneau2018xnli}. Hence, we also compare the effects of mono- and cross-lingual tuning using the translated benchmarks.

\subsection{Impact of Data Quality on LLM Performance}

The scarcity of data is a significant barrier to improving LLM performance \cite{villalobosposition}. In an effort to reduce data barriers, automated methods have been presented to identify high \textit{quality} data samples within an existing corpus \cite{longpre2024pretrainer}. 
% Classical methods include language model generated perplexity measures of candidate text \cite{wenzek2020ccnet}, domain matching based on cosine similarity of embedding representations \cite{lee2024nv}, and evaluating divergences of text translations from a reference \cite{zouhar2021backtranslation}.  
Recent methods have focused on using LLMs to assess the quality of data \cite{tan2024large}, which includes annotating data based on the robustness of question-answer pairs (QA pairs) to variations in the question \cite{chen2024automated}, rating data quality on a Likert scale \cite{zhou2024lima}, evaluating the correctness of an answer \cite{cole2023selectively}, or generating a binary ``preference" score when selecting between two texts, such as the LLM-as-a-Judge framework \cite{zheng2023judging}. In this study, we apply an LLM-as-an-Annotator of data quality across two translated benchmarks in eight African languages.

\section{Methods}

\subsection{Benchmark Translation (Aim 1)}

We translated the popular multiple choice reasoning benchmark Winogrande, as well as three clinical sections of MMLU (college medicine, clinical knowledge, and virology) into 8 low-resourced and under-studied African languages (Amharic, Bambara, Igbo, Sepedi, Shona, Sesotho, Setswana, and Tsonga). These benchmarks were selected because they are multiple choice (and thus easy to compare across languages) and widely used (300+ citations/year). A summary of the translation process is described below, and we provide additional information on the procedures, translator profiles, remuneration approach, and other details in Appendix Section \ref{sec:appendix-benchmark-translation}.

\subsubsection{Winogrande Translation Process} \label{sec:winogrande-translation-process}

The translation of Winogrande (3,674 QA pairs, 73,742 words) occurred in three steps: (i) a \textit{translator} translated QA-pairs from English to the target language, (ii) an independent \textit{validator} checked each of the translations and corrected any identified translation errors, and (iii) two independent \textit{evaluators} assessed the final quality of the validated/corrected translation. 

All individuals were recruited from Upwork.com; we recruited the most qualified available individuals (\textit{translators, validators, and evaluators}) for each language and always paid above the average wage (\$7.30 USD) in South Africa \cite{statsSA2020labour}. All \textit{translators} and \textit{validators} were allowed to use a machine translation tool, but had to manually validate the output and correct errors. The \textit{evaluators} were presented with the \textit{validators'} translations (which were corrections of the \textit{translators'} translations) and asked to rate the \textit{quality} of the translation according to three options: (i) ``Good translation" (``good"), (ii) ``Incorrect, but someone could understand the idea" (``understandable"), and (iii) ``Completely wrong" (``wrong") (Benjamin 2019; Bapna et al. 2022). To aid the research community, we have made the initial translations, corrections, and evaluations publicly available along with the translated benchmarks. Examples of the translation and validation survey forms are available in Figures \ref{fig:translation-survey} and \ref{fig:review-survey}.

\subsubsection{MMLU-Clinical Translation Process}
Unlike Winogrande, specialized domain-knowledge was required to translate our selected MMLU sections: \textit{virology}\footnote{The virology section for Afrikaans, Zulu, Xhosa were also translated, extending the work of \cite{winmmluza2024}.} (189 QA pairs / 5,452 words), \textit{clinical knowledge} (299 QA pairs / 8,744 words), and \textit{college medicine} (200 QA pairs / 12,911 words); thus, we hired a professional translation firm (Translated.com) to perform the translations of the three MMLU sections. The professional translation firm guaranteed that only translators with prior translation experience as well as subject domain translation experience were used. The translation firm was paid \$11,232.29 USD to translate the three MMLU sections over a 15-day period.

% Translators were requested to (i) maintain original punctuation in translations (commas, periods), and (ii) utilize clinical expressions in the target language, where possible. 

% \footnote{A professional translation service was used given the specialized domain knowledge (medical) required to complete the task.} into Amharic, Bambara, Igbo, Sepedi (Northern Sotho), Shona, Sesotho (Southern Sotho), Setswana, and Tsonga. The specific clinical sections translated were ``clinical knowledge" (299 QA pairs / 8,744 words), ``college medicine" (200 QA pairs / 12,911 words), and ``virology"\footnote{The virology section for Afrikaans, Zulu, Xhosa were also translated, extending the work of \cite{winmmluza2024}} (189 QA pairs / 5,452 words) sections from the ``dev", ``val", and ``test" splits of \texttt{MMLU}. Translators were allowed to use a machine translation tool, but had to manually validate the output and correct errors. Translators were requested to (i) maintain original punctuation in translations (commas, periods), and (ii) utilize clinical expressions in the target language, where possible. Translators were selected if they had prior translation experience as well as subject domain translation experience. Translators were paid \$0.08 USD per-word to translate a total of 21,655 words over a 4-day period. 

\subsection{Evaluation of LLM Performance (Aim 2)}

We applied several state-of-the-art LLMs to the newly translated benchmarks from Aim 1 and measured the performance gap between English and each of the African languages on the benchmarks. We also assessed the performance gap between culturally appropriate and inappropriate benchmark questions in Winogrande. Additional details on the models assessed and the means by which cultural appropriateness labels were generated are provided below.

\subsubsection{The LLM Performance Gap}
We performed evaluations of SOTA LLMs on the translated benchmarks. We also compared LLM performance to English as a reference, as well as pre-existing benchmark translations for Afrikaans, Zulu, and Xhosa. The LLMs evaluated represented models that were either private: gpt-3.5-turbo-1106 (GPT-3.5) \cite{gpt35}, gpt-4-turbo-2024-04-09 (GPT-4) \cite{achiam2023gpt}, gpt-4o-2024-05-13 (GPT-4o) \cite{gpt4o}; public: Llama 3 70B / 8B instruction-tuned \cite{llama3}; small for edge-devices: Phi 3 Mini 4K instruction-tuned \cite{abdin2024phi}; or specialized multilingual: Aya 23 \cite{aryabumi2024aya}, Aya 101 \cite{ustun2024aya}, BLOOMZ 7b1 \cite{muennighoff2023crosslingual}. See Appendix \ref{sec:hyperparamters} for model hyperparameters. For both translated benchmarks, \texttt{Winogrande} (binary choice co-reference resolution task) and the three clinical sections of \texttt{MMLU} (multiple choice medical domain knowledge task), 5-shot accuracy was reported, mimicking \cite{achiam2023gpt}. \texttt{Belebele} (multiple choice reading comprehension task) results were also reported to serve as an additional benchmark (which also covers the languages in this study), and for which 0-shot accuracy was used \cite{bandarkar-etal-2024-belebele}. See Figures \ref{fig:winogrande-prompt}-\ref{fig:belebele-prompt} for the evaluation prompts used for each benchmark.

\subsubsection{Measuring Impact of Cultural Appropriateness} \label{sec:main-cultural-approp}

Although our selected MMLU sections have a cross-cultural focus (i.e. health-related), Winogrande may contain a set of questions that could be considered strange, incoherent, or disrespectful in some cultural contexts (e.g. ``\textit{Jessica thought Sandstorm was the greatest song ever written but Patricia hated it. [Patricia / Jessica] bought a ticket to the jazz concert.}").

We assessed how LLM performance in African languages changed when assessed on culturally appropriate vs. culturally inappropriate questions. More specifically, we generated annotations of translation \textit{appropriateness} using the same \textit{evaluators} of Winogrande translation \textit{quality} (described earlier). The \textit{evaluators} were provided with the translated QA pair and asked ``\textit{For a typical native speaker in a typical conversational context (casual or professional), could the translated sentence be considered strange, incoherent, or disrespectful?"}, with the following options: \textit{(i) ``No, the sentence is typical", (ii) ``Maybe, I'm not sure", (iii) ``Yes, the sentence is strange, incoherent, or disrespectful", or (iv) ``I don't understand the sentence}" \cite{sap2020social}. Of the QA pairs labeled as a ``good" or ``understandable" translation (i.e. QA pairs with at least decent translation \textit{quality}), a QA pair was considered ``culturally appropriate" if both \textit{evaluators} labeled it as ``No, the sentence is typical"; otherwise, it was considered ``culturally inappropriate". To clarify, a ``culturally appropriate" QA pair requires labels for both decent translation \textit{quality} and translation \textit{appropriateness}. Following our definition, 74.2\% of translated QA pairs were labeled culturally appropriate, while 20.6\% of translated QA pairs were labeled culturally inappropriate. An example of the task can be seen in Figure \ref{fig:cultural-appropriateness-survey} and the cultural annotation results are shown in Table \ref{table:annotator_confusion_2}. In addition, QA pair examples of translation \textit{quality} and \textit{appropriateness} annotation combinations are shown in Tables \ref{table:annotation_examples_af}-\ref{table:annotation_examples_zu}\footnote{Languages that use non-Latin characters (Amharic, Bambara, Igbo) cannot be rendered; however, all annotations are available in the GitHub repository associated with this work.}. We provide a detailed discussion of the inter-annotator agreements, as well as the Cohen's Kappa and Fleiss' Kappa scores in Section \ref{sec:cultural_appropriateness} of the Appendix.

To determine the effect of cultural appropriateness on LLM performance, the GPT-family models were evaluated out-of-the-box on the Winogrande test set, split by the collected human annotations of appropriateness in each language. Additionally, to account for the possibility of the human annotations capturing appropriateness in English and not just the target language, the same splits of data for each language were evaluated in English to be used as a baseline. Then, if we observe a greater lift achieved in the target language than in English for a given split of the data, we can conclude that the human annotations do indeed reflect a difference in culture-specific appropriateness and also impact the LLM's performance.

% , whereby a QA pair was considered ``appropriate"  if either both annotators marked the translation as ``Yes, the sentence is strange, incoherent, or disrespectful".

%were also asked to assess the cultural appropriateness of the Winogrande QA pairs. We assumed that the annotation task would take 20 seconds per QA pair, or 20.41 working hours per language. Thus, we allocated \$200 per annotator (\$9.80 / hour).

% Annotators were asked to evaluate the previously obtained human translations of Winogrande for quality and appropriateness. An example of the task can be found in Figure \ref{fig:cultural-appropriateness-survey}. Speakers were presented with the original English and human-translated Winogrande QA pair with the correct answers filled in and asked two questions. The first question was to rate the quality of the translation according to three options: (i) ``Good translation", (ii) ``Incorrect, but someone could the understand the idea", and (iii) ``Completely wrong" \cite{gt_tarzan}. %\cite{bapna2022building}. 

%Annotating the perceived translation quality (i.e. the first question), allows us to account for quality as a potentially confounding factor in the cultural appropriateness annotations (i.e. the second question).

\subsection{Enhancement of LLM Performance (Aim 3)}
After measuring the LLM performance gap in African languages (Aim 2), we explored various fine-tuning strategies to reduce it. This included adjusting the fine-tuning data based on domain (Winogrande vs. MMLU college medicine), language (mono-lingual vs. cross-lingual contexts),  data quality (low vs. high), and the volume of training samples (25\%, 50\%, 75\%, 100\%). Additional details on the specific experiments follow. All experiments in this section were performed with Llama 3 70B, which was: (i) the best performing open-source model, (ii) was possible to fine-tune, and (iii) was fiscally feasible. Moreover, all experiments in this section used the same fine-tuning prompts, which can be found in Figures \ref{fig:winogrande-ft-prompt}-\ref{fig:mmlu-ft-prompt}.

\subsubsection{Fine-tuning with Varying Languages and Domains}
We determined the effects of mono- vs. cross-lingual fine-tuning on LLM performance by comparing performance gains of Lllama 3 70B on our benchmarks after tuning. More specifically, we report the mono- and cross-lingual performance of Llama 3 70B after fine-tuning using two of the translated benchmarks (Winogrande small train split, or MMLU ``college medicine" section) and evaluating on four test sets (Winogrande test split, MMLU ``clinical knowledge" section, MMLU ``virology" section, and Belebele).

\subsubsection{Fine-tuning with Varying Data Quality and Quantity}
To determine the effects of data quality and quantity on LLM performance in low-resource settings, we utilized GPT-4o to generate quality scores for each QA pair in our fine-tuning datasets (i.e. LLM-as-an-Annotator). More specifically, for each QA pair in the fine-tuning dataset, GPT-4o was prompted to provide a score of 1 to 10 based on the usefulness of a QA pair for fine-tuning an LLM to improve its performance on a target evaluation benchmark (1 is the least useful, while 10 is the most useful. See Figure \ref{fig:gpt-4o-quality-prompt}). The LLM-as-an-Annotator was run three times to capture potential variability in outputs. The fine-tuning dataset was then divided into tertiles according to the average of three quality scoring runs. The tertile with the highest scores was designated as the ``high quality" set, while the tertile with the lowest scores was labeled as the ``low quality" set. To evaluate the impact of data volume on fine-tuning LLM performance, the quality sets were randomly sampled at increments of 0\% (i.e. no tuning), 25\%, 50\%, 75\%, and 100\%.

% More specifically, for each QA pair in a given fine-tuning dataset, GPT-4o was prompted to provide a score from 1-10 based on the row's estimated usefulness for improving performance on a target evaluation benchmark 
% that the fine-tuned model would later be evaluated on. GPT-4o was prompted to score each row three times, from which the average score was taken. 
% The fine-tuning dataset was then split into thirds based on these average quality scores and took the third with the highest scores as the ``High Quality set" as well as the third with the lowest scores as the ``Low Quality set" (for the GPT-4o prompt used, see Figure \ref{fig:gpt-4o-quality-prompt}). 

% The quality sets were then randomly sampled at increments of 0\% (no fine-tuning), 25\%, 50\%, 75\%, and 100\% (using all data in that set) to observe the effects of quantity and quality on fine-tuning performance simultaneously. Note that for these ``quality vs. quantity" experiments we only performed mono-lingual experiments, where the fine-tuning dataset and evaluation benchmark were provided in the same language.

% \hl{When fine-tuning, the following hyper-parameters were applied to X, Y, X ...}.

\section{Results}
In this section, we provide results of five experiments that support our three aims: (Aim 1) benchmark translation, (Aim 2) evaluation of LLM performance, and (Aim 3) enhancement of LLM performance. More specifically, the sub-sections below provide assessments of: (1) benchmark translation fidelity, (2) ``out-of-the-box" LLM performance on the translated benchmarks, (3) ``out-of-the-box" LLM performance on the culturally ``appropriate" vs. ``inappropriate" subsets, (4) fine-tuned LLM performance using mono- and cross-lingual data, and (5) fine-tuned LLM performance using varying data quality and quantity. 

\begin{table}[b!]
\centering
% \resizebox{\linewidth}{!}{%
\begin{tabular}{|lccccc|}
\hline
% \textbf{Model (params)}  
\multirow{2}{*}{\textbf{Model}}
% \begin{tabular}[c]{@{}l@{}} \textbf{Model}\end{tabular}}
& \multirow{2}{*}{\textbf{Bele}} % \textbf{Bele}                       
& \multirow{2}{*}{\textbf{Wino}} % \textbf{Wino}
& \multicolumn{3}{c|}{\textbf{MMLU}} \\

& & 
& \begin{tabular}[c]{@{}c@{}}CM\end{tabular}
& \begin{tabular}[c]{@{}c@{}}CK\end{tabular}
& \begin{tabular}[c]{@{}c@{}}Vir.\end{tabular} \\ \hline

% \textbf{GPT-4o (-) Eng.}    
\multicolumn{6}{|c|}{\textbf{Baseline Performance} (English language)}  \\ \hline
\begin{tabular}[c]{@{}l@{}}GPT-4o  \end{tabular}
& \cellcolor[HTML]{58BC8B}\textbf{95.9} 
& \cellcolor[HTML]{57BB8A}\textbf{83.9} 
& \cellcolor[HTML]{57BB8A}\textbf{84.4}                                    
& \cellcolor[HTML]{57BB8A}\textbf{89.8}                                      
& \cellcolor[HTML]{57BB8A}\textbf{60.2}  \\ \hline

\multicolumn{6}{|c|}{\textbf{Average Performance} (all 11 African languages)}  \\ \hline
GPT-4o 
& \cellcolor[HTML]{87CFAB}76.0  
& \cellcolor[HTML]{B4E1CB}64.8  
& \cellcolor[HTML]{8AD0AD}66.6  
& \cellcolor[HTML]{89D0AD}70.6 
& \cellcolor[HTML]{91D3B2}48.2  \\

GPT-4                                               
& \cellcolor[HTML]{96D5B6}69.6          
& \cellcolor[HTML]{C7E9D8}60.9          
& \cellcolor[HTML]{A7DCC2}56.2                                             
& \cellcolor[HTML]{A3DABF}60.7                                               
& \cellcolor[HTML]{9BD7BA}46.0  \\

Aya 101                                                    
& \cellcolor[HTML]{B0DFC8}58.4          
& \cellcolor[HTML]{F6FBF9}50.5          
& \cellcolor[HTML]{E0F3EA}35.7                                             
& \cellcolor[HTML]{D7EFE3}36.1                                               
& \cellcolor[HTML]{DAF0E6}32.0  \\

Llama 3 70B \,\,\,\,\,\,
& \cellcolor[HTML]{D4EEE1}41.2          
& \cellcolor[HTML]{F5FBF8}50.6          
& \cellcolor[HTML]{E4F5ED}35.9                                             
& \cellcolor[HTML]{E0F3EA}40.6                                               
& \cellcolor[HTML]{DCF1E7}32.3 \\

Aya 23                                                       
& \cellcolor[HTML]{DEF2E8}38.8          
& \cellcolor[HTML]{F8FCFA}51.2          
& \cellcolor[HTML]{E5F5ED}34.3                                             
& \cellcolor[HTML]{E5F5ED}34.9                                               
& \cellcolor[HTML]{DEF2E8}28.4 \\

GPT-3.5                                                
& \cellcolor[HTML]{E5F5ED}36.2         
& \cellcolor[HTML]{F9FDFB}51.2          
& \cellcolor[HTML]{E1F3EA}34.6                                             
& \cellcolor[HTML]{E6F5EE}37.0                                               
& \cellcolor[HTML]{EFF9F4}32.8  \\
Llama 3 8B     
& \cellcolor[HTML]{E5F5ED}36.3          
& \cellcolor[HTML]{F9FDFB}50.4          
& \cellcolor[HTML]{ECF8F2}31.9                                             
& \cellcolor[HTML]{E3F4EB}35.3                                               
& \cellcolor[HTML]{F2FAF6}27.3  \\
Bloomz 7B                                                       
& \cellcolor[HTML]{EAF7F0}34.2          
& \cellcolor[HTML]{F9FDFB}49.1          
& \cellcolor[HTML]{F0F9F5}28.9                                             
& \cellcolor[HTML]{ECF8F2}31.0                                               
& \cellcolor[HTML]{F5FBF8}25.8  \\

Phi 3 3B                                       
& \cellcolor[HTML]{EEF8F3}32.2          
& \cellcolor[HTML]{FBFEFD}50.7          
& \cellcolor[HTML]{F4FBF8}30.3                                             
& \cellcolor[HTML]{F0F9F5}32.4                                               
& \cellcolor[HTML]{FCFEFD}27.8  \\

\textit{Random}                                                   
& \cellcolor[HTML]{FFFFFF}\textit{25.0} 
& \cellcolor[HTML]{FFFFFF}\textit{50.0} 
& \cellcolor[HTML]{FFFFFF}\textit{25.0}                                    
& \cellcolor[HTML]{FFFFFF}\textit{25.0}                                      
& \cellcolor[HTML]{FFFFFF}\textit{25.0}\\

\hline
 \multicolumn{6}{|c|}{{ \textbf{Performance Gap} (English - African languages)}} \\ \hline
% { {GPT 4o (Eng. - Afr.)}} 
\begin{tabular}[c]{@{}l@{}}GPT-4o\end{tabular}
& { {19.9}} 
& { {19.1}}  
& {{17.8}}                                     
& {{19.2}}                                       
& {{12.0}} \\ \hline
\end{tabular}
\caption{\textbf{LLM Performance Gap Between English and African Languages}. \textit{The table displays SOTA out-of-the-box model performance averaged across 11 African languages. The best performing model (GPT-4o) yields between 12.0\% and 19.9\% absolute difference in performance between English and the average of 11 African languages. Bele: Belebele, Wino: Winogrande, CM: College Medicine, CK: Clinical Knowledge, Vir.: Virology}} \label{table:perf-results-ootb}
\end{table}

\begin{table*}[p]
\centering
% \small
\begin{tabular}{l|llllllllllll@{}l@{}}
\hline \hline
\rule{0pt}{4mm}& \multicolumn{12}{c}{\texttt{Winogrande}}  \\
                & \textbf{en} & \textbf{af} & \textbf{zu} & \textbf{xh} & \textbf{am} & \textbf{bm} & \textbf{ig} & \textbf{nso} & \textbf{sn} & \textbf{st} & \textbf{tn} & \textbf{ts} \\ \hline
\rule{0pt}{4mm}{GPT-4o} & \ul{83.9} & \ul{79.7} & \ul{68.3} & \ul{65.9} & \ul{59.4} & {50.2} & \ul{60.7} & \ul{64.1} & \ul{69.5} & \ul{67.4} & \ul{64.7} & \ul{62.6} \\
{GPT-4} & {83.5} & {77.0} & {64.2} & {62.3} & {51.0} & {50.7} & {58.7} & {58.8} & {65.6} & {63.8} & {59.9} & {57.7} \\
{GPT-3.5} & {59.6} & {55.0} & {50.8} & {52.2} & {51.3} & {50.4} & {51.9} & {50.2} & {51.6} & {49.2} & {51.5} & {49.6} \\
{Llama 3 70B IT} & {61.2} & {51.0} & {50.4} & {50.8} & {50.8} & {50.5} & {50.5} & {50.5} & {50.4} & {50.4} & {50.5} & {50.4} \\
{Llama 3 8B IT} & {52.0} & {50.7} & {50.4} & {50.4} & {50.3} & {50.4} & {50.4} & {50.4} & {50.4} & {50.4} & {50.4} & {50.4} \\
{Phi 3 Mini 4K IT} & {64.7} & {51.6} & {50.2} & {51.3} & {50.2} & {50.0} & {51.0} & {49.7} & {51.9} & {49.5} & {50.9} & {50.9} \\
{Aya 23 35B} & {68.7} & {56.5} & {49.6} & {51.8} & {51.3} & {50.8} & {50.6} & {51.4} & {50.6} & {50.0} & {49.8} & {50.5} \\
{Aya 101} & {49.5} & {51.1} & {49.1} & {51.2} & {51.2} & \ul{52.0} & {50.5} & {49.0} & {51.0} & {50.5} & {50.8} & {49.6} \\
{BLOOMZ 7b1} & {48.6} & {50.3} & {48.7} & {48.8} & {49.3} & {50.0} & {48.8} & {47.9} & {48.6} & {49.6} & {49.2} & {49.1} \\
\hline
\rule{0pt}{4mm}& \multicolumn{12}{c}{\texttt{MMLU College Medicine}} \\
{GPT-4o} & \ul{84.4} & \ul{84.4} & \ul{72.8} & \ul{78.0} & \ul{67.1} & \ul{38.2} & \ul{58.4} & \ul{66.5} & \ul{76.9} & \ul{67.1} & \ul{63.6} & \ul{60.1} \\
{GPT-4} & {78.0} & {79.8} & {61.3} & {61.3} & {46.8} & {30.1} & {50.9} & {53.8} & {68.8} & {54.9} & {57.8} & {53.2} \\
{GPT-3.5} & {63.6} & {56.1} & {32.9} & {37.6} & {24.3} & {30.6} & {28.9} & {34.1} & {35.8} & {32.4} & {32.9} & {34.7} \\
{Llama 3 70B IT} & {76.9} & {68.2} & {35.8} & {40.5} & {36.9} & {24.3} & {33.5} & {27.2} & {38.2} & {28.3} & {34.1} & {29.5} \\
{Llama 3 8B IT} & {60.1} & {44.5} & {31.8} & {37.0} & {16.7} & {30.1} & {24.9} & {32.4} & {38.7} & {34.7} & {27.2} & {32.9} \\
{Phi 3 Mini 4K IT} & {66.5} & {39.9} & {30.6} & {28.9} & {28.0} & {31.2} & {28.9} & {28.3} & {28.3} & {27.7} & {34.1} & {30.1} \\
{Aya 23 35B} & {62.4} & {49.1} & {35.8} & {32.9} & {34.7} & {28.3} & {31.8} & {35.8} & {32.4} & {36.4} & {34.7} & {24.9} \\
{Aya 101} & {42.8} & {40.5} & {35.3} & {32.9} & {35.8} & {31.2} & {31.2} & {34.7} & {42.2} & {38.2} & {34.7} & {35.8} \\
{BLOOMZ 7b1} & {36.4} & {34.1} & {30.1} & {28.3} & {26.0} & {28.9} & {26.6} & {27.2} & {30.1} & {29.5} & {27.7} & {29.5} \\
\hline
\rule{0pt}{4mm}& \multicolumn{12}{c}{\texttt{MMLU Clinical Knowledge}} \\
{GPT-4o} & \ul{89.8} & \ul{87.2} & \ul{79.6} & \ul{78.5} & \ul{70.6} & \ul{40.0} & \ul{63.0} & \ul{72.5} & \ul{80.4} & \ul{69.4} & \ul{71.3} & \ul{64.5} \\
{GPT-4} & {84.2} & {81.9} & {70.2} & {68.3} & {54.0} & {34.7} & {54.7} & {58.1} & {67.2} & {64.2} & {58.9} & {55.8} \\
{GPT-3.5} & {72.5} & {62.6} & {39.2} & {37.4} & {30.2} & {31.3} & {34.3} & {33.2} & {41.1} & {34.3} & {30.6} & {32.8} \\
{Llama 3 70B IT} & {82.3} & {71.3} & {39.2} & {38.5} & {33.6} & {32.5} & {33.6} & {37.0} & {43.4} & {43.8} & {39.2} & {34.0} \\
{Llama 3 8B IT} & {69.1} & {45.3} & {36.6} & {34.7} & {24.9} & {32.1} & {36.2} & {35.5} & {39.2} & {39.2} & {33.2} & {31.7} \\
{Phi 3 Mini 4K IT} & {70.6} & {40.8} & {29.1} & {30.6} & {32.1} & {27.2} & {33.6} & {32.8} & {33.2} & {30.9} & {35.8} & {29.8} \\
{Aya 23 35B} & {69.4} & {55.1} & {32.5} & {33.6} & {29.1} & {33.2} & {32.5} & {37.4} & {32.5} & {34.3} & {30.9} & {32.5} \\
{Aya 101} & {45.3} & {42.6} & {38.1} & {35.5} & {38.9} & {28.3} & {34.7} & {36.6} & {44.2} & {37.0} & {30.6} & {30.9} \\
{BLOOMZ 7b1} & {44.9} & {33.2} & {30.6} & {33.2} & {28.3} & {32.5} & {26.8} & {34.7} & {30.6} & {32.1} & {27.2} & {31.3} \\
\hline
\rule{0pt}{4mm}& \multicolumn{12}{c}{\texttt{MMLU Virology}} \\
{GPT-4o} & \ul{60.2} & {55.4} & \ul{51.2} & \ul{50.6} & \ul{51.2} & {32.5} & {44.0} & \ul{50.0} & \ul{53.0} & \ul{47.6} & \ul{48.8} & \ul{45.8} \\
{GPT-4} & {59.6} & \ul{59.0} & {49.4} & \ul{50.6} & {45.8} & {31.9} & \ul{44.6} & {46.4} & {49.4} & {45.2} & {44.0} & {39.8} \\
{GPT-3.5} & {51.8} & {42.8} & {30.7} & {37.3} & {29.5} & \ul{36.1} & {28.3} & {33.7} & {31.3} & {30.1} & {30.1} & {30.7} \\
{Llama 3 70B IT} & {53.6} & {46.4} & {28.9} & {27.1} & {36.1} & {35.5} & {34.9} & {26.5} & {31.3} & {29.5} & {28.3} & {31.3} \\
{Llama 3 8B IT} & {51.8} & {38.6} & {28.9} & {27.1} & {28.3} & {20.5} & {24.1} & {27.7} & {22.3} & {25.3} & {27.7} & {29.5} \\
{Phi 3 Mini 4K IT} & {48.8} & {30.1} & {27.1} & {25.3} & {27.1} & {25.3} & {32.5} & {27.7} & {28.3} & {24.1} & {25.3} & {32.5} \\
{Aya 23 35B} & {49.4} & {42.2} & {24.7} & {26.5} & {39.2} & {27.7} & {19.9} & {28.3} & {28.3} & {26.5} & {25.9} & {23.5} \\
{Aya 101} & {33.1} & {34.9} & {33.7} & {31.3} & {36.1} & {21.7} & {30.1} & {35.5} & {31.3} & {34.9} & {33.1} & {29.5} \\
{BLOOMZ 7b1} & {38.0} & {26.5} & {26.5} & {24.7} & {21.1} & {21.1} & {26.5} & {24.1} & {31.3} & {27.7} & {27.7} & {26.5} \\
\hline
\rule{0pt}{4mm}& \multicolumn{12}{c}{\texttt{Belebele}} \\
{GPT-4o} & {95.9} & \ul{94.4} & \ul{79.7} & \ul{82.8} & \ul{78.0} & {38.3} & \ul{71.3} & \ul{77.2} & \ul{81.0} & \ul{80.0} & \ul{77.0} & \ul{76.8} \\
{GPT-4} & \ul{96.1} & {93.6} & {75.9} & {76.3} & {61.3} & {37.9} & {64.4} & {64.6} & {77.6} & {76.3} & {67.6} & {70.0} \\
{GPT-3.5} & {86.2} & {75.7} & {33.8} & {31.8} & {30.0} & {30.6} & {29.2} & {32.3} & {34.6} & {32.1} & {31.7} & {36.3} \\
{Llama 3 70B IT} & {94.6} & {84.8} & {36.7} & {36.8} & {35.1} & {35.6} & {35.9} & {37.1} & {38.1} & {35.9} & {36.4} & {40.7} \\
{Llama 3 8B IT} & {80.0} & {69.4} & {32.6} & {33.0} & {30.2} & {32.0} & {35.7} & {32.9} & {33.9} & {32.2} & {34.0} & {33.4} \\
{Phi 3 Mini 4K IT} & {89.2} & {52.6} & {28.3} & {29.1} & {28.7} & {31.7} & {28.8} & {31.6} & {29.1} & {30.6} & {29.2} & {34.3} \\
{Aya 23 35B} & {93.6} & {83.6} & {33.2} & {35.6} & {29.0} & {34.6} & {28.7} & {34.6} & {37.8} & {35.6} & {36.8} & {37.4} \\
{Aya 101} & {79.4} & {77.4} & {59.3} & {60.7} & {67.7} & \ul{40.7} & {50.7} & {59.3} & {57.8} & {59.9} & {57.8} & {51.0} \\
{BLOOMZ 7b1} & {79.0} & {36.7} & {35.8} & {35.0} & {24.3} & {31.6} & {31.2} & {34.1} & {36.7} & {34.7} & {35.9} & {39.7} \\ \hline \hline
\end{tabular}
\caption{\textbf{Results of State-of-the-Art Models on Human-Translated Benchmarks}. \textit{Top state-of-the-art models were evaluated out-of-the-box (no fine-tuning) on the translated \texttt{Winogrande} (binary choice co-reference resolution task), the translated three clinical sections of \texttt{MMLU} (multiple choice medical domain knowledge task), as well a pre-existing benchmark \texttt{Belebele} (reading comprehension task). Results are provided for 11 low-resource African languages of focus: Afrikaans (af), Zulu (zu), Xhosa (xh) (datasets for these three languages were sourced from \cite{winmmluza2024}), Amharic (am), Bambara (bm), Igbo (ig), Sepedi (nso), Shona (sn), Sesotho (st), Setswana (tn), and Tsonga (ts). Results on English (en) are also provided as a reference. All numbers are performance accuracy, where \texttt{MMLU} and \texttt{Winogrande} numbers are 5-shot while \texttt{Belebele} numbers are 0-shot. Best performance is indicated with an \ul{underline}. Overall, GPT-4o is best performing model across benchmarks.}} 
\label{table:perf-results}
\end{table*}

\subsection{Benchmark Translation Fidelity}
MMLU translations (by Translated.com) took 15 days to complete and cost \$11,232.29 USD. Winogrande translation (by Upwork.com \textit{translators}) took between 3 days (Setswana) and 9 days (Sesotho) and had a cumulative cost of \$15,126.40 USD. Winogrande translation verification (by Upwork.com \textit{validators}) took between 4 days (Setswana) and 9 days (Shona) and had a cumulative cost of \$9,179.37 USD. Winogrande translation assessment surveys for quality and appropriateness (by Upwork.com \textit{evaluators}) took between 3 days (Shona) and 8 days (Bambara) and had a cumulative cost of \$4,680.72 USD. 

As seen in Figure \ref{fig:translation-checks}, corrections to the original translations varied by language: from 4.9\% (for Sesotho) to 65.3\% (for Shona) of the QA pairs. As seen in Table \ref{table:annotator_confusion_1}, 94.7\% of the validated / corrected translations were considered a good / understandable translation (``Good translation" or ``Incorrect, but someone could understand the idea") by at least one \textit{evaluator}, only 5.3\% were considered wrong (``Completely wrong") by either \textit{evaluator}, and a mere 0.2\% were considered wrong (``Completely wrong") by both \textit{evaluators}. Thus, we have reason to believe that (while not perfect) the translations are sufficient to measure the LLM performance gap between English and the African languages.

% \subsection{SOTA, Mono- and Cross- lingual Reasoning Evaluations}
% Tuka: Adam had mentioned prompt templates and N-shot evaluations. Need to check.
\subsection{Measuring the LLM Performance Gap}
Here, we provide baseline performance results for SOTA LLM models on three benchmarks: \texttt{Winogrande}, the three clinical sections of \texttt{MMLU}, and \texttt{Belebele}. The average 5-shot (0-shot for Belebele) accuracy scores across all languages and benchmarks are reported in Table \ref{table:perf-results-ootb}, with English for reference. The best performing LLM (GPT-4o) had a performance gap ranging from 12.0\% to 19.9\% absolute between English and the average of the 11 African languages. For individual languages, a breakdown of the baseline performance of SOTA LLM models on the same three benchmarks can be seen in Table \ref{table:perf-results}. We observed that among African languages, GPT-4o consistently performed the best in Afrikaans and the worst in Bambara, with performance gaps between the two ranging from 22.9\% (MMLU Virology) to 56.1\% (Belebele) absolute across benchmarks, suggesting that there is considerable variance in the performance of SOTA models across individual African languages.

Additionally, we provide baseline performance results for SOTA models when using machine-translated versions of the benchmarks (see Tables \ref{table:perf-results-mt-wino}-\ref{table:perf-results-mt-bele} for a breakdown of performance by language and benchmark). The performance difference between machine-translated queries and directly using an LLM in non-English languages was not always significant. Hence, our findings suggest that native language LLMs may not be needed in some contexts (see Appendix Section \ref{sec:ootb-eval-mt} for a more detailed discussion). 

In Figure \ref{fig:out-of-box-correlation}, we present correlations between LLM performance across language pairs. English was the least correlated with the other languages. The seven Bantu languages and Igbo (Volta-Niger) had the highest correlation values (see Figure \ref{fig:languagemap} for language families). Bambara and Amharic were least correlated with the other languages, reflecting some combination of the different grammar paradigms of the Mande and Semitic language families, the difference in data quality, and (Amharic) script characters.

%Moreover, large multilingual models like GPT-4o, which have likely been trained on significantly more low-resource language data (e.g. data in native African languages), performed well at language understanding and question answering in low-resource languages, but, for many types of questions, performed even better when the question posed was machine-translated back into English (see Appendix Section \ref{sec:ootb-eval-mt}). 

% We observe that GPT-4o is the best performing model on both Winogrande and MMLU, with 5-shot accuracy ranging between 59.4\% (Amharic) and 69.5\% (Shona) on Winogrande (excluding Bambara), and between 45.8\% (Tsonga - virology section) and 80.4\% (Shona - clinical knowledge section) on MMLU (excluding Bambara, and Igbo). Overall, models showed the lowest performance on Bambara, underperforming by 9.2\% absolute from the next lowest performing language on Winogrande (59.4\%: Amharic - 50.2\%: Igbo), and underperforming by at least 11.5\% on MMLU-Clinical (44.0\%: Igbo - 32.5\%: Bambara). 

% of \hl{59.4\%} (Amharic) and \hl{77.3\%}, respectively (excluding Bambara). On the African languages, GPT-4o performed best on \hl{Afrikaans, 78.9\% and 85.8\%, respectively on \texttt{Winogrande-ZA} and \texttt{MMLU-Clinical-ZA}. For \texttt{Winogrande-ZA}, Xhosa was the worst performing (64.4\%), while for \texttt{MMLU-Clinical-ZA} Zulu was the worst performing (76.9\%)}.

\begin{figure}[b!]
    \centering
    % \includesvg[width=0.95\columnwidth]{./images/cultural_appropriateness_plot_v2.svg}
    \includegraphics[width=0.95\columnwidth]{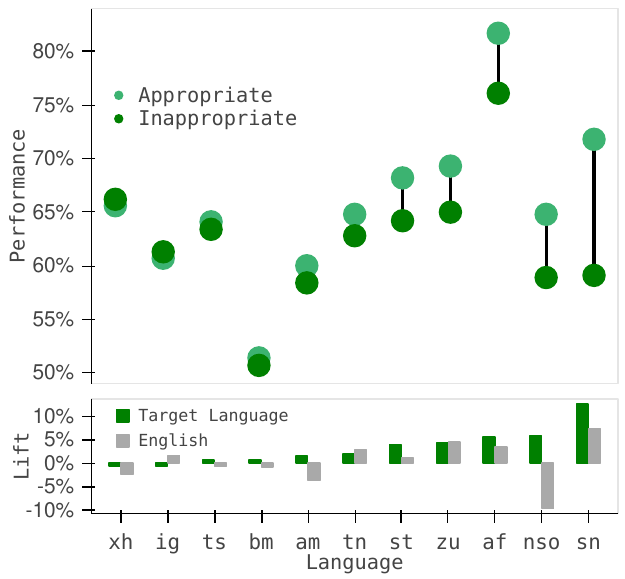}
    \caption{\textbf{GPT-4o Winogrande Performance on ``appropriate" vs. ``inappropriate" Data.} \textit{GPT-4o was evaluated on Winogrande (test set) out-of-the-box in each target language and in English. Top plot: the absolute performance on QA pairs considered culturally ``appropriate" and ``inappropriate" according to native speakers. Bottom plot: performance lifts for each language (green) and in English (grey), using the same annotations. QA Pair was defined as ``appropriate" when either annotator marked the cultural appropriateness of the question as ``typical". Only QA pairs where both annotators reported that the translation quality was ``good" or ``understandable" were considered. Language codes are: Xhosa (xh), Igbo (ig), Tsonga (ts),  Bambara (bm), Amharic (am), Setswana (tn), Sesotho (st), Zulu (zu), Afrikaans (af), Sepedi (nso), Shona (sn). See Table \ref{table:appropriateness-3} for a breakdown of performance by language.  See Figure \ref{fig:appropriateness_boxplots} distributions when repeated random samples of the same size as the appropriate and inappropriate counts for each target language are drawn.}}
    \label{fig:cultural-appropriateness-v2}
\end{figure}

\subsection{Impact of ``Appropriateness" on Performance Gap}

Here, we assess the impact of cultural appropriateness on LLM performance; we evaluated the best-performing model (GPT-4o) out-of-the box on the Winogrande test set, splitting the dataset by the human annotations in each language, and observing the lift obtained for the ``appropriate" QA pairs compared to the ``inappropriate" QA pairs (see Figure \ref{fig:cultural-appropriateness-v2}). When comparing the performance lift of GPT-4o in English on the same annotations, we see that 7/11 African languages have a higher lift in the target language, indicating that GPT-4o performs better on culturally appropriate QA pairs for those languages (that is, the lift is not due to the question itself being odd, independent of language; see Figure \ref{fig:inappropriatenss_kappa_across_languages}). The performance of GPT-4o on appropriate over inappropriate Winogrande subsets ranged from -0.6\% (Xhosa) to +12.6\% (Shona), with an average difference of +3.3\% across all languages. The performance lift of GPT-4o when using English evaluations as a baseline ranged from -2.3\% (Igbo) to +15.6\% (Sepedi), with an average difference of +3.0\% across all languages. A breakdown of the performance by language can be found in Appendix Tables \ref{table:appropriateness-interrater} to \ref{table:appropriateness-3}. 

% \ref{table:appropriateness-interrater}, \ref{table:appropriateness-1}, \ref{table:appropriateness-2}, and \ref{table:appropriateness-3}.

% For each language, the difference between the target language lift and the English lift is as follows (using GPT-4o, from Table \ref{table:appropriateness-3}): 1.8\% (xh), -2.3\% (ig), 1.3\% (ts), 1.6\% (bm), 5.4\% (am). -0.8\% (tn), 2.9\% (st), -0.2\% (zu), 2.2\% (af), 15.7\% (nso), 5.3\% (sn).

\subsection{Mono- and Cross-lingual Evaluations}

%The second set of experiments report the performance of Llama 3 70B when fine-tuned using the translated benchmarks (Winogrande training small and MMLU ``college medicine" section) and evaluate mono- and cross-lingual performance across four test sets (Winogrande test, MMLU ``clinical knowledge" section, MMLU ``virology" section, and Belebele test). The results are shown in Figure \ref{fig:boxplot-crosslingual} (see more detailed breakdowns in Appendix Tables \ref{table:perf-crosslingual-llama3-tr-wino-ts-wino} to \ref{table:perf-crosslingual-llama3-tr-mmlu-ts-bele}). Across the 11 languages, the average mono-lingual gain was 5.6\%. Mono-lingual performance gains when fine-tuning with MMLU (college medicine) were greatest on MMLU clinical knowledge (17.4\% on average), followed by Belebele (9.4\% on average), MMLU virology (3.5\% on average), and Winogrande (1.2\% on average). When fine-tuning with Winogrande, the greatest gains were observed when evaluating LLMs on Belebele (6.4\% on average), followed by Winogrande (2.7\% on average), MMLU virology (2.6\% on average), and MMLU clinical knowledge (1.2\% on average).

Here, we assess how mono- and cross-lingual fine-tuning impacts LLM performance on our selected benchmarks. The results are shown in Figure \ref{fig:boxplot-crosslingual} (see more detailed breakdowns in Appendix Tables \ref{table:perf-crosslingual-llama3-tr-wino-ts-wino} to \ref{table:perf-crosslingual-llama3-tr-mmlu-ts-bele}). Across the 11 languages, the average mono-lingual gain was 5.6\%\footnote{Average individual gains of all fine-tuned models above the baselines in the \emph{same} language as the fine-tuning language.}. Mono-lingual performance gains when fine-tuning with MMLU college medicine were greatest when evaluating on MMLU clinical knowledge (17.4\% on average), followed by Belebele (9.4\% on average), MMLU virology (3.5\% on average), and Winogrande (1.2\% on average). When fine-tuning with Winogrande, the greatest gains were observed when evaluating LLMs on Belebele (6.4\% on average), followed by Winogrande (2.7\% on average), MMLU virology (2.6\% on average), and MMLU clinical knowledge (1.2\% on average).

Cross-lingual evaluations yielded a similar trend as mono-lingual results; that is, mono-lingual gains (5.6\% on average) also yielded corresponding cross-lingual gains (2.9\% on average)\footnote{Average individual gains of all fine-tuned models above the baselines in languages \emph{other than} the fine-tuning language.}, although to a lesser degree (see Figure \ref{fig:boxplot-crosslingual}). At the language-level, when mono-lingual gains for a given language was observed, then most models trained in other languages provided gains (i.e. cross-lingual transfer).

\begin{figure}[t!] % 'h' for "here", other options: t (top), b (bottom), p (page of floats), etc.
    \centering  % Center the figure
    \includegraphics[width=0.49\textwidth]{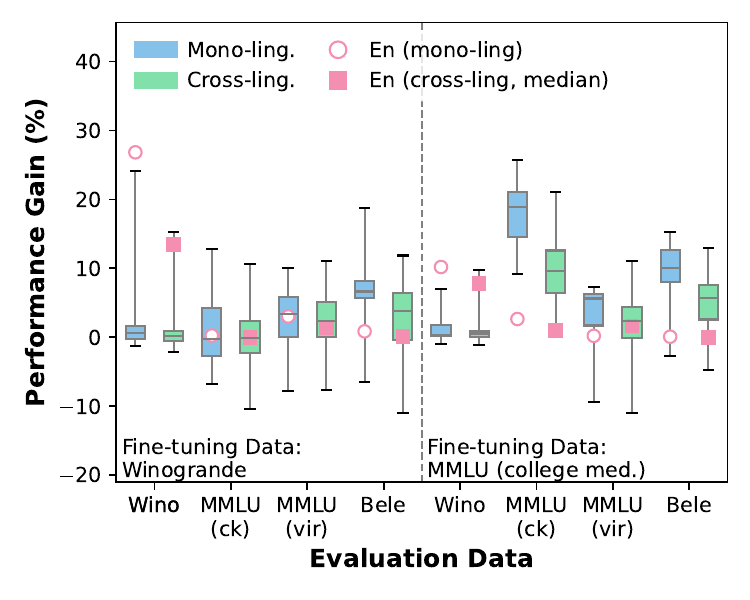}
    \caption{\textbf{Mono- and Cross-lingual LLM Performance Gains.} \textit{The figure displays boxplots of performance gains when fine-tuning with either the translated Winogrande train set (left) or MMLU college medicine section (right). The fine-tuned models were evaluated across 4 datasests (x-axis) for mono-lingual gains (blue) across 11 African languages, and cross-lingual gains (green) across 110 African language pairs. The most significant gains were with models fine-tuned with MMLU college medicine and evaluated on MMLU clinical knowledge. Wino: Winogrande, ck: clinical knowledge, vir: virology, Bele: Belebele. En: English.}}
    \label{fig:boxplot-crosslingual}
\end{figure}

\subsection{Data Quality and Quantity Evaluations}
\begin{figure}[t!] % 'h' for "here", other options: t (top), b (bottom), p (page of floats), etc.
    \centering  % Center the figure
    \includegraphics[width=0.49\textwidth]{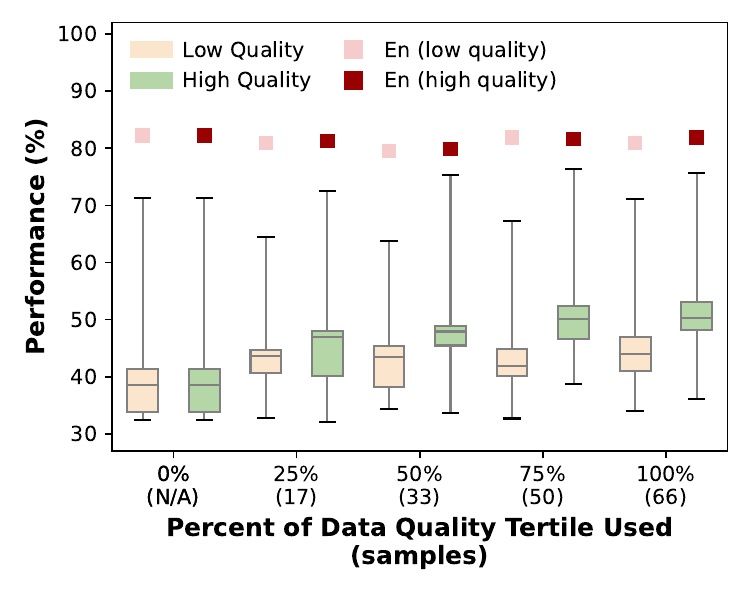}
    \caption{\textbf{LLM Performance Across Quality and Quantity Combinations.} \textit{The figure displays LLM performance when fine-tuning by data quality and quantity, using MMLU college medicine and evaluating on MMLU clinical knowledge (which had the greatest mono-lingual gains from Figure \ref{fig:boxplot-crosslingual}). The quality of samples was rated using GPT-4o LLM-as-an-Annotator scores. The lowest tertile and highest tertile were defined as low (yellow) and high (green) quality samples, respectively, and were used to fine-tune Llama 3 70B IT. Boxplots display performance across 11 African languages. English (En) is provided as a reference (red). Overall, the use of high-quality fine-tuning data over low-quality fine-tuning data improved performance for African languages. See Table \ref{table:quality-x-quantity-ts-mmlu-ck} for a breakdown by language.}}
    \label{fig:quality-x-quantity-summary}
\end{figure}

Here, we evaluate the impact of both the quality and quantity of fine-tuning data on African language LLM performance. Results across the 11 African languages for the highest performing mono-lingual fine-tuning experiments (fine-tuning on MMLU college medicine and evaluating on MMLU clinical knowledge) are shown in Figure \ref{fig:quality-x-quantity-summary}. There were increasing performance gains with increasing data sizes of 2.3\% on average when data sizes were doubled (from 33 to 66 samples), in either quality sets. When data were split between high-quality and low-quality data, the average gain from the data quality split was 5.4\% on average. The complete breakdown of results by language for fine-tuning on Winogrande and MMLU, and evaluating on the four benchmarks is listed in Appendix Tables \ref{table:quality-x-quantity-ts-wino} to \ref{table:quality-x-quantity-ts-bele}.

% if a language-task pair did not yield mono-lingual gains, it usually did not yield cross-lingual gains either language (see Appendix Table \ref{table:perf-crosslingual-llama3}). 

% Exceptions to this were \hl{language(s)} with cross-lingual transfer relative to the baseline on \hl{Zulu (51.4\% rel. 50.3\%), Afrikaans (56.3\% rel. 55.9\%), and English (62.9\% rel. 59.3\%)}, as well as \hl{Zulu} cross-lingual transfer relative to the baseline on \hl{English (64.3\% rel. 59.3\%) and Afrikaans (56.3\% rel. 55,9\%)}. Notably the \hl{Zulu transfer to English was higher than English mono-lingual performance (60.7\%)}. The results suggest that certain attributes of the fine-tuning and test data influence performance gains across tasks and languages, for example: \hl{(i) translations of \texttt{MMLU} might contain more shared vocabulary in English (as seen earlier in the 0.15 - 0.2 ROUGE-1 scores between \texttt{MMLU} translations and English, Table \ref{table:translation-quality}), making it a little easier for a model to improve mono-lingually and cross-lingually on a task from a similar domain (\texttt{MMLU} showed consistent mono- and cross-lingual transfer compared to \texttt{Winogrande-ZA}), (ii) models with a certain threshold of baseline performance, beyond random chance, were able to improve with fine-tuning, as was seen when fine-tuning with \texttt{Winogrande-ZA} on Zulu and Xhosa where the baseline was close to ~50\% on the binary choice task, suggesting an initial ability to perform the task was a prerequisite for mono-lingual or cross-lingual transfer}.

\section{Discussion}
This study aimed to measure (and explore means to address) the performance gap of Large Language Models (LLMs) in English and African languages by translating popular benchmarks, assessing performance on those benchmarks, and exploring fine-tuning strategies that close the gap. The performance gap is not only a technical challenge but also a matter of equity, as many native African languages are low-resource, affecting the accessibility and effectiveness of LLMs for over 160 million speakers\footnote{Ethical Statement of this work is viewable in the Appendix.}. 

%\subsection{Significance of the Benchmark Creation}
The creation of benchmarks in low-resourced African languages is a critical step toward achieving equitable advancements in natural language processing. By translating popular benchmarks such as Winogrande and sections of the MMLU into eight under-studied African languages, we provide essential tools for evaluating and improving LLM performance in these languages. Our work not only highlights existing performance gaps, but also lays the groundwork for future research and development aimed at improving language technologies for native African language speakers. The benchmarks we translated may enable a more accurate assessment of LLM capabilities and drive progress toward more inclusive and effective language models.

% 12.0\% and 19.9\%

%\subsection{Implications of the Performance Gap}
Our study revealed a significant LLM performance gap between English and African languages: 12.0\%-19.9\% absolute. This performance gap has profound implications, as it exacerbates the digital divide and limits the accessibility and utility of LLMs for millions of speakers of low-resource languages. Addressing this gap is crucial for ensuring that advances in AI benefit all language communities equitably, thereby promoting greater inclusivity.

Moreover, our study also revealed a wide LLM performance gap between individual African languages, with Afrikaans (the highest-performing language) performing between 22.9\%-56.1\% (absolute) better than Bambara (the lowest-performing language). This disparity is likely due to Afrikaans and English both belonging to the Germanic language family (indicating greater similarity of Afrikaans to English) and Afrikaans having the highest resource level (indicating greater data availability for pre-training LLMs) across all African languages included in our study \cite{joshi2020state}. In contrast, Bambara is part of the Mande language family (see Figure \ref{fig:languagemap}), which does not have a single language that is considered high-resource (or even at the resource level of Afrikaans) \cite{hammarstrom2015ethnologue, joshi2020state}. This implies that Bambara and its relatives have considerably poorer data availability for pre-training LLMs than Afrikaans. Taken together, these findings underscore the importance of measuring the LLM performance gap of individual African languages to guide and prioritize efforts in creating language resources effectively (i.e. by targeting languages with lower performance). 

%\subsection{Implications of Cultural Appropriateness on Gap}
Models showed up to 15.6\% better performance on culturally appropriate questions, indicating that cultural relevance significantly influences model accuracy. This underscores the importance of incorporating cultural context in model training and evaluation, particularly for diverse and low-resource languages. Ignoring cultural nuances can lead to biases and inaccuracies, further marginalizing underrepresented language communities.

%\subsection{Implications of Fine-tuning Experiments}
Fine-tuning was shown to be effective in enhancing model performance. Across all languages and benchmarks, an average improvement of 5.6\% was observed for mono-lingual experiments (evaluating in the same language as fine-tuning). The quality of the dataset significantly influences performance, with higher quality data leading to improvements of up to 14.5\%, averaging 5.4\% over lower quality data of the same size. In addition, the alignment of the training domain with the target domain yielded the strongest gains. For example, fine-tuning a model with \textit{college medicine} data resulted in notable improvements in \textit{clinical knowledge} tasks: 17.4\% on average.

In scenarios where target language data are scarce, utilizing data from related languages may help. Cross-lingual transfer methods provided gains of up to 21.1\%, with an average improvement of 2.9\% across all languages and benchmarks. Our results highlight the potential benefit of using linguistically similar resources to enhance model performance in low-resource languages.

% \subsection{Mono / Cross-lingual Performance}

% \subsection{Cultural Appropriateness}
% Our results indicate that, GPT-family LLMs perform better out-of-the-box on African-language Winogrande QA-pairs when accounting for their culturally appropriatness in a given language. 
% % providing yet another possible avenue for improvement of evaluation techniques. 
% The observed performance lift may be due to culturally appropriate language representing a greater proportion of available text used to train the GPT-family models. 

% We also observe a decrease in performance on the ``appropriate" subset when evaluated in English for Sepedi annotations. 
% % The source of the observed performance decrease is unclear and may require additional study, however 
% We hypothesize that this may be due to high variability in the quality of translations provided by the workers from Upwork. For example, if the translators spoke a different dialect of Sepedi than the annotators, the annotators may mark QA pairs as inappropriate due to slight differences in connotation of concepts expressed in English, corresponding to a better performance on these ``inappropriate" QA pairs in English.

\section{Limitations}

Our study has several limitations that should be addressed or extended in future work:

\begin{enumerate}
    \item \textbf{Choice of Fine-tuning Model:} Although Llama 3 was the best open-source solution at the time of this study, it does not outperform GPT-4o out-of-the-box, even after fine-tuning. Future studies should consider evaluating other models (e.g. Llama 3.1, released on July 24th, 2024).
    \item \textbf{Fine-Tuning Scope:} Our fine-tuning experiments were conducted on individual languages in isolation. We did not explore the potential gains from tuning LLMs using data from multiple languages. Future research should investigate the effects of grouping African or related high-resource languages to enhance performance.
    \item \textbf{Benchmark Relevance:} The translation of established LLM benchmarks, while valuable, may not fully capture the depth and breadth of African-language specific use-cases. Future benchmark creation efforts should consider generating content that directly supports and aligns with use-cases most relevant to African language speakers.
\item \textbf{Variability in Translation Quality:} Recruiting experienced translators for all our chosen languages proved difficult. The number of speakers available on Upwork was limited for many languages, most notably Xhosa, Sesotho, Shona, Setswana, Bambara, Sepedi, and Tsonga (see Figure \ref{fig:upworker_pool} for more details). In some cases, it was not possible to find workers with prior experience. We also encountered variability in the dialects spoken by the workers, a factor which was challenging to control given the aforementioned lack of available speakers.

\item \textbf{Language Coverage and Scalability:}
Although we were able to cover 11 diverse African languages, there are naturally many other African languages presumably with LLM performance gaps which are not covered in this work. We believe that our presented framework is scalable, but acknowledge that the most costly part, the human translation of text, is a significant barrier. To alleviate this, we suggest incorporating a step to automatically identify translated text that may require additional human verification, rather than verifying all translations multiple times (which we performed to ensure that translations were high-fidelity). This should make it easier to extend our work to other languages.

 % A comprehensive list of profiles for the hired workers can be found in Table \ref{table:upwork-profiles}. Unexpectedly, 
\end{enumerate}

\section{Conclusion}
This study takes steps toward addressing the performance gap in Large Language Models (LLMs) for African languages. As part of this work, we created approximately 1 million human-translated words of new benchmark data in 8 African languages, covering a population of 160 million speakers. This effort involved the translation of established benchmarks and the fine-tuning of more than 400+ models. The benchmarks, translations, and all the code needed to recreate the results of this study are publicly available, supporting ongoing efforts to improve LLM performance in African languages and beyond. Future work should continue to explore and address the challenges identified, ensuring that advancements in AI benefit all language communities equitably.

\newpage

\section*{Acknowledgments}
All research-related costs (experiments, translation costs, etc.) were generously provided by the Bill and Melinda Gates Foundation.

\appendix

% Prepend 'A.' to the figure numbers
\renewcommand{\thefigure}{A.\arabic{figure}}

% Prepend 'A.' to the table numbers
\renewcommand{\thetable}{A.\arabic{table}}

% \section{Acknowledgments}
% % All research related costs (experiments, costs of the translations, etc.) were generously provided by the Bill and Melinda Gates Foundation. 

% % \footnote{PROJECT SPONSOR and AUTHOR N remains anonymous for blinded peer-review and will be substituted with entity names on acceptance.}

% All research related costs (experiments, costs of the translations, etc.) were generously provided by the \<PROJECT SPONSOR\>. \<AUTHOR 1\> contributed to the study design, led the experimental design, led the interpretation of the results, and wrote the manuscript. \<AUTHOR 2\> implemented the mono- and cross-lingual experiments described herein, performed the data quality and quanitity analyses, assisted with writing the manuscript, and lead the preparation of the GitHub repository associated with this work. \<AUTHOR 3\> implemented the cultural appropriateness experiments described herein, assisted with writing the manuscript, and assisted with the preparation of the Github repository associated with this work. \<AUTHOR 4\> contributed to the study design, co-lead the experimental design, and assisted with writing the manuscript and interpretation of the results. \<AUTHOR 5\> contributed to the study design, experimental design, and assisted with interpretation of the results. \<AUTHOR 6\> led the study design, contributed to the experimental design, and assisted with interpretation of the results. 

\bibliography{aaai25}

\pagebreak
\section{Appendix}

% Add seciton numbers in the appendix
\setcounter{secnumdepth}{1}

% reset table and figure counter
\setcounter{figure}{0}
\setcounter{table}{0}

\section*{Ethical Statement}
% You can write a statement about the potential ethical impact of your work, including its broad societal implications, both positive and negative. If included, such statement must be written in an unnumbered section titled \emph{Ethical Statement}.
The lack of publicly available benchmarks for low-resource languages exacerbates the challenge of developing performant LLMs for these languages. By providing the first known translations of the Winogrande and MMLU benchmarks into Amharic, Bambara, Igbo, Sepedi (Northern Sotho), Shona, Sesotho (Southern Sotho), Setswana, and Tsonga, this work addresses a critical gap in linguistic resources. Our evaluations of SOTA models provide a quantitative measure of existing gaps in model performance; critically, the ability to measure such gaps provides researchers and developers with accessible resources to help them reduce such gaps. The work highlights the importance of high-quality data in enhancing model performance, even when data is limited, motivating further efforts for the curation of high-quality data. Additionally, annotation of ``culturally appropriate" data and the demonstration of LLM sensitivity to such annotation help promote linguistic equity and inclusivity in the development of LLMs.

\section{Benchmark Translation} \label{sec:appendix-benchmark-translation}
\subsection{Winogrande Translation Process} 
\begin{figure*}[h!]
    \centering
    \includegraphics[width=0.99\textwidth]{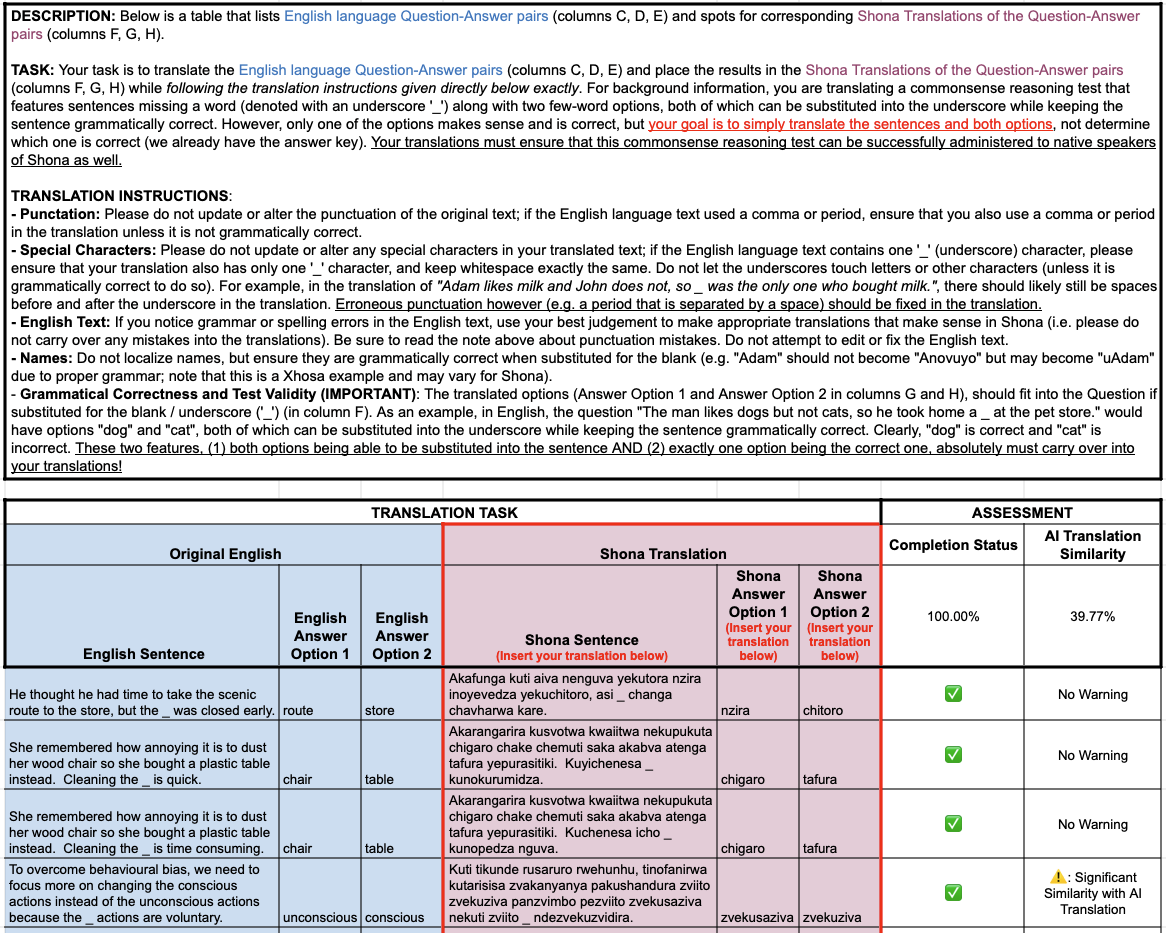}
    \caption{\textbf{Example Translation Task Form Given to Workers Hired on Upwork.com.} Workers were tasked to translate each Winogrande QA pair into eight African languages. The translator was able to see the percentage of the task completed to aid in time management. The translator was also able to see any warnings regarding translation similarity to Google Translate output (see Appendix Section \ref{sec:appendix-benchmark-translation} for more information). The figure above shows an example (in Shona) with four rows filled in by a translator.}
    \label{fig:translation-survey}
\end{figure*}

We recruited a total of 23 initial \textit{translators} (using Upwork.com) to translate Winogrande QA pairs into Amharic, Bambara, Igbo, Sepedi (Northern Sotho), Shona, Sesotho (Southern Sotho), Setswana, and Tsonga (see translator profiles in Table \ref{table:upwork-profiles}, and translation instructions in Figure \ref{fig:translation-survey}). Specifically, the ``small" training (640 QA pairs / 12,650 words), as well as the ``dev" (1,267 QA pairs / 25,470 words) and ``test" (1,767 QA pairs / 35,622 words) sets were translated: a total of 3,674 QA pairs consisting of 73,742 words per language. Translators were requested to (i) maintain original punctuation in translations (commas, periods), (ii) preserve special characters exactly as they appeared in the original text (e.g., $``\_"$ underscore), and (iii) retain original names in translations following the approach of \citet{de2019winograd}. 

Translators were prioritized according to the following metrics: (1) self-reported proficiency in English, (2) self-reported proficiency in the target language, (3) presence of a ``Top Rated" or ``Top Rated Plus" badge on Upwork.com, (4) prior job success rate on Upwork.com, (5) prior income earned reported on Upwork.com, and (6) any history of translation jobs in the target language. Translators were paid \$0.0237 USD per word to translate a portion (from 20\% to 100\% of the dataset, depending on availability) for up to 5 days, which translates into a (minimum) wage of \$8.75 USD per hour, above the average wage (\$7.30 USD) in South Africa \cite{statsSA2020labour}, where four out of eight of our target languages are spoken (Sepedi, Sesotho, Setswana, and Tsonga). Translators were allowed to use a machine translation tool, but with the requirement that they manually validate the outputs and correct any errors. 

\begin{figure*}[h!]
    \centering
    \includegraphics[width=0.99\textwidth]{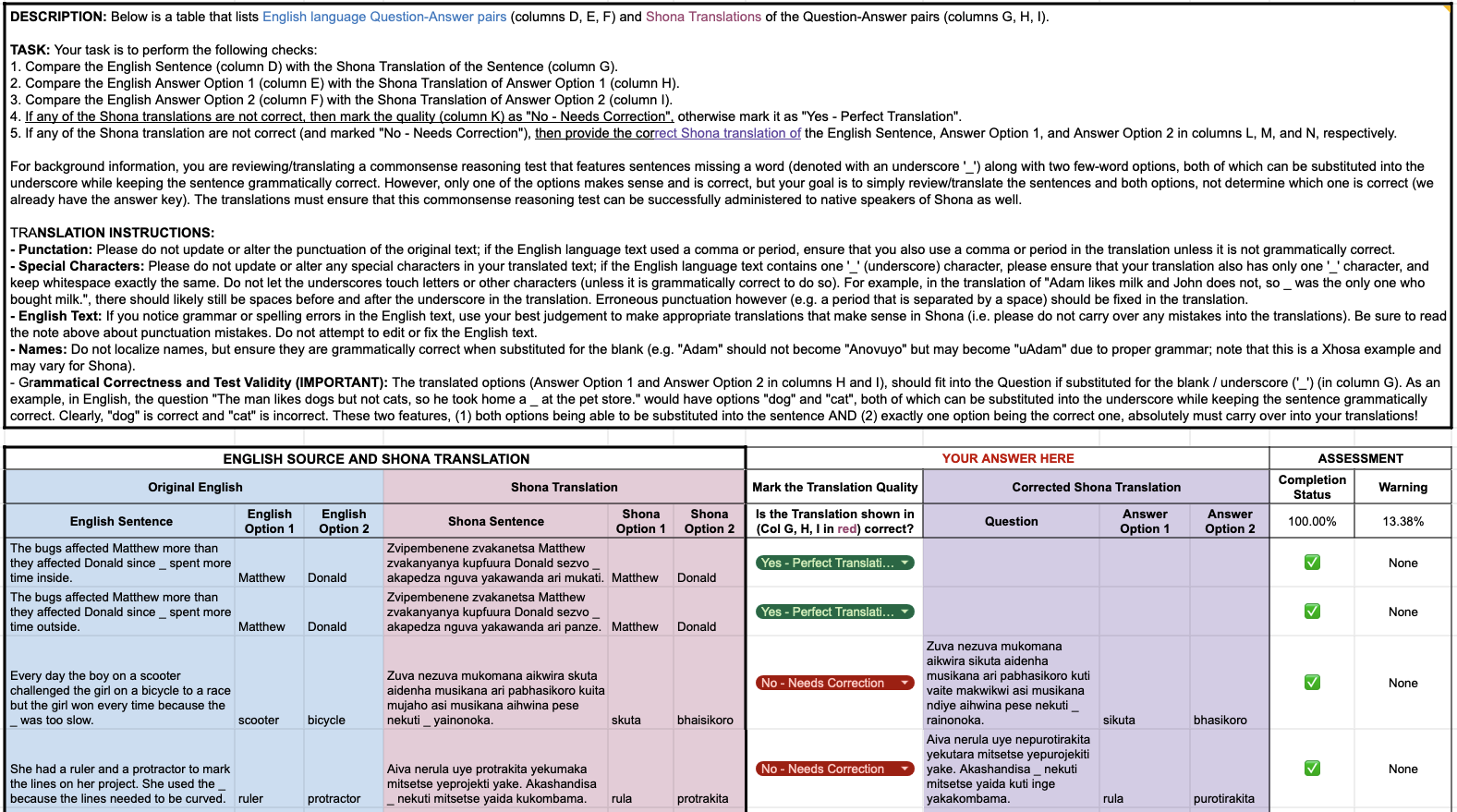}
    \caption{\textbf{Example Translation Review Task Form Given to Workers Hired on Upwork.com.} Workers were tasked to review the Winogrande translations provided by a previous worker. The reviewer was able to see the percentage of the task completed to aid in time management. The reviewer was also able to see any warnings regarding translation similarity to Google Translate output (see Appendix Section \ref{sec:appendix-benchmark-translation} for more information). The figure above shows an example (in Shona) with four rows filled in by a reviewer.}
    \label{fig:review-survey}
\end{figure*}

\begin{figure*}[h]
    \centering
    \includegraphics[width=0.99\textwidth]{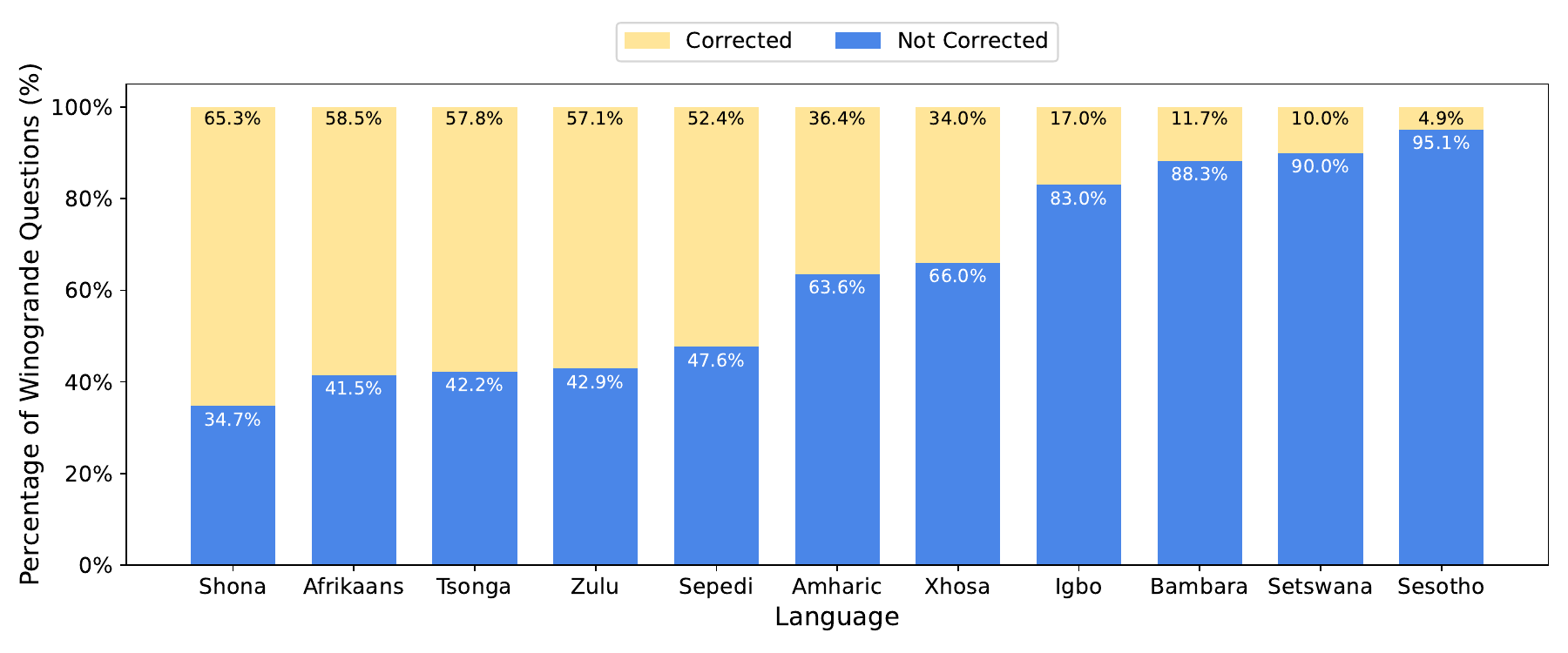} % Replace 'example-image' with the path to your image file
    \caption{\textbf{Results of the Translation Verification Task}. Translators were tasked to verify translations of the Winogrande dataset into 11 African languages, and provide corrections if needed. This table shows the proportion of the Winogrande dataset that needed corrections for each of the 11 African languages. The proportion that needed corrections (orange/top bars) for a given language ranged from 65.3\% on the higher end (Shona) to 4.9\% on the lower end (Sesotho).}
    \label{fig:translation-checks}
\end{figure*}

To ensure that the translations were correct, a second set of 28 \textit{validators} was hired to (i) validate the quality of the initial translation and (ii) provide corrections to the translations if any issues were found\footnote{9 translators were hired from the group of initial translators, although they did not work on their own translations.} (see \textit{validator} profiles in Table \ref{table:upwork-profiles}, validation instructions in Figure \ref{fig:review-survey}). Similarly to the translation task, we paid above the average wage (\$7.30 USD) in South Africa \cite{statsSA2020labour}. The amendments to the original translations varied by language: from 4.9\% (for Sesotho) to 65.2\% (for Shona) of the QA pairs (see Figure \ref{fig:translation-checks}).

%We estimated that at most half of the translations might require amendments, and hence, half the time would be required (in the worse case) for the verification task; therefore, workers were paid half the amount which they would have received for the full translation task. 

\subsection{Translation Quality Assessment} \label{sec:main-trans-qa}

\begin{figure*}[h] % 'h' for "here", other options: t (top), b (bottom), p (page of floats), etc.
    \centering  % Center the figure
    \includegraphics[width=0.99\textwidth]{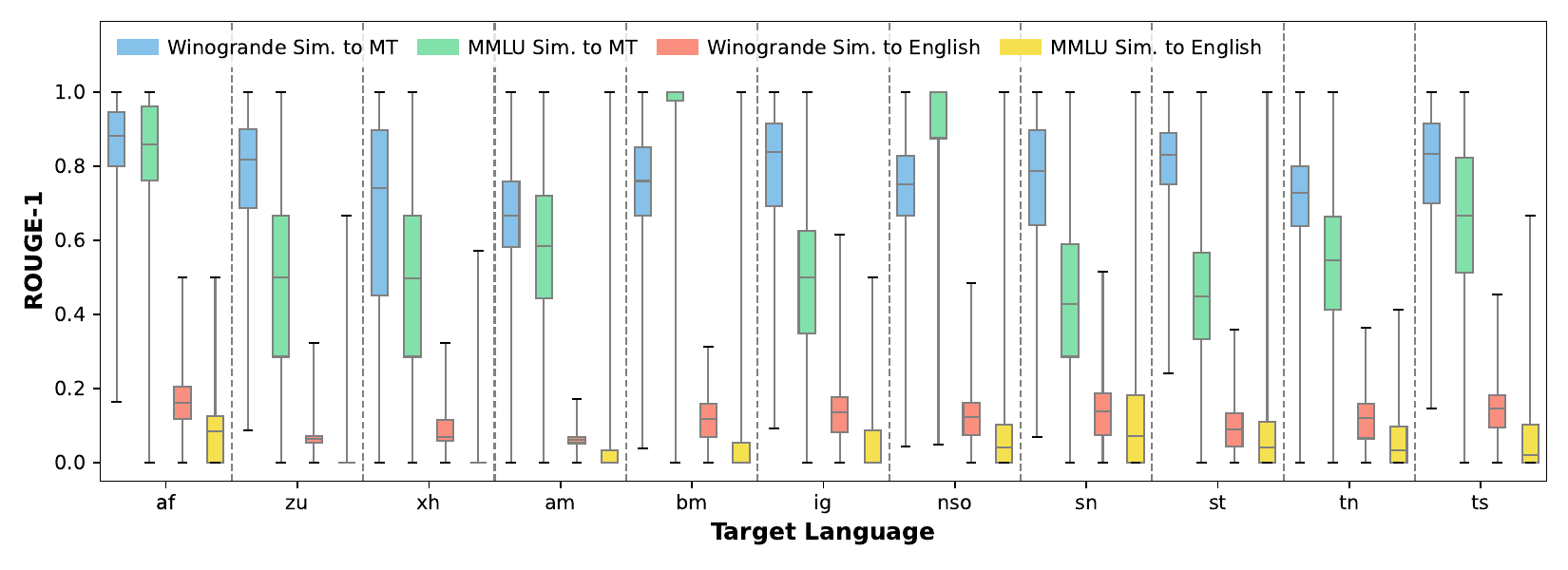}
    \caption{\textbf{ROUGE-1 Scores of Winogrande and MMLU Translations.} The figure displays boxplots of ROUGE-1 distributions for the similarity between the Winogrande and MMLU (college medicine, clinical knowledge, and virology sections combined) rows and their machine translation equivalents, as well as the original English versions, in 11 African languages. Google Translate was used to produce machine translations for all languages except Setswana (tn), in which case GPT-4o was used since Google Translate did not support Setswana on its API. MT: Machine Translation, Sim.: Similarity. Language codes are as follows: Afrikaans (af), Zulu (zu), Xhosa (xh), Amharic (am), Bambara (bm), Igbo (ig), Sepedi (nso), Shona (sn), Sesotho (st), Setswana (tn), Tsonga (ts).}
    \label{fig:rouge1-dists}
\end{figure*}

\begin{figure*}[h] % 'h' for "here", other options: t (top), b (bottom), p (page of floats), etc.
    \centering  % Center the figure
    \includegraphics[width=0.99\textwidth]{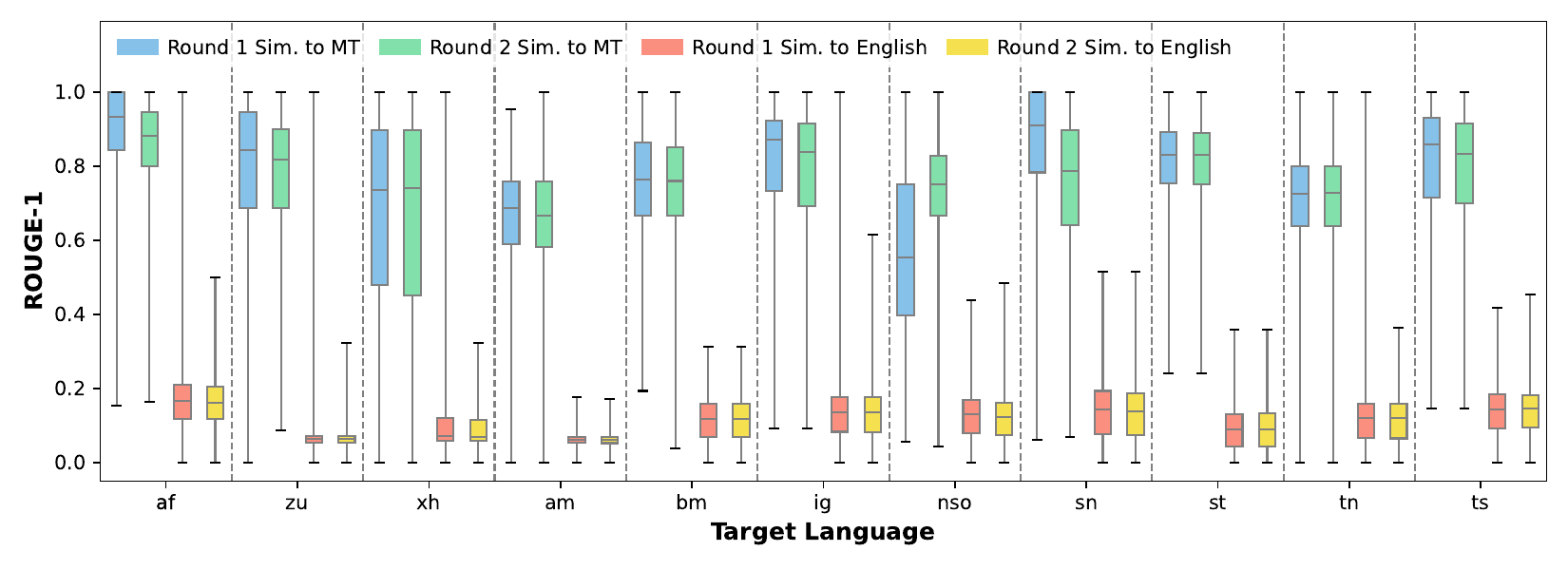}
    \caption{\textbf{ROUGE-1 Scores of Winogrande Translations by Translation Round.} The figure displays boxplots of ROUGE-1 distributions for the similarity between the Winogrande rows and their machine translation equivalents, as well as the original English versions, in 11 African languages. The ROUGE-1 scores before corrections (Round 1) and after corrections (Round 2) are given. Google Translate was used to produce machine translations for all languages except Setswana (tn), in which case GPT-4o was used since Google Translate did not support Setswana on its API. MT: Machine Translation, Sim.: Similarity. Language codes are as follows: Afrikaans (af), Zulu (zu), Xhosa (xh), Amharic (am), Bambara (bm), Igbo (ig), Sepedi (nso), Shona (sn), Sesotho (st), Setswana (tn), Tsonga (ts).}
    \label{fig:rouge1-dists-rounds}
\end{figure*}

\begin{figure*}[h] % 'h' for "here", other options: t (top), b (bottom), p (page of floats), etc.
    \centering  % Center the figure
    \includegraphics[width=0.99\textwidth]{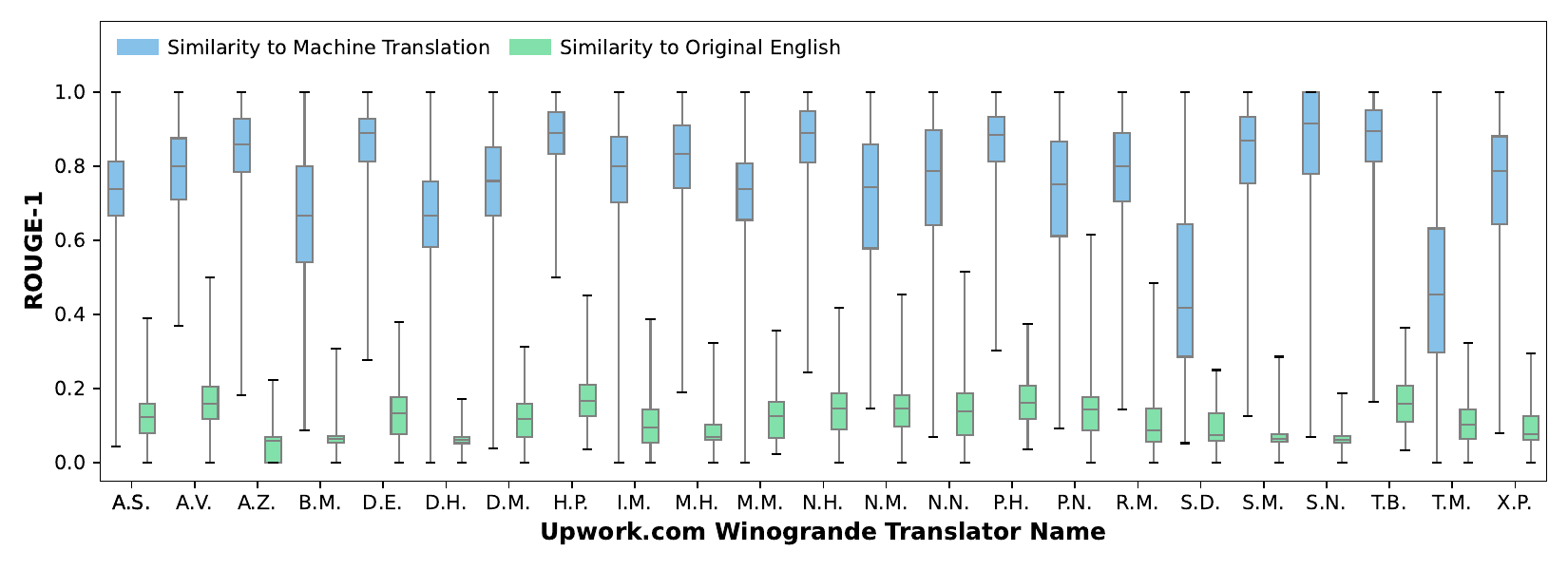}
    \caption{\textbf{ROUGE-1 Scores of Winogrande Translations by Translator.} The figure displays boxplots of ROUGE-1 distributions for the similarity between the Winogrande rows and their machine translation equivalents, as well as the original English versions, in 11 African languages grouped by the initial Upwork.com human translator that translated the rows. Google Translate was used to produce machine translations for all languages except Setswana, in which case GPT-4o was used since Google Translate did not support Setswana on its API. See Table \ref{table:upwork-profiles} for translator profiles.}
    \label{fig:rouge1-dists-by-translator}
\end{figure*}

We evaluated all translations against (i) the original English text, and (ii) Google Translate\footnote{At the time of writing, Google Translate did not support Setswana on its API, so GPT-4o was prompted to machine-translate the MMLU and Winogrande datasets into Setswana instead.} results in the target language. To compare translations with a reference text, we used standard translation similarity metrics, specifically Recall-Oriented Understudy for Gisting Evaluation (ROUGE-1). The aim was to identify if reliance on machine translation outputs was exceedingly high (represented as high similarity scores against Google Translate in the target language $>$ 0.95; this was the threshold used to warn Winogrande translators that their translations were similar to machine translations), or contained a concerningly high number of untranslated English words (a high similarity score against the original English text $>$ 0.95), or were generally poor translations in the target language (represented with low similarity scores against Google Translate into the target language $<$ 0.5). For MMLU, we observed ROUGE-1 scores compared to Google Translations $>$ 0.95 for at most 79\% of the data (Bambara/bm), and ROUGE-1 scores $<$ 0.5 for at most 60\% of the data (Sesotho/st). For Winogrande, we observed ROUGE-1 scores compared to Google Translations $>$ 0.95 for at most 22\% of the data (Afrikaans/af), and ROUGE-1 scores $<$ 0.5 for at most 29\% of the data (Xhosa/xh). See Figures \ref{fig:rouge1-dists}, \ref{fig:rouge1-dists-rounds}, and \ref{fig:rouge1-dists-by-translator} for more detailed ROUGE-1 scores on our translated datasets.  

\begin{figure*}[h!]
    \centering
    \includegraphics[width=0.99\textwidth]{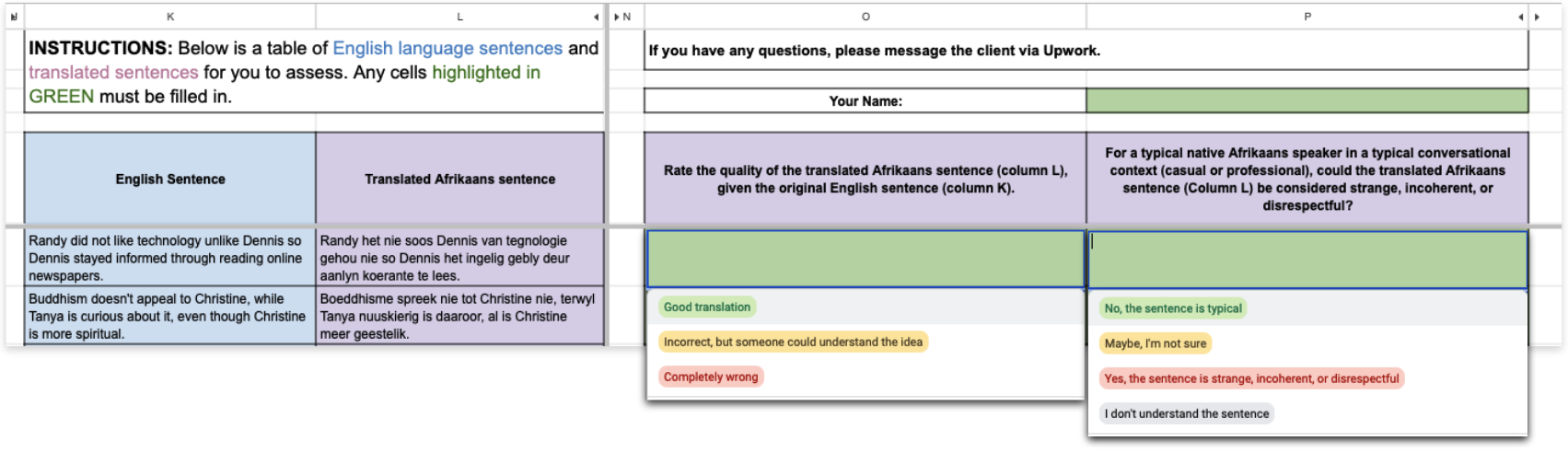}
    \caption{\textbf{Example Translation Quality and Cultural Appropriateness Survey Form Given to Workers Hired on Upwork.com.} Workers were tasked to assess the \textit{quality} and the \textit{appropriateness} of previously obtained human translations of the Winogrande dataset in 11 African languages. Each QA pair in Winogrande was provided (in the blue/leftmost column) with the correct answer filled in, along with the corresponding human translation (in the purple/second-from-the-left column). The \textit{evaluator} was able to see the percentage of the task completed to aid in time management. The figure above shows an example (in Afrikaans), depicting the dropdown options provided for the \textit{evaluator} to select from.}
    \label{fig:cultural-appropriateness-survey}
\end{figure*}

For the Winogrande data, a set of 48 translation quality \textit{evaluators} were recruited from Upwork, using the same recruitment approach described earlier (see evaluator profiles in Table \ref{table:upwork-profiles}). The \textit{evaluators} were asked to assess the previously obtained and verified human translations of Winogrande for translation quality. The \textit{evaluators} were presented with the original English and human-translated Winogrande QA pair with the correct answers filled in and asked to rate the quality of the translation according to three options: (i) ``Good translation", (ii) ``Incorrect, but someone could understand the idea", and (iii) ``Completely wrong" \cite{gt_tarzan, bapna2022building}. An example of the task can be seen in Figure \ref{fig:cultural-appropriateness-survey}, while the evaluation results are shown in Table \ref{table:annotator_confusion_1}. 94.6\% of the translations were considered good/understandable (``Good translation" or ``Incorrect, but someone could understand the idea") by either evaluator, 5.3\% were evaluated as wrong (``Completely wrong") by either evaluator, and 0.2\% were considered wrong (``Completely wrong") by both evaluators.

\begin{figure*}[h] % 'h' for "here", other options: t (top), b (bottom), p (page of floats), etc.
    \centering  % Center the figure
    \includegraphics[width=0.99\textwidth]{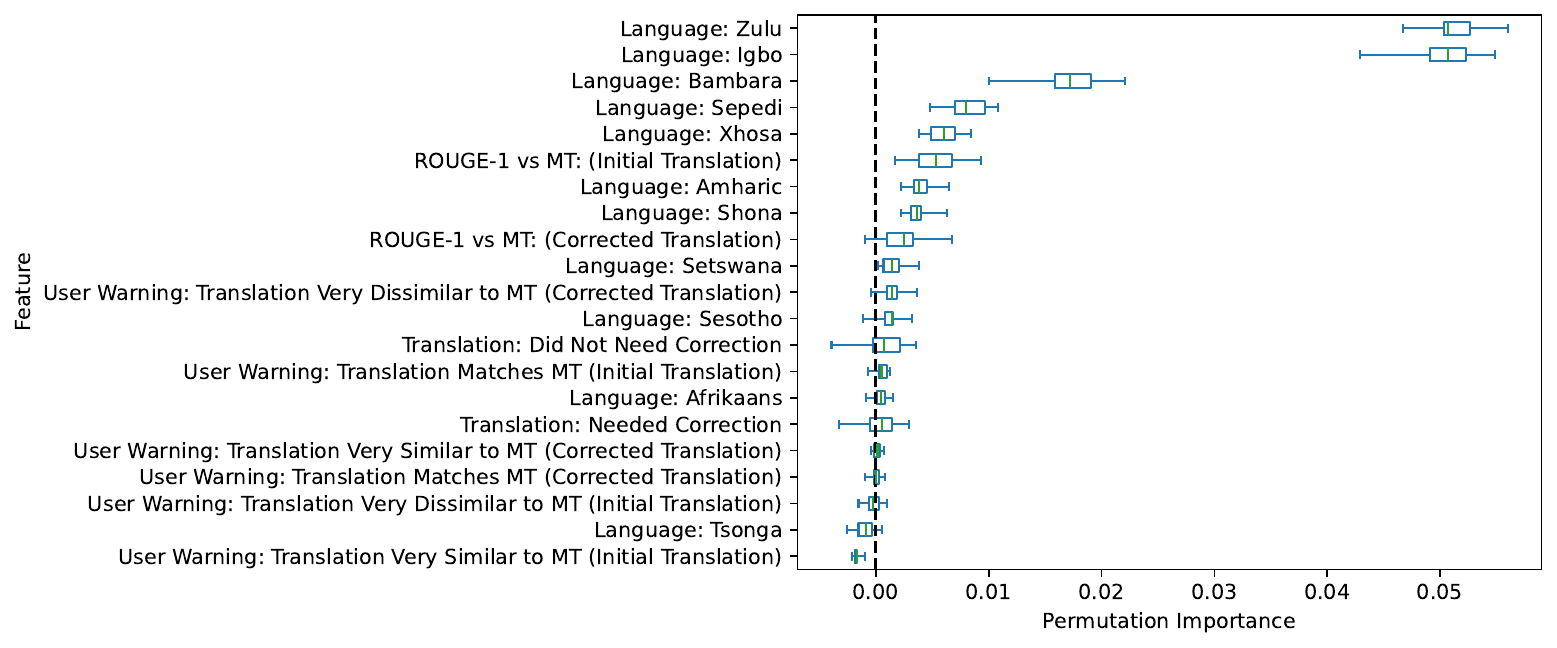}
    \caption{\textbf{Quality Feature Importances}. The plot shows the distribution of feature importances (using the permutation feature importance measure with n=20) from a random forest model to predict if two independent evaluators annotated the translation of Winogrande as a ``Good translation". The three most important features were the target languages Zulu, Igbo, and Bambara. Note that ``MT" stands for ``Machine Translation". ``ROUGE-1 vs MT" denotes the ROUGE-1 score (a number between 0 and 1) of the human translation with the machine translation as the reference. Warnings were displayed to translators according to three possible options: if a translation was ``Very Dissimilar" to the corresponding machine translation (ROUGE-1 $<$ 0.50), ``Very Similar" (ROUGE-1 $>$ 0.95), or ``Matches" (ROUGE-1 $=$ 1.00). A translation ``Needed Correction" if and only if the second translator made changes to the initial translation.
     }
    \label{fig:quality_feature_importance_permutation}
\end{figure*}

We modeled the probability of a ``Good translation" rating from both evaluators; an accuracy of 67\% (Precision=0.67, Recall=0.66, F1=0.65) was achieved using a random forest classifier. The feature importances (permutation importance; n=20), indicating the features that improved accuracy the most, include the Zulu, Igbo, and Bambara languages as the target language of translation (see Figure \ref{fig:quality_feature_importance_permutation}).
  
\section{Model Hyperparameters} \label{sec:hyperparamters}

The hyperparameters used to fine-tune Llama 3 70B IT in all experiments were Seed=42, Epochs=3, Batch Size=1, Learning Rate=2e-4, Optimizer=8-Bit AdamW, Weight Decay=0.01, Learning Rate Scheduler=Linear, LoRA Alpha=8, LoRA Dropout=0, LoRA Matrix Dimension=8. These parameters were assigned based on \cite{polignano2024advanced}. The Unsloth framework was used for fine-tuning\footnote{https://github.com/unslothai/unsloth}. We used one A100 80GB GPU for fine-tuning/evaluation in all experiments.

In all LLM inferences performed for our experiments, as well as GPT-4o prompting for quality scoring, we used Temperature=0.7 and Top\_P=0.9 (selected to mimic what was used in \cite{shaham2024multilingual}, which also conducted cross-lingual fine-tuning experiments; these settings allowed for more variance in LLM responses). Given that the outputs for our experiments were always supposed to be a single letter or number, we also truncated the outputs to three tokens. 

For Setswana translations using GPT-4o, we used a Temperature=0 and Top\_P=1 to minimize stochasticity similar to the consistent responses of Google Translate.

% Only the hyperparameters given above in this section were tried for our experiments. We believe that the results we obtained using them were satisfactory (e.g. we experienced $>$20\% lifts from fine-tuning in some cases), and our goal was not necessarily to optimize benchmark performance but rather make performance comparisons using consistent hyperparameter settings.

\section{Cultural Appropriateness Survey Details} \label{sec:cultural_appropriateness}

We used both Cohen's Kappa and Fleiss' Kappa to determine the inter-rater reliability between the two \textit{evaluators} for each language, finding that the highest inter-rater agreement was achieved in Shona with a Cohen's Kappa score of 0.211 for translation quality (Table \ref{table:appropriateness-interrater}). Although Cohen's Kappa and Fleiss' Kappa are sensitive to highly skewed data (see Table \ref{table:annotator_confusion_2}), comparing scores between languages and annotator questions provides a relative measure of agreement. 
We observed that the inter-rater agreement for appropriateness annotations was very low even when agreement was present for translation quality annotations, indicating that different annotators may be sensitive to different aspects of appropriateness. 

\begin{figure*}[h] % 'h' for "here", other options: t (top), b (bottom), p (page of floats), etc.
    \centering  % Center the figure
    \includegraphics[width=0.99\textwidth]{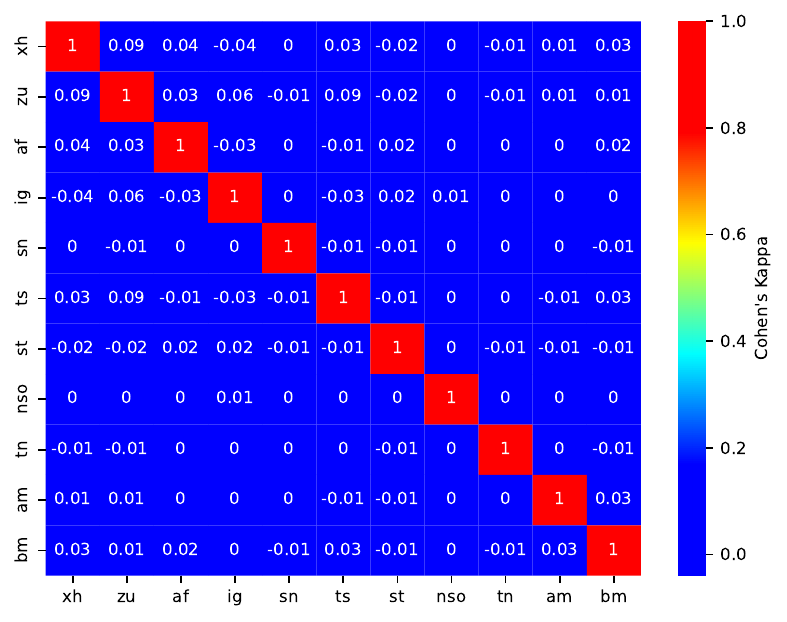}
    \caption{\textbf{Inter-rater Agreement on Cultural Inappropriateness across Different Target Languages.} For each pair of languages, we provide the Cohen's Kappa ($\kappa$) measuring the inter-rater agreement of the cultural inappropriateness labels. Importantly, the analysis only includes translations marked as ``Good translation" by both annotators in both languages. Each cell in the matrix represents the agreement level (ranging from 0 to 1) between pairs of languages, with 1 indicating the highest level of agreement. The Cohen’s Kappa scores do not show strong agreement on the definition of ``cultural appropriateness" across languages. However, it should be noted that among the languages listed, Xhosa and Zulu are the most closely related (linguistically), and are also those for which the inter-annotator agreement was the greatest ($\kappa = 0.09$). Language codes are as follows: Xhosa (xh), Zulu (zu), Afrikaans (af), Igbo (ig), Shona (sn), Tsonga (ts), Sesotho (st), Sepedi (nso), Setswana (tn), Amharic (am), Bambara (bm).}
    \label{fig:inappropriatenss_kappa_across_languages}
\end{figure*}

In Figure \ref{fig:inappropriatenss_kappa_across_languages}, we provide inter-rater agreement on cultural inappropriateness definitions across the different target languages. Importantly, the Cohen’s Kappa scores do not show strong agreement on the definition of ``cultural appropriateness" across languages. However, it should be noted that among the languages in our study, Xhosa and Zulu are the most closely related (linguistically), and are also those for which the inter-annotator agreement was the greatest ($\kappa = 0.09$). 

In order to account for a lack of agreement on appropriateness, we consider a dataset QA pair ``inappropriate" if either \textit{evaluator} did not mark it as ``typical". Additionally, to observe any confounding effect that translation quality may have, we performed a comparative analysis on each language for 3 conditions: (1) All dataset QA pairs included, (2) only QA pairs included at least one \textit{evaluator} considered understandable, and (3) only QA pairs that both \textit{evaluators} considered understandable (``understandable" meaning that the \textit{evaluator} labeled it as either ``good translation" or ``incorrect, but someone could understand the idea"). GPT-3.5, GPT-4, and GPT-4o were evaluated on Winogrande in each target language and in English. Then, for each target language, the performance of each LLM in the sets labeled ``appropriate" and ``inappropriate" was assessed in both the target language and English. Assessment in English was used as a reference because it is possible that inappropriate translations were due to poor dataset QA pairs rather than culture-specific inappropriateness.

\begin{figure*}[h] % 'h' for "here", other options: t (top), b (bottom), p (page of floats), etc.
    \centering  % Center the figure
    \includegraphics[width=0.99\textwidth]{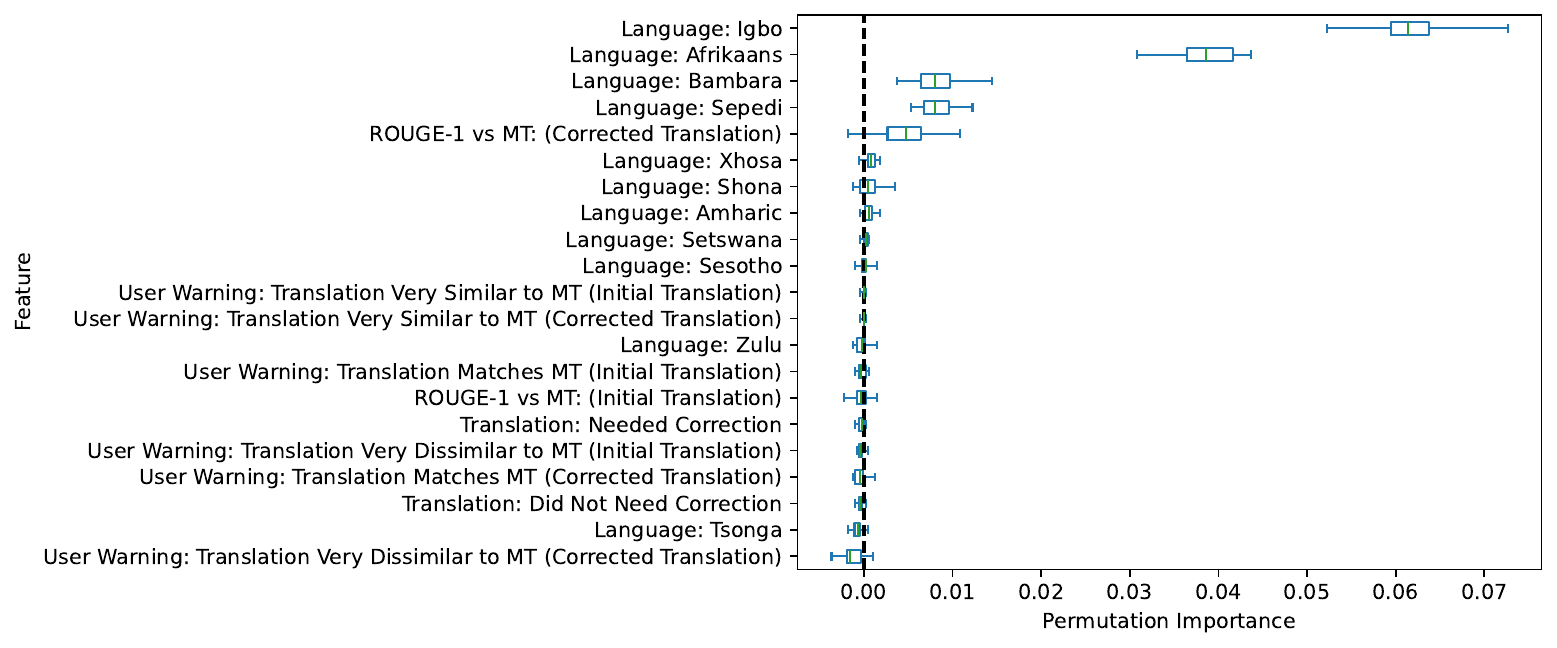}
    \caption{\textbf{Appropriateness Feature Importances without Quality Feature}.  The plot shows the distribution of feature importances (using the permutation feature importance measure with n=20) from a random forest model to predict if either of two independent evaluators annotated the translation of Winogrande as not ``typical" (i.e. ``inappropriate"). The three most important features were the target languages Igbo, Afrikaans, Bambara. Note that ``MT" stands for ``Machine Translation". ``ROUGE-1 vs MT" denotes the ROUGE-1 score (a number between 0 and 1) of the human translation with the machine translation as the reference. Warnings were displayed to translators according to three possible options: if a translation was ``Very Dissimilar" to the corresponding machine translation (ROUGE-1 $<$ 0.50), ``Very Similar" (ROUGE-1 $>$ 0.95), or ``Matches MT" (ROUGE-1 $=$ 1.00). A translation ``Needed Correction" if and only if the second translator made changes to the initial translation.}
    \label{fig:appropriate_feature_importance_permutation}
\end{figure*}

We modeled the probability of inappropriate text using a random forest classifier, with and without a \emph{quality} feature, achieving accuracies of 66\% (Precision=0.67, Recall=0.66, F1=0.65) and 80\% (Precision=0.80, Recall=0.80, F1=0.80), respectively. The \emph{quality} feature was True if and only if both evaluators gave a ``Good translation" rating. For the experiment without the quality feature, the feature importances with the largest contributions toward improving accuracy were: Igbo, Afrikaans, Bambara, and Sepedi as the target language of translation, as well as the ROUGE-1 score against machine translation of the corrected translation (see Figure \ref{fig:appropriate_feature_importance_permutation}).  

\begin{figure*}[h] % 'h' for "here", other options: t (top), b (bottom), p (page of floats), etc.
    \centering  % Center the figure
    \includegraphics[width=0.99\textwidth]{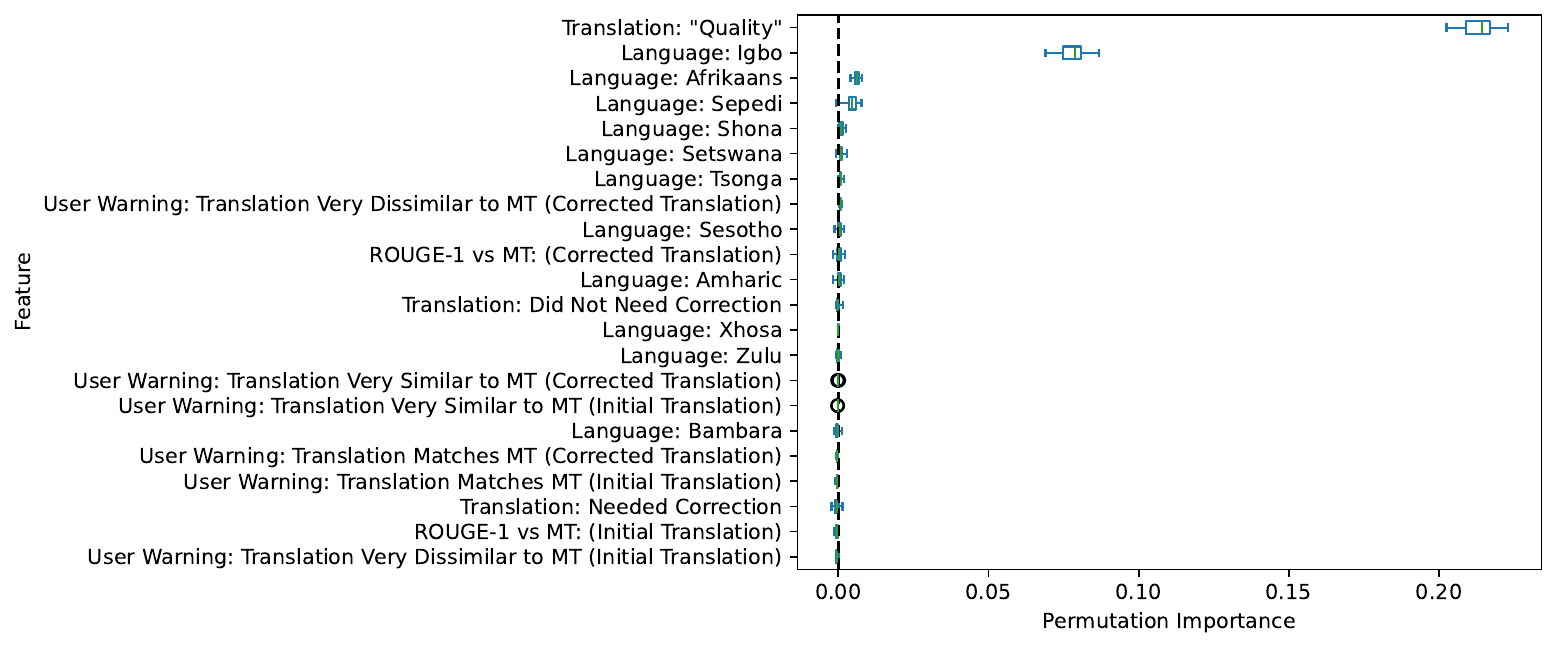}
    \caption{\textbf{Appropriateness Feature Importances with Quality Feature}. The plot shows the distribution of feature importances (using the permutation feature importance measure with n=20) from a random forest model to predict if either of two independent evaluators annotated the translation of Winogrande as not ``typical" (i.e. ``inappropriate"). Unlike Figure \ref{fig:appropriate_feature_importance_permutation}, this figure includes a translation \emph{quality} feature (both evaluators gave a ``Good translation" rating for the Winogrande translation). The three most important features were the \emph{quality} feature, and the target languages Igbo and Afrikaans. Note that ``MT" stands for ``Machine Translation". ``ROUGE-1 vs MT" denotes the ROUGE-1 score (a number between 0 and 1) of the human translation with the machine translation as the reference. Warnings were displayed to translators according to three possible options: if a translation was ``Very Dissimilar" to the corresponding machine translation (ROUGE-1 $<$ 0.50), ``Very Similar" (ROUGE-1 $>$ 0.95), or ``Matches" (ROUGE-1 $=$ 1.00). A translation ``Needed Correction" if and only if the second translator made changes to the initial translation.}
    \label{fig:appropriate_feature_quality_importance_permutation}
\end{figure*}

For the experiment with the quality feature, the feature importances with the largest contributions toward improving accuracy were: the \emph{quality} feature and Igbo as the target language of translation. (see Figure \ref{fig:appropriate_feature_quality_importance_permutation}).  

\section{Out-of-the-Box Evaluations on Machine-Translated Benchmarks}
\label{sec:ootb-eval-mt}

We performed the same evaluations of SOTA LLMs (Aim 2) on our benchmarks using (1) machine translations of the human translations from the target African language back into English, and (2) machine translations of the text instead of human translations from English to the target African language. Google Translate was used to translate all languages except Setswana, for which GPT-4o was employed using a simple \emph{``Translate the following from \{language X\} into \{language Y\}"} prompt.

Specifically, we took the original English versions of \texttt{Winogrande} (binary choice co-reference resolution task), the three clinical sections of \texttt{MMLU} (multiple choice medical domain knowledge task), as well as \texttt{Belebele} (multiple choice reading comprehension task) and machine-translated them into each of Afrikaans, Zulu, Xhosa, Amharic, Bambara, Igbo, Sepedi, Shona, Sesotho, Setswana, and Tsonga. Moreover, we took the human-translated versions of these benchmarks and machine-translated them back into English using the same machine translation tools (i.e. ``backtranslations"). We used the same LLM evaluation prompts as were used for Aim 2 (see Figures \ref{fig:winogrande-prompt}-\ref{fig:belebele-prompt}).

The SOTA LLM evaluations on machine-translated benchmarks are shown in Tables \ref{table:perf-results-mt-wino}-\ref{table:perf-results-mt-bele}. We generally observed that backtranslation to English showed better performance of English-centric LLMs (i.e. LLMs trained on limited amounts of low-resource language text) such as Llama 3 \cite{llama3} or Phi 3 \cite{abdin2024phi}, with gains ranging between 20\%-30\% and reaching upwards of 40\% for Belebele, MMLU College Medicine, and MMLU Clinical Knowledge. This finding suggests that when attempting to leverage the knowledge of an English-centric LLM for multiple choice question answering, it is highly beneficial to first backtranslate the question into English, allowing the LLM to work in its primary language of training. In contrast, for the large multilingual models GPT-4 and GPT-4o, the gains from backtranslation into English are considerably smaller (0\%-10\% for most languages), with evaluations on the human-translated benchmarks in the native language sometimes even yielding greater performance (between 0\%-10\%). This is especially evident for Belebele (Table \ref{table:perf-results-mt-bele}), suggesting that for certain tasks like reading comprehension or language understanding, there are already LLMs that perform about as well in native African languages as they do in English.

For machine translations from English into the target language, we observed that our SOTA LLMs across the board performed better or about the same on machine-translated versions of the benchmarks for certain languages (e.g. Igbo for MMLU College Medicine, MMLU Clinical Knowledge, and Belebele), suggesting SOTA LLM reliance on machine-translated training data or an affinity for ``translationese" versions of certain African languages instead of genuine, human-made text. However, for Winogrande (Table \ref{table:perf-results-mt-wino}), the differences between machine-translated and human-translated benchmark performance were generally 
within 0-3\%, suggesting that machine translation from English is as good as human translation from English for LLM benchmark performance.

\begin{figure*}[h] % 'h' for "here", other options: t (top), b (bottom), p (page of floats), etc.
    \centering  % Center the figure
    \includegraphics[width=0.99\textwidth]{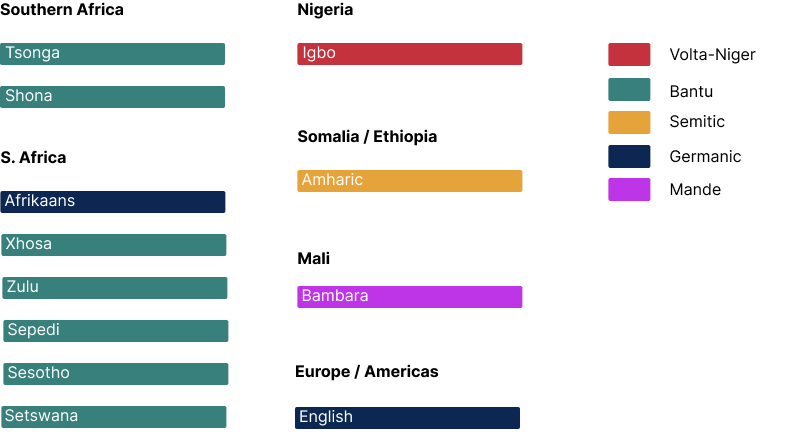}
    \caption{\textbf{Language Groupings.} This figure shows the 12 languages of this study grouped by where they are most spoken, geographically. The legend shows the language families.}
    \label{fig:languagemap}
\end{figure*}

Finally, when observing the trend between LLM performance in English (the original datasets) against African language backtranslations (datasets machine-translated into English but originally a human translation from English into an African language; see Tables \ref{table:perf-results-mt-wino}-\ref{table:perf-results-mt-bele}), we observe degradations in performance (from the original English to the backtranslation) generally between 5\%-25\%, with the lowest degradations occurring consistently when backtranslating from Afrikaans. This indicates that the machine translation quality is higher for Afrikaans than for other African languages, which may be due to Afrikaans and English sharing Germanic language family membership (see Figure \ref{fig:languagemap}) and Afrikaans having a higher resource level (implying greater availability of data for pre-training LLMs) than all other African languages included in our experiments \cite{joshi2020state}. 
% In the future, we expect the aforementioned degradations to shrink as machine translation tools improve, especially for technical content.

\section{Reproducibility Checklist}

This paper:

\begin{itemize}
    \item Includes a conceptual outline and/or pseudocode description of AI methods introduced: \textbf{Yes}
    \item Clearly delineates statements that are opinions, hypothesis, and speculation from objective facts and results: \textbf{Yes}
    \item Provides well marked pedagogical references for less-familiar readers to gain background necessary to replicate the paper: \textbf{Yes}

\end{itemize}

\noindent Does this paper make theoretical contributions? \textbf{No}\\

\noindent If yes, please complete the list below.

\begin{itemize}
\item All assumptions and restrictions are stated clearly and formally. \textbf{No theoretical contributions made.}
\item All novel claims are stated formally (e.g., in theorem statements). \textbf{No theoretical contributions made.}
\item Proofs of all novel claims are included. \textbf{No theoretical contributions made.}
\item Proof sketches or intuitions are given for complex and/or novel results. \textbf{No theoretical contributions made.}
\item Appropriate citations to theoretical tools used are given. \textbf{No theoretical contributions made.}
\item All theoretical claims are demonstrated empirically to hold. \textbf{No theoretical contributions made.}
\item All experimental code used to eliminate or disprove claims is included. \textbf{No theoretical contributions made.}
\end{itemize}

\noindent Does this paper rely on one or more datasets? \textbf{Yes, public datasets.}\\

\noindent If yes, please complete the list below.

\begin{itemize}
\item A motivation is given for why the experiments are conducted on the selected datasets. \textbf{Yes}
\item All novel datasets introduced in this paper are included in a data appendix. \textbf{Yes, these are included in the Supplementary Materials.}
\item All novel datasets introduced in this paper will be made publicly available upon publication of the paper with a license that allows free usage for research purposes. \textbf{Yes}
\item All datasets drawn from the existing literature (potentially including authors’ own previously published work) are accompanied by appropriate citations. \textbf{Yes}
\item All datasets drawn from the existing literature (potentially including authors’ own previously published work) are publicly available. \textbf{Yes}
\item All datasets that are not publicly available are described in detail, with explanation why publicly available alternatives are not scientifically satisficing. \textbf{N/A}
\end{itemize}

\noindent Does this paper include computational experiments? \textbf{Yes}\\

\noindent If yes, please complete the list below.

\begin{itemize}
\item Any code required for pre-processing data is included in the appendix. \textbf{Yes, uploaded in Supplementary Materials}
\item All source code required for conducting and analyzing the experiments is included in a code appendix. \textbf{Yes, uploaded in Supplementary Materials}
\item All source code required for conducting and analyzing the experiments will be made publicly available upon publication of the paper with a license that allows free usage for research purposes. \textbf{Yes}
\item All source code implementing new methods have comments detailing the implementation, with references to the paper where each step comes from. \textbf{Yes}
\item If an algorithm depends on randomness, then the method used for setting seeds is described in a way sufficient to allow replication of results. \textbf{Yes, seed values are always provided.}
\item This paper specifies the computing infrastructure used for running experiments (hardware and software), including GPU/CPU models; amount of memory; operating system; names and versions of relevant software libraries and frameworks. \textbf{Yes}
\item This paper formally describes evaluation metrics used and explains the motivation for choosing these metrics. \textbf{Yes}
\item This paper states the number of algorithm runs used to compute each reported result. \textbf{Yes}
\item Analysis of experiments goes beyond single-dimensional summaries of performance (e.g., average; median) to include measures of variation, confidence, or other distributional information. \textbf{Yes}
\item The significance of any improvement or decrease in performance is judged using appropriate statistical tests (e.g., Wilcoxon signed-rank). \textbf{No}
\item This paper lists all final (hyper-)parameters used for each model/algorithm in the paper’s experiments. \textbf{Yes}
\item This paper states the number and range of values tried per (hyper-) parameter during development of the paper, along with the criterion used for selecting the final parameter setting. \textbf{Yes}
\end{itemize}

\begin{table*}[h]
\centering
\setlength\extrarowheight{4pt}
\begin{tabular}{|cc|ccccccccccc|c|}
\hline
\textbf{Eval. 1} & \textbf{Eval. 2} & \textbf{xh} & \textbf{zu} & \textbf{af} & \textbf{ig} & \textbf{sn} & \textbf{ts} & \textbf{st} & \textbf{nso} & \textbf{tn} & \textbf{am} & \textbf{bm} & \textbf{Avg. (\%)} \\
\hline
Typical            & Typical           & 2755 & 2832 & 2229 & 1738 & 3226 & 2946 & 3035 & 3535 & 2998 & 2830 & 2035 & 74.6\%       \\
Typical            & Not sure          & 26   & 70   & 60   & 3    & 306  & 5    & 1    & 14   & 0    & 0    & 286  & 1.9\%        \\
Typical            & Strange           & 6    & 267  & 335  & 965  & 68   & 160  & 3    & 76   & 1    & 34   & 69   & 4.9\%        \\
Typical            & D.U.              & 1    & 43   & 3    & 0    & 18   & 2    & 1    & 0    & 0    & 1    & 1    & 0.2\%        \\
Not sure           & Typical           & 394  & 51   & 246  & 514  & 26   & 337  & 106  & 6    & 467  & 787  & 498  & 8.5\%        \\
Not sure           & Not sure          & 7    & 1    & 12   & 1    & 5    & 3    & 0    & 0    & 0    & 0    & 107  & 0.3\%        \\
Not sure           & Strange           & 0    & 0    & 41   & 300  & 5    & 23   & 0    & 2    & 0    & 8    & 30   & 1.0\%        \\
Not sure           & D.U.              & 1    & 0    & 1    & 0    & 1    & 0    & 0    & 0    & 0    & 0    & 1    & $<$0.1\%     \\
Strange            & Typical           & 208  & 356  & 593  & 87   & 14   & 158  & 514  & 36   & 205  & 14   & 431  & 6.5\%        \\
Strange            & Not sure          & 3    & 0    & 24   & 0    & 3    & 0    & 0    & 0    & 0    & 0    & 97   & 0.3\%        \\
Strange            & Strange           & 4    & 26   & 85   & 61   & 2    & 20   & 3    & 2    & 1    & 0    & 48   & 0.6\%        \\
Strange            & D.U.              & 0    & 0    & 1    & 0    & 0    & 0    & 0    & 0    & 0    & 0    & 7    & $<$0.1\%     \\
D.U.               & Typical           & 248  & 28   & 34   & 3    & 0    & 11   & 11   & 1    & 2    & 0    & 47   & 1.0\%        \\
D.U.               & Not sure          & 10   & 0    & 3    & 0    & 0    & 0    & 0    & 0    & 0    & 0    & 11   & 0.1\%        \\
D.U.               & Strange           & 10   & 0    & 7    & 2    & 0    & 9    & 0    & 2    & 0    & 0    & 6    & 0.1\%        \\
D.U.               & D.U.              & 1    & 0    & 0    & 0    & 0    & 0    & 0    & 0    & 0    & 0    & 0    & $<$0.1\%     \\
\hline
\multicolumn{14}{|c|}{Study Definitions of Cultural Appropriateness Given Translation Appropriateness and Quality}   \\ \hline
\textbf{Approp.} & \textbf{Quality} & \textbf{xh} & \textbf{zu} & \textbf{af} & \textbf{ig} & \textbf{sn} & \textbf{ts} & \textbf{st} & \textbf{nso} & \textbf{tn} & \textbf{am} & \textbf{bm} & \textbf{Avg. (\%)}  \\ \hline
Approp.  & Good  & 2732 & 2811 & 2229 & 1737 & 3224 & 2863 & 3019 & 3506 & 2995 & 2829 & 2028 & 74.2 \\
Inapprop. & Good & 723  & 659  & 1240 & 1520 & 327  & 547  & 548  & 64   & 617  & 831  & 1238 & 20.6 \\
\hline
\end{tabular}
\caption{\textbf{Human Evaluation of Winogrande Translation Appropriateness.} The table lists the total number of all possible combinations of \textit{evaluator} responses for the second survey question, rating whether the translation can be considered ``strange, incoherent, or disrespectful" in a typical conversational context. Possible answers were ``No, the sentence is typical" (Typical), ``Maybe, I'm not sure" (Not sure), ``Yes, the sentence is strange, incoherent, or disrespectful" (Strange), or ``I don't understand the sentence" (D.U.). The bottom section, labeled ``Study Definitions of Cultural Appropriateness Given Translation Appropriateness and Quality" shows two definitions of cultural appropriateness/inappropriateness used in our study. First are QA pairs considered ``appropriate", where both evaluators labeled the QA pair as ``No, the sentence is typical" (Approp.). Second are QA pairs considered ``inappropriate" where either evaluator labeled the QA pair as anything other than ``No, the sentence is typical" (Inapprop.). In both cases, QA pairs were only considered if both \textit{evaluators} labeled them as a ``good" or ``understandable" translation (see Table \ref{table:annotator_confusion_1}). Language codes are as follows: Xhosa (xh), Zulu (zu), Afrikaans (af), Igbo (ig), Shona (sn), Tsonga (ts), Sesotho (st), Sepedi (nso), Setswana (tn), Amharic (am), Bambara (bm).}
\label{table:annotator_confusion_2}
\end{table*}

\FloatBarrier

\begin{table*}[h!]
\centering
\footnotesize
\setlength\extrarowheight{2pt}
\begin{tabularx}{\textwidth}{|c|c|X|c|c|X|c|c|c|}
\hline
\textbf{} & \textbf{} & \multicolumn{3}{c|}{\textbf{Original English Version}} & \multicolumn{3}{c|}{\textbf{Translated Afrikaans Version}} & \textbf{} \\
\cline{3-8} % Draws horizontal line only under grouped columns
\multirow{-2}{*}{\textbf{Approp.}} & \multirow{-2}{*}{\textbf{Quality}} 
& \centering\textbf{Sentence} & \textbf{Opt. 1} & \textbf{Opt. 2} 
& \centering\textbf{Sentence} & \textbf{Opt. 1} & \textbf{Opt. 2} 
& \multirow{-2}{*}{\textbf{Ans.}} \\
\hline
Approp. & Good / Unders. & Robert wanted to paint his bedroom black. Brett said he would reget it soon because it's a poor choice. \_ preferred the lighter colors. & Robert & Brett & Robert wou sy slaapkamer swart verf. Brett het gesê hy sal dit binnekort berou, omdat dit 'n swak keuse is. \_ het die ligter kleure verkies. & Robert & Brett & 2 \\ \hline
Inapprop. & Good / Unders. & A transmission is what I am in need of today Joseph tells Dennis, \_ has wrecked many cars. & Joseph & Dennis & n Ratkas is wat ek vandag nodig het sê Joseph vir Dennis, \_ het al baie motors verwoes. & Joseph & Dennis & 1 \\ \hline
Inapprop. & Wrong & Jessica often experiences severe nausea, Victoria does not therefore \_ often rides big roller coasters. & Jessica & Victoria & Jessica ervaar dikwels erge naarheid en Victoria nie. \_ ry dus nie dikwels groot wipwaritte nie. & Jessica & Victoria & 2 \\ \hline
\end{tabularx}
\caption{\textbf{Examples of Winogrande Translation Appropriateness and Quality Annotation Combinations (Afrikaans).} The table gives example Winogrande QA pairs for each of the possible combinations of translation appropriateness and quality according to annotations made by evaluators on our Winogrande translations to Afrikaans. QA pairs were considered ``appropriate" when both \textit{evaluators} labeled the QA pair as ``No, the sentence is typical" (Approp.). Conversely, QA pairs were considered ``inappropriate" when either evaluator labeled the QA pair as anything other than ``No, the sentence is typical" (Inapprop.). QA pairs were considered ``culturally appropriate" if they were ``appropriate" and if both \textit{evaluators} labeled them as a ``good" or ``understandable" translation (see Table \ref{table:annotator_confusion_1}), hence why we group ``good" and ``understandable" quality annotations together (Good / Unders.), separating them from ``wrong" quality annotations (Wrong). As such, the top row in the table is an example of a culturally appropriate Afrikaans Winogrande QA pair while the other rows are examples of culturally inappropriate Afrikaans Winogrande QA pairs. Note that for Afrikaans, there was no ``appropriate" yet ``wrong" QA pair.}
\label{table:annotation_examples_af}
\end{table*}

\FloatBarrier

\begin{table*}[h!]
\centering
\footnotesize
\setlength\extrarowheight{2pt}
\begin{tabularx}{\textwidth}{|c|c|X|c|c|X|c|c|c|}
\hline
\textbf{} & \textbf{} & \multicolumn{3}{c|}{\textbf{Original English Version}} & \multicolumn{3}{c|}{\textbf{Translated Sepedi Version}} & \textbf{} \\
\cline{3-8} % Draws horizontal line only under grouped columns
\multirow{-2}{*}{\textbf{Approp.}} & \multirow{-2}{*}{\textbf{Quality}} 
& \centering\textbf{Sentence} & \textbf{Opt. 1} & \textbf{Opt. 2} 
& \centering\textbf{Sentence} & \textbf{Opt. 1} & \textbf{Opt. 2} 
& \multirow{-2}{*}{\textbf{Ans.}} \\
\hline
Approp. & Good / Unders. & Betty avoided getting attacked in the dark alley where Elena was mugged, because \_ ignored their intuition. & Betty & Elena & Betty o ile a phema go hlaselwa mokgotheng wo o bego o le lefsifsing moo Elena a ilego a hlaselwa gona, ka gobane \_ o ile a hlokomologa maikwelo a bona. & Betty & Elena & 2 \\ \hline
Inapprop. & Good / Unders. & Michael ran track as a teenager in high school, but Lawrence joined the math club, because \_ was more studious. & Michael & Lawrence & Michael o be a kitima dipapading ge e be e sa le mofsa yo a lego mahlalagading sekolong se se phagamego, eupša Lawrence o ile a tsenela sehlopha sa dipalo, ka gobane \_ o be a le mafolofolo kudu. & Michael & Lawrence & 2 \\ \hline
Approp. & Wrong & John had to bow to walk through the door but have to crawl to enter the tent. The \_ is shorter. & door & tent & Johane o ile a swanelwa ke go khunama gore a tsene ka mojako eupša o ile a swanelwa ke go khukhuna gore a tsene ka tenteng. \_ ke e kopana. & monyako & tente & 2 \\ \hline
Inapprop. & Wrong & Victoria was disgusted because of the odor of Sarah, and \_ did not want to be around them. & Victoria & Sarah & Ditšhelete tšeo di ile tša lefša ke Jason go Dennis, ka ge \_ a ile a adima tšhelete e ntši kudu nakong e fetilego. & Jason & Dennis & 1 \\ \hline
\end{tabularx}
\caption{\textbf{Examples of Winogrande Translation Appropriateness and Quality Annotation Combinations (Sepedi).} The table gives example Winogrande QA pairs for each of the possible combinations of translation appropriateness and quality according to annotations made by evaluators on our Winogrande translations to Sepedi. QA pairs were considered ``appropriate" when both \textit{evaluators} labeled the QA pair as ``No, the sentence is typical" (Approp.). Conversely, QA pairs were considered ``inappropriate" when either evaluator labeled the QA pair as anything other than ``No, the sentence is typical" (Inapprop.). QA pairs were considered ``culturally appropriate" if they were ``appropriate" and if both \textit{evaluators} labeled them as a ``good" or ``understandable" translation (see Table \ref{table:annotator_confusion_1}), hence why we group ``good" and ``understandable" quality annotations together (Good / Unders.), separating them from ``wrong" quality annotations (Wrong). As such, the top row in the table is an example of a culturally appropriate Sepedi Winogrande QA pair while the other rows are examples of culturally inappropriate Sepedi Winogrande QA pairs.}
\label{table:annotation_examples_nso}
\end{table*}

\FloatBarrier

\begin{table*}[h!]
\centering
\footnotesize
\setlength\extrarowheight{2pt}
\begin{tabularx}{\textwidth}{|c|c|X|c|c|X|c|c|c|}
\hline
\textbf{} & \textbf{} & \multicolumn{3}{c|}{\textbf{Original English Version}} & \multicolumn{3}{c|}{\textbf{Translated Sesotho Version}} & \textbf{} \\
\cline{3-8} % Draws horizontal line only under grouped columns
\multirow{-2}{*}{\textbf{Approp.}} & \multirow{-2}{*}{\textbf{Quality}} 
& \centering\textbf{Sentence} & \textbf{Opt. 1} & \textbf{Opt. 2} 
& \centering\textbf{Sentence} & \textbf{Opt. 1} & \textbf{Opt. 2} 
& \multirow{-2}{*}{\textbf{Ans.}} \\
\hline
Approp. & Good / Unders. & The horse prefers the pasture over the barn because there is more room in the \_ for the horse to stretch. & pasture & barn & Pere e rata lekhulo ho feta moliko hobane ho na le sebaka se sengata \_seo pere e ka ikotlollang. & lekhulo & molikong & 1 \\ \hline
Inapprop. & Good / Unders. & Cynthia had to pee after one large coffee but Betty did not as \_ had a very big bladder. & Cynthia & Betty & Cynthia o ile a tlameha ho rota ka mor'a kofi e le 'ngoe e kholo empa Betty ha aa ka a etsa joalo kaha \_o ne a e na le senya se seholo haholo. & Cynthia & Betty & 2 \\ \hline
Approp. & Wrong & Her boyfriend regretted buying her a jacket and got her a new hat instead, as he knew she thought the \_ was a bad match for her wardrobe. & jacket & hat & Mohlankana oa hae o ile a ikoahlaela ho mo rekela baki le ho mo rekela katiba e ncha, kaha o ne a tseba hore o ne a nahana hore \_ha e tšoanele liaparo tsa hae. & baki & katiba & 1 \\ \hline
Inapprop. & Wrong & While in Tampa, Laura lingered at a pirate ship and Lindsey hurried through a museum, so \_ was early to their meeting. & Laura & Lindsey & Ha a ntse a le Tampa, Laura o ile a lieha ho kena sekepeng sa masholu ’me Lindsey a potlakela ho haola musiamo, kahoo \_o ne a sa le hoseng ho ea sebokeng sa hae. & Laura & Lindsey & 2 \\ \hline
\end{tabularx}
\caption{\textbf{Examples of Winogrande Translation Appropriateness and Quality Annotation Combinations (Sesotho).} The table gives example Winogrande QA pairs for each of the possible combinations of translation appropriateness and quality according to annotations made by evaluators on our Winogrande translations to Sesotho. QA pairs were considered ``appropriate" when both \textit{evaluators} labeled the QA pair as ``No, the sentence is typical" (Approp.). Conversely, QA pairs were considered ``inappropriate" when either evaluator labeled the QA pair as anything other than ``No, the sentence is typical" (Inapprop.). QA pairs were considered ``culturally appropriate" if they were ``appropriate" and if both \textit{evaluators} labeled them as a ``good" or ``understandable" translation (see Table \ref{table:annotator_confusion_1}), hence why we group ``good" and ``understandable" quality annotations together (Good / Unders.), separating them from ``wrong" quality annotations (Wrong). As such, the top row in the table is an example of a culturally appropriate Sesotho Winogrande QA pair while the other rows are examples of culturally inappropriate Sesotho Winogrande QA pairs.}
\label{table:annotation_examples_st}
\end{table*}

\FloatBarrier

\begin{table*}[h!]
\centering
\footnotesize
\setlength\extrarowheight{2pt}
\begin{tabularx}{\textwidth}{|c|c|X|c|c|X|Q|c|c|}
\hline
\textbf{} & \textbf{} & \multicolumn{3}{c|}{\textbf{Original English Version}} & \multicolumn{3}{c|}{\textbf{Translated Setswana Version}} & \textbf{} \\
\cline{3-8} % Draws horizontal line only under grouped columns
\multirow{-2}{*}{\textbf{Approp.}} & \multirow{-2}{*}{\textbf{Quality}} 
& \centering\textbf{Sentence} & \textbf{Opt. 1} & \textbf{Opt. 2} 
& \centering\textbf{Sentence} & \centering\textbf{Opt. 1} & \textbf{Opt. 2} 
& \multirow{-2}{*}{\textbf{Ans.}} \\
\hline
Approp. & Good / Unders. & Brown rice is today's special explains Sarah to Elena, \_ is buying lunch for today. & Sarah & Elena & Raese e e borokwa ke e e kgethegileng ya gompieno Sarah o tlhalosetsa Elena, \_ o reka dijo tsa motshegare tsa gompieno. & \centering{Sarah} & Elena & 2 \\ \hline
Inapprop. & Good / Unders. & I wanted to build a bathroom on the third floor of the house but I couldn't because the \_ would be too full. & bathroom & floor & Ke ne ke batla go aga ntlwana ya botlhapelo mo boalong jwa boraro jwa ntlo mme ke ne ke sa kgone ka gonne \_ e ne e tla tlala thata. & \centering{phaposi ya botlhapelo} & boalo & 2 \\ \hline
Approp. & Wrong & Adam was law-abiding and not a crook like Joseph, so it was surprising that \_ was the one with a restraining order against him. & Adam & Joseph & Adam e ne e le leferefere mme e seng moagi yo o ikobelang molao jaaka Joseph, ka jalo go ne go gakgamatsa gore \_ e ne e le ene yo o nang le taelo ya go mo thibela. & \centering{Adam} & Joseph & 1 \\ \hline
Inapprop. & Wrong & Emily looked up and saw Patricia racing by overhead, as \_ was under the ramp . & Emily & Patricia & Emily o ne a leba kwa godimo mme a bona Patricia a taboga ka fa godimo ga tlhogo, ka \_ a ne a le ka fa tlase ga tsela . & \centering{Emily} & Patricia & 1 \\ \hline
\end{tabularx}
\caption{\textbf{Examples of Winogrande Translation Appropriateness and Quality Annotation Combinations (Setswana).} The table gives example Winogrande QA pairs for each of the possible combinations of translation appropriateness and quality according to annotations made by evaluators on our Winogrande translations to Setswana. QA pairs were considered ``appropriate" when both \textit{evaluators} labeled the QA pair as ``No, the sentence is typical" (Approp.). Conversely, QA pairs were considered ``inappropriate" when either evaluator labeled the QA pair as anything other than ``No, the sentence is typical" (Inapprop.). QA pairs were considered ``culturally appropriate" if they were ``appropriate" and if both \textit{evaluators} labeled them as a ``good" or ``understandable" translation (see Table \ref{table:annotator_confusion_1}), hence why we group ``good" and ``understandable" quality annotations together (Good / Unders.), separating them from ``wrong" quality annotations (Wrong). As such, the top row in the table is an example of a culturally appropriate Setswana Winogrande QA pair while the other rows are examples of culturally inappropriate Setswana Winogrande QA pairs.}
\label{table:annotation_examples_tn}
\end{table*}

\FloatBarrier

\begin{table*}[h!]
\centering
\footnotesize
\setlength\extrarowheight{2pt}
\begin{tabularx}{\textwidth}{|c|c|X|c|c|X|c|Q|c|}
\hline
\textbf{} & \textbf{} & \multicolumn{3}{c|}{\textbf{Original English Version}} & \multicolumn{3}{c|}{\textbf{Translated Shona Version}} & \textbf{} \\
\cline{3-8} % Draws horizontal line only under grouped columns
\multirow{-2}{*}{\textbf{Approp.}} & \multirow{-2}{*}{\textbf{Quality}} 
& \centering\textbf{Sentence} & \textbf{Opt. 1} & \textbf{Opt. 2} 
& \centering\textbf{Sentence} & \textbf{Opt. 1} & \centering\textbf{Opt. 2} 
& \multirow{-2}{*}{\textbf{Ans.}} \\
\hline
Approp. & Good / Unders. & Sue wanted to practice meditation so she went to the gym with her class but the \_ was too late. & class & gym & Sue aida kudzidzira kudzikamisa pfungwa saka akaenda kugym nevaayidzidza navo asi \_ vaaidzidza navo vakanonoka. & muclass & \centering{mugym} & 1 \\ \hline
Inapprop. & Good / Unders. & Angela noticed the lumps on Amy 's arms that she had failed to notice, \_ is just oblivious that way. & Angela & Amy & Angela akaona mapundu aive mumaoko aAmy aainge atadza kuona, \_ anongoshaya hanya nenzira iyoyo. & Angela & \centering{Amy} & 2 \\ \hline
Approp. & Wrong & After the loss of the raft, the crew took the lifeboat as the \_ was older anyway. & raft & lifeboat & Munguva yechamupupuri, gwenzi rakadonha muti usati wadonha, nokuti \_ makanga muine midzi dzakanyudza. & igwa & \centering{chikepe chekununura} & 1 \\ \hline
Inapprop. & Wrong & Cosplay really interested Matthew, but Donald didn't know much about it, so \_ tried to learn about it. & Matthew & Donald & Cosplay aifarira chaizvo Matthew, asi Donald aisaziva zvakawanda nezvazvo, saka \_ akaedza kudzidza nezvazvo. & Matthew & \centering{Donald} & 2 \\ \hline
\end{tabularx}
\caption{\textbf{Examples of Winogrande Translation Appropriateness and Quality Annotation Combinations (Shona).} The table gives example Winogrande QA pairs for each of the possible combinations of translation appropriateness and quality according to annotations made by evaluators on our Winogrande translations to Shona. QA pairs were considered ``appropriate" when both \textit{evaluators} labeled the QA pair as ``No, the sentence is typical" (Approp.). Conversely, QA pairs were considered ``inappropriate" when either evaluator labeled the QA pair as anything other than ``No, the sentence is typical" (Inapprop.). QA pairs were considered ``culturally appropriate" if they were ``appropriate" and if both \textit{evaluators} labeled them as a ``good" or ``understandable" translation (see Table \ref{table:annotator_confusion_1}), hence why we group ``good" and ``understandable" quality annotations together (Good / Unders.), separating them from ``wrong" quality annotations (Wrong). As such, the top row in the table is an example of a culturally appropriate Shona Winogrande QA pair while the other rows are examples of culturally inappropriate Shona Winogrande QA pairs.}
\label{table:annotation_examples_sn}
\end{table*}

\FloatBarrier

\begin{table*}[h!]
\centering
\footnotesize
\setlength\extrarowheight{2pt}
\begin{tabularx}{\textwidth}{|c|c|X|c|c|X|c|c|c|}
\hline
\textbf{} & \textbf{} & \multicolumn{3}{c|}{\textbf{Original English Version}} & \multicolumn{3}{c|}{\textbf{Translated Tsonga Version}} & \textbf{} \\
\cline{3-8} % Draws horizontal line only under grouped columns
\multirow{-2}{*}{\textbf{Approp.}} & \multirow{-2}{*}{\textbf{Quality}} 
& \centering\textbf{Sentence} & \textbf{Opt. 1} & \textbf{Opt. 2} 
& \centering\textbf{Sentence} & \textbf{Opt. 1} & \textbf{Opt. 2} 
& \multirow{-2}{*}{\textbf{Ans.}} \\
\hline
Approp. & Good / Unders. & Victoria is makeup artist and Cynthia is a beauty queen, \_ knows a lot about makeup. & Victoria & Cynthia & Victoria i makeup artist kasi Cynthia i hosikati ya vumbhuri, \_ u tiva swo tala hi makeup. & Victoria  & Cynthia  & 1 \\ \hline
Inapprop. & Good / Unders. & James got the rod broken under his grip when he was trying to bend it to shape because the \_ is strong. & rod & grip & James u tshovile nhonga hi nkhomo wa yena loko a karhi a ringeta ku yi petsa hikuva \_ a yi tiyile. & nhonga & nkhomo & 2 \\ \hline
Approp. & Wrong & The doctor gave no fluids to Neil and gave them to Jason since \_ was not dehydrated. & Neil & Jason & Dokodela a nga nyikanga Neil mati kutani a ma nyika Jason tanihi leswi \_ a nga heleriwangi hi mati ya mirhi. & Neil & Jason & 1 \\ \hline
Inapprop. & Wrong & Lawrence planned to steal the valuable painting from Michael, because \_ wanted to own something beautiful. & Lawrence & Michael & Lawrence u kunguhate ku yiva swilo swa nkoka swa swifaniso swo vatliwa, hikuva\_a lava ku va na wanchumu wo saseka. & Lawrence & Michael & 1 \\ \hline
\end{tabularx}
\caption{\textbf{Examples of Winogrande Translation Appropriateness and Quality Annotation Combinations (Tsonga).} The table gives example Winogrande QA pairs for each of the possible combinations of translation appropriateness and quality according to annotations made by evaluators on our Winogrande translations to Tsonga. QA pairs were considered ``appropriate" when both \textit{evaluators} labeled the QA pair as ``No, the sentence is typical" (Approp.). Conversely, QA pairs were considered ``inappropriate" when either evaluator labeled the QA pair as anything other than ``No, the sentence is typical" (Inapprop.). QA pairs were considered ``culturally appropriate" if they were ``appropriate" and if both \textit{evaluators} labeled them as a ``good" or ``understandable" translation (see Table \ref{table:annotator_confusion_1}), hence why we group ``good" and ``understandable" quality annotations together (Good / Unders.), separating them from ``wrong" quality annotations (Wrong). As such, the top row in the table is an example of a culturally appropriate Tsonga Winogrande QA pair while the other rows are examples of culturally inappropriate Tsonga Winogrande QA pairs.}
\label{table:annotation_examples_ts}
\end{table*}

\FloatBarrier

\begin{table*}[h!]
\centering
\footnotesize
\setlength\extrarowheight{2pt}
\begin{tabularx}{\textwidth}{|c|c|X|c|c|X|c|c|c|}
\hline
\textbf{} & \textbf{} & \multicolumn{3}{c|}{\textbf{Original English Version}} & \multicolumn{3}{c|}{\textbf{Translated Xhosa Version}} & \textbf{} \\
\cline{3-8} % Draws horizontal line only under grouped columns
\multirow{-2}{*}{\textbf{Approp.}} & \multirow{-2}{*}{\textbf{Quality}} 
& \centering\textbf{Sentence} & \textbf{Opt. 1} & \textbf{Opt. 2} 
& \centering\textbf{Sentence} & \textbf{Opt. 1} & \textbf{Opt. 2} 
& \multirow{-2}{*}{\textbf{Ans.}} \\
\hline
Approp. & Good / Unders. & When it came to taking care of elderly people, Adam was suited for the job more than Brian because \_ lived with younger people longer. & Adam & Brian & Xa kufikelelwa kumba wokunyamekela abantu abadala, uAdam wayewufanelekela lo msebenzi ngaphezu kukaBrian kuba \_ wayehlala nabantu abatsha ixesha elide. & Adam & Brian & 2 \\ \hline
Inapprop. & Good / Unders. & Randy has a lot of wrinkles on their face but Matthew does not. \_ went to the dermatologist for treatment. & Randy & Matthew & URandy unemibimbi emininzi ebusweni kodwa uMatthew akanayo. \_ waya kwi-dermatologist ukufumana unyango. & uRandy & uMatthew & 2 \\ \hline
Approp. & Wrong & Cathy felt a stronger bond with her dog than her cat because her \_ was very aloof. & cat & dog & U Cathy waziva esondelelene nenja yakhe ngaphezu kwekati yakhe kuba \_ wayezikhethele. & ikati & inja & 1 \\ \hline
Inapprop. & Wrong & I made more mistakes in the history test than errors in the math test, so the \_ were fewer because I love math. & errors & mistakes & Ndenze iimpazamo ezininzi kuvavanyo lwembali kuneziphoso kuvavanyo lwezibalo, ngoko ke \_ ibimbalwa kuba ndiyazithanda izibalo. & iimpazamo & iimpazamo & 1 \\ \hline
\end{tabularx}
\caption{\textbf{Examples of Winogrande Translation Appropriateness and Quality Annotation Combinations (Xhosa).} The table gives example Winogrande QA pairs for each of the possible combinations of translation appropriateness and quality according to annotations made by evaluators on our Winogrande translations to Xhosa. QA pairs were considered ``appropriate" when both \textit{evaluators} labeled the QA pair as ``No, the sentence is typical" (Approp.). Conversely, QA pairs were considered ``inappropriate" when either evaluator labeled the QA pair as anything other than ``No, the sentence is typical" (Inapprop.). QA pairs were considered ``culturally appropriate" if they were ``appropriate" and if both \textit{evaluators} labeled them as a ``good" or ``understandable" translation (see Table \ref{table:annotator_confusion_1}), hence why we group ``good" and ``understandable" quality annotations together (Good / Unders.), separating them from ``wrong" quality annotations (Wrong). As such, the top row in the table is an example of a culturally appropriate Xhosa Winogrande QA pair while the other rows are examples of culturally inappropriate Xhosa Winogrande QA pairs.}
\label{table:annotation_examples_xh}
\end{table*}

\FloatBarrier

\begin{table*}[h!]
\centering
\footnotesize
\setlength\extrarowheight{2pt}
\begin{tabularx}{\textwidth}{|c|c|X|c|c|X|Q|Q|c|}
\hline
\textbf{} & \textbf{} & \multicolumn{3}{c|}{\textbf{Original English Version}} & \multicolumn{3}{c|}{\textbf{Translated Zulu Version}} & \textbf{} \\
\cline{3-8} % Draws horizontal line only under grouped columns
\multirow{-2}{*}{\textbf{Approp.}} & \multirow{-2}{*}{\textbf{Quality}} 
& \centering\textbf{Sentence} & \textbf{Opt. 1} & \textbf{Opt. 2} 
& \centering\textbf{Sentence} & \centering{\textbf{Opt. 1}} & \centering{\textbf{Opt. 2}}
& \multirow{-2}{*}{\textbf{Ans.}} \\
\hline
Approp. & Good / Unders. & Brian was charged with a felony for stealing the diamonds even though he hadn't touched the gold bars, because the original owner cared more about the \_ . & bars & diamonds & UBrian wabekwa icala lokweba amadayimane nakuba ayengazange azithinte izigxobo zegolide, ngoba umnikazi wangempela wayenendaba n\_ . & \centering{izingcezu} & \centering{amaday- imane} & 2 \\ \hline
Inapprop. & Good / Unders. & Monica was taking a knitting class, Elena giggled at this. This made \_ feel bad. & Monica & Elena & UMonica wayefunda ekilasini lokuluka, u-Elena wagigitheka ngalokhu. Lokhu kwenza \_ waphatheka kabi. & \centering{uMonica} & \centering{u-Elena} & 1 \\ \hline
Approp. & Wrong & He try hard jerk decided to lift kettle bells instead of free weights, because he heard the \_ were more macho. & kettle bells & free weights & Uzama jerk kanzima wanquma ukuphakamisa izinsimbi iketela esikhundleni izisindo khulula, ngoba wezwa \_ babe macho ngaphezulu. & \centering{Amak- hombothi ezibhengu} & \centering{Izisindo zamahhala} & 1 \\ \hline
Inapprop. & Wrong & Erin ordered a martini when Tanya only ordered a tonic water with lemon, because \_ was pregnant. & Erin & Tanya & uErin waOda iMartini lapho uTanya eyala kuphela amanzi eTonic anolamula, ngoba \_ wayekhulelwe. & \centering{uErin} & \centering{uTanya} & 2 \\ \hline
\end{tabularx}
\caption{\textbf{Examples of Winogrande Translation Appropriateness and Quality Annotation Combinations (Zulu).} The table gives example Winogrande QA pairs for each of the possible combinations of translation appropriateness and quality according to annotations made by evaluators on our Winogrande translations to Zulu. QA pairs were considered ``appropriate" when both \textit{evaluators} labeled the QA pair as ``No, the sentence is typical" (Approp.). Conversely, QA pairs were considered ``inappropriate" when either evaluator labeled the QA pair as anything other than ``No, the sentence is typical" (Inapprop.). QA pairs were considered ``culturally appropriate" if they were ``appropriate" and if both \textit{evaluators} labeled them as a ``good" or ``understandable" translation (see Table \ref{table:annotator_confusion_1}), hence why we group ``good" and ``understandable" quality annotations together (Good / Unders.), separating them from ``wrong" quality annotations (Wrong). As such, the top row in the table is an example of a culturally appropriate Zulu Winogrande QA pair while the other rows are examples of culturally inappropriate Zulu Winogrande QA pairs.}
\label{table:annotation_examples_zu}
\end{table*}

\FloatBarrier

\begin{table*}[h]
\centering
\setlength\extrarowheight{4pt}
\begin{tabular}{|cc|ccccccccccc|c|}
\hline
\textbf{Eval. 1} & \textbf{Eval. 2} & \textbf{xh} & \textbf{zu} & \textbf{af} & \textbf{ig} & \textbf{sn} & \textbf{ts} & \textbf{st} & \textbf{nso} & \textbf{tn} & \textbf{am} & \textbf{bm} & \textbf{Avg. (\%)} \\
\hline
Good       & Good      & 2406 & 1349 & 2522 & 1357 & 2561 & 2474 & 2952 & 3076 & 2941 & 2813 & 1751 & 64.8\%       \\
Good       & Unders.   & 552  & 771  & 353  & 1637 & 126  & 136  & 4    & 167  & 59   & 51   & 306  & 10.3\%       \\
Good       & Wrong     & 24   & 39   & 36   & 369  & 36   & 11   & 0    & 42   & 3    & 1    & 27   & 1.5\%        \\
Unders.    & Good      & 312  & 898  & 455  & 106  & 689  & 674  & 605  & 290  & 597  & 778  & 967  & 15.8\%       \\
Unders.    & Unders.   & 185  & 452  & 139  & 157  & 175  & 126  & 6    & 37   & 15   & 18   & 242  & 3.8\%        \\
Unders.    & Wrong     & 3    & 33   & 20   & 32   & 39   & 6    & 0    & 21   & 2    & 2    & 15   & 0.4\%        \\
Wrong      & Good      & 88   & 68   & 80   & 4    & 26   & 186  & 106  & 12   & 52   & 11   & 249  & 2.2\%        \\
Wrong      & Unders.   & 100  & 61   & 56   & 8    & 7    & 49   & 1    & 2    & 2    & 0    & 99   & 1.0\%        \\
Wrong      & Wrong     & 4    & 3    & 13   & 4    & 15   & 12   & 0    & 27   & 3    & 0    & 18   & 0.2\%        \\
\hline
\multicolumn{14}{|c|}{Study Definitions of Translation Quality}   \\ \hline
\multicolumn{2}{|c|}{\textbf{Quality}} & \textbf{xh} & \textbf{zu} & \textbf{af} & \textbf{ig} & \textbf{sn} & \textbf{ts} & \textbf{st} & \textbf{nso} & \textbf{tn} & \textbf{am} & \textbf{bm} & \textbf{Avg. (\%)}  \\ \hline
\multicolumn{2}{|c|}{Good / Unders.}  & 3455 & 3470 & 3469 & 3257 & 3551 & 3410 & 3567 & 3570 & 3612 & 3660 & 3266 & 94.7\%       \\
\multicolumn{2}{|c|}{Wrong} & 219  & 204  & 205  & 417  & 123  & 264  & 107  & 104  & 62   & 14   & 408  & 5.3\%        \\ \hline
\end{tabular}
\caption{\textbf{Human Evaluation of Winogrande Translation Quality.} The table above lists the total number of all possible combinations of \textit{evaluator} responses for the first survey question on the full Winogrande dataset, rating the quality of the translations as ``Good translation" (Good), ``Incorrect, but someone could understand the idea" (Unders.), and ``Completely wrong" (Wrong). The bottom section, labeled ``Study Definitions of Translation Quality", shows two definitions of translation quality used in our study. First are QA pairs considered ``Good/Understandable", where both \textit{evaluators} labeled the QA pair as ``Good translation" or ``Incorrect, but someone could understand the idea". Second are QA pairs considered ``Wrong" where either \textit{evaluator} labeled the QA pair as ``Completely wrong". Language codes are as follows: Xhosa (xh), Zulu (zu), Afrikaans (af), Igbo (ig), Shona (sn), Tsonga (ts), Sesotho (st), Sepedi (nso), Setswana (tn), Amharic (am), Bambara (bm).}
\label{table:annotator_confusion_1}
\end{table*}

\begin{table*}[t!]
\centering
\setlength\extrarowheight{2pt}
% \small
\begin{tabular}{|l|llllllllllll|}
\hline \hline
                 \multicolumn{13}{c}{\texttt{Winogrande}}  \\ \hline
                & \textbf{en} & \textbf{af} & \textbf{zu} & \textbf{xh} & \textbf{am} & \textbf{bm} & \textbf{ig} & \textbf{nso} & \textbf{sn} & \textbf{st} & \textbf{tn} & \textbf{ts} \\ \hline 
                \multicolumn{13}{c}{} \\ \hline
                \multicolumn{13}{|c|}{Machine Translation (en $\rightarrow$ $x$)}  \\ \hline
{GPT-4o} & {83.9} & {77.9} & {66.5} & {65.0} & {56.9} & {51.1} & {62.3} & {61.5} & {65.7} & {65.4} & {61.7} & {61.6} \\
{GPT-4} & {83.5} & {74.8} & {63.2} & {62.5} & {50.9} & {51.2} & {60.2} & {56.0} & {62.7} & {60.0} & {57.7} & {57.0} \\
{GPT-3.5} & {59.6} & {55.9} & {51.3} & {51.6} & {50.0} & {51.5} & {51.9} & {49.7} & {49.9} & {50.2} & {52.0} & {51.0} \\
{Llama 3 70B IT} & {61.2} & {51.7} & {50.5} & {50.5} & {50.4} & {50.7} & {50.4} & {50.3} & {50.4} & {50.5} & {50.4} & {50.4} \\
{Llama 3 8B IT} & {52.0} & {50.9} & {50.5} & {50.4} & {50.4} & {50.4} & {50.4} & {50.4} & {50.3} & {50.4} & {50.5} & {50.4} \\
{Phi 3 Mini 4K IT} & {64.7} & {53.0} & {49.7} & {49.8} & {50.7} & {51.2} & {51.0} & {51.3} & {50.9} & {50.7} & {50.3} & {51.0} \\
{Aya 23 35B} & {68.7} & {55.2} & {50.4} & {51.2} & {48.3} & {52.8} & {52.0} & {49.6} & {49.6} & {50.8} & {52.0} & {51.2} \\
{Aya 101} & {49.5} & {49.1} & {51.1} & {49.3} & {50.0} & {49.0} & {50.0} & {49.2} & {49.0} & {50.1} & {49.9} & {50.5} \\
{BLOOMZ 7b1} & {48.6} & {50.3} & {45.4} & {48.6} & {48.7} & {51.5} & {48.3} & {49.8} & {48.7} & {49.3} & {49.6} & {50.0} \\
\hline
                \multicolumn{13}{|c|}{Difference Between Human (Table \ref{table:perf-results}) and Machine Translation}\\ \hline
{GPT-4o} & {0.0} & {1.8} & {1.8} & {0.9} & {2.5} & {-0.9} & {-1.6} & {2.6} & {3.8} & {2.0} & {3.0} & {1.0} \\
{GPT-4} & {0.0} & {2.2} & {1.0} & {-0.2} & {0.1} & {-0.5} & {-1.5} & {2.8} & {2.9} & {3.8} & {2.2} & {0.7} \\
{GPT-3.5} & {0.0} & {-0.9} & {-0.5} & {0.6} & {1.3} & {-1.1} & {0.0} & {0.5} & {1.7} & {-1.0} & {-0.5} & {-1.4} \\
{Llama 3 70B IT} & {0.0} & {-0.7} & {-0.1} & {0.3} & {0.4} & {-0.2} & {0.1} & {0.2} & {0.0} & {-0.1} & {0.1} & {0.0} \\
{Llama 3 8B IT} & {0.0} & {-0.2} & {-0.1} & {0.0} & {-0.1} & {0.0} & {0.0} & {0.0} & {0.1} & {0.0} & {-0.1} & {0.0} \\
{Phi 3 Mini 4K IT} & {0.0} & {-1.4} & {0.5} & {1.5} & {-0.5} & {-1.2} & {0.0} & {-1.6} & {1.0} & {-1.2} & {0.6} & {-0.1} \\
{Aya 23 35B} & {0.0} & {1.3} & {-0.8} & {0.6} & {3.0} & {-2.0} & {-1.4} & {1.8} & {1.0} & {-0.8} & {-2.2} & {-0.7} \\
{Aya 101} & {0.0} & {2.0} & {-2.0} & {1.9} & {1.2} & {3.0} & {0.5} & {-0.2} & {2.0} & {0.4} & {0.9} & {-0.9} \\
{BLOOMZ 7b1} & {0.0} & {0.0} & {3.3} & {0.2} & {0.6} & {-1.5} & {0.5} & {-1.9} & {-0.1} & {0.3} & {-0.4} & {-0.9} \\
\hline
                \multicolumn{13}{c}{} \\ \hline
                \multicolumn{13}{|c|}{Machine Backtranslation ($x$ $\rightarrow$ en)} \\ \hline
{GPT-4o} & {83.9} & {79.3} & {67.9} & {68.4} & {66.5} & {59.8} & {65.3} & {66.6} & {66.5} & {68.1} & {67.0} & {65.9} \\
{GPT-4} & {83.5} & {78.3} & {67.3} & {65.3} & {62.4} & {58.7} & {65.1} & {67.1} & {66.8} & {67.2} & {64.3} & {66.2} \\
{GPT-3.5} & {59.6} & {58.4} & {53.7} & {54.1} & {53.9} & {52.6} & {54.6} & {55.0} & {54.4} & {54.6} & {55.6} & {53.0} \\
{Llama 3 70B IT} & {61.2} & {59.7} & {54.4} & {54.5} & {53.1} & {53.9} & {53.5} & {53.2} & {54.5} & {53.8} & {53.9} & {52.6} \\
{Llama 3 8B IT} & {52.0} & {51.2} & {52.1} & {51.0} & {52.1} & {50.9} & {51.5} & {51.5} & {51.3} & {51.3} & {51.9} & {50.7} \\
{Phi 3 Mini 4K IT} & {64.7} & {63.2} & {57.0} & {57.2} & {55.4} & {54.6} & {55.1} & {54.7} & {57.0} & {54.9} & {58.0} & {55.6} \\
{Aya 23 35B} & {68.7} & {66.8} & {58.9} & {58.7} & {55.8} & {56.0} & {57.7} & {59.3} & {57.3} & {57.8} & {58.9} & {58.2} \\
{Aya 101} & {49.5} & {51.0} & {50.9} & {49.3} & {48.2} & {51.2} & {52.1} & {49.4} & {48.2} & {50.4} & {49.5} & {51.0} \\
{BLOOMZ 7b1} & {48.6} & {50.0} & {48.8} & {51.0} & {49.6} & {50.0} & {50.3} & {49.6} & {50.4} & {50.6} & {49.7} & {50.7} \\ \hline
% {BLOOMZ 7b1}    & 79.0 & 37.0 & 35.8 & 35.1 & ? & ?  & ? & ? & ? & ? & ? & ? 
% \\ \hline \hline

                \multicolumn{13}{|c|}{Difference Between Human (Table \ref{table:perf-results}) and Machine Translation}\\ \hline
{GPT-4o} & {0.0} & {0.4} & {0.4} & {-2.5} & {-7.1} & {-9.6} & {-4.6} & {-2.5} & {3.0} & {-0.7} & {-2.3} & {-3.3} \\
{GPT-4} & {0.0} & {-1.3} & {-3.1} & {-3.0} & {-11.4} & {-8.0} & {-6.4} & {-8.3} & {-1.2} & {-3.4} & {-4.4} & {-8.5} \\
{GPT-3.5} & {0.0} & {-3.4} & {-2.9} & {-1.9} & {-2.6} & {-2.2} & {-2.7} & {-4.8} & {-2.8} & {-5.4} & {-4.1} & {-3.4} \\
{Llama 3 70B IT} & {0.0} & {-8.7} & {-4.0} & {-3.7} & {-2.3} & {-3.4} & {-3.0} & {-2.7} & {-4.1} & {-3.4} & {-3.4} & {-2.2} \\
{Llama 3 8B IT} & {0.0} & {-0.5} & {-1.7} & {-0.6} & {-1.8} & {-0.5} & {-1.1} & {-1.1} & {-0.9} & {-0.9} & {-1.5} & {-0.3} \\
{Phi 3 Mini 4K IT} & {0.0} & {-11.6} & {-6.8} & {-5.9} & {-5.2} & {-4.6} & {-4.1} & {-5.0} & {-5.1} & {-5.4} & {-7.1} & {-4.7} \\
{Aya 23 35B} & {0.0} & {-10.3} & {-9.3} & {-6.9} & {-4.5} & {-5.2} & {-7.1} & {-7.9} & {-6.7} & {-7.8} & {-9.1} & {-7.7} \\
{Aya 101} & {0.0} & {0.1} & {-1.8} & {1.9} & {3.0} & {0.8} & {-1.6} & {-0.4} & {2.8} & {0.1} & {1.3} & {-1.4} \\
{BLOOMZ 7b1} & {0.0} & {0.3} & {-0.1} & {-2.2} & {-0.3} & {0.0} & {-1.5} & {-1.7} & {-1.8} & {-1.0} & {-0.5} & {-1.6} \\
\hline
\end{tabular}
\caption{\textbf{Results of State-of-the-Art Models on Machine-Translated Winogrande}. Top state-of-the-art models were evaluated out-of-the-box (no fine-tuning) on Winogrande (binary choice co-reference resolution task) test datasets that were machine-translated from English into the target language or backtranslated from the human-translated version back into English. Google Translate was used for machine translations in all languages except Setswana (tn), where GPT-4o was used. Results are provided for 11 low-resource African languages of focus: Afrikaans (af), Zulu (zu), Xhosa (xh) (datasets for these three languages were sourced from \cite{winmmluza2024}), Amharic (am), Bambara (bm), Igbo (ig), Sepedi (nso), Shona (sn), Sesotho (st), Setswana (tn), and Tsonga (ts). Results on English (en) without any translation are also provided as a reference. Differences with corresponding numbers from Table \ref{table:perf-results} are given, with negative values indicating that machine translation led to greater performance than human-translation. All numbers are 5-shot performance accuracy. 
} \label{table:perf-results-mt-wino}
\end{table*}

\begin{table*}[t!]
\centering
\setlength\extrarowheight{2pt}
% \small
\begin{tabular}{|l|llllllllllll|}
\hline \hline
                 \multicolumn{13}{c}{\texttt{MMLU College Medicine}}  \\ \hline
                & \textbf{en} & \textbf{af} & \textbf{zu} & \textbf{xh} & \textbf{am} & \textbf{bm} & \textbf{ig} & \textbf{nso} & \textbf{sn} & \textbf{st} & \textbf{tn} & \textbf{ts} \\ \hline 
                \multicolumn{13}{c}{} \\ \hline
                \multicolumn{13}{|c|}{Machine Translation (en $\rightarrow$ $x$)}  \\ \hline
{GPT-4o} & {84.4} & {84.4} & {72.8} & {74.6} & {56.6} & {42.8} & {68.2} & {64.2} & {72.8} & {73.4} & {61.8} & {67.6} \\
{GPT-4} & {78.0} & {76.9} & {64.7} & {65.9} & {39.3} & {36.4} & {61.3} & {53.2} & {66.5} & {58.4} & {54.9} & {53.8} \\
{GPT-3.5} & {63.6} & {56.6} & {36.4} & {37.0} & {26.6} & {30.1} & {34.1} & {30.6} & {41.0} & {32.9} & {31.8} & {30.1} \\
{Llama 3 70B IT} & {76.9} & {66.5} & {37.0} & {38.2} & {24.4} & {27.2} & {38.2} & {27.2} & {41.6} & {34.1} & {27.7} & {33.5} \\
{Llama 3 8B IT} & {60.1} & {41.6} & {32.9} & {37.0} & {23.2} & {28.3} & {36.4} & {35.3} & {35.3} & {34.7} & {29.5} & {33.5} \\
{Phi 3 Mini 4K IT} & {66.5} & {38.2} & {34.1} & {27.2} & {22.3} & {27.2} & {38.2} & {34.7} & {37.6} & {36.4} & {26.6} & {32.9} \\
{Aya 23 35B} & {62.4} & {51.4} & {34.1} & {33.5} & {37.6} & {30.6} & {32.4} & {28.9} & {35.3} & {35.8} & {30.6} & {31.8} \\
{Aya 101} & {42.8} & {42.8} & {37.0} & {41.0} & {39.3} & {26.0} & {41.0} & {36.4} & {32.9} & {38.2} & {32.4} & {31.2} \\
{BLOOMZ 7b1} & {36.4} & {27.7} & {29.5} & {29.5} & {25.4} & {30.1} & {33.5} & {31.2} & {35.3} & {29.5} & {28.3} & {30.1} \\
\hline
                \multicolumn{13}{|c|}{Difference Between Human (Table \ref{table:perf-results}) and Machine Translation}\\ \hline
{GPT-4o} & {0.0} & {0.0} & {0.0} & {3.4} & {10.5} & {-4.6} & {-9.8} & {2.3} & {4.1} & {-6.3} & {1.8} & {-7.5} \\
{GPT-4} & {0.0} & {2.9} & {-3.4} & {-4.6} & {7.5} & {-6.3} & {-10.4} & {0.6} & {2.3} & {-3.5} & {2.9} & {-0.6} \\
{GPT-3.5} & {0.0} & {-0.5} & {-3.5} & {0.6} & {-2.3} & {0.5} & {-5.2} & {3.5} & {-5.2} & {-0.5} & {1.1} & {4.6} \\
{Llama 3 70B IT} & {0.0} & {1.7} & {-1.2} & {2.3} & {12.5} & {-2.9} & {-4.7} & {0.0} & {-3.4} & {-5.8} & {6.4} & {-4.0} \\
{Llama 3 8B IT} & {0.0} & {2.9} & {-1.1} & {0.0} & {-6.5} & {1.8} & {-11.5} & {-2.9} & {3.4} & {0.0} & {-2.3} & {-0.6} \\
{Phi 3 Mini 4K IT} & {0.0} & {1.7} & {-3.5} & {1.7} & {5.7} & {4.0} & {-9.3} & {-6.4} & {-9.3} & {-8.7} & {7.5} & {-2.8} \\
{Aya 23 35B} & {0.0} & {-2.3} & {1.7} & {-0.6} & {-2.9} & {-2.3} & {-0.6} & {6.9} & {-2.9} & {0.6} & {4.1} & {-6.9} \\
{Aya 101} & {0.0} & {-2.3} & {-1.7} & {-8.1} & {-3.5} & {5.2} & {-9.8} & {-1.7} & {9.3} & {0.0} & {2.3} & {4.6} \\
{BLOOMZ 7b1} & {0.0} & {6.4} & {0.6} & {-1.2} & {0.6} & {-1.2} & {-6.9} & {-4.0} & {-5.2} & {0.0} & {-0.6} & {-0.6} \\
\hline
                \multicolumn{13}{c}{} \\ \hline
                \multicolumn{13}{|c|}{Machine Backtranslation ($x$ $\rightarrow$ en)} \\ \hline
{GPT-4o} & {84.4} & {81.5} & {69.9} & {76.9} & {76.3} & {63.6} & {59.5} & {78.6} & {76.9} & {67.1} & {65.9} & {62.4} \\
{GPT-4} & {78.0} & {72.8} & {65.3} & {67.6} & {67.6} & {60.1} & {54.9} & {69.9} & {68.8} & {59.0} & {54.9} & {55.5} \\
{GPT-3.5} & {63.6} & {60.1} & {52.0} & {53.8} & {57.2} & {53.2} & {46.2} & {56.6} & {57.2} & {52.0} & {53.2} & {42.2} \\
{Llama 3 70B IT} & {76.9} & {75.1} & {62.4} & {67.6} & {68.2} & {56.6} & {56.1} & {68.2} & {71.7} & {60.1} & {57.8} & {53.2} \\
{Llama 3 8B IT} & {60.1} & {59.5} & {50.3} & {54.9} & {49.7} & {49.7} & {31.2} & {54.3} & {54.3} & {39.3} & {45.1} & {37.6} \\
{Phi 3 Mini 4K IT} & {66.5} & {64.2} & {49.7} & {57.8} & {59.0} & {51.4} & {44.5} & {58.4} & {60.1} & {52.6} & {49.1} & {46.2} \\
{Aya 23 35B} & {62.4} & {60.1} & {55.5} & {56.1} & {50.9} & {47.4} & {50.3} & {53.8} & {52.6} & {50.3} & {52.0} & {49.7} \\
{Aya 101} & {42.8} & {42.8} & {41.6} & {41.0} & {38.7} & {34.7} & {40.5} & {38.7} & {45.1} & {42.8} & {41.0} & {37.6} \\
{BLOOMZ 7b1} & {36.4} & {35.8} & {30.6} & {39.9} & {35.3} & {33.5} & {34.7} & {38.2} & {39.3} & {32.4} & {38.7} & {29.5} \\ \hline
% {BLOOMZ 7b1}    & 79.0 & 37.0 & 35.8 & 35.1 & ? & ?  & ? & ? & ? & ? & ? & ? 
% \\ \hline \hline

                \multicolumn{13}{|c|}{Difference Between Human (Table \ref{table:perf-results}) and Machine Translation}\\ \hline
{GPT-4o} & {0.0} & {2.9} & {2.9} & {1.1} & {-9.2} & {-25.4} & {-1.1} & {-12.1} & {0.0} & {0.0} & {-2.3} & {-2.3} \\
{GPT-4} & {0.0} & {7.0} & {-4.0} & {-6.3} & {-20.8} & {-30.0} & {-4.0} & {-16.1} & {0.0} & {-4.1} & {2.9} & {-2.3} \\
{GPT-3.5} & {0.0} & {-4.0} & {-19.1} & {-16.2} & {-32.9} & {-22.6} & {-17.3} & {-22.5} & {-21.4} & {-19.6} & {-20.3} & {-7.5} \\
{Llama 3 70B IT} & {0.0} & {-6.9} & {-26.6} & {-27.1} & {-31.3} & {-32.3} & {-22.6} & {-41.0} & {-33.5} & {-31.8} & {-23.7} & {-23.7} \\
{Llama 3 8B IT} & {0.0} & {-15.0} & {-18.5} & {-17.9} & {-33.0} & {-19.6} & {-6.3} & {-21.9} & {-15.6} & {-4.6} & {-17.9} & {-4.7} \\
{Phi 3 Mini 4K IT} & {0.0} & {-24.3} & {-19.1} & {-28.9} & {-31.0} & {-20.2} & {-15.6} & {-30.1} & {-31.8} & {-24.9} & {-15.0} & {-16.1} \\
{Aya 23 35B} & {0.0} & {-11.0} & {-19.7} & {-23.2} & {-16.2} & {-19.1} & {-18.5} & {-18.0} & {-20.2} & {-13.9} & {-17.3} & {-24.8} \\
{Aya 101} & {0.0} & {-2.3} & {-6.3} & {-8.1} & {-2.9} & {-3.5} & {-9.3} & {-4.0} & {-2.9} & {-4.6} & {-6.3} & {-1.8} \\
{BLOOMZ 7b1} & {0.0} & {-1.7} & {-0.5} & {-11.6} & {-9.3} & {-4.6} & {-8.1} & {-11.0} & {-9.2} & {-2.9} & {-11.0} & {0.0} \\
\hline
\end{tabular}
\caption{\textbf{Results of State-of-the-Art Models on Machine-Translated MMLU College Medicine}. Top state-of-the-art models were evaluated out-of-the-box (no fine-tuning) on MMLU College Medicine (medical domain knowledge task) test datasets that were machine-translated from English into the target language or backtranslated from the human-translated version back into English. Google Translate was used for machine translations in all languages except Setswana (tn), where GPT-4o was used. Results are provided for 11 low-resource African languages of focus: Afrikaans (af), Zulu (zu), Xhosa (xh) (datasets for these three languages were sourced from \cite{winmmluza2024}), Amharic (am), Bambara (bm), Igbo (ig), Sepedi (nso), Shona (sn), Sesotho (st), Setswana (tn), and Tsonga (ts). Results on English (en) without any translation are also provided as a reference. Differences with corresponding numbers from Table \ref{table:perf-results} are given, with negative values indicating that machine translation led to greater performance than human-translation. All numbers are 5-shot performance accuracy. 
} \label{table:perf-results-mt-mmlu-cm}
\end{table*}

\begin{table*}[t!]
\centering
\setlength\extrarowheight{2pt}
% \small
\begin{tabular}{|l|llllllllllll|}
\hline \hline
                 \multicolumn{13}{c}{\texttt{MMLU Clinical Knowledge}}  \\ \hline
                & \textbf{en} & \textbf{af} & \textbf{zu} & \textbf{xh} & \textbf{am} & \textbf{bm} & \textbf{ig} & \textbf{nso} & \textbf{sn} & \textbf{st} & \textbf{tn} & \textbf{ts} \\ \hline 
                \multicolumn{13}{c}{} \\ \hline
                \multicolumn{13}{|c|}{Machine Translation (en $\rightarrow$ $x$)}  \\ \hline
{GPT-4o} & {89.8} & {88.3} & {80.4} & {79.2} & {64.5} & {41.1} & {72.8} & {76.6} & {81.9} & {80.4} & {69.8} & {72.5} \\
{GPT-4} & {84.2} & {80.0} & {71.3} & {69.1} & {42.6} & {36.2} & {64.9} & {58.9} & {74.7} & {70.9} & {62.3} & {61.1} \\
{GPT-3.5} & {72.5} & {63.8} & {40.4} & {40.0} & {30.9} & {34.7} & {35.8} & {33.6} & {41.1} & {33.2} & {34.0} & {32.5} \\
{Llama 3 70B IT} & {82.3} & {69.4} & {41.5} & {42.3} & {27.5} & {32.1} & {44.5} & {34.7} & {44.2} & {38.1} & {36.2} & {37.0} \\
{Llama 3 8B IT} & {69.1} & {50.2} & {37.0} & {31.3} & {24.2} & {34.3} & {40.0} & {37.7} & {38.1} & {35.8} & {34.7} & {32.1} \\
{Phi 3 Mini 4K IT} & {70.6} & {41.1} & {30.6} & {30.9} & {24.9} & {32.1} & {37.7} & {32.1} & {33.2} & {27.9} & {30.6} & {27.9} \\
{Aya 23 35B} & {69.4} & {55.5} & {37.4} & {33.2} & {30.6} & {29.1} & {35.8} & {35.8} & {38.1} & {35.1} & {30.6} & {37.0} \\
{Aya 101} & {45.3} & {42.6} & {40.8} & {38.9} & {37.0} & {29.1} & {35.5} & {37.4} & {39.2} & {39.6} & {40.0} & {35.1} \\
{BLOOMZ 7b1} & {44.9} & {28.7} & {32.1} & {32.8} & {26.0} & {30.2} & {29.4} & {30.9} & {37.4} & {30.6} & {29.4} & {33.2} \\
\hline
                \multicolumn{13}{|c|}{Difference Between Human (Table \ref{table:perf-results}) and Machine Translation}\\ \hline
{GPT-4o} & {0.0} & {-1.1} & {-0.8} & {-0.7} & {6.1} & {-1.1} & {-9.8} & {-4.1} & {-1.5} & {-11.0} & {1.5} & {-8.0} \\
{GPT-4} & {0.0} & {1.9} & {-1.1} & {-0.8} & {11.4} & {-1.5} & {-10.2} & {-0.8} & {-7.5} & {-6.7} & {-3.4} & {-5.3} \\
{GPT-3.5} & {0.0} & {-1.2} & {-1.2} & {-2.6} & {-0.7} & {-3.4} & {-1.5} & {-0.4} & {0.0} & {1.1} & {-3.4} & {0.3} \\
{Llama 3 70B IT} & {0.0} & {1.9} & {-2.3} & {-3.8} & {6.1} & {0.4} & {-10.9} & {2.3} & {-0.8} & {5.7} & {3.0} & {-3.0} \\
{Llama 3 8B IT} & {0.0} & {-4.9} & {-0.4} & {3.4} & {0.7} & {-2.2} & {-3.8} & {-2.2} & {1.1} & {3.4} & {-1.5} & {-0.4} \\
{Phi 3 Mini 4K IT} & {0.0} & {-0.3} & {-1.5} & {-0.3} & {7.2} & {-4.9} & {-4.1} & {0.7} & {0.0} & {3.0} & {5.2} & {1.9} \\
{Aya 23 35B} & {0.0} & {-0.4} & {-4.9} & {0.4} & {-1.5} & {4.1} & {-3.3} & {1.6} & {-5.6} & {-0.8} & {0.3} & {-4.5} \\
{Aya 101} & {0.0} & {0.0} & {-2.7} & {-3.4} & {1.9} & {-0.8} & {-0.8} & {-0.8} & {5.0} & {-2.6} & {-9.4} & {-4.2} \\
{BLOOMZ 7b1} & {0.0} & {4.5} & {-1.5} & {0.4} & {2.3} & {2.3} & {-2.6} & {3.8} & {-6.8} & {1.5} & {-2.2} & {-1.9} \\
\hline
                \multicolumn{13}{c}{} \\ \hline
                \multicolumn{13}{|c|}{Machine Backtranslation ($x$ $\rightarrow$ en)} \\ \hline
{GPT-4o} & {89.8} & {87.5} & {78.9} & {77.7} & {86.4} & {66.4} & {58.9} & {83.4} & {75.8} & {65.7} & {67.2} & {64.9} \\
{GPT-4} & {84.2} & {81.5} & {72.5} & {68.3} & {76.6} & {59.6} & {50.9} & {73.6} & {68.3} & {63.0} & {61.9} & {60.4} \\
{GPT-3.5} & {72.5} & {69.1} & {60.4} & {61.9} & {63.8} & {55.1} & {49.8} & {67.5} & {61.5} & {55.5} & {55.5} & {56.6} \\
{Llama 3 70B IT} & {82.3} & {81.5} & {66.4} & {68.7} & {74.0} & {56.2} & {54.0} & {73.2} & {67.5} & {61.9} & {60.0} & {57.4} \\
{Llama 3 8B IT} & {69.1} & {65.7} & {58.1} & {60.0} & {61.1} & {51.3} & {43.8} & {59.6} & {55.1} & {55.8} & {54.0} & {50.9} \\
{Phi 3 Mini 4K IT} & {70.6} & {70.9} & {57.4} & {56.2} & {66.0} & {52.8} & {49.8} & {64.5} & {60.4} & {56.2} & {58.5} & {52.5} \\
{Aya 23 35B} & {69.4} & {70.6} & {58.9} & {59.6} & {67.2} & {52.5} & {47.9} & {64.5} & {57.4} & {54.0} & {51.7} & {52.5} \\
{Aya 101} & {45.3} & {43.8} & {43.0} & {38.9} & {41.9} & {35.5} & {39.6} & {42.6} & {42.6} & {38.9} & {38.1} & {36.2} \\
{BLOOMZ 7b1} & {44.9} & {49.1} & {42.6} & {42.3} & {43.4} & {38.9} & {40.8} & {45.7} & {41.9} & {40.4} & {38.1} & {40.8} \\ \hline
% {BLOOMZ 7b1}    & 79.0 & 37.0 & 35.8 & 35.1 & ? & ?  & ? & ? & ? & ? & ? & ? 
% \\ \hline \hline

                \multicolumn{13}{|c|}{Difference Between Human (Table \ref{table:perf-results}) and Machine Translation}\\ \hline
{GPT-4o} & {0.0} & {-0.3} & {0.7} & {0.8} & {-15.8} & {-26.4} & {4.1} & {-10.9} & {4.6} & {3.7} & {4.1} & {-0.4} \\
{GPT-4} & {0.0} & {0.4} & {-2.3} & {0.0} & {-22.6} & {-24.9} & {3.8} & {-15.5} & {-1.1} & {1.2} & {-3.0} & {-4.6} \\
{GPT-3.5} & {0.0} & {-6.5} & {-21.2} & {-24.5} & {-33.6} & {-23.8} & {-15.5} & {-34.3} & {-20.4} & {-21.2} & {-24.9} & {-23.8} \\
{Llama 3 70B IT} & {0.0} & {-10.2} & {-27.2} & {-30.2} & {-40.4} & {-23.7} & {-20.4} & {-36.2} & {-24.1} & {-18.1} & {-20.8} & {-23.4} \\
{Llama 3 8B IT} & {0.0} & {-20.4} & {-21.5} & {-25.3} & {-36.2} & {-19.2} & {-7.6} & {-24.1} & {-15.9} & {-16.6} & {-20.8} & {-19.2} \\
{Phi 3 Mini 4K IT} & {0.0} & {-30.1} & {-28.3} & {-25.6} & {-33.9} & {-25.6} & {-16.2} & {-31.7} & {-27.2} & {-25.3} & {-22.7} & {-22.7} \\
{Aya 23 35B} & {0.0} & {-15.5} & {-26.4} & {-26.0} & {-38.1} & {-19.3} & {-15.4} & {-27.1} & {-24.9} & {-19.7} & {-20.8} & {-20.0} \\
{Aya 101} & {0.0} & {-1.2} & {-4.9} & {-3.4} & {-3.0} & {-7.2} & {-4.9} & {-6.0} & {1.6} & {-1.9} & {-7.5} & {-5.3} \\
{BLOOMZ 7b1} & {0.0} & {-15.9} & {-12.0} & {-9.1} & {-15.1} & {-6.4} & {-14.0} & {-11.0} & {-11.3} & {-8.3} & {-10.9} & {-9.5} \\
\hline
\end{tabular}
\caption{\textbf{Results of State-of-the-Art Models on Machine-Translated MMLU Clinical Knowledge}. Top state-of-the-art models were evaluated out-of-the-box (no fine-tuning) on MMLU Clinical Knowledge (medical domain knowledge task) test datasets that were machine-translated from English into the target language or backtranslated from the human-translated version back into English. Google Translate was used for machine translations in all languages except Setswana (tn), where GPT-4o was used. Results are provided for 11 low-resource African languages of focus: Afrikaans (af), Zulu (zu), Xhosa (xh) (datasets for these three languages were sourced from \cite{winmmluza2024}), Amharic (am), Bambara (bm), Igbo (ig), Sepedi (nso), Shona (sn), Sesotho (st), Setswana (tn), and Tsonga (ts). Results on English (en) without any translation are also provided as a reference. Differences with corresponding numbers from Table \ref{table:perf-results} are given, with negative values indicating that machine translation led to greater performance than human-translation. All numbers are 5-shot performance accuracy. 
} \label{table:perf-results-mt-mmlu-ck}
\end{table*}

\begin{table*}[t!]
\centering
\setlength\extrarowheight{2pt}
% \small
\begin{tabular}{|l|llllllllllll|}
\hline \hline
                 \multicolumn{13}{c}{\texttt{MMLU Virology}}  \\ \hline
                & \textbf{en} & \textbf{af} & \textbf{zu} & \textbf{xh} & \textbf{am} & \textbf{bm} & \textbf{ig} & \textbf{nso} & \textbf{sn} & \textbf{st} & \textbf{tn} & \textbf{ts} \\ \hline 
                \multicolumn{13}{c}{} \\ \hline
                \multicolumn{13}{|c|}{Machine Translation (en $\rightarrow$ $x$)}  \\ \hline
{GPT-4o} & {60.2} & {57.8} & {54.2} & {51.8} & {49.4} & {28.9} & {48.2} & {50.0} & {51.2} & {54.8} & {54.2} & {51.2} \\
{GPT-4} & {59.6} & {57.2} & {53.6} & {53.0} & {37.3} & {32.5} & {50.0} & {45.8} & {50.6} & {48.8} & {50.0} & {46.4} \\
{GPT-3.5} & {51.8} & {44.6} & {31.3} & {31.9} & {31.9} & {35.5} & {26.5} & {34.3} & {34.3} & {31.9} & {31.3} & {35.5} \\
{Llama 3 70B IT} & {53.6} & {47.0} & {35.5} & {31.9} & {30.1} & {31.3} & {30.7} & {26.5} & {28.9} & {28.3} & {29.5} & {33.1} \\
{Llama 3 8B IT} & {51.8} & {42.8} & {24.1} & {26.5} & {27.7} & {29.5} & {28.9} & {27.7} & {24.7} & {25.3} & {28.9} & {26.5} \\
{Phi 3 Mini 4K IT} & {48.8} & {39.8} & {26.5} & {25.9} & {22.9} & {27.7} & {25.9} & {26.5} & {28.9} & {26.5} & {27.9} & {32.5} \\
{Aya 23 35B} & {49.4} & {42.2} & {29.5} & {26.5} & {27.7} & {23.5} & {29.5} & {27.1} & {32.5} & {26.5} & {27.1} & {24.7} \\
{Aya 101} & {33.1} & {38.6} & {39.2} & {30.7} & {30.1} & {25.3} & {31.3} & {28.3} & {36.1} & {31.3} & {34.3} & {29.5} \\
{BLOOMZ 7b1} & {38.0} & {27.7} & {28.9} & {24.7} & {25.9} & {22.9} & {27.7} & {27.1} & {30.1} & {27.1} & {24.1} & {27.1} \\
\hline
                \multicolumn{13}{|c|}{Difference Between Human (Table \ref{table:perf-results}) and Machine Translation}\\ \hline
{GPT-4o} & {0.0} & {-2.4} & {-3.0} & {-1.2} & {1.8} & {3.6} & {-4.2} & {0.0} & {1.8} & {-7.2} & {-5.4} & {-5.4} \\
{GPT-4} & {0.0} & {1.8} & {-4.2} & {-2.4} & {8.5} & {-0.6} & {-5.4} & {0.6} & {-1.2} & {-3.6} & {-6.0} & {-6.6} \\
{GPT-3.5} & {0.0} & {-1.8} & {-0.6} & {5.4} & {-2.4} & {0.6} & {1.8} & {-0.6} & {-3.0} & {-1.8} & {-1.2} & {-4.8} \\
{Llama 3 70B IT} & {0.0} & {-0.6} & {-6.6} & {-4.8} & {6.0} & {4.2} & {4.2} & {0.0} & {2.4} & {1.2} & {-1.2} & {-1.8} \\
{Llama 3 8B IT} & {0.0} & {-4.2} & {4.8} & {0.6} & {0.6} & {-9.0} & {-4.8} & {0.0} & {-2.4} & {0.0} & {-1.2} & {3.0} \\
{Phi 3 Mini 4K IT} & {0.0} & {-9.7} & {0.6} & {-0.6} & {4.2} & {-2.4} & {6.6} & {1.2} & {-0.6} & {-2.4} & {-2.6} & {0.0} \\
{Aya 23 35B} & {0.0} & {0.0} & {-4.8} & {0.0} & {11.5} & {4.2} & {-9.6} & {1.2} & {-4.2} & {0.0} & {-1.2} & {-1.2} \\
{Aya 101} & {0.0} & {-3.7} & {-5.5} & {0.6} & {6.0} & {-3.6} & {-1.2} & {7.2} & {-4.8} & {3.6} & {-1.2} & {0.0} \\
{BLOOMZ 7b1} & {0.0} & {-1.2} & {-2.4} & {0.0} & {-4.8} & {-1.8} & {-1.2} & {-3.0} & {1.2} & {0.6} & {3.6} & {-0.6} \\
\hline
                \multicolumn{13}{c}{} \\ \hline
                \multicolumn{13}{|c|}{Machine Backtranslation ($x$ $\rightarrow$ en)} \\ \hline
{GPT-4o} & {60.2} & {55.4} & {50.6} & {50.6} & {57.2} & {47.6} & {50.0} & {53.0} & {55.4} & {45.8} & {46.4} & {50.0} \\
{GPT-4} & {59.6} & {56.0} & {49.4} & {50.0} & {54.8} & {47.0} & {47.0} & {53.0} & {52.4} & {41.0} & {39.2} & {50.6} \\
{GPT-3.5} & {51.8} & {47.6} & {44.6} & {42.8} & {48.8} & {48.2} & {40.4} & {45.2} & {44.0} & {44.6} & {39.2} & {45.2} \\
{Llama 3 70B IT} & {53.6} & {54.2} & {46.4} & {43.4} & {55.4} & {44.0} & {41.6} & {46.4} & {48.2} & {41.0} & {39.2} & {47.0} \\
{Llama 3 8B IT} & {51.8} & {48.2} & {41.0} & {43.4} & {47.0} & {44.0} & {39.2} & {44.0} & {46.4} & {37.3} & {41.0} & {41.0} \\
{Phi 3 Mini 4K IT} & {48.8} & {47.0} & {42.2} & {42.2} & {50.0} & {40.4} & {39.8} & {39.8} & {38.6} & {39.2} & {41.0} & {43.4} \\
{Aya 23 35B} & {49.4} & {49.4} & {43.4} & {44.0} & {50.0} & {45.2} & {41.6} & {45.2} & {43.4} & {36.7} & {42.8} & {44.6} \\
{Aya 101} & {33.1} & {36.1} & {37.3} & {35.5} & {30.7} & {35.5} & {34.3} & {29.5} & {30.7} & {34.9} & {35.5} & {32.5} \\
{BLOOMZ 7b1} & {38.0} & {41.0} & {38.0} & {33.7} & {33.1} & {30.1} & {27.7} & {34.9} & {33.1} & {28.9} & {31.9} & {28.3} \\ \hline
% {BLOOMZ 7b1}    & 79.0 & 37.0 & 35.8 & 35.1 & ? & ?  & ? & ? & ? & ? & ? & ? 
% \\ \hline \hline

                \multicolumn{13}{|c|}{Difference Between Human (Table \ref{table:perf-results}) and Machine Translation}\\ \hline
{GPT-4o} & {0.0} & {0.0} & {0.6} & {0.0} & {-6.0} & {-15.1} & {-6.0} & {-3.0} & {-2.4} & {1.8} & {2.4} & {-4.2} \\
{GPT-4} & {0.0} & {3.0} & {0.0} & {0.6} & {-9.0} & {-15.1} & {-2.4} & {-6.6} & {-3.0} & {4.2} & {4.8} & {-10.8} \\
{GPT-3.5} & {0.0} & {-4.8} & {-13.9} & {-5.5} & {-19.3} & {-12.1} & {-12.1} & {-11.5} & {-12.7} & {-14.5} & {-9.1} & {-14.5} \\
{Llama 3 70B IT} & {0.0} & {-7.8} & {-17.5} & {-16.3} & {-19.3} & {-8.5} & {-6.7} & {-19.9} & {-16.9} & {-11.5} & {-10.9} & {-15.7} \\
{Llama 3 8B IT} & {0.0} & {-9.6} & {-12.1} & {-16.3} & {-18.7} & {-23.5} & {-15.1} & {-16.3} & {-24.1} & {-12.0} & {-13.3} & {-11.5} \\
{Phi 3 Mini 4K IT} & {0.0} & {-16.9} & {-15.1} & {-16.9} & {-22.9} & {-15.1} & {-7.3} & {-12.1} & {-10.3} & {-15.1} & {-15.7} & {-10.9} \\
{Aya 23 35B} & {0.0} & {-7.2} & {-18.7} & {-17.5} & {-10.8} & {-17.5} & {-21.7} & {-16.9} & {-15.1} & {-10.2} & {-16.9} & {-21.1} \\
{Aya 101} & {0.0} & {-1.2} & {-3.6} & {-4.2} & {5.4} & {-13.8} & {-4.2} & {6.0} & {0.6} & {0.0} & {-2.4} & {-3.0} \\
{BLOOMZ 7b1} & {0.0} & {-14.5} & {-11.5} & {-9.0} & {-12.0} & {-9.0} & {-1.2} & {-10.8} & {-1.8} & {-1.2} & {-4.2} & {-1.8} \\
\hline
\end{tabular}
\caption{\textbf{Results of State-of-the-Art Models on Machine-Translated MMLU Virology}. Top state-of-the-art models were evaluated out-of-the-box (no fine-tuning) on MMLU Virology (medical domain knowledge task) test datasets that were machine-translated from English into the target language or backtranslated from the human-translated version back into English. Google Translate was used for machine translations in all languages except Setswana (tn), where GPT-4o was used. Results are provided for 11 low-resource African languages of focus: Afrikaans (af), Zulu (zu), Xhosa (xh) (datasets for these three languages were sourced from \cite{winmmluza2024}), Amharic (am), Bambara (bm), Igbo (ig), Sepedi (nso), Shona (sn), Sesotho (st), Setswana (tn), and Tsonga (ts). Results on English (en) without any translation are also provided as a reference. Differences with corresponding numbers from Table \ref{table:perf-results} are given, with negative values indicating that machine translation led to greater performance than human-translation. All numbers are 5-shot performance accuracy. 
} \label{table:perf-results-mt-mmlu-vir}
\end{table*}

\begin{table*}[t!]
\centering
\setlength\extrarowheight{2pt}
% \small
\begin{tabular}{|l|llllllllllll|}
\hline \hline
                 \multicolumn{13}{c}{\texttt{Belebele}}  \\ \hline
                & \textbf{en} & \textbf{af} & \textbf{zu} & \textbf{xh} & \textbf{am} & \textbf{bm} & \textbf{ig} & \textbf{nso} & \textbf{sn} & \textbf{st} & \textbf{tn} & \textbf{ts} \\ \hline 
                \multicolumn{13}{c}{} \\ \hline
                \multicolumn{13}{|c|}{Machine Translation (en $\rightarrow$ $x$)}  \\ \hline
{GPT-4o} & {95.9} & {94.8} & {86.0} & {86.8} & {76.8} & {41.4} & {81.4} & {86.3} & {85.7} & {88.0} & {80.4} & {81.4} \\
{GPT-4} & {96.1} & {93.4} & {82.2} & {82.8} & {62.7} & {37.0} & {74.2} & {72.7} & {82.0} & {81.8} & {75.6} & {73.6} \\
{GPT-3.5} & {86.2} & {75.4} & {35.1} & {32.9} & {32.4} & {30.4} & {32.0} & {35.8} & {32.1} & {33.3} & {33.0} & {34.2} \\
{Llama 3 70B IT} & {94.6} & {85.4} & {42.0} & {38.7} & {33.9} & {35.0} & {45.3} & {39.1} & {37.7} & {37.7} & {38.8} & {41.0} \\
{Llama 3 8B IT} & {80.0} & {69.6} & {36.3} & {36.1} & {33.0} & {34.1} & {40.1} & {33.0} & {31.4} & {31.1} & {33.7} & {33.3} \\
{Phi 3 Mini 4K IT} & {89.2} & {52.6} & {27.6} & {28.0} & {27.1} & {26.9} & {28.2} & {30.4} & {28.3} & {31.0} & {28.0} & {31.0} \\
{Aya 23 35B} & {93.6} & {84.8} & {39.3} & {37.6} & {28.9} & {31.4} & {38.9} & {35.7} & {36.3} & {37.4} & {36.2} & {36.4} \\
{Aya 101} & {79.7} & {76.9} & {62.2} & {62.9} & {66.7} & {47.1} & {61.1} & {66.1} & {63.3} & {64.1} & {63.9} & {56.1} \\
{BLOOMZ 7b1} & {79.0} & {35.4} & {36.6} & {36.7} & {24.1} & {28.7} & {37.2} & {38.3} & {34.3} & {33.1} & {37.6} & {38.8} \\
\hline
                \multicolumn{13}{|c|}{Difference Between Human (Table \ref{table:perf-results}) and Machine Translation}\\ \hline
{GPT-4o} & {0.0} & {-0.4} & {-6.3} & {-4.0} & {1.2} & {-3.1} & {-10.1} & {-9.1} & {-4.7} & {-8.0} & {-3.4} & {-4.6} \\
{GPT-4} & {0.0} & {0.2} & {-6.3} & {-6.5} & {-1.4} & {0.9} & {-9.8} & {-8.1} & {-4.4} & {-5.5} & {-8.0} & {-3.6} \\
{GPT-3.5} & {0.0} & {0.3} & {-1.3} & {-1.1} & {-2.4} & {0.2} & {-2.8} & {-3.5} & {2.5} & {-1.2} & {-1.3} & {2.1} \\
{Llama 3 70B IT} & {0.0} & {-0.6} & {-5.3} & {-1.9} & {1.2} & {0.6} & {-9.4} & {-2.0} & {0.4} & {-1.8} & {-2.4} & {-0.3} \\
{Llama 3 8B IT} & {0.0} & {-0.2} & {-3.7} & {-3.1} & {-2.8} & {-2.1} & {-4.4} & {-0.1} & {2.5} & {1.1} & {0.3} & {0.1} \\
{Phi 3 Mini 4K IT} & {0.0} & {0.0} & {0.7} & {1.1} & {1.6} & {4.8} & {0.6} & {1.2} & {0.8} & {-0.4} & {1.2} & {3.3} \\
{Aya 23 35B} & {0.0} & {-1.2} & {-6.1} & {-2.0} & {0.1} & {3.2} & {-10.2} & {-1.1} & {1.5} & {-1.8} & {0.6} & {1.0} \\
{Aya 101} & {0.0} & {0.5} & {-2.9} & {-2.2} & {1.0} & {-6.4} & {-10.4} & {-6.8} & {-5.5} & {-4.2} & {-6.1} & {-5.1} \\
{BLOOMZ 7b1} & {0.0} & {1.3} & {-0.8} & {-1.7} & {0.2} & {2.9} & {-6.0} & {-4.2} & {2.4} & {1.6} & {-1.7} & {0.9} \\
\hline
                \multicolumn{13}{c}{} \\ \hline
                \multicolumn{13}{|c|}{Machine Backtranslation ($x$ $\rightarrow$ en)} \\ \hline
{GPT-4o} & {95.9} & {92.0} & {71.0} & {77.0} & {81.2} & {49.1} & {62.8} & {77.9} & {75.4} & {76.8} & {67.7} & {75.0} \\
{GPT-4} & {96.1} & {92.0} & {69.2} & {76.4} & {81.1} & {48.1} & {62.2} & {77.0} & {73.2} & {75.8} & {66.3} & {72.7} \\
{GPT-3.5} & {86.2} & {80.4} & {62.1} & {66.0} & {69.1} & {41.8} & {52.7} & {66.9} & {63.2} & {64.8} & {60.4} & {64.1} \\
{Llama 3 70B IT} & {94.6} & {89.8} & {65.7} & {72.7} & {77.2} & {40.0} & {53.4} & {71.6} & {68.3} & {69.8} & {61.8} & {68.4} \\
{Llama 3 8B IT} & {80.0} & {76.0} & {54.3} & {58.3} & {62.9} & {38.8} & {48.2} & {60.0} & {56.6} & {58.7} & {50.7} & {57.4} \\
{Phi 3 Mini 4K IT} & {89.2} & {84.0} & {62.2} & {64.7} & {72.2} & {43.3} & {50.3} & {68.8} & {62.0} & {65.3} & {58.2} & {62.9} \\
{Aya 23 35B} & {93.6} & {89.8} & {66.6} & {72.1} & {76.6} & {43.9} & {52.2} & {74.2} & {68.3} & {71.8} & {63.4} & {69.1} \\
{Aya 101} & {79.7} & {76.7} & {57.7} & {61.4} & {67.9} & {39.6} & {48.9} & {62.1} & {59.2} & {59.1} & {57.7} & {60.1} \\
{BLOOMZ 7b1} & {79.0} & {74.6} & {56.9} & {58.1} & {61.9} & {37.4} & {45.7} & {58.3} & {55.4} & {57.0} & {56.3} & {58.1} \\ \hline
% {BLOOMZ 7b1}    & 79.0 & 37.0 & 35.8 & 35.1 & ? & ?  & ? & ? & ? & ? & ? & ? 
% \\ \hline \hline

                \multicolumn{13}{|c|}{Difference Between Human (Table \ref{table:perf-results}) and Machine Translation}\\ \hline
{GPT-4o} & {0.0} & {2.4} & {8.7} & {5.8} & {-3.2} & {-10.8} & {8.5} & {-0.7} & {5.6} & {3.2} & {9.3} & {1.8} \\
{GPT-4} & {0.0} & {1.6} & {6.7} & {-0.1} & {-19.8} & {-10.2} & {2.2} & {-12.4} & {4.4} & {0.5} & {1.3} & {-2.7} \\
{GPT-3.5} & {0.0} & {-4.7} & {-28.3} & {-34.2} & {-39.1} & {-11.2} & {-23.5} & {-34.6} & {-28.6} & {-32.7} & {-28.7} & {-27.8} \\
{Llama 3 70B IT} & {0.0} & {-5.0} & {-29.0} & {-35.9} & {-42.1} & {-4.4} & {-17.5} & {-34.5} & {-30.2} & {-33.9} & {-25.4} & {-27.7} \\
{Llama 3 8B IT} & {0.0} & {-6.6} & {-21.7} & {-25.3} & {-32.7} & {-6.8} & {-12.5} & {-27.1} & {-22.7} & {-26.5} & {-16.7} & {-24.0} \\
{Phi 3 Mini 4K IT} & {0.0} & {-31.4} & {-33.9} & {-35.6} & {-43.5} & {-11.6} & {-21.5} & {-37.2} & {-32.9} & {-34.7} & {-29.0} & {-28.6} \\
{Aya 23 35B} & {0.0} & {-6.2} & {-33.4} & {-36.5} & {-47.6} & {-9.3} & {-23.5} & {-39.6} & {-30.5} & {-36.2} & {-26.6} & {-31.7} \\
{Aya 101} & {0.0} & {0.7} & {1.6} & {-0.7} & {-0.2} & {1.1} & {1.8} & {-2.8} & {-1.4} & {0.8} & {0.1} & {-9.1} \\
{BLOOMZ 7b1} & {0.0} & {-37.9} & {-21.1} & {-23.1} & {-37.6} & {-5.8} & {-14.5} & {-24.2} & {-18.7} & {-22.3} & {-20.4} & {-18.4} \\
\hline
\end{tabular}
\caption{\textbf{Results of State-of-the-Art Models on Machine-Translated Belebele}. Top state-of-the-art models were evaluated out-of-the-box (no fine-tuning) on Belebele (reading comprehension task) datasets that were machine-translated from English into the target language or backtranslated from the human-translated version back into English. Google Translate was used for machine translations in all languages except Setswana (tn), where GPT-4o was used. Results are provided for 11 low-resource African languages of focus: Afrikaans (af), Zulu (zu), Xhosa (xh) (datasets for these three languages were sourced from \cite{winmmluza2024}), Amharic (am), Bambara (bm), Igbo (ig), Sepedi (nso), Shona (sn), Sesotho (st), Setswana (tn), and Tsonga (ts). Results on English (en) without any translation are also provided as a reference. Differences with corresponding numbers from Table \ref{table:perf-results} are given, with negative values indicating that machine translation led to greater performance than human-translation. All numbers are 0-shot performance accuracy. 
} \label{table:perf-results-mt-bele}
\end{table*}

\begin{table*}[h]
\centering
\setlength\extrarowheight{4pt}
\begin{tabular}{|c|cc|cc|}
    \hline
    & \multicolumn{2}{c|}{\begin{tabular}{@{}c@{}}Question 1 \\ \textbf{(quality)}\end{tabular}}
    & \multicolumn{2}{c|}{\begin{tabular}{@{}c@{}}Question 2 \\ \textbf{(appropriateness)}\end{tabular}} \\ 
    \hline
    Lang. & Cohen's $\kappa$ & Fleiss' $\kappa$ & Cohen's $\kappa$ & Fleiss' $\kappa$ \\
    \hline
af  & 0.136  & 0.134  & 0.008   & -0.009  \\
zu  & -0.023 & -0.025 & -0.037  & -0.038  \\
xh  & 0.158  & 0.150   &  0.027  & -0.042  \\
am  & 0.008  & -0.085 & -0.002  & -0.111  \\
bm  & 0.060   & 0.014  & 0.074   & 0.051   \\
ig  & 0.013  & -0.214 & 0.011   & -0.073  \\
nso & 0.157  & 0.154  & 0.041   & 0.039   \\
sn  & 0.211  & 0.179  & 0.029   & -0.001  \\
st  & 0.012  & -0.076 & 0.006   & -0.070   \\
tn  & 0.016  & -0.046 & 0.002   & -0.075  \\
ts  & 0.133  & 0.093  & 0.045   & 0.023   \\ 
\hline
\end{tabular}
\caption{\textbf{Inter-rater Reliability between Evaluators.} The above table shows Cohen's Kappa ($\kappa$) and Fleiss' Kappa ($\kappa$) scores for each question asked of the human \textit{evaluators} (translation quality and appropriateness). The translation quality question asked the \textit{evaluators} to rate the quality of the translation as ``Good translation", ``Incorrect, but someone could understand the idea", or ``Completely wrong". The translation appropriateness question asked \textit{evaluators} to determine whether the translation could be considered ``strange, incoherent, or disrespectful" in a typical conversational context, with possible options of ``No, the sentence is typical", ``Maybe, I'm not sure", ``Yes, the sentence is strange, incoherent, or disrespectful", or ``I don't understand the sentence". We observe that there was little agreement on appropriateness even when agreement was present for translation quality, indicating that different \textit{evaluators} may be sensitive to different aspects of appropriateness. See Tables \ref{table:annotator_confusion_2} and \ref{table:annotator_confusion_1} for the full agreement matrices for each language. Language codes are as follows: Afrikaans (af), Zulu (zu), Xhosa (xh), Amharic (am), Bambara (bm), Igbo (ig), Sepedi (nso), Shona (sn), Sesotho (st),  Setswana (tn), Tsonga (ts).}
\label{table:appropriateness-interrater}
\end{table*}

\begin{table*}[]
\centering
\setlength\extrarowheight{4pt}
\begin{tabular}{|lccccccccccc|}
\hline
\multicolumn{1}{|l|}{\textbf{Language}}& \textbf{xh}  & \textbf{zu}   & \textbf{af} & \textbf{ig}   & \textbf{sn}  & \textbf{ts} & \textbf{st} & \textbf{nso}  & \textbf{tn} & \textbf{bm} & \textbf{am} \\ \hline
\multicolumn{12}{|c|}{A. Target Language Performance on ``appropriate" subset}                                                                                         \\ \hline
\multicolumn{1}{|l|}{Sample Size}      & 1379   & 1355   & 1086      & 809    & 1548   & 1422   & 1461    & 1711    & 1433     & 1019    & 1370    \\ \hline
\multicolumn{1}{|l|}{\textbf{\textit{GPT 4o}}}  & \textbf{65.5\%} & \textbf{69.1\%} & \textbf{81.7\%}    & \textbf{60.7\%} & \textbf{71.7\%} & \textbf{63.5\%} & \textbf{68.2\%}  & \textbf{64.7\%}  & \textbf{64.8\%}   & 51.4\%  & \textbf{60.0\%}  \\
\multicolumn{1}{|l|}{\textit{GPT 4}}   & 61.4\% & 65.3\% & 80.0\%    & \textbf{60.7\%} & 67.4\% & 58.7\% & 65.2\%  & 58.9\%  & 60.1\%   & \textbf{52.2\%}  & 51.7\%  \\
\multicolumn{1}{|l|}{\textit{GPT 3.5}} & 51.6\% & 49.7\% & 56.7\%    & 51.9\% & 51.1\% & 50.4\% & 49.9\%  & 50.8\%  & 50.7\%   & 49.6\%  & 50.4\%  \\ \hline
\multicolumn{12}{|c|}{B. Target Language Performance on ``inappropriate" subset}                                                                                     \\ \hline
\multicolumn{1}{|l|}{Sample Size}      & 388    & 412    & 681       & 958    & 219    & 345    & 306     & 56      & 334      & 748     & 397     \\ \hline
\multicolumn{1}{|l|}{\textbf{\textit{GPT 4o}}}  & \textbf{65.7\%} & \textbf{64.3\%} & \textbf{74.9\%}    & \textbf{61.7\%} & \textbf{54.5\%} & \textbf{60.2\%} & \textbf{64.5\%}  & 56.0\%  & \textbf{62.7\%}   & 49.4\%  & \textbf{58.0\%}  \\
\multicolumn{1}{|l|}{\textit{GPT 4}}   & 60.7\% & 59.2\% & 72.3\%    & 57.5\% & 53.6\% & 53.3\% & 58.7\%  & \textbf{59.5\%}  & 62.1\%   & 49.9\%  & 49.3\%  \\
\multicolumn{1}{|l|}{\textit{GPT 3.5}} & 53.5\% & 50.6\% & 51.9\%    & 50.0\% & 53.9\% & 51.3\% & 52.7\%  & 56.5\%  & 53.9\%   & \textbf{51.4\%}  & 51.4\%  \\ \hline
\multicolumn{12}{|c|}{C. English Performance on ``appropriate" subset}                                                                                 \\ \hline
\multicolumn{1}{|l|}{Sample Size}      & 1379   & 1355   & 1086      & 809    & 1548   & 1422   & 1461    & 1711    & 1433     & 1019    & 1370    \\ \hline
\multicolumn{1}{|l|}{\textbf{\textit{GPT 4o}}}  & \textbf{83.9\%} & \textbf{85.1\%} & \textbf{85.3\%}    & \textbf{84.9\%} & \textbf{84.3\%} & \textbf{84.2\%} & \textbf{84.0\%}  & \textbf{83.6\%}  & \textbf{84.6\%}   & \textbf{83.4\%}  & \textbf{83.2\%}  \\
\multicolumn{1}{|l|}{\textit{GPT 4}}   & 83.1\% & 83.1\% & 84.8\%    & 83.0\% & 83.3\% & 83.8\% & 83.5\%  & 82.9\%  & 83.2\%   & 82.1\%  & 82.5\%  \\
\multicolumn{1}{|l|}{\textit{GPT 3.5}} & 58.7\% & 58.5\% & 59.9\%    & 61.0\% & 59.0\% & 59.0\% & 59.4\%  & 58.8\%  & 58.6\%   & 57.0\%  & 58.7\%  \\ \hline
\multicolumn{12}{|c|}{D. English Performance on ``inappropriate" subset}                                                                             \\ \hline
\multicolumn{1}{|l|}{Sample Size}      & 388    & 412    & 681       & 958    & 219    & 345    & 306     & 56      & 334      & 748     & 397     \\ \hline
\multicolumn{1}{|l|}{\textbf{\textit{GPT 4o}}}  & \textbf{84.2\%} & 80.0\% & \textbf{81.8\%}    & \textbf{83.2\%} & 81.3\% & \textbf{82.7\%} & \textbf{83.9\%}  & \textbf{94.0\%}  & 81.3\%   & \textbf{84.7\%}  & \textbf{86.6\%}  \\
\multicolumn{1}{|l|}{\textit{GPT 4}}   & 82.9\% & \textbf{82.8\%} & 80.2\%    & 83.1\% & \textbf{81.4\%} & 80.1\% & 80.7\%  & 86.3\%  & \textbf{82.4\%}   & 84.3\%  & 85.1\%  \\
\multicolumn{1}{|l|}{\textit{GPT 3.5}} & 59.0\% & 59.7\% & 57.0\%    & 56.9\% & 57.4\% & 57.9\% & 55.9\%  & 59.5\%  & 59.7\%   & 61.1\%  & 59.2\%  \\ \hline
\multicolumn{12}{|c|}{E. Difference in Target Language Performance on ``appropriate" subset vs. ``inappropriate" subset (A.-B.)}                                                               \\ \hline
\multicolumn{1}{|l|}{\textbf{\textit{GPT 4o}}}  & -0.2\% & 4.8\%  & 6.8\%    & -1.0\% & \textbf{17.2\%} & 3.3\%  & 3.7\%   & \textbf{8.8\%}   & \textbf{2.2\%}    & 2.0\%   & 2.0\%   \\
\multicolumn{1}{|l|}{\textit{GPT 4}}   & \textbf{0.8\%}  & \textbf{6.1\%}  & \textbf{7.8\%}     & \textbf{3.2\%}  & 13.8\% & \textbf{5.4\%}  & \textbf{6.5\%}   & -0.6\%  & -2.0\%   & \textbf{2.4\%}   & \textbf{2.4\%}   \\
\multicolumn{1}{|l|}{GPT 3.5}          & -1.9\% & -1.0\% & 4.7\%     & 1.8\%  & -2.7\% & -0.9\% & -2.8\%  & -5.8\%  & -3.2\%   & -1.8\%  & -1.0\%  \\ \hline
\multicolumn{12}{|c|}{F. Difference in English Performance on ``appropriate" subset vs. ``inappropriate" subset (C.-D.)}                                                                       \\ \hline
\multicolumn{1}{|l|}{\textit{GPT 4o}}  & -0.3\% & \textbf{5.1\%}  & 3.5\%     & 1.7\%  & \textbf{3.0\%}  & 1.5\%  & 0.1\%   & -10.4\% & \textbf{3.2\%}    & \textbf{-1.3\%}  & -3.5\%  \\
\multicolumn{1}{|l|}{\textbf{\textit{GPT 4}}}   & \textbf{0.2\%}  & 0.3\%  & \textbf{4.7\%}     & 0.0\%  & 1.8\%  & \textbf{3.7\%}  & 2.8\%   & -3.4\%  & 0.7\%    & -2.2\%  & -2.6\%  \\
\multicolumn{1}{|l|}{GPT 3.5}          & -0.3\% & -1.2\% & 2.9\%     & \textbf{4.1\%}  & 1.6\%  & 1.1\%  & \textbf{3.5\%}   & \textbf{-0.8\%}  & -1.1\%   & -4.1\%  & \textbf{-0.5\%}  \\ \hline
\multicolumn{12}{|c|}{G. Comparative Diff. Between Target Lang. and Eng. Performance on ``appropriate" vs. ``inappropriate" subsets (E.-F.)}                                        \\ \hline
\multicolumn{1}{|l|}{\textbf{\textit{GPT 4o}}}  & -0.2\% & -0.4\% & \textbf{3.3\%}     & -2.7\% & \textbf{14.2\%} & 1.7\%  & 3.6\%   & \textbf{19.2\%}  & \textbf{-1.1\%}   & 3.2\%   & \textbf{5.4\%}   \\
\multicolumn{1}{|l|}{\textit{GPT 4}}   & \textbf{0.6\%}  & \textbf{5.9\%}  & 3.1\%     & \textbf{3.2\%}  & 12.0\% & \textbf{1.8\%}  & \textbf{3.7\%}   & 2.7\%   & -2.7\%   & \textbf{4.6\%}   & 5.0\%   \\
\multicolumn{1}{|l|}{GPT 3.5}          & -1.6\% & 0.3\%  & 1.9\%     & -2.3\% & -4.3\% & -2.1\% & -6.3\%  & -5.0\%  & -2.1\%   & 2.3\%   & -0.5\% \\ \hline
\end{tabular}
\caption{\textbf{GPT-Family Performance on Winogrande Split by Cultural Appropriateness (All QA Pairs).} The table above shows the performance of GPT-3.5, GPT-4, and GPT-4o on Winogrande when split by human annotations of cultural appropriateness for each language. \textbf{Bold} values indicate the highest-performing values or model (overall across all languages, determined by the largest sum of performance scores between the models) for the given subtable. \textit{Evaluators} were asked if each Winogrande QA pair could be considered ``strange, incoherent, or disrespectful in a typical conversational context", with the following possible options: (i) ``No, the sentence is typical", (ii) ``Not sure", (iii) ``Yes, the sentence is strange, incoherent, or disrespectful", or (iv) ``I don't understand the sentence". A dataset QA pair is considered ``inappropriate" if at least one \textit{evaluator} marked it as not ``typical".  Language codes are as follows: Xhosa (xh), Zulu (zu), Afrikaans (af), Igbo (ig), Shona (sn), Tsonga (ts), Sesotho (st), Sepedi (nso), Setswana (tn), Bambara (bm), Amharic (am).}
\label{table:appropriateness-1}
\end{table*}

\begin{table*}[]
\centering
\setlength\extrarowheight{4pt}
\begin{tabular}{|lccccccccccc|}
\hline
\multicolumn{1}{|l|}{\textbf{Language}}& \textbf{xh}  & \textbf{zu}   & \textbf{af} & \textbf{ig}   & \textbf{sn}  & \textbf{ts} & \textbf{st} & \textbf{nso}  & \textbf{tn} & \textbf{bm} & \textbf{am} \\ \hline
\multicolumn{12}{|c|}{A. Target Language Performance on ``appropriate" subset}                                                                                 \\ \hline
\multicolumn{1}{|l|}{Sample Size}      & 1379   & 1355   & 1086   & 809    & 1548   & 1422   & 1461   & 1710    & 1433   & 1019   & 1370   \\ \hline
\multicolumn{1}{|l|}{\textbf{\textit{GPT 4o}}}  & \textbf{65.5\%} & \textbf{69.1\%} & \textbf{81.7\%} & \textbf{60.7\%} & \textbf{71.7\%} & \textbf{63.5\%} & \textbf{68.2\%} & \textbf{64.8\%}  & \textbf{64.8\%} & 51.4\% & \textbf{60.0\%} \\
\multicolumn{1}{|l|}{\textit{GPT 4}}   & 61.4\% & 65.3\% & 80.0\% & \textbf{60.7\%} & 67.4\% & 58.7\% & 65.2\% & 58.9\%  & 60.1\% & \textbf{52.2\%} & 51.7\% \\
\multicolumn{1}{|l|}{\textit{GPT 3.5}} & 51.6\% & 49.7\% & 56.7\% & 51.9\% & 51.1\% & 50.4\% & 49.9\% & 50.8\%  & 50.7\% & 49.6\% & 50.4\% \\ \hline
\multicolumn{12}{|c|}{B. Target Language Performance on ``inappropriate" subset}                                                                             \\ \hline
\multicolumn{1}{|l|}{Sample Size}      & 385    & 409    & 676    & 956    & 207    & 339    & 306    & 56      & 331    & 737    & 397    \\ \hline
\multicolumn{1}{|l|}{\textbf{\textit{GPT 4o}}}  & \textbf{65.7\%} & \textbf{64.1\%} & \textbf{75.2\%} & \textbf{61.7\%} & \textbf{55.4\%} & \textbf{60.6\%} & \textbf{64.5\%} & 56.0\%  & \textbf{62.8\%} & 49.4\% & \textbf{58.0\%} \\
\multicolumn{1}{|l|}{\textit{GPT 4}}   & 60.9\% & 59.4\% & 72.5\% & 57.4\% & 54.4\% & 53.6\% & 58.7\% & \textbf{59.5\%}  & 62.1\% & 50.0\% & 49.3\% \\
\multicolumn{1}{|l|}{\textit{GPT 3.5}} & 53.9\% & 50.7\% & 52.0\% & 50.0\% & 53.9\% & 51.0\% & 52.7\% & 56.5\%  & 53.7\% & \textbf{51.5\%} & 51.4\% \\ \hline
\multicolumn{12}{|c|}{C. English Performance on ``appropriate" subset}                                                                         \\ \hline
\multicolumn{1}{|l|}{Sample Size}      & 1379   & 1355   & 1086   & 809    & 1548   & 1422   & 1461   & 1710    & 1433   & 1019   & 1370   \\ \hline
\multicolumn{1}{|l|}{\textbf{\textit{GPT 4o}}}  & \textbf{83.9\%} & \textbf{85.1\%} & \textbf{85.3\%} & \textbf{84.9\%} & \textbf{84.3\%} & \textbf{84.2\%} & \textbf{84.0\%} & \textbf{83.7\%}  & \textbf{84.6\%} & \textbf{83.4\%} & \textbf{83.2\%} \\
\multicolumn{1}{|l|}{\textit{GPT 4}}   & 83.1\% & 83.1\% & 84.8\% & 83.0\% & 83.3\% & 83.8\% & 83.5\% & 83.0\%  & 83.2\% & 82.1\% & 82.5\% \\
\multicolumn{1}{|l|}{\textit{GPT 3.5}} & 58.7\% & 58.5\% & 59.9\% & 61.0\% & 59.0\% & 59.0\% & 59.4\% & 58.8\%  & 58.6\% & 57.0\% & 58.7\% \\ \hline
\multicolumn{12}{|c|}{D. English Performance on ``inappropriate" subset}                                                                     \\ \hline
\multicolumn{1}{|l|}{Sample Size}      & 385    & 409    & 676    & 956    & 207    & 339    & 306    & 56      & 331    & 737    & 397    \\ \hline
\multicolumn{1}{|l|}{\textbf{\textit{GPT 4o}}}  & \textbf{84.1\%} & 79.9\% & \textbf{81.8\%} & \textbf{83.2\%} & 81.2\% & \textbf{82.5\%} & \textbf{83.9\%} & \textbf{94.0\%}  & 81.2\% & \textbf{84.5\%} & \textbf{86.6\%} \\
\multicolumn{1}{|l|}{\textit{GPT 4}}   & 83.0\% & \textbf{82.7\%} & 80.0\% & 83.1\% & \textbf{81.6\%} & 79.7\% & 80.7\% & 86.3\%  & \textbf{82.3\%} & 84.3\% & 85.1\% \\
\multicolumn{1}{|l|}{\textit{GPT 3.5}} & 59.2\% & 59.7\% & 57.1\% & 56.9\% & 57.3\% & 57.7\% & 55.9\% & 59.5\%  & 59.4\% & 61.0\% & 59.2\% \\ \hline
\multicolumn{12}{|c|}{E. Difference in Target Language Performance on ``appropriate" subset vs. ``inappropriate" subset (A.-B.)}                                                       \\ \hline
\multicolumn{1}{|l|}{\textbf{\textit{GPT 4o}}}  & -0.2\% & 4.9\%  & 6.5\%  & -1.0\% & \textbf{16.3\%} & 2.9\%  & 3.7\%  & \textbf{8.8\%}  & \textbf{2.0\%}  & 2.0\%  & 2.0\%  \\
\multicolumn{1}{|l|}{\textit{GPT 4}}   & \textbf{0.6\%}  & \textbf{5.9\%}  & \textbf{7.6\%}  & \textbf{3.3\%}  & 12.9\% & \textbf{5.2\%}  & \textbf{6.5\%}  & -0.6\%  & -2.1\% & \textbf{2.2\%}  & \textbf{2.4\%}  \\
\multicolumn{1}{|l|}{GPT 3.5}          & -2.2\% & -1.0\% & 4.6\%  & 1.8\%  & -2.8\% & -0.7\% & -2.8\% & -5.8\%  & -3.0\% & -1.9\% & -1.0\% \\ \hline
\multicolumn{12}{|c|}{F. Difference in English Performance on ``appropriate" subset vs. ``inappropriate" subset (C.-D.)}                                                               \\ \hline
\multicolumn{1}{|l|}{\textit{GPT 4o}}  & -0.2\% & \textbf{5.3\%}  & 3.5\%  & 1.6\%  & \textbf{3.2\%}  & 1.7\%  & 0.1\%  & -10.4\% & \textbf{3.4\%}  & \textbf{-1.1\%} & -3.5\% \\
\multicolumn{1}{|l|}{\textbf{\textit{GPT 4}}}   & \textbf{0.0\%}  & 0.4\%  & \textbf{4.8\%}  & -0.1\% & 1.6\%  & \textbf{4.0\%}  & 2.8\%  & -3.3\%  & 0.9\%  & -2.2\% & -2.6\% \\
\multicolumn{1}{|l|}{GPT 3.5}          & -0.5\% & -1.2\% & 2.7\%  & \textbf{4.1\%}  & 1.7\%  & 1.3\%  & \textbf{3.5\%}  & \textbf{-0.7\%}  & -0.8\% & -4.0\% & \textbf{-0.5\%} \\ \hline
\multicolumn{12}{|c|}{G. Comparative Diff. Between Target Lang. and Eng. Performance on ``appropriate" vs. ``inappropriate" subsets (E.-F.)}                                \\ \hline
\multicolumn{1}{|l|}{\textbf{\textit{GPT 4o}}}  & 0.0\%  & -0.3\% & \textbf{3.0\%}  & -2.6\% & \textbf{13.2\%} & \textbf{1.2\%}  & 3.6\%  & \textbf{19.2\%}  & \textbf{-1.4\%} & 3.1\%  & \textbf{5.4\%}  \\
\multicolumn{1}{|l|}{\textit{GPT 4}}   & \textbf{0.5\%}  & \textbf{5.5\%}  & 2.8\%  & \textbf{3.4\%}  & 11.3\% & 1.1\%  & \textbf{3.7\%}  & 2.7\%   & -3.0\% & \textbf{4.4\%}  & 5.0\%  \\
\multicolumn{1}{|l|}{GPT 3.5}          & -1.7\% & 0.2\%  & 1.9\%  & -2.3\% & -4.5\% & -1.9\% & -6.3\% & -5.1\%  & -2.1\% & 2.0\%  & -0.5\% \\ \hline
\end{tabular}
\caption{\textbf{GPT-Family Performance on Winogrande Split by Cultural Appropriateness (QA Pairs with at Least One ``good" Quality Annotation).} The table above shows the performance of GPT-3.5, GPT-4, and GPT-4o on Winogrande when split by human annotations of cultural appropriateness for each language. \textbf{Bold} values indicate the highest-performing values or model (overall across all languages, determined by the largest sum of performance scores between the models) for the given subtable. Evaluators were asked if each Winogrande QA pair could be considered ``strange, incoherent, or disrespectful in a typical conversational context", with the following possible options: (i) ``No, the sentence is typical", (ii) ``Not sure", (iii) ``Yes, the sentence is strange, incoherent, or disrespectful", or (iv) ``I don't understand the sentence". A dataset QA pair is considered ``inappropriate" if at least one \textit{evaluator} marked it as not ``typical". QA pairs are only included in this calculation if \textbf{at least one \textit{evaluator} marked it as a ``good" or ``understandable" translation}. Language codes are as follows: Xhosa (xh), Zulu (zu), Afrikaans (af), Igbo (ig), Shona (sn), Tsonga (ts), Sesotho (st), Sepedi (nso), Setswana (tn), Bambara (bm), Amharic (am).}
\label{table:appropriateness-2}
\end{table*}

\begin{table*}[]
\centering
\setlength\extrarowheight{4pt}
\begin{tabular}{|lccccccccccc|}
\hline
\multicolumn{1}{|l|}{\textbf{Language}}& \textbf{xh}  & \textbf{zu}   & \textbf{af} & \textbf{ig}   & \textbf{sn}  & \textbf{ts} & \textbf{st} & \textbf{nso}  & \textbf{tn} & \textbf{bm} & \textbf{am} \\ \hline
\multicolumn{12}{|c|}{A. Target Language Performance on ``appropriate" subset}                                                                                \\ \hline
\multicolumn{1}{|l|}{Sample Size}      & 1371   & 1339   & 1086   & 809    & 1546   & 1379   & 1452   & 1701   & 1433   & 1016   & 1369   \\ \hline
\multicolumn{1}{|l|}{\textbf{\textit{GPT 4o}}}  & \textbf{65.6\%} & \textbf{69.3\%} & \textbf{81.7\%} & \textbf{60.7\%} & \textbf{71.8\%} & \textbf{64.1\%} & \textbf{68.2\%} & \textbf{64.8\%} & \textbf{64.8\%} & 51.4\% & \textbf{60.0\%} \\
\multicolumn{1}{|l|}{\textit{GPT 4}}   & 61.6\% & 65.5\% & 80.0\% & \textbf{60.7\%} & 67.4\% & 59.3\% & 65.4\% & 58.8\% & 60.1\% & \textbf{52.2\%} & 51.6\% \\
\multicolumn{1}{|l|}{\textit{GPT 3.5}} & 51.6\% & 49.8\% & 56.7\% & 51.9\% & 51.1\% & 50.3\% & 49.8\% & 50.6\% & 50.7\% & 49.7\% & 50.4\% \\ \hline
\multicolumn{12}{|c|}{B. Target Language Performance on ``inappropriate" subset}                                                                            \\ \hline
\multicolumn{1}{|l|}{Sample Size}      & 292    & 324    & 609    & 758    & 146    & 255    & 259    & 30     & 303    & 569    & 390    \\ \hline
\multicolumn{1}{|l|}{\textbf{\textit{GPT 4o}}}  & \textbf{66.2\%} & \textbf{65.0\%} & \textbf{76.1\%} & \textbf{61.3\%} & 59.1\% & \textbf{63.4\%} & \textbf{64.2\%} & \textbf{58.9\%} & \textbf{62.8\%} & 50.7\% & \textbf{58.4\%} \\
\multicolumn{1}{|l|}{\textit{GPT 4}}   & 61.5\% & 59.3\% & 72.7\% & 57.8\% & \textbf{59.6\%} & 55.0\% & 58.4\% & \textbf{58.9\%} & 62.2\% & 50.3\% & 49.8\% \\
\multicolumn{1}{|l|}{\textit{GPT 3.5}} & 54.2\% & 50.0\% & 51.6\% & 49.1\% & 51.1\% & 52.0\% & 53.5\% & 54.4\% & 53.8\% & \textbf{54.2\%} & 51.8\% \\ \hline
\multicolumn{12}{|c|}{C. English Performance on ``appropriate" subset}                                                                        \\ \hline
\multicolumn{1}{|l|}{Sample Size}      & 1371   & 1339   & 1086   & 809    & 1546   & 1379   & 1452   & 1701   & 1433   & 1016   & 1369   \\ \hline
\multicolumn{1}{|l|}{\textbf{\textit{GPT 4o}}}  & \textbf{83.8\%} & \textbf{85.3\%} & \textbf{85.3\%} & \textbf{84.9\%} & \textbf{84.3\%} & \textbf{84.3\%} & \textbf{84.0\%} & \textbf{83.6\%} & \textbf{84.6\%} & \textbf{83.4\%} & \textbf{83.2\%} \\
\multicolumn{1}{|l|}{\textit{GPT 4}}   & 83.1\% & 83.1\% & 84.8\% & 83.0\% & 83.3\% & 83.9\% & 83.7\% & 83.0\% & 83.2\% & 82.1\% & 82.4\% \\
\multicolumn{1}{|l|}{\textit{GPT 3.5}} & 58.8\% & 58.7\% & 59.9\% & 61.0\% & 59.0\% & 58.9\% & 59.4\% & 58.7\% & 58.6\% & 57.1\% & 58.7\% \\ \hline
\multicolumn{12}{|c|}{D. English Performance on ``inappropriate" subset}                                                                    \\ \hline
\multicolumn{1}{|l|}{Sample Size}      & 292    & 324    & 609    & 758    & 146    & 255    & 259    & 30     & 303    & 569    & 390    \\ \hline
\multicolumn{1}{|l|}{\textbf{\textit{GPT 4o}}}  & \textbf{86.2\%} & 80.8\% & \textbf{81.8\%} & \textbf{83.2\%} & 76.9\% & \textbf{85.0\%} & \textbf{82.9\%} & \textbf{93.3\%} & 81.7\% & \textbf{84.2\%} & \textbf{86.9\%} \\
\multicolumn{1}{|l|}{\textit{GPT 4}}   & 83.2\% & \textbf{83.1\%} & 79.6\% & 82.6\% & \textbf{79.7\%} & 79.7\% & 80.2\% & 82.2\% & \textbf{83.3\%} & 83.9\% & 85.0\% \\
\multicolumn{1}{|l|}{\textit{GPT 3.5}} & 58.3\% & 60.7\% & 57.3\% & 56.6\% & 57.8\% & 58.7\% & 56.0\% & 55.6\% & 59.3\% & 62.8\% & 59.7\% \\ \hline
\multicolumn{12}{|c|}{E. Difference in Target Language Performance on ``appropriate" subset vs. ``inappropriate" subset (A.-B.)}                                                      \\ \hline
\multicolumn{1}{|l|}{\textit{GPT 4o}}  & -0.6\% & 4.3\%  & 5.6\%  & -0.6\% & \textbf{12.6\%} & 0.7\%  & 4.0\%  & \textbf{5.9\%}  & \textbf{2.0\%}  & 0.7\%  & 1.6\%  \\
\multicolumn{1}{|l|}{\textbf{\textit{GPT 4}}}   & \textbf{0.0\%}  & \textbf{6.3\%}  & \textbf{7.3\%}  & \textbf{2.9\%}  & 7.8\%  & \textbf{4.2\%}  & \textbf{7.0\%}  & -0.1\% & -2.1\% & \textbf{2.0\%}  & \textbf{1.8\%}  \\
\multicolumn{1}{|l|}{GPT 3.5}          & -2.7\% & -0.2\% & 5.1\%  & 2.8\%  & 0.0\%  & -1.7\% & -3.7\% & -3.8\% & -3.1\% & -4.5\% & -1.4\% \\ \hline
\multicolumn{12}{|c|}{F. Difference in English Performance on ``appropriate" subset vs. ``inappropriate" subset (C.-D.)}                                                              \\ \hline
\multicolumn{1}{|l|}{\textit{GPT 4o}}  & -2.4\% & \textbf{4.5\%}  & 3.5\%  & 1.7\%  & \textbf{7.4\%}  & -0.7\% & 1.1\%  & -9.8\% & \textbf{2.8\%}  & \textbf{-0.9\%} & -3.8\% \\
\multicolumn{1}{|l|}{\textbf{\textit{GPT 4}}}   & -0.2\% & -0.1\% & \textbf{5.3\%}  & 0.4\%  & 3.6\%  & \textbf{4.2\%}  & \textbf{3.5\%}  & 0.7\%  & -0.1\% & -1.9\% & -2.6\% \\
\multicolumn{1}{|l|}{GPT 3.5}          & \textbf{0.4\%}  & -2.0\% & 2.6\%  & \textbf{4.4\%}  & 1.3\%  & 0.2\%  & \textbf{3.5\%}  & \textbf{3.1\%}  & -0.7\% & -5.7\% & \textbf{-1.0\%} \\ \hline
\multicolumn{12}{|c|}{G. Comparative Diff. Between Target Lang. and Eng. Performance on ``appropriate" vs. ``inappropriate" subsets (E.-F.)}                               \\ \hline
\multicolumn{1}{|l|}{\textbf{\textit{GPT 4o}}}  & \textbf{1.8\%}  & -0.2\% & 2.2\%  & -2.3\% & \textbf{5.3\%}  & \textbf{1.3\%}  & 2.9\%  & \textbf{15.7\%} & \textbf{-0.8\%} & 1.6\%  & \textbf{5.4\%}  \\
\multicolumn{1}{|l|}{\textit{GPT 4}}   & 0.2\%  & \textbf{6.3\%}  & 2.1\%  & \textbf{2.5\%}  & 4.2\%  & 0.1\%  & \textbf{3.4\%}  & -0.8\% & -2.0\% & \textbf{3.9\%}  & 4.4\%  \\
\multicolumn{1}{|l|}{GPT 3.5}          & -3.1\% & 1.8\%  & \textbf{2.5\%}  & -1.6\% & -1.3\% & -1.9\% & -7.2\% & -6.9\% & -2.4\% & 1.2\%  & -0.3\% \\ \hline
\end{tabular}
\caption{\textbf{GPT-Family Performance on Winogrande Split by Cultural Appropriateness (QA Pairs with Only ``good" Quality Annotations).} The table above shows the performance of GPT-3.5, GPT-4, and GPT-4o on Winogrande when split by human annotations of cultural appropriateness for each language. \textbf{Bold} values indicate the highest-performing values or model (overall across all languages, determined by the largest sum of performance scores between the models) for the given subtable. Evaluators were asked if each Winogrande QA pair could be considered ``strange, incoherent, or disrespectful in a typical conversational context", with the following possible options: (i) ``No, the sentence is typical", (ii) ``Not sure", (iii) ``Yes, the sentence is strange, incoherent, or disrespectful", or (iv) ``I don't understand the sentence". A dataset QA pair is considered ``inappropriate" if at least one \textit{evaluator} marked it as not ``typical". QA pairs are only included in this calculation if \textbf{both evaluators marked it as a ``good" or ``understandable" translation}. Language codes are as follows: Xhosa (xh), Zulu (zu), Afrikaans (af), Igbo (ig), Shona (sn), Tsonga (ts), Sesotho (st), Sepedi (nso), Setswana (tn), Bambara (bm), Amharic (am).}
\label{table:appropriateness-3}
\end{table*}

\begin{table*}[t!]
\centering
% \small

\setlength{\tabcolsep}{5.5pt}
\begin{tabular}{r|llllllllllll}
                    \hline \hline
                     & \multicolumn{12}{c}{\texttt{Winogrande}} \\
                     % & \multicolumn{4}{c|}{\begin{tabular}[c]{@{}c@{}}\texttt{MMLU-Clinical-ZA} \\ "clinical knowledge"\end{tabular}} 
                     % & \multicolumn{4}{c}{\texttt{Belebele}} \\
                     & \textbf{en} & \textbf{af} & \textbf{zu} & \textbf{xh} 
                     & \textbf{am} & \textbf{bm} & \textbf{ig} & \textbf{nso} 
                     & \textbf{sn} & \textbf{st} & \textbf{tn} & \textbf{ts} \\ \hline 

\begin{tabular}[c]{@{}c@{}}Base \\ \tiny{Llama 3} \vspace{-0.1cm}\\ \tiny{70B IT} \end{tabular} 
& 61.2 & 51.0 & 50.4 & 50.8 & 50.8 & 50.5 & 50.5 & 50.5 & 50.4 & 50.4 & 50.5 & 50.4
\\ \hline
\multicolumn{12}{l}{\begin{tabular}[c]{@{}c@{}}\texttt{Winogrande} \\ train (small)\end{tabular}}  \\
{en$\rightarrow$}
& \begin{tabular}[c]{@{}c@{}} \colorbox{myblue}{88.0} \vspace{-0.1cm} \\ \scriptsize{±0.3}\end{tabular}
& \begin{tabular}[c]{@{}c@{}} \colorbox{mygreen}{73.4} \vspace{-0.1cm} \\ \scriptsize{±1.6}\end{tabular}
& \begin{tabular}[c]{@{}c@{}} {50.7} \vspace{-0.1cm} \\ \scriptsize{±0.8}\end{tabular}
& \begin{tabular}[c]{@{}c@{}} {52.2} \vspace{-0.1cm} \\ \scriptsize{±1.8}\end{tabular}
& \begin{tabular}[c]{@{}c@{}} {50.1} \vspace{-0.1cm} \\ \scriptsize{±0.9}\end{tabular}
& \begin{tabular}[c]{@{}c@{}} {51.0} \vspace{-0.1cm} \\ \scriptsize{±0.7}\end{tabular}
& \begin{tabular}[c]{@{}c@{}} \colorbox{mygreen}{52.6} \vspace{-0.1cm} \\ \scriptsize{±0.9}\end{tabular}
& \begin{tabular}[c]{@{}c@{}} {50.8} \vspace{-0.1cm} \\ \scriptsize{±1.2}\end{tabular}
& \begin{tabular}[c]{@{}c@{}} {52.6} \vspace{-0.1cm} \\ \scriptsize{±1.5}\end{tabular}
& \begin{tabular}[c]{@{}c@{}} {51.1} \vspace{-0.1cm} \\ \scriptsize{±0.6}\end{tabular}
& \begin{tabular}[c]{@{}c@{}} {51.6} \vspace{-0.1cm} \\ \scriptsize{±0.9}\end{tabular}
& \begin{tabular}[c]{@{}c@{}} {49.7} \vspace{-0.1cm} \\ \scriptsize{±0.6}\end{tabular}
\\
{af$\rightarrow$}
& \begin{tabular}[c]{@{}c@{}} \colorbox{mygreen}{87.3} \vspace{-0.1cm} \\ \scriptsize{±0.2}\end{tabular}
& \begin{tabular}[c]{@{}c@{}} \colorbox{myblue}{75.2} \vspace{-0.1cm} \\ \scriptsize{±0.9}\end{tabular}
& \begin{tabular}[c]{@{}c@{}} \colorbox{mygreen}{52.7} \vspace{-0.1cm} \\ \scriptsize{±1.1}\end{tabular}
& \begin{tabular}[c]{@{}c@{}} {52.2} \vspace{-0.1cm} \\ \scriptsize{±1.1}\end{tabular}
& \begin{tabular}[c]{@{}c@{}} {49.4} \vspace{-0.1cm} \\ \scriptsize{±1.6}\end{tabular}
& \begin{tabular}[c]{@{}c@{}} {48.8} \vspace{-0.1cm} \\ \scriptsize{±1.1}\end{tabular}
& \begin{tabular}[c]{@{}c@{}} \colorbox{mygreen}{53.2} \vspace{-0.1cm} \\ \scriptsize{±0.5}\end{tabular}
& \begin{tabular}[c]{@{}c@{}} {52.1} \vspace{-0.1cm} \\ \scriptsize{±1.2}\end{tabular}
& \begin{tabular}[c]{@{}c@{}} {52.3} \vspace{-0.1cm} \\ \scriptsize{±1.2}\end{tabular}
& \begin{tabular}[c]{@{}c@{}} \colorbox{mygreen}{51.3} \vspace{-0.1cm} \\ \scriptsize{±0.1}\end{tabular}
& \begin{tabular}[c]{@{}c@{}} {51.3} \vspace{-0.1cm} \\ \scriptsize{±1.2}\end{tabular}
& \begin{tabular}[c]{@{}c@{}} {50.6} \vspace{-0.1cm} \\ \scriptsize{±0.9}\end{tabular}
\\
{zu$\rightarrow$}
& \begin{tabular}[c]{@{}c@{}} \colorbox{mygreen}{74.4} \vspace{-0.1cm} \\ \scriptsize{±0.9}\end{tabular}
& \begin{tabular}[c]{@{}c@{}} \colorbox{mygreen}{61.0} \vspace{-0.1cm} \\ \scriptsize{±0.5}\end{tabular}
& \begin{tabular}[c]{@{}c@{}} {51.2} \vspace{-0.1cm} \\ \scriptsize{±0.9}\end{tabular}
& \begin{tabular}[c]{@{}c@{}} {51.2} \vspace{-0.1cm} \\ \scriptsize{±0.3}\end{tabular}
& \begin{tabular}[c]{@{}c@{}} {49.1} \vspace{-0.1cm} \\ \scriptsize{±0.5}\end{tabular}
& \begin{tabular}[c]{@{}c@{}} {50.0} \vspace{-0.1cm} \\ \scriptsize{±0.7}\end{tabular}
& \begin{tabular}[c]{@{}c@{}} {51.9} \vspace{-0.1cm} \\ \scriptsize{±0.7}\end{tabular}
& \begin{tabular}[c]{@{}c@{}} {50.2} \vspace{-0.1cm} \\ \scriptsize{±0.6}\end{tabular}
& \begin{tabular}[c]{@{}c@{}} {51.1} \vspace{-0.1cm} \\ \scriptsize{±1.9}\end{tabular}
& \begin{tabular}[c]{@{}c@{}} {50.1} \vspace{-0.1cm} \\ \scriptsize{±1.3}\end{tabular}
& \begin{tabular}[c]{@{}c@{}} {49.4} \vspace{-0.1cm} \\ \scriptsize{±0.3}\end{tabular}
& \begin{tabular}[c]{@{}c@{}} {50.6} \vspace{-0.1cm} \\ \scriptsize{±0.8}\end{tabular}
\\
{xh$\rightarrow$}
& \begin{tabular}[c]{@{}c@{}} \colorbox{mygreen}{70.3} \vspace{-0.1cm} \\ \scriptsize{±1.1}\end{tabular}
& \begin{tabular}[c]{@{}c@{}} \colorbox{mygreen}{59.2} \vspace{-0.1cm} \\ \scriptsize{±0.5}\end{tabular}
& \begin{tabular}[c]{@{}c@{}} {50.6} \vspace{-0.1cm} \\ \scriptsize{±1.4}\end{tabular}
& \begin{tabular}[c]{@{}c@{}} {50.6} \vspace{-0.1cm} \\ \scriptsize{±0.3}\end{tabular}
& \begin{tabular}[c]{@{}c@{}} {49.8} \vspace{-0.1cm} \\ \scriptsize{±1.0}\end{tabular}
& \begin{tabular}[c]{@{}c@{}} {50.4} \vspace{-0.1cm} \\ \scriptsize{±0.4}\end{tabular}
& \begin{tabular}[c]{@{}c@{}} \colorbox{mygreen}{51.6} \vspace{-0.1cm} \\ \scriptsize{±0.2}\end{tabular}
& \begin{tabular}[c]{@{}c@{}} {50.7} \vspace{-0.1cm} \\ \scriptsize{±2.0}\end{tabular}
& \begin{tabular}[c]{@{}c@{}} {50.8} \vspace{-0.1cm} \\ \scriptsize{±0.4}\end{tabular}
& \begin{tabular}[c]{@{}c@{}} {48.3} \vspace{-0.1cm} \\ \scriptsize{±1.8}\end{tabular}
& \begin{tabular}[c]{@{}c@{}} {50.0} \vspace{-0.1cm} \\ \scriptsize{±0.8}\end{tabular}
& \begin{tabular}[c]{@{}c@{}} {50.2} \vspace{-0.1cm} \\ \scriptsize{±2.2}\end{tabular}
\\
{am$\rightarrow$}
& \begin{tabular}[c]{@{}c@{}} \colorbox{mygreen}{77.0} \vspace{-0.1cm} \\ \scriptsize{±0.7}\end{tabular}
& \begin{tabular}[c]{@{}c@{}} \colorbox{mygreen}{66.2} \vspace{-0.1cm} \\ \scriptsize{±0.6}\end{tabular}
& \begin{tabular}[c]{@{}c@{}} {51.1} \vspace{-0.1cm} \\ \scriptsize{±1.5}\end{tabular}
& \begin{tabular}[c]{@{}c@{}} {50.1} \vspace{-0.1cm} \\ \scriptsize{±1.4}\end{tabular}
& \begin{tabular}[c]{@{}c@{}} {50.4} \vspace{-0.1cm} \\ \scriptsize{±1.5}\end{tabular}
& \begin{tabular}[c]{@{}c@{}} {50.2} \vspace{-0.1cm} \\ \scriptsize{±1.4}\end{tabular}
& \begin{tabular}[c]{@{}c@{}} {52.1} \vspace{-0.1cm} \\ \scriptsize{±1.2}\end{tabular}
& \begin{tabular}[c]{@{}c@{}} {50.5} \vspace{-0.1cm} \\ \scriptsize{±1.0}\end{tabular}
& \begin{tabular}[c]{@{}c@{}} {50.3} \vspace{-0.1cm} \\ \scriptsize{±0.3}\end{tabular}
& \begin{tabular}[c]{@{}c@{}} {51.0} \vspace{-0.1cm} \\ \scriptsize{±1.5}\end{tabular}
& \begin{tabular}[c]{@{}c@{}} {50.6} \vspace{-0.1cm} \\ \scriptsize{±0.8}\end{tabular}
& \begin{tabular}[c]{@{}c@{}} \colorbox{mygreen}{51.7} \vspace{-0.1cm} \\ \scriptsize{±0.5}\end{tabular}
\\
{bm$\rightarrow$}
& \begin{tabular}[c]{@{}c@{}} {60.9} \vspace{-0.1cm} \\ \scriptsize{±0.9}\end{tabular}
& \begin{tabular}[c]{@{}c@{}} {52.4} \vspace{-0.1cm} \\ \scriptsize{±1.2}\end{tabular}
& \begin{tabular}[c]{@{}c@{}} {51.1} \vspace{-0.1cm} \\ \scriptsize{±0.5}\end{tabular}
& \begin{tabular}[c]{@{}c@{}} {50.7} \vspace{-0.1cm} \\ \scriptsize{±1.3}\end{tabular}
& \begin{tabular}[c]{@{}c@{}} {50.5} \vspace{-0.1cm} \\ \scriptsize{±0.6}\end{tabular}
& \begin{tabular}[c]{@{}c@{}} {49.2} \vspace{-0.1cm} \\ \scriptsize{±2.4}\end{tabular}
& \begin{tabular}[c]{@{}c@{}} {49.2} \vspace{-0.1cm} \\ \scriptsize{±2.1}\end{tabular}
& \begin{tabular}[c]{@{}c@{}} {50.4} \vspace{-0.1cm} \\ \scriptsize{±0.5}\end{tabular}
& \begin{tabular}[c]{@{}c@{}} {50.9} \vspace{-0.1cm} \\ \scriptsize{±1.2}\end{tabular}
& \begin{tabular}[c]{@{}c@{}} {49.6} \vspace{-0.1cm} \\ \scriptsize{±1.0}\end{tabular}
& \begin{tabular}[c]{@{}c@{}} {49.3} \vspace{-0.1cm} \\ \scriptsize{±1.1}\end{tabular}
& \begin{tabular}[c]{@{}c@{}} {51.1} \vspace{-0.1cm} \\ \scriptsize{±1.6}\end{tabular}
\\
{ig$\rightarrow$}
& \begin{tabular}[c]{@{}c@{}} \colorbox{mygreen}{77.9} \vspace{-0.1cm} \\ \scriptsize{±0.9}\end{tabular}
& \begin{tabular}[c]{@{}c@{}} \colorbox{mygreen}{64.7} \vspace{-0.1cm} \\ \scriptsize{±0.3}\end{tabular}
& \begin{tabular}[c]{@{}c@{}} {50.5} \vspace{-0.1cm} \\ \scriptsize{±1.1}\end{tabular}
& \begin{tabular}[c]{@{}c@{}} {50.0} \vspace{-0.1cm} \\ \scriptsize{±2.0}\end{tabular}
& \begin{tabular}[c]{@{}c@{}} {50.2} \vspace{-0.1cm} \\ \scriptsize{±0.6}\end{tabular}
& \begin{tabular}[c]{@{}c@{}} {50.4} \vspace{-0.1cm} \\ \scriptsize{±0.8}\end{tabular}
& \begin{tabular}[c]{@{}c@{}} \colorbox{myblue}{53.8} \vspace{-0.1cm} \\ \scriptsize{±1.2}\end{tabular}
& \begin{tabular}[c]{@{}c@{}} {49.9} \vspace{-0.1cm} \\ \scriptsize{±1.1}\end{tabular}
& \begin{tabular}[c]{@{}c@{}} {51.3} \vspace{-0.1cm} \\ \scriptsize{±1.0}\end{tabular}
& \begin{tabular}[c]{@{}c@{}} {51.3} \vspace{-0.1cm} \\ \scriptsize{±2.2}\end{tabular}
& \begin{tabular}[c]{@{}c@{}} {50.2} \vspace{-0.1cm} \\ \scriptsize{±0.7}\end{tabular}
& \begin{tabular}[c]{@{}c@{}} {50.5} \vspace{-0.1cm} \\ \scriptsize{±1.0}\end{tabular}
\\
{nso$\rightarrow$}
& \begin{tabular}[c]{@{}c@{}} \colorbox{mygreen}{76.3} \vspace{-0.1cm} \\ \scriptsize{±1.3}\end{tabular}
& \begin{tabular}[c]{@{}c@{}} \colorbox{mygreen}{63.1} \vspace{-0.1cm} \\ \scriptsize{±0.4}\end{tabular}
& \begin{tabular}[c]{@{}c@{}} {51.3} \vspace{-0.1cm} \\ \scriptsize{±1.9}\end{tabular}
& \begin{tabular}[c]{@{}c@{}} {50.7} \vspace{-0.1cm} \\ \scriptsize{±1.2}\end{tabular}
& \begin{tabular}[c]{@{}c@{}} {50.5} \vspace{-0.1cm} \\ \scriptsize{±0.7}\end{tabular}
& \begin{tabular}[c]{@{}c@{}} {50.5} \vspace{-0.1cm} \\ \scriptsize{±1.4}\end{tabular}
& \begin{tabular}[c]{@{}c@{}} {51.4} \vspace{-0.1cm} \\ \scriptsize{±1.2}\end{tabular}
& \begin{tabular}[c]{@{}c@{}} {51.2} \vspace{-0.1cm} \\ \scriptsize{±1.1}\end{tabular}
& \begin{tabular}[c]{@{}c@{}} {49.7} \vspace{-0.1cm} \\ \scriptsize{±1.2}\end{tabular}
& \begin{tabular}[c]{@{}c@{}} {51.2} \vspace{-0.1cm} \\ \scriptsize{±1.5}\end{tabular}
& \begin{tabular}[c]{@{}c@{}} {50.2} \vspace{-0.1cm} \\ \scriptsize{±1.0}\end{tabular}
& \begin{tabular}[c]{@{}c@{}} \colorbox{mygreen}{51.1} \vspace{-0.1cm} \\ \scriptsize{±0.3}\end{tabular}
\\
{sn$\rightarrow$}
& \begin{tabular}[c]{@{}c@{}} \colorbox{mygreen}{76.9} \vspace{-0.1cm} \\ \scriptsize{±1.2}\end{tabular}
& \begin{tabular}[c]{@{}c@{}} \colorbox{mygreen}{65.5} \vspace{-0.1cm} \\ \scriptsize{±0.3}\end{tabular}
& \begin{tabular}[c]{@{}c@{}} {50.6} \vspace{-0.1cm} \\ \scriptsize{±1.5}\end{tabular}
& \begin{tabular}[c]{@{}c@{}} {51.2} \vspace{-0.1cm} \\ \scriptsize{±1.0}\end{tabular}
& \begin{tabular}[c]{@{}c@{}} {50.8} \vspace{-0.1cm} \\ \scriptsize{±1.2}\end{tabular}
& \begin{tabular}[c]{@{}c@{}} {51.0} \vspace{-0.1cm} \\ \scriptsize{±1.4}\end{tabular}
& \begin{tabular}[c]{@{}c@{}} {52.6} \vspace{-0.1cm} \\ \scriptsize{±1.2}\end{tabular}
& \begin{tabular}[c]{@{}c@{}} {52.0} \vspace{-0.1cm} \\ \scriptsize{±1.0}\end{tabular}
& \begin{tabular}[c]{@{}c@{}} {50.7} \vspace{-0.1cm} \\ \scriptsize{±0.4}\end{tabular}
& \begin{tabular}[c]{@{}c@{}} {50.6} \vspace{-0.1cm} \\ \scriptsize{±0.5}\end{tabular}
& \begin{tabular}[c]{@{}c@{}} {49.5} \vspace{-0.1cm} \\ \scriptsize{±0.9}\end{tabular}
& \begin{tabular}[c]{@{}c@{}} {51.3} \vspace{-0.1cm} \\ \scriptsize{±1.0}\end{tabular}
\\
{st$\rightarrow$}
& \begin{tabular}[c]{@{}c@{}} \colorbox{mygreen}{74.6} \vspace{-0.1cm} \\ \scriptsize{±1.7}\end{tabular}
& \begin{tabular}[c]{@{}c@{}} \colorbox{mygreen}{62.2} \vspace{-0.1cm} \\ \scriptsize{±0.9}\end{tabular}
& \begin{tabular}[c]{@{}c@{}} {49.8} \vspace{-0.1cm} \\ \scriptsize{±0.3}\end{tabular}
& \begin{tabular}[c]{@{}c@{}} {50.6} \vspace{-0.1cm} \\ \scriptsize{±1.5}\end{tabular}
& \begin{tabular}[c]{@{}c@{}} {49.8} \vspace{-0.1cm} \\ \scriptsize{±1.4}\end{tabular}
& \begin{tabular}[c]{@{}c@{}} {50.1} \vspace{-0.1cm} \\ \scriptsize{±1.6}\end{tabular}
& \begin{tabular}[c]{@{}c@{}} {52.8} \vspace{-0.1cm} \\ \scriptsize{±1.4}\end{tabular}
& \begin{tabular}[c]{@{}c@{}} {51.1} \vspace{-0.1cm} \\ \scriptsize{±0.7}\end{tabular}
& \begin{tabular}[c]{@{}c@{}} {51.3} \vspace{-0.1cm} \\ \scriptsize{±1.2}\end{tabular}
& \begin{tabular}[c]{@{}c@{}} {52.3} \vspace{-0.1cm} \\ \scriptsize{±1.2}\end{tabular}
& \begin{tabular}[c]{@{}c@{}} {50.9} \vspace{-0.1cm} \\ \scriptsize{±0.9}\end{tabular}
& \begin{tabular}[c]{@{}c@{}} {52.4} \vspace{-0.1cm} \\ \scriptsize{±1.3}\end{tabular}
\\
{tn$\rightarrow$}
& \begin{tabular}[c]{@{}c@{}} \colorbox{mygreen}{74.2} \vspace{-0.1cm} \\ \scriptsize{±1.2}\end{tabular}
& \begin{tabular}[c]{@{}c@{}} \colorbox{mygreen}{61.9} \vspace{-0.1cm} \\ \scriptsize{±0.6}\end{tabular}
& \begin{tabular}[c]{@{}c@{}} {49.6} \vspace{-0.1cm} \\ \scriptsize{±0.6}\end{tabular}
& \begin{tabular}[c]{@{}c@{}} {50.1} \vspace{-0.1cm} \\ \scriptsize{±0.9}\end{tabular}
& \begin{tabular}[c]{@{}c@{}} {49.9} \vspace{-0.1cm} \\ \scriptsize{±0.8}\end{tabular}
& \begin{tabular}[c]{@{}c@{}} {49.5} \vspace{-0.1cm} \\ \scriptsize{±2.1}\end{tabular}
& \begin{tabular}[c]{@{}c@{}} {51.2} \vspace{-0.1cm} \\ \scriptsize{±0.4}\end{tabular}
& \begin{tabular}[c]{@{}c@{}} {50.9} \vspace{-0.1cm} \\ \scriptsize{±1.7}\end{tabular}
& \begin{tabular}[c]{@{}c@{}} {50.9} \vspace{-0.1cm} \\ \scriptsize{±1.5}\end{tabular}
& \begin{tabular}[c]{@{}c@{}} {52.0} \vspace{-0.1cm} \\ \scriptsize{±0.8}\end{tabular}
& \begin{tabular}[c]{@{}c@{}} {51.8} \vspace{-0.1cm} \\ \scriptsize{±2.8}\end{tabular}
& \begin{tabular}[c]{@{}c@{}} {50.5} \vspace{-0.1cm} \\ \scriptsize{±0.6}\end{tabular}
\\
{ts$\rightarrow$}
& \begin{tabular}[c]{@{}c@{}} {63.5} \vspace{-0.1cm} \\ \scriptsize{±1.4}\end{tabular}
& \begin{tabular}[c]{@{}c@{}} \colorbox{mygreen}{55.1} \vspace{-0.1cm} \\ \scriptsize{±0.7}\end{tabular}
& \begin{tabular}[c]{@{}c@{}} {49.1} \vspace{-0.1cm} \\ \scriptsize{±1.7}\end{tabular}
& \begin{tabular}[c]{@{}c@{}} {50.8} \vspace{-0.1cm} \\ \scriptsize{±1.1}\end{tabular}
& \begin{tabular}[c]{@{}c@{}} {49.3} \vspace{-0.1cm} \\ \scriptsize{±0.5}\end{tabular}
& \begin{tabular}[c]{@{}c@{}} {50.5} \vspace{-0.1cm} \\ \scriptsize{±0.6}\end{tabular}
& \begin{tabular}[c]{@{}c@{}} \colorbox{mygreen}{52.2} \vspace{-0.1cm} \\ \scriptsize{±0.8}\end{tabular}
& \begin{tabular}[c]{@{}c@{}} {49.3} \vspace{-0.1cm} \\ \scriptsize{±1.6}\end{tabular}
& \begin{tabular}[c]{@{}c@{}} {49.7} \vspace{-0.1cm} \\ \scriptsize{±0.5}\end{tabular}
& \begin{tabular}[c]{@{}c@{}} \colorbox{mygreen}{50.9} \vspace{-0.1cm} \\ \scriptsize{±0.1}\end{tabular}
& \begin{tabular}[c]{@{}c@{}} {49.4} \vspace{-0.1cm} \\ \scriptsize{±1.4}\end{tabular}
& \begin{tabular}[c]{@{}c@{}} {49.9} \vspace{-0.1cm} \\ \scriptsize{±1.1}\end{tabular}
\\    
    \hline \hline

\end{tabular}
\caption{\textbf{Results of Cross-lingual Transfer using Llama 3 70B Instruct when Tuning on Winogrande and Testing on Winogrande}. Llama 3 70B Instruct was independently fine-tuned on the 11 African languages of focus:  Afrikaans (af), Zulu (zu),
Xhosa (xh) (datasets for these three languages were sourced from (BMGF 2024)), Amharic (am), Bambara (bm), Igbo (ig),
Sepedi (nso), Shona (sn), Sesotho (st), Setswana (tn), and Tsonga (ts). Llama 3 70B Instruct was also fine-tuned on English (en) for reference. The translated \texttt{Winogrande} training set (small) was used for fine-tuning. The fine-tuned models were evaluated for their cross-lingual transfer performance on the translated \texttt{Winogrande} (binary choice co-reference resolution task) test set. Columns indicate the target language of the evaluation, while the rows indicate the source language the models were fine-tuned with (e.g. ``zu$\rightarrow$" indicates models fine-tuned with data in Zulu). All numbers are the mean 5-shot performance accuracy of three evaluations followed by the standard deviation (±). Model fine-tuning that yielded mono-lingual improvements (relative to the baseline) more than two standard deviations are in \colorbox{myblue}{blue}, while cross-lingual improvements (relative to the baseline) more than two standard deviations are in \colorbox{mygreen}{green}.\\} \label{table:perf-crosslingual-llama3-tr-wino-ts-wino}
\end{table*}

\begin{table*}[t!]
\centering
% \small

\setlength{\tabcolsep}{5.5pt}
\begin{tabular}{r|llllllllllll}
                    \hline \hline
                     & \multicolumn{12}{c}{\texttt{Winogrande}} \\
                     % & \multicolumn{4}{c|}{\begin{tabular}[c]{@{}c@{}}\texttt{MMLU-Clinical-ZA} \\ "clinical knowledge"\end{tabular}} 
                     % & \multicolumn{4}{c}{\texttt{Belebele}} \\
                     & \textbf{en} & \textbf{af} & \textbf{zu} & \textbf{xh} 
                     & \textbf{am} & \textbf{bm} & \textbf{ig} & \textbf{nso} 
                     & \textbf{sn} & \textbf{st} & \textbf{tn} & \textbf{ts} \\ \hline 

\begin{tabular}[c]{@{}c@{}}Base \\ \tiny{Llama 3} \vspace{-0.1cm}\\ \tiny{70B IT} \end{tabular} 
& 61.2 & 51.0 & 50.4 & 50.8 & 50.8 & 50.5 & 50.5 & 50.5 & 50.4 & 50.4 & 50.5 & 50.4
\\ \hline
\multicolumn{12}{l}{\begin{tabular}[c]{@{}c@{}}\texttt{MMLU College Medicine} \\ dev + test + val\end{tabular}} \\
{en$\rightarrow$}
& \begin{tabular}[c]{@{}c@{}} \colorbox{myblue}{71.4} \vspace{-0.1cm} \\ \scriptsize{±0.5}\end{tabular}
& \begin{tabular}[c]{@{}c@{}} \colorbox{mygreen}{56.9} \vspace{-0.1cm} \\ \scriptsize{±0.4}\end{tabular}
& \begin{tabular}[c]{@{}c@{}} \colorbox{mygreen}{51.0} \vspace{-0.1cm} \\ \scriptsize{±0.2}\end{tabular}
& \begin{tabular}[c]{@{}c@{}} {50.9} \vspace{-0.1cm} \\ \scriptsize{±0.2}\end{tabular}
& \begin{tabular}[c]{@{}c@{}} {50.0} \vspace{-0.1cm} \\ \scriptsize{±0.2}\end{tabular}
& \begin{tabular}[c]{@{}c@{}} {50.4} \vspace{-0.1cm} \\ \scriptsize{±0.1}\end{tabular}
& \begin{tabular}[c]{@{}c@{}} {50.4} \vspace{-0.1cm} \\ \scriptsize{±0.6}\end{tabular}
& \begin{tabular}[c]{@{}c@{}} {50.6} \vspace{-0.1cm} \\ \scriptsize{±0.3}\end{tabular}
& \begin{tabular}[c]{@{}c@{}} {50.3} \vspace{-0.1cm} \\ \scriptsize{±0.3}\end{tabular}
& \begin{tabular}[c]{@{}c@{}} {50.5} \vspace{-0.1cm} \\ \scriptsize{±0.2}\end{tabular}
& \begin{tabular}[c]{@{}c@{}} \colorbox{mygreen}{51.3} \vspace{-0.1cm} \\ \scriptsize{±0.3}\end{tabular}
& \begin{tabular}[c]{@{}c@{}} {50.7} \vspace{-0.1cm} \\ \scriptsize{±0.3}\end{tabular}
\\
{af$\rightarrow$}
& \begin{tabular}[c]{@{}c@{}} \colorbox{mygreen}{68.3} \vspace{-0.1cm} \\ \scriptsize{±0.1}\end{tabular}
& \begin{tabular}[c]{@{}c@{}} \colorbox{myblue}{58.0} \vspace{-0.1cm} \\ \scriptsize{±0.6}\end{tabular}
& \begin{tabular}[c]{@{}c@{}} {50.8} \vspace{-0.1cm} \\ \scriptsize{±0.2}\end{tabular}
& \begin{tabular}[c]{@{}c@{}} {51.0} \vspace{-0.1cm} \\ \scriptsize{±0.3}\end{tabular}
& \begin{tabular}[c]{@{}c@{}} {50.2} \vspace{-0.1cm} \\ \scriptsize{±0.9}\end{tabular}
& \begin{tabular}[c]{@{}c@{}} {50.9} \vspace{-0.1cm} \\ \scriptsize{±0.4}\end{tabular}
& \begin{tabular}[c]{@{}c@{}} {51.3} \vspace{-0.1cm} \\ \scriptsize{±0.9}\end{tabular}
& \begin{tabular}[c]{@{}c@{}} {50.4} \vspace{-0.1cm} \\ \scriptsize{±0.8}\end{tabular}
& \begin{tabular}[c]{@{}c@{}} {50.4} \vspace{-0.1cm} \\ \scriptsize{±0.1}\end{tabular}
& \begin{tabular}[c]{@{}c@{}} {50.4} \vspace{-0.1cm} \\ \scriptsize{±0.7}\end{tabular}
& \begin{tabular}[c]{@{}c@{}} {50.2} \vspace{-0.1cm} \\ \scriptsize{±0.4}\end{tabular}
& \begin{tabular}[c]{@{}c@{}} {50.0} \vspace{-0.1cm} \\ \scriptsize{±0.6}\end{tabular}
\\
{zu$\rightarrow$}
& \begin{tabular}[c]{@{}c@{}} \colorbox{mygreen}{71.5} \vspace{-0.1cm} \\ \scriptsize{±0.2}\end{tabular}
& \begin{tabular}[c]{@{}c@{}} \colorbox{mygreen}{60.0} \vspace{-0.1cm} \\ \scriptsize{±0.2}\end{tabular}
& \begin{tabular}[c]{@{}c@{}} \colorbox{myblue}{52.4} \vspace{-0.1cm} \\ \scriptsize{±0.3}\end{tabular}
& \begin{tabular}[c]{@{}c@{}} {51.4} \vspace{-0.1cm} \\ \scriptsize{±0.4}\end{tabular}
& \begin{tabular}[c]{@{}c@{}} {50.8} \vspace{-0.1cm} \\ \scriptsize{±0.4}\end{tabular}
& \begin{tabular}[c]{@{}c@{}} {50.8} \vspace{-0.1cm} \\ \scriptsize{±0.4}\end{tabular}
& \begin{tabular}[c]{@{}c@{}} \colorbox{mygreen}{52.5} \vspace{-0.1cm} \\ \scriptsize{±0.3}\end{tabular}
& \begin{tabular}[c]{@{}c@{}} {50.9} \vspace{-0.1cm} \\ \scriptsize{±1.3}\end{tabular}
& \begin{tabular}[c]{@{}c@{}} {50.8} \vspace{-0.1cm} \\ \scriptsize{±0.5}\end{tabular}
& \begin{tabular}[c]{@{}c@{}} \colorbox{mygreen}{52.3} \vspace{-0.1cm} \\ \scriptsize{±0.3}\end{tabular}
& \begin{tabular}[c]{@{}c@{}} {51.0} \vspace{-0.1cm} \\ \scriptsize{±0.3}\end{tabular}
& \begin{tabular}[c]{@{}c@{}} {51.5} \vspace{-0.1cm} \\ \scriptsize{±0.8}\end{tabular}
\\
{xh$\rightarrow$}
& \begin{tabular}[c]{@{}c@{}} \colorbox{mygreen}{69.0} \vspace{-0.1cm} \\ \scriptsize{±0.3}\end{tabular}
& \begin{tabular}[c]{@{}c@{}} \colorbox{mygreen}{59.2} \vspace{-0.1cm} \\ \scriptsize{±0.2}\end{tabular}
& \begin{tabular}[c]{@{}c@{}} \colorbox{mygreen}{51.0} \vspace{-0.1cm} \\ \scriptsize{±0.2}\end{tabular}
& \begin{tabular}[c]{@{}c@{}} {51.1} \vspace{-0.1cm} \\ \scriptsize{±0.3}\end{tabular}
& \begin{tabular}[c]{@{}c@{}} {50.8} \vspace{-0.1cm} \\ \scriptsize{±1.2}\end{tabular}
& \begin{tabular}[c]{@{}c@{}} {50.3} \vspace{-0.1cm} \\ \scriptsize{±0.4}\end{tabular}
& \begin{tabular}[c]{@{}c@{}} {51.0} \vspace{-0.1cm} \\ \scriptsize{±0.7}\end{tabular}
& \begin{tabular}[c]{@{}c@{}} {50.9} \vspace{-0.1cm} \\ \scriptsize{±0.6}\end{tabular}
& \begin{tabular}[c]{@{}c@{}} {50.8} \vspace{-0.1cm} \\ \scriptsize{±0.4}\end{tabular}
& \begin{tabular}[c]{@{}c@{}} \colorbox{mygreen}{51.3} \vspace{-0.1cm} \\ \scriptsize{±0.3}\end{tabular}
& \begin{tabular}[c]{@{}c@{}} {51.1} \vspace{-0.1cm} \\ \scriptsize{±0.5}\end{tabular}
& \begin{tabular}[c]{@{}c@{}} {51.2} \vspace{-0.1cm} \\ \scriptsize{±0.5}\end{tabular}
\\
{am$\rightarrow$}
& \begin{tabular}[c]{@{}c@{}} \colorbox{mygreen}{70.5} \vspace{-0.1cm} \\ \scriptsize{±0.3}\end{tabular}
& \begin{tabular}[c]{@{}c@{}} \colorbox{mygreen}{60.0} \vspace{-0.1cm} \\ \scriptsize{±0.4}\end{tabular}
& \begin{tabular}[c]{@{}c@{}} \colorbox{mygreen}{50.9} \vspace{-0.1cm} \\ \scriptsize{±0.2}\end{tabular}
& \begin{tabular}[c]{@{}c@{}} {50.8} \vspace{-0.1cm} \\ \scriptsize{±0.2}\end{tabular}
& \begin{tabular}[c]{@{}c@{}} {49.9} \vspace{-0.1cm} \\ \scriptsize{±0.2}\end{tabular}
& \begin{tabular}[c]{@{}c@{}} {50.1} \vspace{-0.1cm} \\ \scriptsize{±0.2}\end{tabular}
& \begin{tabular}[c]{@{}c@{}} \colorbox{mygreen}{51.5} \vspace{-0.1cm} \\ \scriptsize{±0.2}\end{tabular}
& \begin{tabular}[c]{@{}c@{}} {50.4} \vspace{-0.1cm} \\ \scriptsize{±0.2}\end{tabular}
& \begin{tabular}[c]{@{}c@{}} \colorbox{mygreen}{50.9} \vspace{-0.1cm} \\ \scriptsize{±0.1}\end{tabular}
& \begin{tabular}[c]{@{}c@{}} \colorbox{mygreen}{50.9} \vspace{-0.1cm} \\ \scriptsize{±0.1}\end{tabular}
& \begin{tabular}[c]{@{}c@{}} {50.7} \vspace{-0.1cm} \\ \scriptsize{±0.4}\end{tabular}
& \begin{tabular}[c]{@{}c@{}} {50.6} \vspace{-0.1cm} \\ \scriptsize{±0.2}\end{tabular}
\\
{bm$\rightarrow$}
& \begin{tabular}[c]{@{}c@{}} {60.2} \vspace{-0.1cm} \\ \scriptsize{±0.2}\end{tabular}
& \begin{tabular}[c]{@{}c@{}} \colorbox{mygreen}{53.4} \vspace{-0.1cm} \\ \scriptsize{±0.1}\end{tabular}
& \begin{tabular}[c]{@{}c@{}} {50.5} \vspace{-0.1cm} \\ \scriptsize{±0.1}\end{tabular}
& \begin{tabular}[c]{@{}c@{}} {50.6} \vspace{-0.1cm} \\ \scriptsize{±0.2}\end{tabular}
& \begin{tabular}[c]{@{}c@{}} {50.4} \vspace{-0.1cm} \\ \scriptsize{±0.1}\end{tabular}
& \begin{tabular}[c]{@{}c@{}} {50.4} \vspace{-0.1cm} \\ \scriptsize{±0.1}\end{tabular}
& \begin{tabular}[c]{@{}c@{}} {50.3} \vspace{-0.1cm} \\ \scriptsize{±0.1}\end{tabular}
& \begin{tabular}[c]{@{}c@{}} {50.4} \vspace{-0.1cm} \\ \scriptsize{±0.1}\end{tabular}
& \begin{tabular}[c]{@{}c@{}} {50.5} \vspace{-0.1cm} \\ \scriptsize{±0.1}\end{tabular}
& \begin{tabular}[c]{@{}c@{}} {50.5} \vspace{-0.1cm} \\ \scriptsize{±0.1}\end{tabular}
& \begin{tabular}[c]{@{}c@{}} {50.5} \vspace{-0.1cm} \\ \scriptsize{±0.0}\end{tabular}
& \begin{tabular}[c]{@{}c@{}} {50.4} \vspace{-0.1cm} \\ \scriptsize{±0.0}\end{tabular}
\\
{ig$\rightarrow$}
& \begin{tabular}[c]{@{}c@{}} \colorbox{mygreen}{69.0} \vspace{-0.1cm} \\ \scriptsize{±0.3}\end{tabular}
& \begin{tabular}[c]{@{}c@{}} \colorbox{mygreen}{59.2} \vspace{-0.1cm} \\ \scriptsize{±0.3}\end{tabular}
& \begin{tabular}[c]{@{}c@{}} {51.5} \vspace{-0.1cm} \\ \scriptsize{±0.9}\end{tabular}
& \begin{tabular}[c]{@{}c@{}} \colorbox{mygreen}{51.6} \vspace{-0.1cm} \\ \scriptsize{±0.1}\end{tabular}
& \begin{tabular}[c]{@{}c@{}} {50.2} \vspace{-0.1cm} \\ \scriptsize{±0.1}\end{tabular}
& \begin{tabular}[c]{@{}c@{}} {50.7} \vspace{-0.1cm} \\ \scriptsize{±0.1}\end{tabular}
& \begin{tabular}[c]{@{}c@{}} \colorbox{myblue}{52.0} \vspace{-0.1cm} \\ \scriptsize{±0.3}\end{tabular}
& \begin{tabular}[c]{@{}c@{}} \colorbox{mygreen}{51.1} \vspace{-0.1cm} \\ \scriptsize{±0.1}\end{tabular}
& \begin{tabular}[c]{@{}c@{}} {50.6} \vspace{-0.1cm} \\ \scriptsize{±0.6}\end{tabular}
& \begin{tabular}[c]{@{}c@{}} \colorbox{mygreen}{51.2} \vspace{-0.1cm} \\ \scriptsize{±0.4}\end{tabular}
& \begin{tabular}[c]{@{}c@{}} {50.9} \vspace{-0.1cm} \\ \scriptsize{±0.2}\end{tabular}
& \begin{tabular}[c]{@{}c@{}} {50.8} \vspace{-0.1cm} \\ \scriptsize{±0.6}\end{tabular}
\\
{nso$\rightarrow$}
& \begin{tabular}[c]{@{}c@{}} \colorbox{mygreen}{68.8} \vspace{-0.1cm} \\ \scriptsize{±0.4}\end{tabular}
& \begin{tabular}[c]{@{}c@{}} \colorbox{mygreen}{59.5} \vspace{-0.1cm} \\ \scriptsize{±0.1}\end{tabular}
& \begin{tabular}[c]{@{}c@{}} {50.9} \vspace{-0.1cm} \\ \scriptsize{±0.3}\end{tabular}
& \begin{tabular}[c]{@{}c@{}} {50.8} \vspace{-0.1cm} \\ \scriptsize{±0.2}\end{tabular}
& \begin{tabular}[c]{@{}c@{}} {50.3} \vspace{-0.1cm} \\ \scriptsize{±0.6}\end{tabular}
& \begin{tabular}[c]{@{}c@{}} {50.5} \vspace{-0.1cm} \\ \scriptsize{±0.1}\end{tabular}
& \begin{tabular}[c]{@{}c@{}} \colorbox{mygreen}{52.0} \vspace{-0.1cm} \\ \scriptsize{±0.3}\end{tabular}
& \begin{tabular}[c]{@{}c@{}} {50.7} \vspace{-0.1cm} \\ \scriptsize{±0.2}\end{tabular}
& \begin{tabular}[c]{@{}c@{}} \colorbox{mygreen}{50.9} \vspace{-0.1cm} \\ \scriptsize{±0.1}\end{tabular}
& \begin{tabular}[c]{@{}c@{}} \colorbox{mygreen}{51.3} \vspace{-0.1cm} \\ \scriptsize{±0.3}\end{tabular}
& \begin{tabular}[c]{@{}c@{}} {51.0} \vspace{-0.1cm} \\ \scriptsize{±0.6}\end{tabular}
& \begin{tabular}[c]{@{}c@{}} \colorbox{mygreen}{50.8} \vspace{-0.1cm} \\ \scriptsize{±0.1}\end{tabular}
\\
{sn$\rightarrow$}
& \begin{tabular}[c]{@{}c@{}} \colorbox{mygreen}{72.0} \vspace{-0.1cm} \\ \scriptsize{±0.3}\end{tabular}
& \begin{tabular}[c]{@{}c@{}} \colorbox{mygreen}{60.8} \vspace{-0.1cm} \\ \scriptsize{±0.1}\end{tabular}
& \begin{tabular}[c]{@{}c@{}} \colorbox{mygreen}{51.3} \vspace{-0.1cm} \\ \scriptsize{±0.2}\end{tabular}
& \begin{tabular}[c]{@{}c@{}} {51.4} \vspace{-0.1cm} \\ \scriptsize{±0.3}\end{tabular}
& \begin{tabular}[c]{@{}c@{}} {49.7} \vspace{-0.1cm} \\ \scriptsize{±1.1}\end{tabular}
& \begin{tabular}[c]{@{}c@{}} \colorbox{mygreen}{50.8} \vspace{-0.1cm} \\ \scriptsize{±0.1}\end{tabular}
& \begin{tabular}[c]{@{}c@{}} \colorbox{mygreen}{52.0} \vspace{-0.1cm} \\ \scriptsize{±0.5}\end{tabular}
& \begin{tabular}[c]{@{}c@{}} {50.8} \vspace{-0.1cm} \\ \scriptsize{±0.4}\end{tabular}
& \begin{tabular}[c]{@{}c@{}} {51.3} \vspace{-0.1cm} \\ \scriptsize{±0.6}\end{tabular}
& \begin{tabular}[c]{@{}c@{}} {51.1} \vspace{-0.1cm} \\ \scriptsize{±1.0}\end{tabular}
& \begin{tabular}[c]{@{}c@{}} {50.8} \vspace{-0.1cm} \\ \scriptsize{±0.6}\end{tabular}
& \begin{tabular}[c]{@{}c@{}} {50.9} \vspace{-0.1cm} \\ \scriptsize{±0.5}\end{tabular}
\\
{st$\rightarrow$}
& \begin{tabular}[c]{@{}c@{}} \colorbox{mygreen}{70.2} \vspace{-0.1cm} \\ \scriptsize{±0.8}\end{tabular}
& \begin{tabular}[c]{@{}c@{}} \colorbox{mygreen}{59.8} \vspace{-0.1cm} \\ \scriptsize{±0.4}\end{tabular}
& \begin{tabular}[c]{@{}c@{}} \colorbox{mygreen}{51.4} \vspace{-0.1cm} \\ \scriptsize{±0.3}\end{tabular}
& \begin{tabular}[c]{@{}c@{}} {50.9} \vspace{-0.1cm} \\ \scriptsize{±0.1}\end{tabular}
& \begin{tabular}[c]{@{}c@{}} {50.8} \vspace{-0.1cm} \\ \scriptsize{±0.9}\end{tabular}
& \begin{tabular}[c]{@{}c@{}} \colorbox{mygreen}{51.2} \vspace{-0.1cm} \\ \scriptsize{±0.1}\end{tabular}
& \begin{tabular}[c]{@{}c@{}} \colorbox{mygreen}{52.0} \vspace{-0.1cm} \\ \scriptsize{±0.3}\end{tabular}
& \begin{tabular}[c]{@{}c@{}} {51.7} \vspace{-0.1cm} \\ \scriptsize{±0.8}\end{tabular}
& \begin{tabular}[c]{@{}c@{}} {51.1} \vspace{-0.1cm} \\ \scriptsize{±0.7}\end{tabular}
& \begin{tabular}[c]{@{}c@{}} \colorbox{myblue}{52.6} \vspace{-0.1cm} \\ \scriptsize{±0.4}\end{tabular}
& \begin{tabular}[c]{@{}c@{}} {51.7} \vspace{-0.1cm} \\ \scriptsize{±0.6}\end{tabular}
& \begin{tabular}[c]{@{}c@{}} {51.1} \vspace{-0.1cm} \\ \scriptsize{±0.4}\end{tabular}
\\
{tn$\rightarrow$}
& \begin{tabular}[c]{@{}c@{}} \colorbox{mygreen}{65.2} \vspace{-0.1cm} \\ \scriptsize{±0.4}\end{tabular}
& \begin{tabular}[c]{@{}c@{}} \colorbox{mygreen}{58.1} \vspace{-0.1cm} \\ \scriptsize{±0.3}\end{tabular}
& \begin{tabular}[c]{@{}c@{}} {51.0} \vspace{-0.1cm} \\ \scriptsize{±0.3}\end{tabular}
& \begin{tabular}[c]{@{}c@{}} {50.7} \vspace{-0.1cm} \\ \scriptsize{±0.2}\end{tabular}
& \begin{tabular}[c]{@{}c@{}} {50.8} \vspace{-0.1cm} \\ \scriptsize{±0.4}\end{tabular}
& \begin{tabular}[c]{@{}c@{}} {50.5} \vspace{-0.1cm} \\ \scriptsize{±0.1}\end{tabular}
& \begin{tabular}[c]{@{}c@{}} \colorbox{mygreen}{51.4} \vspace{-0.1cm} \\ \scriptsize{±0.4}\end{tabular}
& \begin{tabular}[c]{@{}c@{}} {50.5} \vspace{-0.1cm} \\ \scriptsize{±0.1}\end{tabular}
& \begin{tabular}[c]{@{}c@{}} \colorbox{mygreen}{50.7} \vspace{-0.1cm} \\ \scriptsize{±0.1}\end{tabular}
& \begin{tabular}[c]{@{}c@{}} {50.8} \vspace{-0.1cm} \\ \scriptsize{±0.4}\end{tabular}
& \begin{tabular}[c]{@{}c@{}} {50.9} \vspace{-0.1cm} \\ \scriptsize{±0.4}\end{tabular}
& \begin{tabular}[c]{@{}c@{}} {50.6} \vspace{-0.1cm} \\ \scriptsize{±0.3}\end{tabular}
\\
{ts$\rightarrow$}
& \begin{tabular}[c]{@{}c@{}} \colorbox{mygreen}{70.7} \vspace{-0.1cm} \\ \scriptsize{±0.0}\end{tabular}
& \begin{tabular}[c]{@{}c@{}} \colorbox{mygreen}{59.3} \vspace{-0.1cm} \\ \scriptsize{±0.0}\end{tabular}
& \begin{tabular}[c]{@{}c@{}} \colorbox{mygreen}{50.7} \vspace{-0.1cm} \\ \scriptsize{±0.1}\end{tabular}
& \begin{tabular}[c]{@{}c@{}} {50.9} \vspace{-0.1cm} \\ \scriptsize{±0.1}\end{tabular}
& \begin{tabular}[c]{@{}c@{}} {50.5} \vspace{-0.1cm} \\ \scriptsize{±0.1}\end{tabular}
& \begin{tabular}[c]{@{}c@{}} {50.5} \vspace{-0.1cm} \\ \scriptsize{±0.0}\end{tabular}
& \begin{tabular}[c]{@{}c@{}} \colorbox{mygreen}{51.6} \vspace{-0.1cm} \\ \scriptsize{±0.2}\end{tabular}
& \begin{tabular}[c]{@{}c@{}} {50.4} \vspace{-0.1cm} \\ \scriptsize{±0.6}\end{tabular}
& \begin{tabular}[c]{@{}c@{}} \colorbox{mygreen}{51.0} \vspace{-0.1cm} \\ \scriptsize{±0.2}\end{tabular}
& \begin{tabular}[c]{@{}c@{}} {51.4} \vspace{-0.1cm} \\ \scriptsize{±0.7}\end{tabular}
& \begin{tabular}[c]{@{}c@{}} {50.8} \vspace{-0.1cm} \\ \scriptsize{±0.3}\end{tabular}
& \begin{tabular}[c]{@{}c@{}} {50.7} \vspace{-0.1cm} \\ \scriptsize{±0.2}\end{tabular}
\\    
    \hline \hline

\end{tabular}
\caption{\textbf{Results of Cross-lingual Transfer using Llama 3 70B Instruct when Tuning on MMLU College Medicine and Testing on Winogrande}. Llama 3 70B Instruct was independently fine-tuned on the 11 African languages of focus:  Afrikaans (af), Zulu (zu),
Xhosa (xh) (datasets for these three languages were sourced from (BMGF 2024)), Amharic (am), Bambara (bm), Igbo (ig),
Sepedi (nso), Shona (sn), Sesotho (st), Setswana (tn), and Tsonga (ts). Llama 3 70B Instruct was also fine-tuned on English (en) for reference. The translated \texttt{MMLU College Medicine} section (dev + test + val sets) was used for fine-tuning. The fine-tuned models were evaluated for their cross-lingual transfer performance on the translated \texttt{Winogrande} (binary choice co-reference resolution task) test set. Columns indicate the target language of the evaluation, while the rows indicate the source language the models were fine-tuned with (e.g. ``zu$\rightarrow$" indicates models fine-tuned with data in Zulu). All numbers are the mean 5-shot performance accuracy of three evaluations followed by the standard deviation (±). Model fine-tuning that yielded mono-lingual improvements (relative to the baseline) more than two standard deviations are in \colorbox{myblue}{blue}, while cross-lingual improvements (relative to the baseline) more than two standard deviations are in \colorbox{mygreen}{green}.\\}\label{table:perf-crosslingual-llama3-tr-mmlu-ts-wino}
\end{table*}

\begin{table*}[t!]
\centering
% \small

\setlength{\tabcolsep}{5.5pt}
\begin{tabular}{r|llllllllllll}
                    \hline \hline
                     & \multicolumn{12}{c}{\texttt{MMLU Clinical Knowledge}} \\
                     % & \multicolumn{4}{c|}{\begin{tabular}[c]{@{}c@{}}\texttt{MMLU-Clinical-ZA} \\ "clinical knowledge"\end{tabular}} 
                     % & \multicolumn{4}{c}{\texttt{Belebele}} \\
                     & \textbf{en} & \textbf{af} & \textbf{zu} & \textbf{xh} 
                     & \textbf{am} & \textbf{bm} & \textbf{ig} & \textbf{nso} 
                     & \textbf{sn} & \textbf{st} & \textbf{tn} & \textbf{ts} \\ \hline 

\begin{tabular}[c]{@{}c@{}}Base \\ \tiny{Llama 3} \vspace{-0.1cm}\\ \tiny{70B IT} \end{tabular} 
& 82.3 & 71.3 & 39.2 & 38.5 & 33.6 & 32.5 & 33.6 & 37.0 & 43.4 & 43.8 & 39.2 & 34.0
\\ \hline
\multicolumn{12}{l}{\begin{tabular}[c]{@{}c@{}}\texttt{Winogrande} \\ train (small)\end{tabular}}  \\
{en$\rightarrow$}
& \begin{tabular}[c]{@{}c@{}} {82.5} \vspace{-0.1cm} \\ \scriptsize{±0.8}\end{tabular}
& \begin{tabular}[c]{@{}c@{}} {71.5} \vspace{-0.1cm} \\ \scriptsize{±0.8}\end{tabular}
& \begin{tabular}[c]{@{}c@{}} {36.5} \vspace{-0.1cm} \\ \scriptsize{±4.4}\end{tabular}
& \begin{tabular}[c]{@{}c@{}} {37.6} \vspace{-0.1cm} \\ \scriptsize{±1.7}\end{tabular}
& \begin{tabular}[c]{@{}c@{}} {34.8} \vspace{-0.1cm} \\ \scriptsize{±2.3}\end{tabular}
& \begin{tabular}[c]{@{}c@{}} {31.1} \vspace{-0.1cm} \\ \scriptsize{±1.9}\end{tabular}
& \begin{tabular}[c]{@{}c@{}} {35.6} \vspace{-0.1cm} \\ \scriptsize{±1.8}\end{tabular}
& \begin{tabular}[c]{@{}c@{}} {37.2} \vspace{-0.1cm} \\ \scriptsize{±0.8}\end{tabular}
& \begin{tabular}[c]{@{}c@{}} {41.0} \vspace{-0.1cm} \\ \scriptsize{±2.5}\end{tabular}
& \begin{tabular}[c]{@{}c@{}} {40.0} \vspace{-0.1cm} \\ \scriptsize{±1.3}\end{tabular}
& \begin{tabular}[c]{@{}c@{}} {37.3} \vspace{-0.1cm} \\ \scriptsize{±2.1}\end{tabular}
& \begin{tabular}[c]{@{}c@{}} {35.2} \vspace{-0.1cm} \\ \scriptsize{±2.9}\end{tabular}
\\
{af$\rightarrow$}
& \begin{tabular}[c]{@{}c@{}} {83.6} \vspace{-0.1cm} \\ \scriptsize{±0.8}\end{tabular}
& \begin{tabular}[c]{@{}c@{}} \colorbox{myblue}{74.2} \vspace{-0.1cm} \\ \scriptsize{±0.8}\end{tabular}
& \begin{tabular}[c]{@{}c@{}} {40.2} \vspace{-0.1cm} \\ \scriptsize{±3.3}\end{tabular}
& \begin{tabular}[c]{@{}c@{}} {39.5} \vspace{-0.1cm} \\ \scriptsize{±2.2}\end{tabular}
& \begin{tabular}[c]{@{}c@{}} \colorbox{mygreen}{38.1} \vspace{-0.1cm} \\ \scriptsize{±1.3}\end{tabular}
& \begin{tabular}[c]{@{}c@{}} {30.3} \vspace{-0.1cm} \\ \scriptsize{±1.5}\end{tabular}
& \begin{tabular}[c]{@{}c@{}} \colorbox{mygreen}{38.7} \vspace{-0.1cm} \\ \scriptsize{±2.2}\end{tabular}
& \begin{tabular}[c]{@{}c@{}} {40.4} \vspace{-0.1cm} \\ \scriptsize{±3.6}\end{tabular}
& \begin{tabular}[c]{@{}c@{}} {43.2} \vspace{-0.1cm} \\ \scriptsize{±2.5}\end{tabular}
& \begin{tabular}[c]{@{}c@{}} {44.3} \vspace{-0.1cm} \\ \scriptsize{±1.9}\end{tabular}
& \begin{tabular}[c]{@{}c@{}} {39.1} \vspace{-0.1cm} \\ \scriptsize{±1.0}\end{tabular}
& \begin{tabular}[c]{@{}c@{}} {38.1} \vspace{-0.1cm} \\ \scriptsize{±2.7}\end{tabular}
\\
{zu$\rightarrow$}
& \begin{tabular}[c]{@{}c@{}} {81.3} \vspace{-0.1cm} \\ \scriptsize{±2.3}\end{tabular}
& \begin{tabular}[c]{@{}c@{}} \colorbox{mygreen}{73.1} \vspace{-0.1cm} \\ \scriptsize{±0.8}\end{tabular}
& \begin{tabular}[c]{@{}c@{}} {37.1} \vspace{-0.1cm} \\ \scriptsize{±3.5}\end{tabular}
& \begin{tabular}[c]{@{}c@{}} {39.3} \vspace{-0.1cm} \\ \scriptsize{±2.3}\end{tabular}
& \begin{tabular}[c]{@{}c@{}} {39.1} \vspace{-0.1cm} \\ \scriptsize{±3.0}\end{tabular}
& \begin{tabular}[c]{@{}c@{}} {31.6} \vspace{-0.1cm} \\ \scriptsize{±2.3}\end{tabular}
& \begin{tabular}[c]{@{}c@{}} \colorbox{mygreen}{41.3} \vspace{-0.1cm} \\ \scriptsize{±0.6}\end{tabular}
& \begin{tabular}[c]{@{}c@{}} \colorbox{mygreen}{38.6} \vspace{-0.1cm} \\ \scriptsize{±0.2}\end{tabular}
& \begin{tabular}[c]{@{}c@{}} {42.1} \vspace{-0.1cm} \\ \scriptsize{±0.6}\end{tabular}
& \begin{tabular}[c]{@{}c@{}} {40.9} \vspace{-0.1cm} \\ \scriptsize{±2.6}\end{tabular}
& \begin{tabular}[c]{@{}c@{}} {39.1} \vspace{-0.1cm} \\ \scriptsize{±1.5}\end{tabular}
& \begin{tabular}[c]{@{}c@{}} {33.7} \vspace{-0.1cm} \\ \scriptsize{±2.1}\end{tabular}
\\
{xh$\rightarrow$}
& \begin{tabular}[c]{@{}c@{}} {81.5} \vspace{-0.1cm} \\ \scriptsize{±0.7}\end{tabular}
& \begin{tabular}[c]{@{}c@{}} {70.6} \vspace{-0.1cm} \\ \scriptsize{±2.3}\end{tabular}
& \begin{tabular}[c]{@{}c@{}} {37.0} \vspace{-0.1cm} \\ \scriptsize{±3.3}\end{tabular}
& \begin{tabular}[c]{@{}c@{}} {36.8} \vspace{-0.1cm} \\ \scriptsize{±2.9}\end{tabular}
& \begin{tabular}[c]{@{}c@{}} {32.8} \vspace{-0.1cm} \\ \scriptsize{±0.4}\end{tabular}
& \begin{tabular}[c]{@{}c@{}} {27.9} \vspace{-0.1cm} \\ \scriptsize{±1.9}\end{tabular}
& \begin{tabular}[c]{@{}c@{}} {37.5} \vspace{-0.1cm} \\ \scriptsize{±3.1}\end{tabular}
& \begin{tabular}[c]{@{}c@{}} {36.3} \vspace{-0.1cm} \\ \scriptsize{±3.0}\end{tabular}
& \begin{tabular}[c]{@{}c@{}} {41.1} \vspace{-0.1cm} \\ \scriptsize{±1.7}\end{tabular}
& \begin{tabular}[c]{@{}c@{}} {37.9} \vspace{-0.1cm} \\ \scriptsize{±2.5}\end{tabular}
& \begin{tabular}[c]{@{}c@{}} {38.4} \vspace{-0.1cm} \\ \scriptsize{±3.6}\end{tabular}
& \begin{tabular}[c]{@{}c@{}} {32.1} \vspace{-0.1cm} \\ \scriptsize{±0.4}\end{tabular}
\\
{am$\rightarrow$}
& \begin{tabular}[c]{@{}c@{}} {83.6} \vspace{-0.1cm} \\ \scriptsize{±0.8}\end{tabular}
& \begin{tabular}[c]{@{}c@{}} \colorbox{mygreen}{76.6} \vspace{-0.1cm} \\ \scriptsize{±1.7}\end{tabular}
& \begin{tabular}[c]{@{}c@{}} {41.3} \vspace{-0.1cm} \\ \scriptsize{±3.0}\end{tabular}
& \begin{tabular}[c]{@{}c@{}} {40.3} \vspace{-0.1cm} \\ \scriptsize{±1.2}\end{tabular}
& \begin{tabular}[c]{@{}c@{}} \colorbox{myblue}{46.4} \vspace{-0.1cm} \\ \scriptsize{±2.0}\end{tabular}
& \begin{tabular}[c]{@{}c@{}} {31.1} \vspace{-0.1cm} \\ \scriptsize{±2.9}\end{tabular}
& \begin{tabular}[c]{@{}c@{}} \colorbox{mygreen}{44.2} \vspace{-0.1cm} \\ \scriptsize{±0.8}\end{tabular}
& \begin{tabular}[c]{@{}c@{}} {41.6} \vspace{-0.1cm} \\ \scriptsize{±5.5}\end{tabular}
& \begin{tabular}[c]{@{}c@{}} {45.3} \vspace{-0.1cm} \\ \scriptsize{±1.3}\end{tabular}
& \begin{tabular}[c]{@{}c@{}} {42.6} \vspace{-0.1cm} \\ \scriptsize{±0.4}\end{tabular}
& \begin{tabular}[c]{@{}c@{}} {41.6} \vspace{-0.1cm} \\ \scriptsize{±1.7}\end{tabular}
& \begin{tabular}[c]{@{}c@{}} {37.6} \vspace{-0.1cm} \\ \scriptsize{±2.5}\end{tabular}
\\
{bm$\rightarrow$}
& \begin{tabular}[c]{@{}c@{}} {79.9} \vspace{-0.1cm} \\ \scriptsize{±1.4}\end{tabular}
& \begin{tabular}[c]{@{}c@{}} {67.9} \vspace{-0.1cm} \\ \scriptsize{±1.7}\end{tabular}
& \begin{tabular}[c]{@{}c@{}} {31.4} \vspace{-0.1cm} \\ \scriptsize{±2.6}\end{tabular}
& \begin{tabular}[c]{@{}c@{}} {31.2} \vspace{-0.1cm} \\ \scriptsize{±3.6}\end{tabular}
& \begin{tabular}[c]{@{}c@{}} {29.7} \vspace{-0.1cm} \\ \scriptsize{±1.9}\end{tabular}
& \begin{tabular}[c]{@{}c@{}} {25.7} \vspace{-0.1cm} \\ \scriptsize{±1.9}\end{tabular}
& \begin{tabular}[c]{@{}c@{}} {34.2} \vspace{-0.1cm} \\ \scriptsize{±2.9}\end{tabular}
& \begin{tabular}[c]{@{}c@{}} {31.2} \vspace{-0.1cm} \\ \scriptsize{±1.4}\end{tabular}
& \begin{tabular}[c]{@{}c@{}} {35.3} \vspace{-0.1cm} \\ \scriptsize{±2.1}\end{tabular}
& \begin{tabular}[c]{@{}c@{}} {33.3} \vspace{-0.1cm} \\ \scriptsize{±2.3}\end{tabular}
& \begin{tabular}[c]{@{}c@{}} {31.8} \vspace{-0.1cm} \\ \scriptsize{±3.5}\end{tabular}
& \begin{tabular}[c]{@{}c@{}} {30.7} \vspace{-0.1cm} \\ \scriptsize{±0.8}\end{tabular}
\\
{ig$\rightarrow$}
& \begin{tabular}[c]{@{}c@{}} {81.8} \vspace{-0.1cm} \\ \scriptsize{±1.4}\end{tabular}
& \begin{tabular}[c]{@{}c@{}} \colorbox{mygreen}{74.0} \vspace{-0.1cm} \\ \scriptsize{±1.2}\end{tabular}
& \begin{tabular}[c]{@{}c@{}} {36.6} \vspace{-0.1cm} \\ \scriptsize{±2.3}\end{tabular}
& \begin{tabular}[c]{@{}c@{}} {38.1} \vspace{-0.1cm} \\ \scriptsize{±2.0}\end{tabular}
& \begin{tabular}[c]{@{}c@{}} {38.5} \vspace{-0.1cm} \\ \scriptsize{±5.3}\end{tabular}
& \begin{tabular}[c]{@{}c@{}} {26.8} \vspace{-0.1cm} \\ \scriptsize{±1.3}\end{tabular}
& \begin{tabular}[c]{@{}c@{}} \colorbox{myblue}{40.5} \vspace{-0.1cm} \\ \scriptsize{±1.7}\end{tabular}
& \begin{tabular}[c]{@{}c@{}} {37.1} \vspace{-0.1cm} \\ \scriptsize{±3.0}\end{tabular}
& \begin{tabular}[c]{@{}c@{}} {42.9} \vspace{-0.1cm} \\ \scriptsize{±1.8}\end{tabular}
& \begin{tabular}[c]{@{}c@{}} {43.3} \vspace{-0.1cm} \\ \scriptsize{±1.3}\end{tabular}
& \begin{tabular}[c]{@{}c@{}} {37.2} \vspace{-0.1cm} \\ \scriptsize{±1.9}\end{tabular}
& \begin{tabular}[c]{@{}c@{}} {32.2} \vspace{-0.1cm} \\ \scriptsize{±2.5}\end{tabular}
\\
{nso$\rightarrow$}
& \begin{tabular}[c]{@{}c@{}} {82.2} \vspace{-0.1cm} \\ \scriptsize{±0.6}\end{tabular}
& \begin{tabular}[c]{@{}c@{}} {72.3} \vspace{-0.1cm} \\ \scriptsize{±1.2}\end{tabular}
& \begin{tabular}[c]{@{}c@{}} {38.7} \vspace{-0.1cm} \\ \scriptsize{±0.9}\end{tabular}
& \begin{tabular}[c]{@{}c@{}} {40.1} \vspace{-0.1cm} \\ \scriptsize{±2.0}\end{tabular}
& \begin{tabular}[c]{@{}c@{}} \colorbox{mygreen}{36.3} \vspace{-0.1cm} \\ \scriptsize{±1.2}\end{tabular}
& \begin{tabular}[c]{@{}c@{}} {27.8} \vspace{-0.1cm} \\ \scriptsize{±1.4}\end{tabular}
& \begin{tabular}[c]{@{}c@{}} \colorbox{mygreen}{41.7} \vspace{-0.1cm} \\ \scriptsize{±2.9}\end{tabular}
& \begin{tabular}[c]{@{}c@{}} \colorbox{myblue}{42.5} \vspace{-0.1cm} \\ \scriptsize{±2.3}\end{tabular}
& \begin{tabular}[c]{@{}c@{}} {43.3} \vspace{-0.1cm} \\ \scriptsize{±0.6}\end{tabular}
& \begin{tabular}[c]{@{}c@{}} {40.1} \vspace{-0.1cm} \\ \scriptsize{±1.9}\end{tabular}
& \begin{tabular}[c]{@{}c@{}} {38.5} \vspace{-0.1cm} \\ \scriptsize{±1.0}\end{tabular}
& \begin{tabular}[c]{@{}c@{}} {35.5} \vspace{-0.1cm} \\ \scriptsize{±1.6}\end{tabular}
\\
{sn$\rightarrow$}
& \begin{tabular}[c]{@{}c@{}} {83.3} \vspace{-0.1cm} \\ \scriptsize{±1.5}\end{tabular}
& \begin{tabular}[c]{@{}c@{}} {72.7} \vspace{-0.1cm} \\ \scriptsize{±1.2}\end{tabular}
& \begin{tabular}[c]{@{}c@{}} {36.2} \vspace{-0.1cm} \\ \scriptsize{±2.7}\end{tabular}
& \begin{tabular}[c]{@{}c@{}} {39.0} \vspace{-0.1cm} \\ \scriptsize{±3.1}\end{tabular}
& \begin{tabular}[c]{@{}c@{}} {34.1} \vspace{-0.1cm} \\ \scriptsize{±0.8}\end{tabular}
& \begin{tabular}[c]{@{}c@{}} {30.7} \vspace{-0.1cm} \\ \scriptsize{±2.1}\end{tabular}
& \begin{tabular}[c]{@{}c@{}} \colorbox{mygreen}{38.3} \vspace{-0.1cm} \\ \scriptsize{±1.1}\end{tabular}
& \begin{tabular}[c]{@{}c@{}} {39.7} \vspace{-0.1cm} \\ \scriptsize{±1.9}\end{tabular}
& \begin{tabular}[c]{@{}c@{}} {39.4} \vspace{-0.1cm} \\ \scriptsize{±2.3}\end{tabular}
& \begin{tabular}[c]{@{}c@{}} {38.9} \vspace{-0.1cm} \\ \scriptsize{±1.2}\end{tabular}
& \begin{tabular}[c]{@{}c@{}} {40.3} \vspace{-0.1cm} \\ \scriptsize{±1.6}\end{tabular}
& \begin{tabular}[c]{@{}c@{}} {33.0} \vspace{-0.1cm} \\ \scriptsize{±3.6}\end{tabular}
\\
{st$\rightarrow$}
& \begin{tabular}[c]{@{}c@{}} {82.6} \vspace{-0.1cm} \\ \scriptsize{±1.0}\end{tabular}
& \begin{tabular}[c]{@{}c@{}} {73.7} \vspace{-0.1cm} \\ \scriptsize{±2.7}\end{tabular}
& \begin{tabular}[c]{@{}c@{}} {39.1} \vspace{-0.1cm} \\ \scriptsize{±3.9}\end{tabular}
& \begin{tabular}[c]{@{}c@{}} \colorbox{mygreen}{40.1} \vspace{-0.1cm} \\ \scriptsize{±0.5}\end{tabular}
& \begin{tabular}[c]{@{}c@{}} \colorbox{mygreen}{36.4} \vspace{-0.1cm} \\ \scriptsize{±1.1}\end{tabular}
& \begin{tabular}[c]{@{}c@{}} {31.2} \vspace{-0.1cm} \\ \scriptsize{±1.2}\end{tabular}
& \begin{tabular}[c]{@{}c@{}} \colorbox{mygreen}{39.4} \vspace{-0.1cm} \\ \scriptsize{±1.1}\end{tabular}
& \begin{tabular}[c]{@{}c@{}} \colorbox{mygreen}{42.3} \vspace{-0.1cm} \\ \scriptsize{±2.4}\end{tabular}
& \begin{tabular}[c]{@{}c@{}} {42.5} \vspace{-0.1cm} \\ \scriptsize{±0.6}\end{tabular}
& \begin{tabular}[c]{@{}c@{}} {40.5} \vspace{-0.1cm} \\ \scriptsize{±2.9}\end{tabular}
& \begin{tabular}[c]{@{}c@{}} \colorbox{mygreen}{41.8} \vspace{-0.1cm} \\ \scriptsize{±0.5}\end{tabular}
& \begin{tabular}[c]{@{}c@{}} {36.3} \vspace{-0.1cm} \\ \scriptsize{±2.0}\end{tabular}
\\
{tn$\rightarrow$}
& \begin{tabular}[c]{@{}c@{}} {83.9} \vspace{-0.1cm} \\ \scriptsize{±1.9}\end{tabular}
& \begin{tabular}[c]{@{}c@{}} \colorbox{mygreen}{75.2} \vspace{-0.1cm} \\ \scriptsize{±0.2}\end{tabular}
& \begin{tabular}[c]{@{}c@{}} {40.3} \vspace{-0.1cm} \\ \scriptsize{±2.0}\end{tabular}
& \begin{tabular}[c]{@{}c@{}} {39.5} \vspace{-0.1cm} \\ \scriptsize{±2.3}\end{tabular}
& \begin{tabular}[c]{@{}c@{}} \colorbox{mygreen}{38.5} \vspace{-0.1cm} \\ \scriptsize{±2.0}\end{tabular}
& \begin{tabular}[c]{@{}c@{}} {30.2} \vspace{-0.1cm} \\ \scriptsize{±0.6}\end{tabular}
& \begin{tabular}[c]{@{}c@{}} \colorbox{mygreen}{43.4} \vspace{-0.1cm} \\ \scriptsize{±2.6}\end{tabular}
& \begin{tabular}[c]{@{}c@{}} {40.1} \vspace{-0.1cm} \\ \scriptsize{±3.0}\end{tabular}
& \begin{tabular}[c]{@{}c@{}} {45.9} \vspace{-0.1cm} \\ \scriptsize{±1.4}\end{tabular}
& \begin{tabular}[c]{@{}c@{}} {43.0} \vspace{-0.1cm} \\ \scriptsize{±3.0}\end{tabular}
& \begin{tabular}[c]{@{}c@{}} {41.9} \vspace{-0.1cm} \\ \scriptsize{±1.9}\end{tabular}
& \begin{tabular}[c]{@{}c@{}} {38.6} \vspace{-0.1cm} \\ \scriptsize{±4.8}\end{tabular}
\\
{ts$\rightarrow$}
& \begin{tabular}[c]{@{}c@{}} {78.9} \vspace{-0.1cm} \\ \scriptsize{±1.3}\end{tabular}
& \begin{tabular}[c]{@{}c@{}} {65.9} \vspace{-0.1cm} \\ \scriptsize{±1.9}\end{tabular}
& \begin{tabular}[c]{@{}c@{}} {32.7} \vspace{-0.1cm} \\ \scriptsize{±3.3}\end{tabular}
& \begin{tabular}[c]{@{}c@{}} {34.6} \vspace{-0.1cm} \\ \scriptsize{±4.1}\end{tabular}
& \begin{tabular}[c]{@{}c@{}} {30.9} \vspace{-0.1cm} \\ \scriptsize{±2.3}\end{tabular}
& \begin{tabular}[c]{@{}c@{}} {25.9} \vspace{-0.1cm} \\ \scriptsize{±2.9}\end{tabular}
& \begin{tabular}[c]{@{}c@{}} {35.7} \vspace{-0.1cm} \\ \scriptsize{±1.1}\end{tabular}
& \begin{tabular}[c]{@{}c@{}} {36.7} \vspace{-0.1cm} \\ \scriptsize{±3.2}\end{tabular}
& \begin{tabular}[c]{@{}c@{}} {37.7} \vspace{-0.1cm} \\ \scriptsize{±2.1}\end{tabular}
& \begin{tabular}[c]{@{}c@{}} {37.2} \vspace{-0.1cm} \\ \scriptsize{±0.9}\end{tabular}
& \begin{tabular}[c]{@{}c@{}} {34.2} \vspace{-0.1cm} \\ \scriptsize{±1.4}\end{tabular}
& \begin{tabular}[c]{@{}c@{}} {33.7} \vspace{-0.1cm} \\ \scriptsize{±0.5}\end{tabular}
\\    
    \hline \hline

\end{tabular}
\caption{\textbf{Results of Cross-lingual Transfer using Llama 3 70B Instruct when Tuning on Winogrande and Testing on MMLU Clinical Knowledge}. Llama 3 70B Instruct was independently fine-tuned on the 11 African languages of focus:  Afrikaans (af), Zulu (zu),
Xhosa (xh) (datasets for these three languages were sourced from (BMGF 2024)), Amharic (am), Bambara (bm), Igbo (ig),
Sepedi (nso), Shona (sn), Sesotho (st), Setswana (tn), and Tsonga (ts). Llama 3 70B Instruct was also fine-tuned on English (en) for reference. The translated \texttt{Winogrande} training set (small) was used for fine-tuning. The fine-tuned models were evaluated for their cross-lingual transfer performance on the translated \texttt{MMLU Clinical Knowledge} (clinical domain knowledge task) test set. Columns indicate the target language of the evaluation, while the rows indicate the source language the models were fine-tuned with (e.g. ``zu$\rightarrow$" indicates models fine-tuned with data in Zulu). All numbers are the mean 5-shot performance accuracy of three evaluations followed by the standard deviation (±). Model fine-tuning that yielded mono-lingual improvements (relative to the baseline) more than two standard deviations are in \colorbox{myblue}{blue}, while cross-lingual improvements (relative to the baseline) more than two standard deviations are in \colorbox{mygreen}{green}.\\}\label{table:perf-crosslingual-llama3-tr-wino-ts-mmlu}
\end{table*}

\begin{table*}[t!]
\centering
% \small

\setlength{\tabcolsep}{5.5pt}
\begin{tabular}{r|llllllllllll}
                    \hline \hline
                     & \multicolumn{12}{c}{\texttt{MMLU Clinical Knowledge}} \\
                     % & \multicolumn{4}{c|}{\begin{tabular}[c]{@{}c@{}}\texttt{MMLU-Clinical-ZA} \\ "clinical knowledge"\end{tabular}} 
                     % & \multicolumn{4}{c}{\texttt{Belebele}} \\
                     & \textbf{en} & \textbf{af} & \textbf{zu} & \textbf{xh} 
                     & \textbf{am} & \textbf{bm} & \textbf{ig} & \textbf{nso} 
                     & \textbf{sn} & \textbf{st} & \textbf{tn} & \textbf{ts} \\ \hline 

\begin{tabular}[c]{@{}c@{}}Base \\ \tiny{Llama 3} \vspace{-0.1cm}\\ \tiny{70B IT} \end{tabular} 
& 82.3 & 71.3 & 39.2 & 38.5 & 33.6 & 32.5 & 33.6 & 37.0 & 43.4 & 43.8 & 39.2 & 34.0
\\ \hline
\multicolumn{12}{l}{\begin{tabular}[c]{@{}c@{}}\texttt{MMLU College Medicine} \\ dev + test + val\end{tabular}} \\
{en$\rightarrow$}
& \begin{tabular}[c]{@{}c@{}} \colorbox{myblue}{84.9} \vspace{-0.1cm} \\ \scriptsize{±1.0}\end{tabular}
& \begin{tabular}[c]{@{}c@{}} \colorbox{mygreen}{75.1} \vspace{-0.1cm} \\ \scriptsize{±1.6}\end{tabular}
& \begin{tabular}[c]{@{}c@{}} {39.3} \vspace{-0.1cm} \\ \scriptsize{±2.8}\end{tabular}
& \begin{tabular}[c]{@{}c@{}} {38.6} \vspace{-0.1cm} \\ \scriptsize{±1.7}\end{tabular}
& \begin{tabular}[c]{@{}c@{}} \colorbox{mygreen}{36.8} \vspace{-0.1cm} \\ \scriptsize{±1.1}\end{tabular}
& \begin{tabular}[c]{@{}c@{}} {33.6} \vspace{-0.1cm} \\ \scriptsize{±3.0}\end{tabular}
& \begin{tabular}[c]{@{}c@{}} \colorbox{mygreen}{40.7} \vspace{-0.1cm} \\ \scriptsize{±2.5}\end{tabular}
& \begin{tabular}[c]{@{}c@{}} {39.3} \vspace{-0.1cm} \\ \scriptsize{±1.7}\end{tabular}
& \begin{tabular}[c]{@{}c@{}} {44.5} \vspace{-0.1cm} \\ \scriptsize{±1.3}\end{tabular}
& \begin{tabular}[c]{@{}c@{}} {43.8} \vspace{-0.1cm} \\ \scriptsize{±1.0}\end{tabular}
& \begin{tabular}[c]{@{}c@{}} {41.8} \vspace{-0.1cm} \\ \scriptsize{±1.9}\end{tabular}
& \begin{tabular}[c]{@{}c@{}} \colorbox{mygreen}{38.4} \vspace{-0.1cm} \\ \scriptsize{±0.5}\end{tabular}
\\
{af$\rightarrow$}
& \begin{tabular}[c]{@{}c@{}} {85.7} \vspace{-0.1cm} \\ \scriptsize{±2.0}\end{tabular}
& \begin{tabular}[c]{@{}c@{}} \colorbox{myblue}{80.9} \vspace{-0.1cm} \\ \scriptsize{±1.7}\end{tabular}
& \begin{tabular}[c]{@{}c@{}} \colorbox{mygreen}{47.1} \vspace{-0.1cm} \\ \scriptsize{±2.6}\end{tabular}
& \begin{tabular}[c]{@{}c@{}} \colorbox{mygreen}{47.4} \vspace{-0.1cm} \\ \scriptsize{±2.2}\end{tabular}
& \begin{tabular}[c]{@{}c@{}} \colorbox{mygreen}{43.5} \vspace{-0.1cm} \\ \scriptsize{±2.0}\end{tabular}
& \begin{tabular}[c]{@{}c@{}} {34.5} \vspace{-0.1cm} \\ \scriptsize{±1.9}\end{tabular}
& \begin{tabular}[c]{@{}c@{}} \colorbox{mygreen}{47.7} \vspace{-0.1cm} \\ \scriptsize{±1.4}\end{tabular}
& \begin{tabular}[c]{@{}c@{}} \colorbox{mygreen}{45.8} \vspace{-0.1cm} \\ \scriptsize{±1.7}\end{tabular}
& \begin{tabular}[c]{@{}c@{}} \colorbox{mygreen}{49.2} \vspace{-0.1cm} \\ \scriptsize{±2.1}\end{tabular}
& \begin{tabular}[c]{@{}c@{}} \colorbox{mygreen}{48.8} \vspace{-0.1cm} \\ \scriptsize{±1.7}\end{tabular}
& \begin{tabular}[c]{@{}c@{}} \colorbox{mygreen}{50.1} \vspace{-0.1cm} \\ \scriptsize{±2.9}\end{tabular}
& \begin{tabular}[c]{@{}c@{}} {38.6} \vspace{-0.1cm} \\ \scriptsize{±4.4}\end{tabular}
\\
{zu$\rightarrow$}
& \begin{tabular}[c]{@{}c@{}} {83.3} \vspace{-0.1cm} \\ \scriptsize{±0.5}\end{tabular}
& \begin{tabular}[c]{@{}c@{}} \colorbox{mygreen}{78.0} \vspace{-0.1cm} \\ \scriptsize{±1.9}\end{tabular}
& \begin{tabular}[c]{@{}c@{}} \colorbox{myblue}{53.2} \vspace{-0.1cm} \\ \scriptsize{±2.5}\end{tabular}
& \begin{tabular}[c]{@{}c@{}} \colorbox{mygreen}{51.8} \vspace{-0.1cm} \\ \scriptsize{±3.5}\end{tabular}
& \begin{tabular}[c]{@{}c@{}} \colorbox{mygreen}{45.9} \vspace{-0.1cm} \\ \scriptsize{±4.6}\end{tabular}
& \begin{tabular}[c]{@{}c@{}} \colorbox{mygreen}{36.3} \vspace{-0.1cm} \\ \scriptsize{±1.7}\end{tabular}
& \begin{tabular}[c]{@{}c@{}} \colorbox{mygreen}{49.4} \vspace{-0.1cm} \\ \scriptsize{±1.2}\end{tabular}
& \begin{tabular}[c]{@{}c@{}} \colorbox{mygreen}{49.6} \vspace{-0.1cm} \\ \scriptsize{±1.9}\end{tabular}
& \begin{tabular}[c]{@{}c@{}} \colorbox{mygreen}{49.8} \vspace{-0.1cm} \\ \scriptsize{±0.8}\end{tabular}
& \begin{tabular}[c]{@{}c@{}} \colorbox{mygreen}{52.6} \vspace{-0.1cm} \\ \scriptsize{±0.8}\end{tabular}
& \begin{tabular}[c]{@{}c@{}} \colorbox{mygreen}{49.3} \vspace{-0.1cm} \\ \scriptsize{±2.6}\end{tabular}
& \begin{tabular}[c]{@{}c@{}} \colorbox{mygreen}{44.3} \vspace{-0.1cm} \\ \scriptsize{±1.4}\end{tabular}
\\
{xh$\rightarrow$}
& \begin{tabular}[c]{@{}c@{}} {82.3} \vspace{-0.1cm} \\ \scriptsize{±1.4}\end{tabular}
& \begin{tabular}[c]{@{}c@{}} {75.7} \vspace{-0.1cm} \\ \scriptsize{±2.7}\end{tabular}
& \begin{tabular}[c]{@{}c@{}} \colorbox{mygreen}{49.6} \vspace{-0.1cm} \\ \scriptsize{±2.5}\end{tabular}
& \begin{tabular}[c]{@{}c@{}} \colorbox{myblue}{53.7} \vspace{-0.1cm} \\ \scriptsize{±0.8}\end{tabular}
& \begin{tabular}[c]{@{}c@{}} \colorbox{mygreen}{43.7} \vspace{-0.1cm} \\ \scriptsize{±1.7}\end{tabular}
& \begin{tabular}[c]{@{}c@{}} \colorbox{mygreen}{35.4} \vspace{-0.1cm} \\ \scriptsize{±1.0}\end{tabular}
& \begin{tabular}[c]{@{}c@{}} \colorbox{mygreen}{50.7} \vspace{-0.1cm} \\ \scriptsize{±3.4}\end{tabular}
& \begin{tabular}[c]{@{}c@{}} \colorbox{mygreen}{52.2} \vspace{-0.1cm} \\ \scriptsize{±1.6}\end{tabular}
& \begin{tabular}[c]{@{}c@{}} \colorbox{mygreen}{51.6} \vspace{-0.1cm} \\ \scriptsize{±1.7}\end{tabular}
& \begin{tabular}[c]{@{}c@{}} \colorbox{mygreen}{52.3} \vspace{-0.1cm} \\ \scriptsize{±1.2}\end{tabular}
& \begin{tabular}[c]{@{}c@{}} \colorbox{mygreen}{50.7} \vspace{-0.1cm} \\ \scriptsize{±3.7}\end{tabular}
& \begin{tabular}[c]{@{}c@{}} \colorbox{mygreen}{44.0} \vspace{-0.1cm} \\ \scriptsize{±1.6}\end{tabular}
\\
{am$\rightarrow$}
& \begin{tabular}[c]{@{}c@{}} {82.7} \vspace{-0.1cm} \\ \scriptsize{±0.2}\end{tabular}
& \begin{tabular}[c]{@{}c@{}} \colorbox{mygreen}{75.8} \vspace{-0.1cm} \\ \scriptsize{±1.6}\end{tabular}
& \begin{tabular}[c]{@{}c@{}} \colorbox{mygreen}{48.1} \vspace{-0.1cm} \\ \scriptsize{±2.5}\end{tabular}
& \begin{tabular}[c]{@{}c@{}} \colorbox{mygreen}{47.2} \vspace{-0.1cm} \\ \scriptsize{±3.1}\end{tabular}
& \begin{tabular}[c]{@{}c@{}} \colorbox{myblue}{59.2} \vspace{-0.1cm} \\ \scriptsize{±2.3}\end{tabular}
& \begin{tabular}[c]{@{}c@{}} \colorbox{mygreen}{40.1} \vspace{-0.1cm} \\ \scriptsize{±0.8}\end{tabular}
& \begin{tabular}[c]{@{}c@{}} \colorbox{mygreen}{47.3} \vspace{-0.1cm} \\ \scriptsize{±0.4}\end{tabular}
& \begin{tabular}[c]{@{}c@{}} \colorbox{mygreen}{52.5} \vspace{-0.1cm} \\ \scriptsize{±2.1}\end{tabular}
& \begin{tabular}[c]{@{}c@{}} \colorbox{mygreen}{49.6} \vspace{-0.1cm} \\ \scriptsize{±1.2}\end{tabular}
& \begin{tabular}[c]{@{}c@{}} \colorbox{mygreen}{51.6} \vspace{-0.1cm} \\ \scriptsize{±2.3}\end{tabular}
& \begin{tabular}[c]{@{}c@{}} \colorbox{mygreen}{48.2} \vspace{-0.1cm} \\ \scriptsize{±1.2}\end{tabular}
& \begin{tabular}[c]{@{}c@{}} \colorbox{mygreen}{41.5} \vspace{-0.1cm} \\ \scriptsize{±0.6}\end{tabular}
\\
{bm$\rightarrow$}
& \begin{tabular}[c]{@{}c@{}} {81.7} \vspace{-0.1cm} \\ \scriptsize{±1.1}\end{tabular}
& \begin{tabular}[c]{@{}c@{}} {72.3} \vspace{-0.1cm} \\ \scriptsize{±1.0}\end{tabular}
& \begin{tabular}[c]{@{}c@{}} {42.1} \vspace{-0.1cm} \\ \scriptsize{±1.9}\end{tabular}
& \begin{tabular}[c]{@{}c@{}} {42.0} \vspace{-0.1cm} \\ \scriptsize{±2.6}\end{tabular}
& \begin{tabular}[c]{@{}c@{}} \colorbox{mygreen}{39.9} \vspace{-0.1cm} \\ \scriptsize{±2.0}\end{tabular}
& \begin{tabular}[c]{@{}c@{}} \colorbox{myblue}{41.6} \vspace{-0.1cm} \\ \scriptsize{±3.2}\end{tabular}
& \begin{tabular}[c]{@{}c@{}} \colorbox{mygreen}{44.8} \vspace{-0.1cm} \\ \scriptsize{±1.0}\end{tabular}
& \begin{tabular}[c]{@{}c@{}} \colorbox{mygreen}{43.7} \vspace{-0.1cm} \\ \scriptsize{±2.1}\end{tabular}
& \begin{tabular}[c]{@{}c@{}} \colorbox{mygreen}{47.5} \vspace{-0.1cm} \\ \scriptsize{±0.6}\end{tabular}
& \begin{tabular}[c]{@{}c@{}} {45.8} \vspace{-0.1cm} \\ \scriptsize{±2.1}\end{tabular}
& \begin{tabular}[c]{@{}c@{}} {43.8} \vspace{-0.1cm} \\ \scriptsize{±4.0}\end{tabular}
& \begin{tabular}[c]{@{}c@{}} {41.9} \vspace{-0.1cm} \\ \scriptsize{±4.0}\end{tabular}
\\
{ig$\rightarrow$}
& \begin{tabular}[c]{@{}c@{}} {84.5} \vspace{-0.1cm} \\ \scriptsize{±1.3}\end{tabular}
& \begin{tabular}[c]{@{}c@{}} \colorbox{mygreen}{75.8} \vspace{-0.1cm} \\ \scriptsize{±1.2}\end{tabular}
& \begin{tabular}[c]{@{}c@{}} \colorbox{mygreen}{44.5} \vspace{-0.1cm} \\ \scriptsize{±0.4}\end{tabular}
& \begin{tabular}[c]{@{}c@{}} \colorbox{mygreen}{45.9} \vspace{-0.1cm} \\ \scriptsize{±0.8}\end{tabular}
& \begin{tabular}[c]{@{}c@{}} \colorbox{mygreen}{42.1} \vspace{-0.1cm} \\ \scriptsize{±2.1}\end{tabular}
& \begin{tabular}[c]{@{}c@{}} {35.7} \vspace{-0.1cm} \\ \scriptsize{±2.4}\end{tabular}
& \begin{tabular}[c]{@{}c@{}} \colorbox{myblue}{56.0} \vspace{-0.1cm} \\ \scriptsize{±4.2}\end{tabular}
& \begin{tabular}[c]{@{}c@{}} \colorbox{mygreen}{49.2} \vspace{-0.1cm} \\ \scriptsize{±2.5}\end{tabular}
& \begin{tabular}[c]{@{}c@{}} \colorbox{mygreen}{49.3} \vspace{-0.1cm} \\ \scriptsize{±0.9}\end{tabular}
& \begin{tabular}[c]{@{}c@{}} \colorbox{mygreen}{50.9} \vspace{-0.1cm} \\ \scriptsize{±1.4}\end{tabular}
& \begin{tabular}[c]{@{}c@{}} \colorbox{mygreen}{50.6} \vspace{-0.1cm} \\ \scriptsize{±1.4}\end{tabular}
& \begin{tabular}[c]{@{}c@{}} \colorbox{mygreen}{40.8} \vspace{-0.1cm} \\ \scriptsize{±1.2}\end{tabular}
\\
{nso$\rightarrow$}
& \begin{tabular}[c]{@{}c@{}} {83.1} \vspace{-0.1cm} \\ \scriptsize{±2.1}\end{tabular}
& \begin{tabular}[c]{@{}c@{}} \colorbox{mygreen}{76.1} \vspace{-0.1cm} \\ \scriptsize{±0.9}\end{tabular}
& \begin{tabular}[c]{@{}c@{}} \colorbox{mygreen}{48.5} \vspace{-0.1cm} \\ \scriptsize{±1.1}\end{tabular}
& \begin{tabular}[c]{@{}c@{}} \colorbox{mygreen}{51.7} \vspace{-0.1cm} \\ \scriptsize{±2.3}\end{tabular}
& \begin{tabular}[c]{@{}c@{}} \colorbox{mygreen}{46.9} \vspace{-0.1cm} \\ \scriptsize{±2.0}\end{tabular}
& \begin{tabular}[c]{@{}c@{}} {37.4} \vspace{-0.1cm} \\ \scriptsize{±3.1}\end{tabular}
& \begin{tabular}[c]{@{}c@{}} \colorbox{mygreen}{49.4} \vspace{-0.1cm} \\ \scriptsize{±3.0}\end{tabular}
& \begin{tabular}[c]{@{}c@{}} \colorbox{myblue}{58.6} \vspace{-0.1cm} \\ \scriptsize{±1.1}\end{tabular}
& \begin{tabular}[c]{@{}c@{}} \colorbox{mygreen}{52.2} \vspace{-0.1cm} \\ \scriptsize{±1.0}\end{tabular}
& \begin{tabular}[c]{@{}c@{}} \colorbox{mygreen}{54.7} \vspace{-0.1cm} \\ \scriptsize{±1.1}\end{tabular}
& \begin{tabular}[c]{@{}c@{}} \colorbox{mygreen}{53.2} \vspace{-0.1cm} \\ \scriptsize{±1.0}\end{tabular}
& \begin{tabular}[c]{@{}c@{}} \colorbox{mygreen}{46.2} \vspace{-0.1cm} \\ \scriptsize{±2.7}\end{tabular}
\\
{sn$\rightarrow$}
& \begin{tabular}[c]{@{}c@{}} {83.5} \vspace{-0.1cm} \\ \scriptsize{±1.0}\end{tabular}
& \begin{tabular}[c]{@{}c@{}} \colorbox{mygreen}{76.7} \vspace{-0.1cm} \\ \scriptsize{±0.8}\end{tabular}
& \begin{tabular}[c]{@{}c@{}} \colorbox{mygreen}{49.8} \vspace{-0.1cm} \\ \scriptsize{±3.3}\end{tabular}
& \begin{tabular}[c]{@{}c@{}} \colorbox{mygreen}{47.8} \vspace{-0.1cm} \\ \scriptsize{±1.2}\end{tabular}
& \begin{tabular}[c]{@{}c@{}} \colorbox{mygreen}{47.3} \vspace{-0.1cm} \\ \scriptsize{±0.9}\end{tabular}
& \begin{tabular}[c]{@{}c@{}} {38.8} \vspace{-0.1cm} \\ \scriptsize{±3.5}\end{tabular}
& \begin{tabular}[c]{@{}c@{}} \colorbox{mygreen}{50.3} \vspace{-0.1cm} \\ \scriptsize{±2.3}\end{tabular}
& \begin{tabular}[c]{@{}c@{}} \colorbox{mygreen}{54.5} \vspace{-0.1cm} \\ \scriptsize{±1.0}\end{tabular}
& \begin{tabular}[c]{@{}c@{}} \colorbox{myblue}{62.3} \vspace{-0.1cm} \\ \scriptsize{±0.4}\end{tabular}
& \begin{tabular}[c]{@{}c@{}} \colorbox{mygreen}{54.7} \vspace{-0.1cm} \\ \scriptsize{±3.9}\end{tabular}
& \begin{tabular}[c]{@{}c@{}} \colorbox{mygreen}{53.6} \vspace{-0.1cm} \\ \scriptsize{±2.3}\end{tabular}
& \begin{tabular}[c]{@{}c@{}} \colorbox{mygreen}{47.4} \vspace{-0.1cm} \\ \scriptsize{±1.6}\end{tabular}
\\
{st$\rightarrow$}
& \begin{tabular}[c]{@{}c@{}} \colorbox{mygreen}{83.7} \vspace{-0.1cm} \\ \scriptsize{±0.6}\end{tabular}
& \begin{tabular}[c]{@{}c@{}} \colorbox{mygreen}{77.0} \vspace{-0.1cm} \\ \scriptsize{±1.0}\end{tabular}
& \begin{tabular}[c]{@{}c@{}} \colorbox{mygreen}{50.3} \vspace{-0.1cm} \\ \scriptsize{±3.1}\end{tabular}
& \begin{tabular}[c]{@{}c@{}} \colorbox{mygreen}{51.7} \vspace{-0.1cm} \\ \scriptsize{±0.4}\end{tabular}
& \begin{tabular}[c]{@{}c@{}} \colorbox{mygreen}{47.3} \vspace{-0.1cm} \\ \scriptsize{±1.9}\end{tabular}
& \begin{tabular}[c]{@{}c@{}} \colorbox{mygreen}{38.9} \vspace{-0.1cm} \\ \scriptsize{±1.7}\end{tabular}
& \begin{tabular}[c]{@{}c@{}} \colorbox{mygreen}{48.8} \vspace{-0.1cm} \\ \scriptsize{±2.1}\end{tabular}
& \begin{tabular}[c]{@{}c@{}} \colorbox{mygreen}{54.3} \vspace{-0.1cm} \\ \scriptsize{±3.1}\end{tabular}
& \begin{tabular}[c]{@{}c@{}} \colorbox{mygreen}{53.7} \vspace{-0.1cm} \\ \scriptsize{±1.8}\end{tabular}
& \begin{tabular}[c]{@{}c@{}} \colorbox{myblue}{59.2} \vspace{-0.1cm} \\ \scriptsize{±1.0}\end{tabular}
& \begin{tabular}[c]{@{}c@{}} \colorbox{mygreen}{56.2} \vspace{-0.1cm} \\ \scriptsize{±0.7}\end{tabular}
& \begin{tabular}[c]{@{}c@{}} \colorbox{mygreen}{45.5} \vspace{-0.1cm} \\ \scriptsize{±1.5}\end{tabular}
\\
{tn$\rightarrow$}
& \begin{tabular}[c]{@{}c@{}} {84.0} \vspace{-0.1cm} \\ \scriptsize{±1.4}\end{tabular}
& \begin{tabular}[c]{@{}c@{}} \colorbox{mygreen}{76.7} \vspace{-0.1cm} \\ \scriptsize{±0.2}\end{tabular}
& \begin{tabular}[c]{@{}c@{}} \colorbox{mygreen}{49.8} \vspace{-0.1cm} \\ \scriptsize{±2.5}\end{tabular}
& \begin{tabular}[c]{@{}c@{}} \colorbox{mygreen}{50.1} \vspace{-0.1cm} \\ \scriptsize{±2.3}\end{tabular}
& \begin{tabular}[c]{@{}c@{}} \colorbox{mygreen}{47.1} \vspace{-0.1cm} \\ \scriptsize{±0.2}\end{tabular}
& \begin{tabular}[c]{@{}c@{}} \colorbox{mygreen}{39.0} \vspace{-0.1cm} \\ \scriptsize{±1.7}\end{tabular}
& \begin{tabular}[c]{@{}c@{}} \colorbox{mygreen}{50.6} \vspace{-0.1cm} \\ \scriptsize{±0.8}\end{tabular}
& \begin{tabular}[c]{@{}c@{}} \colorbox{mygreen}{58.1} \vspace{-0.1cm} \\ \scriptsize{±0.6}\end{tabular}
& \begin{tabular}[c]{@{}c@{}} \colorbox{mygreen}{52.3} \vspace{-0.1cm} \\ \scriptsize{±2.5}\end{tabular}
& \begin{tabular}[c]{@{}c@{}} \colorbox{mygreen}{55.9} \vspace{-0.1cm} \\ \scriptsize{±1.6}\end{tabular}
& \begin{tabular}[c]{@{}c@{}} \colorbox{myblue}{59.6} \vspace{-0.1cm} \\ \scriptsize{±0.6}\end{tabular}
& \begin{tabular}[c]{@{}c@{}} \colorbox{mygreen}{46.5} \vspace{-0.1cm} \\ \scriptsize{±1.6}\end{tabular}
\\
{ts$\rightarrow$}
& \begin{tabular}[c]{@{}c@{}} {82.5} \vspace{-0.1cm} \\ \scriptsize{±0.2}\end{tabular}
& \begin{tabular}[c]{@{}c@{}} \colorbox{mygreen}{77.8} \vspace{-0.1cm} \\ \scriptsize{±1.0}\end{tabular}
& \begin{tabular}[c]{@{}c@{}} \colorbox{mygreen}{50.3} \vspace{-0.1cm} \\ \scriptsize{±2.0}\end{tabular}
& \begin{tabular}[c]{@{}c@{}} \colorbox{mygreen}{49.2} \vspace{-0.1cm} \\ \scriptsize{±0.6}\end{tabular}
& \begin{tabular}[c]{@{}c@{}} \colorbox{mygreen}{47.8} \vspace{-0.1cm} \\ \scriptsize{±3.4}\end{tabular}
& \begin{tabular}[c]{@{}c@{}} \colorbox{mygreen}{36.2} \vspace{-0.1cm} \\ \scriptsize{±1.7}\end{tabular}
& \begin{tabular}[c]{@{}c@{}} \colorbox{mygreen}{51.3} \vspace{-0.1cm} \\ \scriptsize{±1.2}\end{tabular}
& \begin{tabular}[c]{@{}c@{}} \colorbox{mygreen}{52.2} \vspace{-0.1cm} \\ \scriptsize{±2.5}\end{tabular}
& \begin{tabular}[c]{@{}c@{}} \colorbox{mygreen}{50.9} \vspace{-0.1cm} \\ \scriptsize{±0.8}\end{tabular}
& \begin{tabular}[c]{@{}c@{}} \colorbox{mygreen}{54.6} \vspace{-0.1cm} \\ \scriptsize{±1.3}\end{tabular}
& \begin{tabular}[c]{@{}c@{}} \colorbox{mygreen}{49.0} \vspace{-0.1cm} \\ \scriptsize{±2.0}\end{tabular}
& \begin{tabular}[c]{@{}c@{}} \colorbox{myblue}{53.5} \vspace{-0.1cm} \\ \scriptsize{±1.7}\end{tabular}
\\    
    \hline \hline

\end{tabular}
\caption{\textbf{Results of Cross-lingual Transfer using Llama 3 70B Instruct when Tuning on MMLU College Medicine and Testing on MMLU Clinical Knowledge}. Llama 3 70B Instruct was independently fine-tuned on the 11 African languages of focus:  Afrikaans (af), Zulu (zu),
Xhosa (xh) (datasets for these three languages were sourced from (BMGF 2024)), Amharic (am), Bambara (bm), Igbo (ig),
Sepedi (nso), Shona (sn), Sesotho (st), Setswana (tn), and Tsonga (ts). Llama 3 70B Instruct was also fine-tuned on English (en) for reference. The translated \texttt{MMLU College Medicine} section (dev + test + val sets) was used for fine-tuning. The fine-tuned models were evaluated for their cross-lingual transfer performance on the translated \texttt{MMLU Clinical Knowledge} (clinical domain knowledge task) test set. Columns indicate the target language of the evaluation, while the rows indicate the source language the models were fine-tuned with (e.g. ``zu$\rightarrow$" indicates models fine-tuned with data in Zulu). All numbers are the mean 5-shot performance accuracy of three evaluations followed by the standard deviation (±). Model fine-tuning that yielded mono-lingual improvements (relative to the baseline) more than two standard deviations are in \colorbox{myblue}{blue}, while cross-lingual improvements (relative to the baseline) more than two standard deviations are in \colorbox{mygreen}{green}.\\} \label{table:perf-crosslingual-llama3-tr-mmlu-ts-mmlu-ck}
\end{table*}

\begin{table*}[t!]
\centering
% \small

\setlength{\tabcolsep}{5.5pt}
\begin{tabular}{r|llllllllllll}
                    \hline \hline
                     & \multicolumn{12}{c}{\texttt{MMLU Virology}} \\
                     % & \multicolumn{4}{c|}{\begin{tabular}[c]{@{}c@{}}\texttt{MMLU-Clinical-ZA} \\ "clinical knowledge"\end{tabular}} 
                     % & \multicolumn{4}{c}{\texttt{Belebele}} \\
                     & \textbf{en} & \textbf{af} & \textbf{zu} & \textbf{xh} 
                     & \textbf{am} & \textbf{bm} & \textbf{ig} & \textbf{nso} 
                     & \textbf{sn} & \textbf{st} & \textbf{tn} & \textbf{ts} \\ \hline 

\begin{tabular}[c]{@{}c@{}}Base \\ \tiny{Llama 3} \vspace{-0.1cm}\\ \tiny{70B IT} \end{tabular} 
& 53.6 & 46.4 & 28.9 & 27.1 & 36.1 & 35.5 & 34.9 & 26.5 & 31.3 & 29.5 & 28.3 & 31.3
\\ \hline
\multicolumn{12}{l}{\begin{tabular}[c]{@{}c@{}}\texttt{Winogrande} \\ train (small)\end{tabular}}  \\
{en$\rightarrow$}
& \begin{tabular}[c]{@{}c@{}} {56.6} \vspace{-0.1cm} \\ \scriptsize{±2.1}\end{tabular}
& \begin{tabular}[c]{@{}c@{}} {46.6} \vspace{-0.1cm} \\ \scriptsize{±2.4}\end{tabular}
& \begin{tabular}[c]{@{}c@{}} {34.5} \vspace{-0.1cm} \\ \scriptsize{±3.4}\end{tabular}
& \begin{tabular}[c]{@{}c@{}} {31.1} \vspace{-0.1cm} \\ \scriptsize{±3.3}\end{tabular}
& \begin{tabular}[c]{@{}c@{}} {30.9} \vspace{-0.1cm} \\ \scriptsize{±3.1}\end{tabular}
& \begin{tabular}[c]{@{}c@{}} {32.7} \vspace{-0.1cm} \\ \scriptsize{±5.2}\end{tabular}
& \begin{tabular}[c]{@{}c@{}} {31.3} \vspace{-0.1cm} \\ \scriptsize{±1.0}\end{tabular}
& \begin{tabular}[c]{@{}c@{}} {30.1} \vspace{-0.1cm} \\ \scriptsize{±2.1}\end{tabular}
& \begin{tabular}[c]{@{}c@{}} {31.7} \vspace{-0.1cm} \\ \scriptsize{±1.9}\end{tabular}
& \begin{tabular}[c]{@{}c@{}} {31.9} \vspace{-0.1cm} \\ \scriptsize{±3.3}\end{tabular}
& \begin{tabular}[c]{@{}c@{}} {30.7} \vspace{-0.1cm} \\ \scriptsize{±2.1}\end{tabular}
& \begin{tabular}[c]{@{}c@{}} {32.7} \vspace{-0.1cm} \\ \scriptsize{±7.0}\end{tabular}
\\
{af$\rightarrow$}
& \begin{tabular}[c]{@{}c@{}} \colorbox{mygreen}{55.4} \vspace{-0.1cm} \\ \scriptsize{±0.6}\end{tabular}
& \begin{tabular}[c]{@{}c@{}} {49.8} \vspace{-0.1cm} \\ \scriptsize{±1.9}\end{tabular}
& \begin{tabular}[c]{@{}c@{}} \colorbox{mygreen}{32.9} \vspace{-0.1cm} \\ \scriptsize{±1.2}\end{tabular}
& \begin{tabular}[c]{@{}c@{}} {32.9} \vspace{-0.1cm} \\ \scriptsize{±3.9}\end{tabular}
& \begin{tabular}[c]{@{}c@{}} {37.7} \vspace{-0.1cm} \\ \scriptsize{±2.3}\end{tabular}
& \begin{tabular}[c]{@{}c@{}} {35.3} \vspace{-0.1cm} \\ \scriptsize{±0.9}\end{tabular}
& \begin{tabular}[c]{@{}c@{}} {36.5} \vspace{-0.1cm} \\ \scriptsize{±3.1}\end{tabular}
& \begin{tabular}[c]{@{}c@{}} \colorbox{mygreen}{30.3} \vspace{-0.1cm} \\ \scriptsize{±1.5}\end{tabular}
& \begin{tabular}[c]{@{}c@{}} \colorbox{mygreen}{37.1} \vspace{-0.1cm} \\ \scriptsize{±1.9}\end{tabular}
& \begin{tabular}[c]{@{}c@{}} \colorbox{mygreen}{34.7} \vspace{-0.1cm} \\ \scriptsize{±1.8}\end{tabular}
& \begin{tabular}[c]{@{}c@{}} {33.9} \vspace{-0.1cm} \\ \scriptsize{±3.0}\end{tabular}
& \begin{tabular}[c]{@{}c@{}} {30.9} \vspace{-0.1cm} \\ \scriptsize{±1.2}\end{tabular}
\\
{zu$\rightarrow$}
& \begin{tabular}[c]{@{}c@{}} {52.8} \vspace{-0.1cm} \\ \scriptsize{±3.0}\end{tabular}
& \begin{tabular}[c]{@{}c@{}} {48.0} \vspace{-0.1cm} \\ \scriptsize{±1.5}\end{tabular}
& \begin{tabular}[c]{@{}c@{}} \colorbox{myblue}{39.0} \vspace{-0.1cm} \\ \scriptsize{±3.1}\end{tabular}
& \begin{tabular}[c]{@{}c@{}} \colorbox{mygreen}{33.3} \vspace{-0.1cm} \\ \scriptsize{±1.9}\end{tabular}
& \begin{tabular}[c]{@{}c@{}} {34.7} \vspace{-0.1cm} \\ \scriptsize{±5.1}\end{tabular}
& \begin{tabular}[c]{@{}c@{}} {33.5} \vspace{-0.1cm} \\ \scriptsize{±3.3}\end{tabular}
& \begin{tabular}[c]{@{}c@{}} {35.1} \vspace{-0.1cm} \\ \scriptsize{±0.9}\end{tabular}
& \begin{tabular}[c]{@{}c@{}} {30.7} \vspace{-0.1cm} \\ \scriptsize{±6.4}\end{tabular}
& \begin{tabular}[c]{@{}c@{}} {35.3} \vspace{-0.1cm} \\ \scriptsize{±5.3}\end{tabular}
& \begin{tabular}[c]{@{}c@{}} {33.7} \vspace{-0.1cm} \\ \scriptsize{±2.7}\end{tabular}
& \begin{tabular}[c]{@{}c@{}} {31.7} \vspace{-0.1cm} \\ \scriptsize{±2.7}\end{tabular}
& \begin{tabular}[c]{@{}c@{}} {31.1} \vspace{-0.1cm} \\ \scriptsize{±2.3}\end{tabular}
\\
{xh$\rightarrow$}
& \begin{tabular}[c]{@{}c@{}} {55.4} \vspace{-0.1cm} \\ \scriptsize{±1.6}\end{tabular}
& \begin{tabular}[c]{@{}c@{}} {49.6} \vspace{-0.1cm} \\ \scriptsize{±2.7}\end{tabular}
& \begin{tabular}[c]{@{}c@{}} \colorbox{mygreen}{38.5} \vspace{-0.1cm} \\ \scriptsize{±3.2}\end{tabular}
& \begin{tabular}[c]{@{}c@{}} \colorbox{myblue}{35.3} \vspace{-0.1cm} \\ \scriptsize{±2.3}\end{tabular}
& \begin{tabular}[c]{@{}c@{}} {35.3} \vspace{-0.1cm} \\ \scriptsize{±4.3}\end{tabular}
& \begin{tabular}[c]{@{}c@{}} {27.9} \vspace{-0.1cm} \\ \scriptsize{±1.2}\end{tabular}
& \begin{tabular}[c]{@{}c@{}} {35.9} \vspace{-0.1cm} \\ \scriptsize{±2.8}\end{tabular}
& \begin{tabular}[c]{@{}c@{}} \colorbox{mygreen}{32.7} \vspace{-0.1cm} \\ \scriptsize{±1.5}\end{tabular}
& \begin{tabular}[c]{@{}c@{}} {32.3} \vspace{-0.1cm} \\ \scriptsize{±3.0}\end{tabular}
& \begin{tabular}[c]{@{}c@{}} {32.7} \vspace{-0.1cm} \\ \scriptsize{±1.9}\end{tabular}
& \begin{tabular}[c]{@{}c@{}} {32.1} \vspace{-0.1cm} \\ \scriptsize{±2.5}\end{tabular}
& \begin{tabular}[c]{@{}c@{}} {31.3} \vspace{-0.1cm} \\ \scriptsize{±1.6}\end{tabular}
\\
{am$\rightarrow$}
& \begin{tabular}[c]{@{}c@{}} {52.8} \vspace{-0.1cm} \\ \scriptsize{±0.7}\end{tabular}
& \begin{tabular}[c]{@{}c@{}} \colorbox{mygreen}{51.4} \vspace{-0.1cm} \\ \scriptsize{±0.9}\end{tabular}
& \begin{tabular}[c]{@{}c@{}} \colorbox{mygreen}{37.5} \vspace{-0.1cm} \\ \scriptsize{±4.1}\end{tabular}
& \begin{tabular}[c]{@{}c@{}} \colorbox{mygreen}{34.7} \vspace{-0.1cm} \\ \scriptsize{±3.1}\end{tabular}
& \begin{tabular}[c]{@{}c@{}} {38.4} \vspace{-0.1cm} \\ \scriptsize{±2.6}\end{tabular}
& \begin{tabular}[c]{@{}c@{}} {34.1} \vspace{-0.1cm} \\ \scriptsize{±1.5}\end{tabular}
& \begin{tabular}[c]{@{}c@{}} {36.9} \vspace{-0.1cm} \\ \scriptsize{±2.2}\end{tabular}
& \begin{tabular}[c]{@{}c@{}} {32.7} \vspace{-0.1cm} \\ \scriptsize{±3.3}\end{tabular}
& \begin{tabular}[c]{@{}c@{}} {34.3} \vspace{-0.1cm} \\ \scriptsize{±2.2}\end{tabular}
& \begin{tabular}[c]{@{}c@{}} {30.7} \vspace{-0.1cm} \\ \scriptsize{±2.2}\end{tabular}
& \begin{tabular}[c]{@{}c@{}} \colorbox{mygreen}{32.1} \vspace{-0.1cm} \\ \scriptsize{±0.9}\end{tabular}
& \begin{tabular}[c]{@{}c@{}} {28.3} \vspace{-0.1cm} \\ \scriptsize{±3.7}\end{tabular}
\\
{bm$\rightarrow$}
& \begin{tabular}[c]{@{}c@{}} {52.0} \vspace{-0.1cm} \\ \scriptsize{±2.7}\end{tabular}
& \begin{tabular}[c]{@{}c@{}} \colorbox{mygreen}{49.0} \vspace{-0.1cm} \\ \scriptsize{±0.3}\end{tabular}
& \begin{tabular}[c]{@{}c@{}} {30.1} \vspace{-0.1cm} \\ \scriptsize{±4.9}\end{tabular}
& \begin{tabular}[c]{@{}c@{}} {29.1} \vspace{-0.1cm} \\ \scriptsize{±1.2}\end{tabular}
& \begin{tabular}[c]{@{}c@{}} {30.9} \vspace{-0.1cm} \\ \scriptsize{±3.6}\end{tabular}
& \begin{tabular}[c]{@{}c@{}} {27.7} \vspace{-0.1cm} \\ \scriptsize{±5.7}\end{tabular}
& \begin{tabular}[c]{@{}c@{}} {29.7} \vspace{-0.1cm} \\ \scriptsize{±3.5}\end{tabular}
& \begin{tabular}[c]{@{}c@{}} \colorbox{mygreen}{29.3} \vspace{-0.1cm} \\ \scriptsize{±0.9}\end{tabular}
& \begin{tabular}[c]{@{}c@{}} {31.5} \vspace{-0.1cm} \\ \scriptsize{±0.9}\end{tabular}
& \begin{tabular}[c]{@{}c@{}} {29.7} \vspace{-0.1cm} \\ \scriptsize{±6.1}\end{tabular}
& \begin{tabular}[c]{@{}c@{}} {29.3} \vspace{-0.1cm} \\ \scriptsize{±3.3}\end{tabular}
& \begin{tabular}[c]{@{}c@{}} {30.3} \vspace{-0.1cm} \\ \scriptsize{±3.3}\end{tabular}
\\
{ig$\rightarrow$}
& \begin{tabular}[c]{@{}c@{}} {55.2} \vspace{-0.1cm} \\ \scriptsize{±1.5}\end{tabular}
& \begin{tabular}[c]{@{}c@{}} {50.6} \vspace{-0.1cm} \\ \scriptsize{±2.6}\end{tabular}
& \begin{tabular}[c]{@{}c@{}} \colorbox{mygreen}{38.0} \vspace{-0.1cm} \\ \scriptsize{±1.6}\end{tabular}
& \begin{tabular}[c]{@{}c@{}} {29.5} \vspace{-0.1cm} \\ \scriptsize{±2.4}\end{tabular}
& \begin{tabular}[c]{@{}c@{}} {32.7} \vspace{-0.1cm} \\ \scriptsize{±2.4}\end{tabular}
& \begin{tabular}[c]{@{}c@{}} {33.9} \vspace{-0.1cm} \\ \scriptsize{±5.5}\end{tabular}
& \begin{tabular}[c]{@{}c@{}} {32.3} \vspace{-0.1cm} \\ \scriptsize{±1.5}\end{tabular}
& \begin{tabular}[c]{@{}c@{}} \colorbox{mygreen}{33.1} \vspace{-0.1cm} \\ \scriptsize{±3.2}\end{tabular}
& \begin{tabular}[c]{@{}c@{}} {33.3} \vspace{-0.1cm} \\ \scriptsize{±3.0}\end{tabular}
& \begin{tabular}[c]{@{}c@{}} {31.9} \vspace{-0.1cm} \\ \scriptsize{±4.7}\end{tabular}
& \begin{tabular}[c]{@{}c@{}} \colorbox{mygreen}{31.3} \vspace{-0.1cm} \\ \scriptsize{±1.0}\end{tabular}
& \begin{tabular}[c]{@{}c@{}} {32.1} \vspace{-0.1cm} \\ \scriptsize{±1.2}\end{tabular}
\\
{nso$\rightarrow$}
& \begin{tabular}[c]{@{}c@{}} {55.4} \vspace{-0.1cm} \\ \scriptsize{±1.6}\end{tabular}
& \begin{tabular}[c]{@{}c@{}} {50.2} \vspace{-0.1cm} \\ \scriptsize{±2.3}\end{tabular}
& \begin{tabular}[c]{@{}c@{}} \colorbox{mygreen}{38.3} \vspace{-0.1cm} \\ \scriptsize{±2.9}\end{tabular}
& \begin{tabular}[c]{@{}c@{}} {31.7} \vspace{-0.1cm} \\ \scriptsize{±6.4}\end{tabular}
& \begin{tabular}[c]{@{}c@{}} {29.7} \vspace{-0.1cm} \\ \scriptsize{±2.1}\end{tabular}
& \begin{tabular}[c]{@{}c@{}} {32.1} \vspace{-0.1cm} \\ \scriptsize{±2.7}\end{tabular}
& \begin{tabular}[c]{@{}c@{}} {34.9} \vspace{-0.1cm} \\ \scriptsize{±0.6}\end{tabular}
& \begin{tabular}[c]{@{}c@{}} \colorbox{myblue}{33.1} \vspace{-0.1cm} \\ \scriptsize{±1.8}\end{tabular}
& \begin{tabular}[c]{@{}c@{}} \colorbox{mygreen}{38.4} \vspace{-0.1cm} \\ \scriptsize{±3.1}\end{tabular}
& \begin{tabular}[c]{@{}c@{}} {34.3} \vspace{-0.1cm} \\ \scriptsize{±2.4}\end{tabular}
& \begin{tabular}[c]{@{}c@{}} \colorbox{mygreen}{35.7} \vspace{-0.1cm} \\ \scriptsize{±1.2}\end{tabular}
& \begin{tabular}[c]{@{}c@{}} {32.1} \vspace{-0.1cm} \\ \scriptsize{±1.8}\end{tabular}
\\
{sn$\rightarrow$}
& \begin{tabular}[c]{@{}c@{}} {53.0} \vspace{-0.1cm} \\ \scriptsize{±1.6}\end{tabular}
& \begin{tabular}[c]{@{}c@{}} \colorbox{mygreen}{48.2} \vspace{-0.1cm} \\ \scriptsize{±0.6}\end{tabular}
& \begin{tabular}[c]{@{}c@{}} \colorbox{mygreen}{36.9} \vspace{-0.1cm} \\ \scriptsize{±2.5}\end{tabular}
& \begin{tabular}[c]{@{}c@{}} {33.7} \vspace{-0.1cm} \\ \scriptsize{±3.3}\end{tabular}
& \begin{tabular}[c]{@{}c@{}} {32.1} \vspace{-0.1cm} \\ \scriptsize{±0.9}\end{tabular}
& \begin{tabular}[c]{@{}c@{}} {31.5} \vspace{-0.1cm} \\ \scriptsize{±3.3}\end{tabular}
& \begin{tabular}[c]{@{}c@{}} {32.5} \vspace{-0.1cm} \\ \scriptsize{±3.2}\end{tabular}
& \begin{tabular}[c]{@{}c@{}} {32.7} \vspace{-0.1cm} \\ \scriptsize{±3.6}\end{tabular}
& \begin{tabular}[c]{@{}c@{}} {32.7} \vspace{-0.1cm} \\ \scriptsize{±1.5}\end{tabular}
& \begin{tabular}[c]{@{}c@{}} \colorbox{mygreen}{31.9} \vspace{-0.1cm} \\ \scriptsize{±0.6}\end{tabular}
& \begin{tabular}[c]{@{}c@{}} \colorbox{mygreen}{31.9} \vspace{-0.1cm} \\ \scriptsize{±0.6}\end{tabular}
& \begin{tabular}[c]{@{}c@{}} {30.5} \vspace{-0.1cm} \\ \scriptsize{±3.0}\end{tabular}
\\
{st$\rightarrow$}
& \begin{tabular}[c]{@{}c@{}} {55.0} \vspace{-0.1cm} \\ \scriptsize{±1.5}\end{tabular}
& \begin{tabular}[c]{@{}c@{}} {48.4} \vspace{-0.1cm} \\ \scriptsize{±1.2}\end{tabular}
& \begin{tabular}[c]{@{}c@{}} {34.7} \vspace{-0.1cm} \\ \scriptsize{±3.7}\end{tabular}
& \begin{tabular}[c]{@{}c@{}} \colorbox{mygreen}{35.5} \vspace{-0.1cm} \\ \scriptsize{±3.4}\end{tabular}
& \begin{tabular}[c]{@{}c@{}} {33.3} \vspace{-0.1cm} \\ \scriptsize{±2.1}\end{tabular}
& \begin{tabular}[c]{@{}c@{}} {30.9} \vspace{-0.1cm} \\ \scriptsize{±4.7}\end{tabular}
& \begin{tabular}[c]{@{}c@{}} {35.5} \vspace{-0.1cm} \\ \scriptsize{±2.2}\end{tabular}
& \begin{tabular}[c]{@{}c@{}} \colorbox{mygreen}{31.7} \vspace{-0.1cm} \\ \scriptsize{±1.5}\end{tabular}
& \begin{tabular}[c]{@{}c@{}} \colorbox{mygreen}{37.3} \vspace{-0.1cm} \\ \scriptsize{±1.3}\end{tabular}
& \begin{tabular}[c]{@{}c@{}} {34.5} \vspace{-0.1cm} \\ \scriptsize{±3.4}\end{tabular}
& \begin{tabular}[c]{@{}c@{}} {31.1} \vspace{-0.1cm} \\ \scriptsize{±1.5}\end{tabular}
& \begin{tabular}[c]{@{}c@{}} {32.7} \vspace{-0.1cm} \\ \scriptsize{±1.2}\end{tabular}
\\
{tn$\rightarrow$}
& \begin{tabular}[c]{@{}c@{}} \colorbox{mygreen}{55.6} \vspace{-0.1cm} \\ \scriptsize{±0.7}\end{tabular}
& \begin{tabular}[c]{@{}c@{}} \colorbox{mygreen}{51.2} \vspace{-0.1cm} \\ \scriptsize{±1.6}\end{tabular}
& \begin{tabular}[c]{@{}c@{}} \colorbox{mygreen}{38.6} \vspace{-0.1cm} \\ \scriptsize{±0.6}\end{tabular}
& \begin{tabular}[c]{@{}c@{}} \colorbox{mygreen}{38.2} \vspace{-0.1cm} \\ \scriptsize{±1.6}\end{tabular}
& \begin{tabular}[c]{@{}c@{}} {31.3} \vspace{-0.1cm} \\ \scriptsize{±7.3}\end{tabular}
& \begin{tabular}[c]{@{}c@{}} {32.5} \vspace{-0.1cm} \\ \scriptsize{±2.2}\end{tabular}
& \begin{tabular}[c]{@{}c@{}} {35.9} \vspace{-0.1cm} \\ \scriptsize{±0.7}\end{tabular}
& \begin{tabular}[c]{@{}c@{}} \colorbox{mygreen}{34.3} \vspace{-0.1cm} \\ \scriptsize{±1.0}\end{tabular}
& \begin{tabular}[c]{@{}c@{}} \colorbox{mygreen}{37.3} \vspace{-0.1cm} \\ \scriptsize{±1.6}\end{tabular}
& \begin{tabular}[c]{@{}c@{}} \colorbox{mygreen}{32.1} \vspace{-0.1cm} \\ \scriptsize{±0.3}\end{tabular}
& \begin{tabular}[c]{@{}c@{}} \colorbox{myblue}{32.3} \vspace{-0.1cm} \\ \scriptsize{±1.4}\end{tabular}
& \begin{tabular}[c]{@{}c@{}} {33.1} \vspace{-0.1cm} \\ \scriptsize{±3.3}\end{tabular}
\\
{ts$\rightarrow$}
& \begin{tabular}[c]{@{}c@{}} {52.4} \vspace{-0.1cm} \\ \scriptsize{±2.2}\end{tabular}
& \begin{tabular}[c]{@{}c@{}} {47.4} \vspace{-0.1cm} \\ \scriptsize{±2.4}\end{tabular}
& \begin{tabular}[c]{@{}c@{}} \colorbox{mygreen}{34.1} \vspace{-0.1cm} \\ \scriptsize{±2.3}\end{tabular}
& \begin{tabular}[c]{@{}c@{}} {34.5} \vspace{-0.1cm} \\ \scriptsize{±4.0}\end{tabular}
& \begin{tabular}[c]{@{}c@{}} {31.7} \vspace{-0.1cm} \\ \scriptsize{±1.8}\end{tabular}
& \begin{tabular}[c]{@{}c@{}} {34.5} \vspace{-0.1cm} \\ \scriptsize{±2.8}\end{tabular}
& \begin{tabular}[c]{@{}c@{}} {34.1} \vspace{-0.1cm} \\ \scriptsize{±4.0}\end{tabular}
& \begin{tabular}[c]{@{}c@{}} {30.5} \vspace{-0.1cm} \\ \scriptsize{±2.4}\end{tabular}
& \begin{tabular}[c]{@{}c@{}} {32.7} \vspace{-0.1cm} \\ \scriptsize{±0.9}\end{tabular}
& \begin{tabular}[c]{@{}c@{}} {32.5} \vspace{-0.1cm} \\ \scriptsize{±3.9}\end{tabular}
& \begin{tabular}[c]{@{}c@{}} {30.3} \vspace{-0.1cm} \\ \scriptsize{±1.2}\end{tabular}
& \begin{tabular}[c]{@{}c@{}} {29.9} \vspace{-0.1cm} \\ \scriptsize{±2.8}\end{tabular}
\\    
    \hline \hline

\end{tabular}
\caption{\textbf{Results of Cross-lingual Transfer using Llama 3 70B Instruct when Tuning on Winogrande and Testing on MMLU Virology}. Llama 3 70B Instruct was independently fine-tuned on the 11 African languages of focus:  Afrikaans (af), Zulu (zu),
Xhosa (xh) (datasets for these three languages were sourced from (BMGF 2024)), Amharic (am), Bambara (bm), Igbo (ig),
Sepedi (nso), Shona (sn), Sesotho (st), Setswana (tn), and Tsonga (ts). Llama 3 70B Instruct was also fine-tuned on English (en) for reference. The translated \texttt{Winogrande} training set (small) was used for fine-tuning. The fine-tuned models were evaluated for their cross-lingual transfer performance on the translated \texttt{MMLU Virology} (virology domain knowledge task) test set. Columns indicate the target language of the evaluation, while the rows indicate the source language the models were fine-tuned with (e.g. ``zu$\rightarrow$" indicates models fine-tuned with data in Zulu). All numbers are the mean 5-shot performance accuracy of three evaluations followed by the standard deviation (±). Model fine-tuning that yielded mono-lingual improvements (relative to the baseline) more than two standard deviations are in \colorbox{myblue}{blue}, while cross-lingual improvements (relative to the baseline) more than two standard deviations are in \colorbox{mygreen}{green}.\\} \label{table:perf-crosslingual-llama3-tr-wino-ts-mmlu-vir}
\end{table*}

\begin{table*}[t!]
\centering
% \small

\setlength{\tabcolsep}{5.5pt}
\begin{tabular}{r|llllllllllll}
                    \hline \hline
                     & \multicolumn{12}{c}{\texttt{MMLU Virology}} \\
                     % & \multicolumn{4}{c|}{\begin{tabular}[c]{@{}c@{}}\texttt{MMLU-Clinical-ZA} \\ "clinical knowledge"\end{tabular}} 
                     % & \multicolumn{4}{c}{\texttt{Belebele}} \\
                     & \textbf{en} & \textbf{af} & \textbf{zu} & \textbf{xh} 
                     & \textbf{am} & \textbf{bm} & \textbf{ig} & \textbf{nso} 
                     & \textbf{sn} & \textbf{st} & \textbf{tn} & \textbf{ts} \\ \hline 

\begin{tabular}[c]{@{}c@{}}Base \\ \tiny{Llama 3} \vspace{-0.1cm}\\ \tiny{70B IT} \end{tabular} 
& 53.6 & 46.4 & 28.9 & 27.1 & 36.1 & 35.5 & 34.9 & 26.5 & 31.3 & 29.5 & 28.3 & 31.3
\\ \hline
\multicolumn{12}{l}{\begin{tabular}[c]{@{}c@{}}\texttt{MMLU College Medicine} \\ dev + test + val\end{tabular}} \\
{en$\rightarrow$}
& \begin{tabular}[c]{@{}c@{}} {53.8} \vspace{-0.1cm} \\ \scriptsize{±1.8}\end{tabular}
& \begin{tabular}[c]{@{}c@{}} {46.2} \vspace{-0.1cm} \\ \scriptsize{±0.7}\end{tabular}
& \begin{tabular}[c]{@{}c@{}} \colorbox{mygreen}{31.1} \vspace{-0.1cm} \\ \scriptsize{±0.9}\end{tabular}
& \begin{tabular}[c]{@{}c@{}} {31.9} \vspace{-0.1cm} \\ \scriptsize{±3.1}\end{tabular}
& \begin{tabular}[c]{@{}c@{}} {30.5} \vspace{-0.1cm} \\ \scriptsize{±3.0}\end{tabular}
& \begin{tabular}[c]{@{}c@{}} {30.1} \vspace{-0.1cm} \\ \scriptsize{±2.2}\end{tabular}
& \begin{tabular}[c]{@{}c@{}} {29.9} \vspace{-0.1cm} \\ \scriptsize{±3.1}\end{tabular}
& \begin{tabular}[c]{@{}c@{}} {30.5} \vspace{-0.1cm} \\ \scriptsize{±2.4}\end{tabular}
& \begin{tabular}[c]{@{}c@{}} {27.5} \vspace{-0.1cm} \\ \scriptsize{±1.2}\end{tabular}
& \begin{tabular}[c]{@{}c@{}} {28.1} \vspace{-0.1cm} \\ \scriptsize{±0.9}\end{tabular}
& \begin{tabular}[c]{@{}c@{}} {25.7} \vspace{-0.1cm} \\ \scriptsize{±0.9}\end{tabular}
& \begin{tabular}[c]{@{}c@{}} {28.9} \vspace{-0.1cm} \\ \scriptsize{±0.6}\end{tabular}
\\
{af$\rightarrow$}
& \begin{tabular}[c]{@{}c@{}} {53.8} \vspace{-0.1cm} \\ \scriptsize{±1.8}\end{tabular}
& \begin{tabular}[c]{@{}c@{}} \colorbox{myblue}{48.4} \vspace{-0.1cm} \\ \scriptsize{±0.9}\end{tabular}
& \begin{tabular}[c]{@{}c@{}} {31.1} \vspace{-0.1cm} \\ \scriptsize{±3.0}\end{tabular}
& \begin{tabular}[c]{@{}c@{}} {28.9} \vspace{-0.1cm} \\ \scriptsize{±4.8}\end{tabular}
& \begin{tabular}[c]{@{}c@{}} {29.7} \vspace{-0.1cm} \\ \scriptsize{±1.2}\end{tabular}
& \begin{tabular}[c]{@{}c@{}} {28.3} \vspace{-0.1cm} \\ \scriptsize{±2.6}\end{tabular}
& \begin{tabular}[c]{@{}c@{}} {31.1} \vspace{-0.1cm} \\ \scriptsize{±0.3}\end{tabular}
& \begin{tabular}[c]{@{}c@{}} {28.9} \vspace{-0.1cm} \\ \scriptsize{±1.2}\end{tabular}
& \begin{tabular}[c]{@{}c@{}} {30.3} \vspace{-0.1cm} \\ \scriptsize{±2.8}\end{tabular}
& \begin{tabular}[c]{@{}c@{}} {31.9} \vspace{-0.1cm} \\ \scriptsize{±5.1}\end{tabular}
& \begin{tabular}[c]{@{}c@{}} {24.3} \vspace{-0.1cm} \\ \scriptsize{±0.9}\end{tabular}
& \begin{tabular}[c]{@{}c@{}} {28.1} \vspace{-0.1cm} \\ \scriptsize{±1.9}\end{tabular}
\\
{zu$\rightarrow$}
& \begin{tabular}[c]{@{}c@{}} \colorbox{mygreen}{55.6} \vspace{-0.1cm} \\ \scriptsize{±0.9}\end{tabular}
& \begin{tabular}[c]{@{}c@{}} \colorbox{mygreen}{50.8} \vspace{-0.1cm} \\ \scriptsize{±0.7}\end{tabular}
& \begin{tabular}[c]{@{}c@{}} \colorbox{myblue}{35.3} \vspace{-0.1cm} \\ \scriptsize{±1.4}\end{tabular}
& \begin{tabular}[c]{@{}c@{}} \colorbox{mygreen}{32.5} \vspace{-0.1cm} \\ \scriptsize{±1.6}\end{tabular}
& \begin{tabular}[c]{@{}c@{}} {34.1} \vspace{-0.1cm} \\ \scriptsize{±1.2}\end{tabular}
& \begin{tabular}[c]{@{}c@{}} {26.9} \vspace{-0.1cm} \\ \scriptsize{±1.5}\end{tabular}
& \begin{tabular}[c]{@{}c@{}} \colorbox{mygreen}{36.5} \vspace{-0.1cm} \\ \scriptsize{±0.3}\end{tabular}
& \begin{tabular}[c]{@{}c@{}} \colorbox{mygreen}{30.3} \vspace{-0.1cm} \\ \scriptsize{±0.7}\end{tabular}
& \begin{tabular}[c]{@{}c@{}} {34.3} \vspace{-0.1cm} \\ \scriptsize{±1.8}\end{tabular}
& \begin{tabular}[c]{@{}c@{}} {33.9} \vspace{-0.1cm} \\ \scriptsize{±4.1}\end{tabular}
& \begin{tabular}[c]{@{}c@{}} \colorbox{mygreen}{31.7} \vspace{-0.1cm} \\ \scriptsize{±0.3}\end{tabular}
& \begin{tabular}[c]{@{}c@{}} {31.9} \vspace{-0.1cm} \\ \scriptsize{±3.3}\end{tabular}
\\
{xh$\rightarrow$}
& \begin{tabular}[c]{@{}c@{}} {55.0} \vspace{-0.1cm} \\ \scriptsize{±0.9}\end{tabular}
& \begin{tabular}[c]{@{}c@{}} \colorbox{mygreen}{50.6} \vspace{-0.1cm} \\ \scriptsize{±0.0}\end{tabular}
& \begin{tabular}[c]{@{}c@{}} \colorbox{mygreen}{34.5} \vspace{-0.1cm} \\ \scriptsize{±0.7}\end{tabular}
& \begin{tabular}[c]{@{}c@{}} \colorbox{myblue}{32.7} \vspace{-0.1cm} \\ \scriptsize{±2.1}\end{tabular}
& \begin{tabular}[c]{@{}c@{}} {32.5} \vspace{-0.1cm} \\ \scriptsize{±3.9}\end{tabular}
& \begin{tabular}[c]{@{}c@{}} {28.5} \vspace{-0.1cm} \\ \scriptsize{±3.3}\end{tabular}
& \begin{tabular}[c]{@{}c@{}} {35.1} \vspace{-0.1cm} \\ \scriptsize{±0.9}\end{tabular}
& \begin{tabular}[c]{@{}c@{}} {28.3} \vspace{-0.1cm} \\ \scriptsize{±1.2}\end{tabular}
& \begin{tabular}[c]{@{}c@{}} {34.1} \vspace{-0.1cm} \\ \scriptsize{±1.4}\end{tabular}
& \begin{tabular}[c]{@{}c@{}} {31.7} \vspace{-0.1cm} \\ \scriptsize{±2.4}\end{tabular}
& \begin{tabular}[c]{@{}c@{}} {28.7} \vspace{-0.1cm} \\ \scriptsize{±1.9}\end{tabular}
& \begin{tabular}[c]{@{}c@{}} {33.7} \vspace{-0.1cm} \\ \scriptsize{±1.6}\end{tabular}
\\
{am$\rightarrow$}
& \begin{tabular}[c]{@{}c@{}} \colorbox{mygreen}{55.8} \vspace{-0.1cm} \\ \scriptsize{±0.7}\end{tabular}
& \begin{tabular}[c]{@{}c@{}} \colorbox{mygreen}{52.0} \vspace{-0.1cm} \\ \scriptsize{±0.3}\end{tabular}
& \begin{tabular}[c]{@{}c@{}} \colorbox{mygreen}{38.6} \vspace{-0.1cm} \\ \scriptsize{±2.2}\end{tabular}
& \begin{tabular}[c]{@{}c@{}} \colorbox{mygreen}{33.3} \vspace{-0.1cm} \\ \scriptsize{±1.2}\end{tabular}
& \begin{tabular}[c]{@{}c@{}} {37.6} \vspace{-0.1cm} \\ \scriptsize{±2.5}\end{tabular}
& \begin{tabular}[c]{@{}c@{}} {28.1} \vspace{-0.1cm} \\ \scriptsize{±4.1}\end{tabular}
& \begin{tabular}[c]{@{}c@{}} {33.9} \vspace{-0.1cm} \\ \scriptsize{±2.8}\end{tabular}
& \begin{tabular}[c]{@{}c@{}} \colorbox{mygreen}{30.7} \vspace{-0.1cm} \\ \scriptsize{±1.2}\end{tabular}
& \begin{tabular}[c]{@{}c@{}} {34.7} \vspace{-0.1cm} \\ \scriptsize{±3.1}\end{tabular}
& \begin{tabular}[c]{@{}c@{}} {33.5} \vspace{-0.1cm} \\ \scriptsize{±3.7}\end{tabular}
& \begin{tabular}[c]{@{}c@{}} \colorbox{mygreen}{30.7} \vspace{-0.1cm} \\ \scriptsize{±0.6}\end{tabular}
& \begin{tabular}[c]{@{}c@{}} {32.1} \vspace{-0.1cm} \\ \scriptsize{±1.5}\end{tabular}
\\
{bm$\rightarrow$}
& \begin{tabular}[c]{@{}c@{}} \colorbox{mygreen}{55.2} \vspace{-0.1cm} \\ \scriptsize{±0.7}\end{tabular}
& \begin{tabular}[c]{@{}c@{}} \colorbox{mygreen}{48.8} \vspace{-0.1cm} \\ \scriptsize{±0.6}\end{tabular}
& \begin{tabular}[c]{@{}c@{}} {30.9} \vspace{-0.1cm} \\ \scriptsize{±2.4}\end{tabular}
& \begin{tabular}[c]{@{}c@{}} {30.9} \vspace{-0.1cm} \\ \scriptsize{±2.4}\end{tabular}
& \begin{tabular}[c]{@{}c@{}} {31.7} \vspace{-0.1cm} \\ \scriptsize{±2.7}\end{tabular}
& \begin{tabular}[c]{@{}c@{}} {26.1} \vspace{-0.1cm} \\ \scriptsize{±2.4}\end{tabular}
& \begin{tabular}[c]{@{}c@{}} {28.7} \vspace{-0.1cm} \\ \scriptsize{±3.0}\end{tabular}
& \begin{tabular}[c]{@{}c@{}} \colorbox{mygreen}{29.9} \vspace{-0.1cm} \\ \scriptsize{±0.7}\end{tabular}
& \begin{tabular}[c]{@{}c@{}} {28.9} \vspace{-0.1cm} \\ \scriptsize{±2.7}\end{tabular}
& \begin{tabular}[c]{@{}c@{}} {28.9} \vspace{-0.1cm} \\ \scriptsize{±1.0}\end{tabular}
& \begin{tabular}[c]{@{}c@{}} {28.5} \vspace{-0.1cm} \\ \scriptsize{±2.7}\end{tabular}
& \begin{tabular}[c]{@{}c@{}} {26.5} \vspace{-0.1cm} \\ \scriptsize{±4.3}\end{tabular}
\\
{ig$\rightarrow$}
& \begin{tabular}[c]{@{}c@{}} {54.6} \vspace{-0.1cm} \\ \scriptsize{±1.8}\end{tabular}
& \begin{tabular}[c]{@{}c@{}} \colorbox{mygreen}{51.0} \vspace{-0.1cm} \\ \scriptsize{±0.9}\end{tabular}
& \begin{tabular}[c]{@{}c@{}} \colorbox{mygreen}{35.9} \vspace{-0.1cm} \\ \scriptsize{±1.8}\end{tabular}
& \begin{tabular}[c]{@{}c@{}} {31.9} \vspace{-0.1cm} \\ \scriptsize{±3.2}\end{tabular}
& \begin{tabular}[c]{@{}c@{}} {30.9} \vspace{-0.1cm} \\ \scriptsize{±2.5}\end{tabular}
& \begin{tabular}[c]{@{}c@{}} {24.5} \vspace{-0.1cm} \\ \scriptsize{±2.8}\end{tabular}
& \begin{tabular}[c]{@{}c@{}} {35.9} \vspace{-0.1cm} \\ \scriptsize{±3.0}\end{tabular}
& \begin{tabular}[c]{@{}c@{}} \colorbox{mygreen}{31.7} \vspace{-0.1cm} \\ \scriptsize{±2.1}\end{tabular}
& \begin{tabular}[c]{@{}c@{}} \colorbox{mygreen}{37.8} \vspace{-0.1cm} \\ \scriptsize{±1.4}\end{tabular}
& \begin{tabular}[c]{@{}c@{}} {31.7} \vspace{-0.1cm} \\ \scriptsize{±2.1}\end{tabular}
& \begin{tabular}[c]{@{}c@{}} {30.9} \vspace{-0.1cm} \\ \scriptsize{±3.0}\end{tabular}
& \begin{tabular}[c]{@{}c@{}} \colorbox{mygreen}{33.3} \vspace{-0.1cm} \\ \scriptsize{±0.3}\end{tabular}
\\
{nso$\rightarrow$}
& \begin{tabular}[c]{@{}c@{}} {54.6} \vspace{-0.1cm} \\ \scriptsize{±1.2}\end{tabular}
& \begin{tabular}[c]{@{}c@{}} \colorbox{mygreen}{50.4} \vspace{-0.1cm} \\ \scriptsize{±1.4}\end{tabular}
& \begin{tabular}[c]{@{}c@{}} \colorbox{mygreen}{35.7} \vspace{-0.1cm} \\ \scriptsize{±0.9}\end{tabular}
& \begin{tabular}[c]{@{}c@{}} \colorbox{mygreen}{33.7} \vspace{-0.1cm} \\ \scriptsize{±1.6}\end{tabular}
& \begin{tabular}[c]{@{}c@{}} {38.5} \vspace{-0.1cm} \\ \scriptsize{±3.7}\end{tabular}
& \begin{tabular}[c]{@{}c@{}} {26.3} \vspace{-0.1cm} \\ \scriptsize{±1.5}\end{tabular}
& \begin{tabular}[c]{@{}c@{}} {34.9} \vspace{-0.1cm} \\ \scriptsize{±2.1}\end{tabular}
& \begin{tabular}[c]{@{}c@{}} \colorbox{myblue}{33.7} \vspace{-0.1cm} \\ \scriptsize{±2.4}\end{tabular}
& \begin{tabular}[c]{@{}c@{}} {32.1} \vspace{-0.1cm} \\ \scriptsize{±2.4}\end{tabular}
& \begin{tabular}[c]{@{}c@{}} \colorbox{mygreen}{33.1} \vspace{-0.1cm} \\ \scriptsize{±1.2}\end{tabular}
& \begin{tabular}[c]{@{}c@{}} {30.5} \vspace{-0.1cm} \\ \scriptsize{±2.5}\end{tabular}
& \begin{tabular}[c]{@{}c@{}} {32.9} \vspace{-0.1cm} \\ \scriptsize{±2.3}\end{tabular}
\\
{sn$\rightarrow$}
& \begin{tabular}[c]{@{}c@{}} \colorbox{mygreen}{56.4} \vspace{-0.1cm} \\ \scriptsize{±0.3}\end{tabular}
& \begin{tabular}[c]{@{}c@{}} \colorbox{mygreen}{50.6} \vspace{-0.1cm} \\ \scriptsize{±0.6}\end{tabular}
& \begin{tabular}[c]{@{}c@{}} \colorbox{mygreen}{40.0} \vspace{-0.1cm} \\ \scriptsize{±1.7}\end{tabular}
& \begin{tabular}[c]{@{}c@{}} \colorbox{mygreen}{35.1} \vspace{-0.1cm} \\ \scriptsize{±3.3}\end{tabular}
& \begin{tabular}[c]{@{}c@{}} {34.5} \vspace{-0.1cm} \\ \scriptsize{±1.2}\end{tabular}
& \begin{tabular}[c]{@{}c@{}} {26.5} \vspace{-0.1cm} \\ \scriptsize{±2.7}\end{tabular}
& \begin{tabular}[c]{@{}c@{}} {38.6} \vspace{-0.1cm} \\ \scriptsize{±3.7}\end{tabular}
& \begin{tabular}[c]{@{}c@{}} \colorbox{mygreen}{32.9} \vspace{-0.1cm} \\ \scriptsize{±1.8}\end{tabular}
& \begin{tabular}[c]{@{}c@{}} \colorbox{myblue}{37.5} \vspace{-0.1cm} \\ \scriptsize{±0.4}\end{tabular}
& \begin{tabular}[c]{@{}c@{}} \colorbox{mygreen}{33.7} \vspace{-0.1cm} \\ \scriptsize{±2.1}\end{tabular}
& \begin{tabular}[c]{@{}c@{}} \colorbox{mygreen}{32.3} \vspace{-0.1cm} \\ \scriptsize{±0.3}\end{tabular}
& \begin{tabular}[c]{@{}c@{}} {32.5} \vspace{-0.1cm} \\ \scriptsize{±3.3}\end{tabular}
\\
{st$\rightarrow$}
& \begin{tabular}[c]{@{}c@{}} {55.2} \vspace{-0.1cm} \\ \scriptsize{±0.9}\end{tabular}
& \begin{tabular}[c]{@{}c@{}} \colorbox{mygreen}{50.0} \vspace{-0.1cm} \\ \scriptsize{±1.0}\end{tabular}
& \begin{tabular}[c]{@{}c@{}} {33.5} \vspace{-0.1cm} \\ \scriptsize{±2.4}\end{tabular}
& \begin{tabular}[c]{@{}c@{}} {31.7} \vspace{-0.1cm} \\ \scriptsize{±4.8}\end{tabular}
& \begin{tabular}[c]{@{}c@{}} {35.3} \vspace{-0.1cm} \\ \scriptsize{±1.2}\end{tabular}
& \begin{tabular}[c]{@{}c@{}} {30.7} \vspace{-0.1cm} \\ \scriptsize{±3.3}\end{tabular}
& \begin{tabular}[c]{@{}c@{}} {34.7} \vspace{-0.1cm} \\ \scriptsize{±2.3}\end{tabular}
& \begin{tabular}[c]{@{}c@{}} \colorbox{mygreen}{34.3} \vspace{-0.1cm} \\ \scriptsize{±2.2}\end{tabular}
& \begin{tabular}[c]{@{}c@{}} {33.1} \vspace{-0.1cm} \\ \scriptsize{±3.1}\end{tabular}
& \begin{tabular}[c]{@{}c@{}} {33.7} \vspace{-0.1cm} \\ \scriptsize{±3.2}\end{tabular}
& \begin{tabular}[c]{@{}c@{}} \colorbox{mygreen}{32.7} \vspace{-0.1cm} \\ \scriptsize{±1.7}\end{tabular}
& \begin{tabular}[c]{@{}c@{}} {35.7} \vspace{-0.1cm} \\ \scriptsize{±2.6}\end{tabular}
\\
{tn$\rightarrow$}
& \begin{tabular}[c]{@{}c@{}} \colorbox{mygreen}{55.2} \vspace{-0.1cm} \\ \scriptsize{±0.3}\end{tabular}
& \begin{tabular}[c]{@{}c@{}} \colorbox{mygreen}{51.0} \vspace{-0.1cm} \\ \scriptsize{±1.8}\end{tabular}
& \begin{tabular}[c]{@{}c@{}} \colorbox{mygreen}{36.3} \vspace{-0.1cm} \\ \scriptsize{±1.9}\end{tabular}
& \begin{tabular}[c]{@{}c@{}} \colorbox{mygreen}{32.5} \vspace{-0.1cm} \\ \scriptsize{±1.6}\end{tabular}
& \begin{tabular}[c]{@{}c@{}} {33.1} \vspace{-0.1cm} \\ \scriptsize{±2.4}\end{tabular}
& \begin{tabular}[c]{@{}c@{}} {27.9} \vspace{-0.1cm} \\ \scriptsize{±1.5}\end{tabular}
& \begin{tabular}[c]{@{}c@{}} {37.2} \vspace{-0.1cm} \\ \scriptsize{±3.6}\end{tabular}
& \begin{tabular}[c]{@{}c@{}} \colorbox{mygreen}{35.5} \vspace{-0.1cm} \\ \scriptsize{±1.2}\end{tabular}
& \begin{tabular}[c]{@{}c@{}} {33.1} \vspace{-0.1cm} \\ \scriptsize{±1.2}\end{tabular}
& \begin{tabular}[c]{@{}c@{}} {32.5} \vspace{-0.1cm} \\ \scriptsize{±3.3}\end{tabular}
& \begin{tabular}[c]{@{}c@{}} \colorbox{myblue}{34.5} \vspace{-0.1cm} \\ \scriptsize{±1.8}\end{tabular}
& \begin{tabular}[c]{@{}c@{}} {32.9} \vspace{-0.1cm} \\ \scriptsize{±2.5}\end{tabular}
\\
{ts$\rightarrow$}
& \begin{tabular}[c]{@{}c@{}} \colorbox{mygreen}{56.2} \vspace{-0.1cm} \\ \scriptsize{±0.3}\end{tabular}
& \begin{tabular}[c]{@{}c@{}} \colorbox{mygreen}{49.8} \vspace{-0.1cm} \\ \scriptsize{±1.2}\end{tabular}
& \begin{tabular}[c]{@{}c@{}} \colorbox{mygreen}{37.3} \vspace{-0.1cm} \\ \scriptsize{±2.5}\end{tabular}
& \begin{tabular}[c]{@{}c@{}} \colorbox{mygreen}{33.1} \vspace{-0.1cm} \\ \scriptsize{±0.6}\end{tabular}
& \begin{tabular}[c]{@{}c@{}} \colorbox{mygreen}{38.4} \vspace{-0.1cm} \\ \scriptsize{±0.3}\end{tabular}
& \begin{tabular}[c]{@{}c@{}} {24.7} \vspace{-0.1cm} \\ \scriptsize{±2.6}\end{tabular}
& \begin{tabular}[c]{@{}c@{}} \colorbox{mygreen}{38.0} \vspace{-0.1cm} \\ \scriptsize{±1.3}\end{tabular}
& \begin{tabular}[c]{@{}c@{}} \colorbox{mygreen}{34.3} \vspace{-0.1cm} \\ \scriptsize{±1.6}\end{tabular}
& \begin{tabular}[c]{@{}c@{}} \colorbox{mygreen}{35.3} \vspace{-0.1cm} \\ \scriptsize{±1.2}\end{tabular}
& \begin{tabular}[c]{@{}c@{}} \colorbox{mygreen}{33.9} \vspace{-0.1cm} \\ \scriptsize{±1.9}\end{tabular}
& \begin{tabular}[c]{@{}c@{}} \colorbox{mygreen}{32.9} \vspace{-0.1cm} \\ \scriptsize{±1.2}\end{tabular}
& \begin{tabular}[c]{@{}c@{}} \colorbox{myblue}{38.6} \vspace{-0.1cm} \\ \scriptsize{±0.6}\end{tabular}
\\    
    \hline \hline

\end{tabular}
\caption{\textbf{Results of Cross-lingual Transfer using Llama 3 70B Instruct when Tuning on MMLU College Medicine and Testing on MMLU Virology}. Llama 3 70B Instruct was independently fine-tuned on the 11 African languages of focus:  Afrikaans (af), Zulu (zu),
Xhosa (xh) (datasets for these three languages were sourced from (BMGF 2024)), Amharic (am), Bambara (bm), Igbo (ig),
Sepedi (nso), Shona (sn), Sesotho (st), Setswana (tn), and Tsonga (ts). Llama 3 70B Instruct was also fine-tuned on English (en) for reference. The translated \texttt{MMLU College Medicine} section (dev + test + val sets) was used for fine-tuning. The fine-tuned models were evaluated for their cross-lingual transfer performance on the translated \texttt{MMLU Virology} (virology domain knowledge task) test set. Columns indicate the target language of the evaluation, while the rows indicate the source language the models were fine-tuned with (e.g. ``zu$\rightarrow$" indicates models fine-tuned with data in Zulu). All numbers are the mean 5-shot performance accuracy of three evaluations followed by the standard deviation (±). Model fine-tuning that yielded mono-lingual improvements (relative to the baseline) more than two standard deviations are in \colorbox{myblue}{blue}, while cross-lingual improvements (relative to the baseline) more than two standard deviations are in \colorbox{mygreen}{green}.\\} \label{table:perf-crosslingual-llama3-tr-mmlu-ts-mmlu-vir}
\end{table*}

\begin{table*}[t!]
\centering
% \small

\setlength{\tabcolsep}{5.5pt}
\begin{tabular}{r|llllllllllll}
                    \hline \hline
                     & \multicolumn{12}{c}{\texttt{Belebele}} \\
                     % & \multicolumn{4}{c|}{\begin{tabular}[c]{@{}c@{}}\texttt{MMLU-Clinical-ZA} \\ "clinical knowledge"\end{tabular}} 
                     % & \multicolumn{4}{c}{\texttt{Belebele}} \\
                     & \textbf{en} & \textbf{af} & \textbf{zu} & \textbf{xh} 
                     & \textbf{am} & \textbf{bm} & \textbf{ig} & \textbf{nso} 
                     & \textbf{sn} & \textbf{st} & \textbf{tn} & \textbf{ts} \\ \hline 

\begin{tabular}[c]{@{}c@{}}Base \\ \tiny{Llama 3} \vspace{-0.1cm}\\ \tiny{70B IT} \end{tabular} 
& 94.6 & 84.8 & 36.7 & 36.8 & 35.1 & 35.6 & 35.9 & 37.1 & 38.1 & 35.9 & 36.4 & 40.7
\\ \hline
\multicolumn{12}{l}{\begin{tabular}[c]{@{}c@{}}\texttt{Winogrande} \\ train (small)\end{tabular}}  \\
{en$\rightarrow$}
& \begin{tabular}[c]{@{}c@{}} \colorbox{myblue}{95.4} \vspace{-0.1cm} \\ \scriptsize{±0.3}\end{tabular}
& \begin{tabular}[c]{@{}c@{}} \colorbox{mygreen}{90.1} \vspace{-0.1cm} \\ \scriptsize{±0.4}\end{tabular}
& \begin{tabular}[c]{@{}c@{}} {37.2} \vspace{-0.1cm} \\ \scriptsize{±1.5}\end{tabular}
& \begin{tabular}[c]{@{}c@{}} {37.4} \vspace{-0.1cm} \\ \scriptsize{±0.4}\end{tabular}
& \begin{tabular}[c]{@{}c@{}} {32.4} \vspace{-0.1cm} \\ \scriptsize{±1.2}\end{tabular}
& \begin{tabular}[c]{@{}c@{}} {33.4} \vspace{-0.1cm} \\ \scriptsize{±1.3}\end{tabular}
& \begin{tabular}[c]{@{}c@{}} \colorbox{mygreen}{39.2} \vspace{-0.1cm} \\ \scriptsize{±1.3}\end{tabular}
& \begin{tabular}[c]{@{}c@{}} {35.4} \vspace{-0.1cm} \\ \scriptsize{±0.5}\end{tabular}
& \begin{tabular}[c]{@{}c@{}} {38.8} \vspace{-0.1cm} \\ \scriptsize{±2.0}\end{tabular}
& \begin{tabular}[c]{@{}c@{}} {36.2} \vspace{-0.1cm} \\ \scriptsize{±0.2}\end{tabular}
& \begin{tabular}[c]{@{}c@{}} {35.3} \vspace{-0.1cm} \\ \scriptsize{±1.4}\end{tabular}
& \begin{tabular}[c]{@{}c@{}} {37.7} \vspace{-0.1cm} \\ \scriptsize{±1.2}\end{tabular}
\\
{af$\rightarrow$}
& \begin{tabular}[c]{@{}c@{}} \colorbox{mygreen}{95.2} \vspace{-0.1cm} \\ \scriptsize{±0.1}\end{tabular}
& \begin{tabular}[c]{@{}c@{}} \colorbox{myblue}{91.0} \vspace{-0.1cm} \\ \scriptsize{±0.3}\end{tabular}
& \begin{tabular}[c]{@{}c@{}} \colorbox{mygreen}{44.7} \vspace{-0.1cm} \\ \scriptsize{±2.2}\end{tabular}
& \begin{tabular}[c]{@{}c@{}} \colorbox{mygreen}{42.4} \vspace{-0.1cm} \\ \scriptsize{±0.4}\end{tabular}
& \begin{tabular}[c]{@{}c@{}} {36.4} \vspace{-0.1cm} \\ \scriptsize{±1.1}\end{tabular}
& \begin{tabular}[c]{@{}c@{}} {33.6} \vspace{-0.1cm} \\ \scriptsize{±0.8}\end{tabular}
& \begin{tabular}[c]{@{}c@{}} \colorbox{mygreen}{46.7} \vspace{-0.1cm} \\ \scriptsize{±0.3}\end{tabular}
& \begin{tabular}[c]{@{}c@{}} \colorbox{mygreen}{40.1} \vspace{-0.1cm} \\ \scriptsize{±1.4}\end{tabular}
& \begin{tabular}[c]{@{}c@{}} \colorbox{mygreen}{42.2} \vspace{-0.1cm} \\ \scriptsize{±1.0}\end{tabular}
& \begin{tabular}[c]{@{}c@{}} \colorbox{mygreen}{42.6} \vspace{-0.1cm} \\ \scriptsize{±0.8}\end{tabular}
& \begin{tabular}[c]{@{}c@{}} \colorbox{mygreen}{43.3} \vspace{-0.1cm} \\ \scriptsize{±0.2}\end{tabular}
& \begin{tabular}[c]{@{}c@{}} {42.3} \vspace{-0.1cm} \\ \scriptsize{±1.0}\end{tabular}
\\
{zu$\rightarrow$}
& \begin{tabular}[c]{@{}c@{}} {95.1} \vspace{-0.1cm} \\ \scriptsize{±0.3}\end{tabular}
& \begin{tabular}[c]{@{}c@{}} \colorbox{mygreen}{89.2} \vspace{-0.1cm} \\ \scriptsize{±0.7}\end{tabular}
& \begin{tabular}[c]{@{}c@{}} \colorbox{myblue}{43.3} \vspace{-0.1cm} \\ \scriptsize{±0.9}\end{tabular}
& \begin{tabular}[c]{@{}c@{}} \colorbox{mygreen}{41.6} \vspace{-0.1cm} \\ \scriptsize{±0.3}\end{tabular}
& \begin{tabular}[c]{@{}c@{}} {33.8} \vspace{-0.1cm} \\ \scriptsize{±1.5}\end{tabular}
& \begin{tabular}[c]{@{}c@{}} {30.9} \vspace{-0.1cm} \\ \scriptsize{±1.2}\end{tabular}
& \begin{tabular}[c]{@{}c@{}} \colorbox{mygreen}{44.3} \vspace{-0.1cm} \\ \scriptsize{±1.5}\end{tabular}
& \begin{tabular}[c]{@{}c@{}} {38.9} \vspace{-0.1cm} \\ \scriptsize{±1.2}\end{tabular}
& \begin{tabular}[c]{@{}c@{}} \colorbox{mygreen}{41.1} \vspace{-0.1cm} \\ \scriptsize{±1.3}\end{tabular}
& \begin{tabular}[c]{@{}c@{}} \colorbox{mygreen}{39.5} \vspace{-0.1cm} \\ \scriptsize{±0.3}\end{tabular}
& \begin{tabular}[c]{@{}c@{}} \colorbox{mygreen}{40.9} \vspace{-0.1cm} \\ \scriptsize{±1.2}\end{tabular}
& \begin{tabular}[c]{@{}c@{}} {40.4} \vspace{-0.1cm} \\ \scriptsize{±1.1}\end{tabular}
\\
{xh$\rightarrow$}
& \begin{tabular}[c]{@{}c@{}} {94.3} \vspace{-0.1cm} \\ \scriptsize{±0.4}\end{tabular}
& \begin{tabular}[c]{@{}c@{}} \colorbox{mygreen}{89.0} \vspace{-0.1cm} \\ \scriptsize{±0.1}\end{tabular}
& \begin{tabular}[c]{@{}c@{}} \colorbox{mygreen}{43.0} \vspace{-0.1cm} \\ \scriptsize{±0.7}\end{tabular}
& \begin{tabular}[c]{@{}c@{}} \colorbox{myblue}{42.0} \vspace{-0.1cm} \\ \scriptsize{±1.6}\end{tabular}
& \begin{tabular}[c]{@{}c@{}} {31.7} \vspace{-0.1cm} \\ \scriptsize{±1.1}\end{tabular}
& \begin{tabular}[c]{@{}c@{}} {29.7} \vspace{-0.1cm} \\ \scriptsize{±2.5}\end{tabular}
& \begin{tabular}[c]{@{}c@{}} \colorbox{mygreen}{41.1} \vspace{-0.1cm} \\ \scriptsize{±0.6}\end{tabular}
& \begin{tabular}[c]{@{}c@{}} {37.5} \vspace{-0.1cm} \\ \scriptsize{±1.5}\end{tabular}
& \begin{tabular}[c]{@{}c@{}} {39.7} \vspace{-0.1cm} \\ \scriptsize{±2.0}\end{tabular}
& \begin{tabular}[c]{@{}c@{}} \colorbox{mygreen}{38.0} \vspace{-0.1cm} \\ \scriptsize{±0.6}\end{tabular}
& \begin{tabular}[c]{@{}c@{}} \colorbox{mygreen}{39.4} \vspace{-0.1cm} \\ \scriptsize{±0.8}\end{tabular}
& \begin{tabular}[c]{@{}c@{}} {39.6} \vspace{-0.1cm} \\ \scriptsize{±1.3}\end{tabular}
\\
{am$\rightarrow$}
& \begin{tabular}[c]{@{}c@{}} {95.5} \vspace{-0.1cm} \\ \scriptsize{±0.5}\end{tabular}
& \begin{tabular}[c]{@{}c@{}} \colorbox{mygreen}{91.2} \vspace{-0.1cm} \\ \scriptsize{±0.4}\end{tabular}
& \begin{tabular}[c]{@{}c@{}} \colorbox{mygreen}{45.9} \vspace{-0.1cm} \\ \scriptsize{±1.9}\end{tabular}
& \begin{tabular}[c]{@{}c@{}} \colorbox{mygreen}{43.2} \vspace{-0.1cm} \\ \scriptsize{±0.9}\end{tabular}
& \begin{tabular}[c]{@{}c@{}} \colorbox{myblue}{53.9} \vspace{-0.1cm} \\ \scriptsize{±2.0}\end{tabular}
& \begin{tabular}[c]{@{}c@{}} {34.9} \vspace{-0.1cm} \\ \scriptsize{±0.9}\end{tabular}
& \begin{tabular}[c]{@{}c@{}} \colorbox{mygreen}{47.2} \vspace{-0.1cm} \\ \scriptsize{±0.8}\end{tabular}
& \begin{tabular}[c]{@{}c@{}} \colorbox{mygreen}{42.5} \vspace{-0.1cm} \\ \scriptsize{±0.7}\end{tabular}
& \begin{tabular}[c]{@{}c@{}} \colorbox{mygreen}{42.5} \vspace{-0.1cm} \\ \scriptsize{±1.1}\end{tabular}
& \begin{tabular}[c]{@{}c@{}} \colorbox{mygreen}{44.3} \vspace{-0.1cm} \\ \scriptsize{±1.3}\end{tabular}
& \begin{tabular}[c]{@{}c@{}} \colorbox{mygreen}{44.7} \vspace{-0.1cm} \\ \scriptsize{±1.0}\end{tabular}
& \begin{tabular}[c]{@{}c@{}} \colorbox{mygreen}{44.9} \vspace{-0.1cm} \\ \scriptsize{±1.1}\end{tabular}
\\
{bm$\rightarrow$}
& \begin{tabular}[c]{@{}c@{}} {94.4} \vspace{-0.1cm} \\ \scriptsize{±0.5}\end{tabular}
& \begin{tabular}[c]{@{}c@{}} {84.9} \vspace{-0.1cm} \\ \scriptsize{±0.3}\end{tabular}
& \begin{tabular}[c]{@{}c@{}} {32.7} \vspace{-0.1cm} \\ \scriptsize{±0.8}\end{tabular}
& \begin{tabular}[c]{@{}c@{}} {30.3} \vspace{-0.1cm} \\ \scriptsize{±1.1}\end{tabular}
& \begin{tabular}[c]{@{}c@{}} {26.8} \vspace{-0.1cm} \\ \scriptsize{±1.8}\end{tabular}
& \begin{tabular}[c]{@{}c@{}} {29.0} \vspace{-0.1cm} \\ \scriptsize{±0.3}\end{tabular}
& \begin{tabular}[c]{@{}c@{}} {33.0} \vspace{-0.1cm} \\ \scriptsize{±2.7}\end{tabular}
& \begin{tabular}[c]{@{}c@{}} {31.3} \vspace{-0.1cm} \\ \scriptsize{±1.2}\end{tabular}
& \begin{tabular}[c]{@{}c@{}} {32.1} \vspace{-0.1cm} \\ \scriptsize{±0.5}\end{tabular}
& \begin{tabular}[c]{@{}c@{}} {30.1} \vspace{-0.1cm} \\ \scriptsize{±1.2}\end{tabular}
& \begin{tabular}[c]{@{}c@{}} {32.2} \vspace{-0.1cm} \\ \scriptsize{±0.6}\end{tabular}
& \begin{tabular}[c]{@{}c@{}} {29.7} \vspace{-0.1cm} \\ \scriptsize{±1.8}\end{tabular}
\\
{ig$\rightarrow$}
& \begin{tabular}[c]{@{}c@{}} {94.5} \vspace{-0.1cm} \\ \scriptsize{±0.2}\end{tabular}
& \begin{tabular}[c]{@{}c@{}} \colorbox{mygreen}{90.4} \vspace{-0.1cm} \\ \scriptsize{±0.1}\end{tabular}
& \begin{tabular}[c]{@{}c@{}} \colorbox{mygreen}{40.6} \vspace{-0.1cm} \\ \scriptsize{±1.4}\end{tabular}
& \begin{tabular}[c]{@{}c@{}} \colorbox{mygreen}{41.1} \vspace{-0.1cm} \\ \scriptsize{±0.3}\end{tabular}
& \begin{tabular}[c]{@{}c@{}} \colorbox{mygreen}{37.8} \vspace{-0.1cm} \\ \scriptsize{±0.3}\end{tabular}
& \begin{tabular}[c]{@{}c@{}} {29.8} \vspace{-0.1cm} \\ \scriptsize{±0.8}\end{tabular}
& \begin{tabular}[c]{@{}c@{}} \colorbox{myblue}{48.4} \vspace{-0.1cm} \\ \scriptsize{±1.8}\end{tabular}
& \begin{tabular}[c]{@{}c@{}} {37.7} \vspace{-0.1cm} \\ \scriptsize{±0.4}\end{tabular}
& \begin{tabular}[c]{@{}c@{}} {40.2} \vspace{-0.1cm} \\ \scriptsize{±2.8}\end{tabular}
& \begin{tabular}[c]{@{}c@{}} \colorbox{mygreen}{39.3} \vspace{-0.1cm} \\ \scriptsize{±1.3}\end{tabular}
& \begin{tabular}[c]{@{}c@{}} \colorbox{mygreen}{37.7} \vspace{-0.1cm} \\ \scriptsize{±0.4}\end{tabular}
& \begin{tabular}[c]{@{}c@{}} {38.8} \vspace{-0.1cm} \\ \scriptsize{±1.7}\end{tabular}
\\
{nso$\rightarrow$}
& \begin{tabular}[c]{@{}c@{}} {94.6} \vspace{-0.1cm} \\ \scriptsize{±0.2}\end{tabular}
& \begin{tabular}[c]{@{}c@{}} \colorbox{mygreen}{89.1} \vspace{-0.1cm} \\ \scriptsize{±1.0}\end{tabular}
& \begin{tabular}[c]{@{}c@{}} \colorbox{mygreen}{44.5} \vspace{-0.1cm} \\ \scriptsize{±1.0}\end{tabular}
& \begin{tabular}[c]{@{}c@{}} \colorbox{mygreen}{44.2} \vspace{-0.1cm} \\ \scriptsize{±0.6}\end{tabular}
& \begin{tabular}[c]{@{}c@{}} {33.8} \vspace{-0.1cm} \\ \scriptsize{±0.1}\end{tabular}
& \begin{tabular}[c]{@{}c@{}} {30.2} \vspace{-0.1cm} \\ \scriptsize{±0.5}\end{tabular}
& \begin{tabular}[c]{@{}c@{}} \colorbox{mygreen}{44.5} \vspace{-0.1cm} \\ \scriptsize{±1.0}\end{tabular}
& \begin{tabular}[c]{@{}c@{}} \colorbox{myblue}{44.9} \vspace{-0.1cm} \\ \scriptsize{±0.7}\end{tabular}
& \begin{tabular}[c]{@{}c@{}} \colorbox{mygreen}{42.6} \vspace{-0.1cm} \\ \scriptsize{±1.0}\end{tabular}
& \begin{tabular}[c]{@{}c@{}} \colorbox{mygreen}{45.1} \vspace{-0.1cm} \\ \scriptsize{±1.2}\end{tabular}
& \begin{tabular}[c]{@{}c@{}} \colorbox{mygreen}{45.8} \vspace{-0.1cm} \\ \scriptsize{±1.4}\end{tabular}
& \begin{tabular}[c]{@{}c@{}} {43.6} \vspace{-0.1cm} \\ \scriptsize{±1.5}\end{tabular}
\\
{sn$\rightarrow$}
& \begin{tabular}[c]{@{}c@{}} \colorbox{mygreen}{95.2} \vspace{-0.1cm} \\ \scriptsize{±0.3}\end{tabular}
& \begin{tabular}[c]{@{}c@{}} \colorbox{mygreen}{90.5} \vspace{-0.1cm} \\ \scriptsize{±0.5}\end{tabular}
& \begin{tabular}[c]{@{}c@{}} \colorbox{mygreen}{44.2} \vspace{-0.1cm} \\ \scriptsize{±1.6}\end{tabular}
& \begin{tabular}[c]{@{}c@{}} \colorbox{mygreen}{43.3} \vspace{-0.1cm} \\ \scriptsize{±1.1}\end{tabular}
& \begin{tabular}[c]{@{}c@{}} {34.7} \vspace{-0.1cm} \\ \scriptsize{±1.1}\end{tabular}
& \begin{tabular}[c]{@{}c@{}} {33.3} \vspace{-0.1cm} \\ \scriptsize{±1.6}\end{tabular}
& \begin{tabular}[c]{@{}c@{}} \colorbox{mygreen}{43.0} \vspace{-0.1cm} \\ \scriptsize{±0.8}\end{tabular}
& \begin{tabular}[c]{@{}c@{}} \colorbox{mygreen}{42.6} \vspace{-0.1cm} \\ \scriptsize{±1.5}\end{tabular}
& \begin{tabular}[c]{@{}c@{}} \colorbox{myblue}{46.3} \vspace{-0.1cm} \\ \scriptsize{±0.8}\end{tabular}
& \begin{tabular}[c]{@{}c@{}} \colorbox{mygreen}{43.1} \vspace{-0.1cm} \\ \scriptsize{±0.4}\end{tabular}
& \begin{tabular}[c]{@{}c@{}} \colorbox{mygreen}{45.7} \vspace{-0.1cm} \\ \scriptsize{±1.0}\end{tabular}
& \begin{tabular}[c]{@{}c@{}} \colorbox{mygreen}{45.3} \vspace{-0.1cm} \\ \scriptsize{±1.2}\end{tabular}
\\
{st$\rightarrow$}
& \begin{tabular}[c]{@{}c@{}} {94.6} \vspace{-0.1cm} \\ \scriptsize{±0.2}\end{tabular}
& \begin{tabular}[c]{@{}c@{}} \colorbox{mygreen}{89.7} \vspace{-0.1cm} \\ \scriptsize{±0.4}\end{tabular}
& \begin{tabular}[c]{@{}c@{}} \colorbox{mygreen}{41.7} \vspace{-0.1cm} \\ \scriptsize{±2.3}\end{tabular}
& \begin{tabular}[c]{@{}c@{}} \colorbox{mygreen}{42.4} \vspace{-0.1cm} \\ \scriptsize{±1.1}\end{tabular}
& \begin{tabular}[c]{@{}c@{}} \colorbox{mygreen}{36.1} \vspace{-0.1cm} \\ \scriptsize{±0.4}\end{tabular}
& \begin{tabular}[c]{@{}c@{}} {32.2} \vspace{-0.1cm} \\ \scriptsize{±1.2}\end{tabular}
& \begin{tabular}[c]{@{}c@{}} \colorbox{mygreen}{43.7} \vspace{-0.1cm} \\ \scriptsize{±0.3}\end{tabular}
& \begin{tabular}[c]{@{}c@{}} \colorbox{mygreen}{42.3} \vspace{-0.1cm} \\ \scriptsize{±0.7}\end{tabular}
& \begin{tabular}[c]{@{}c@{}} {43.0} \vspace{-0.1cm} \\ \scriptsize{±3.0}\end{tabular}
& \begin{tabular}[c]{@{}c@{}} \colorbox{myblue}{42.3} \vspace{-0.1cm} \\ \scriptsize{±0.6}\end{tabular}
& \begin{tabular}[c]{@{}c@{}} \colorbox{mygreen}{44.4} \vspace{-0.1cm} \\ \scriptsize{±0.7}\end{tabular}
& \begin{tabular}[c]{@{}c@{}} {42.0} \vspace{-0.1cm} \\ \scriptsize{±2.4}\end{tabular}
\\
{tn$\rightarrow$}
& \begin{tabular}[c]{@{}c@{}} \colorbox{mygreen}{95.1} \vspace{-0.1cm} \\ \scriptsize{±0.2}\end{tabular}
& \begin{tabular}[c]{@{}c@{}} \colorbox{mygreen}{89.3} \vspace{-0.1cm} \\ \scriptsize{±0.2}\end{tabular}
& \begin{tabular}[c]{@{}c@{}} \colorbox{mygreen}{44.3} \vspace{-0.1cm} \\ \scriptsize{±1.6}\end{tabular}
& \begin{tabular}[c]{@{}c@{}} \colorbox{mygreen}{44.4} \vspace{-0.1cm} \\ \scriptsize{±1.5}\end{tabular}
& \begin{tabular}[c]{@{}c@{}} \colorbox{mygreen}{39.7} \vspace{-0.1cm} \\ \scriptsize{±1.0}\end{tabular}
& \begin{tabular}[c]{@{}c@{}} {31.6} \vspace{-0.1cm} \\ \scriptsize{±0.9}\end{tabular}
& \begin{tabular}[c]{@{}c@{}} \colorbox{mygreen}{46.2} \vspace{-0.1cm} \\ \scriptsize{±1.1}\end{tabular}
& \begin{tabular}[c]{@{}c@{}} \colorbox{mygreen}{44.8} \vspace{-0.1cm} \\ \scriptsize{±1.2}\end{tabular}
& \begin{tabular}[c]{@{}c@{}} \colorbox{mygreen}{43.6} \vspace{-0.1cm} \\ \scriptsize{±0.5}\end{tabular}
& \begin{tabular}[c]{@{}c@{}} \colorbox{mygreen}{47.8} \vspace{-0.1cm} \\ \scriptsize{±0.9}\end{tabular}
& \begin{tabular}[c]{@{}c@{}} \colorbox{myblue}{44.5} \vspace{-0.1cm} \\ \scriptsize{±0.4}\end{tabular}
& \begin{tabular}[c]{@{}c@{}} \colorbox{mygreen}{44.0} \vspace{-0.1cm} \\ \scriptsize{±0.2}\end{tabular}
\\
{ts$\rightarrow$}
& \begin{tabular}[c]{@{}c@{}} {93.9} \vspace{-0.1cm} \\ \scriptsize{±0.8}\end{tabular}
& \begin{tabular}[c]{@{}c@{}} {85.1} \vspace{-0.1cm} \\ \scriptsize{±0.3}\end{tabular}
& \begin{tabular}[c]{@{}c@{}} {38.2} \vspace{-0.1cm} \\ \scriptsize{±0.8}\end{tabular}
& \begin{tabular}[c]{@{}c@{}} {37.2} \vspace{-0.1cm} \\ \scriptsize{±0.6}\end{tabular}
& \begin{tabular}[c]{@{}c@{}} {29.4} \vspace{-0.1cm} \\ \scriptsize{±1.0}\end{tabular}
& \begin{tabular}[c]{@{}c@{}} {29.3} \vspace{-0.1cm} \\ \scriptsize{±0.9}\end{tabular}
& \begin{tabular}[c]{@{}c@{}} {37.3} \vspace{-0.1cm} \\ \scriptsize{±0.7}\end{tabular}
& \begin{tabular}[c]{@{}c@{}} {34.0} \vspace{-0.1cm} \\ \scriptsize{±1.7}\end{tabular}
& \begin{tabular}[c]{@{}c@{}} {37.5} \vspace{-0.1cm} \\ \scriptsize{±0.7}\end{tabular}
& \begin{tabular}[c]{@{}c@{}} {36.6} \vspace{-0.1cm} \\ \scriptsize{±0.7}\end{tabular}
& \begin{tabular}[c]{@{}c@{}} {36.8} \vspace{-0.1cm} \\ \scriptsize{±1.8}\end{tabular}
& \begin{tabular}[c]{@{}c@{}} {37.9} \vspace{-0.1cm} \\ \scriptsize{±0.9}\end{tabular}
\\   
    \hline \hline

\end{tabular}
\caption{\textbf{Results of Cross-lingual Transfer using Llama 3 70B Instruct when Tuning on Winogrande and Testing on Belebele}. Llama 3 70B Instruct was independently fine-tuned on the 11 African languages of focus:  Afrikaans (af), Zulu (zu),
Xhosa (xh) (datasets for these three languages were sourced from (BMGF 2024)), Amharic (am), Bambara (bm), Igbo (ig),
Sepedi (nso), Shona (sn), Sesotho (st), Setswana (tn), and Tsonga (ts). Llama 3 70B Instruct was also fine-tuned on English (en) for reference. The translated \texttt{Winogrande} training set (small) was used for fine-tuning. The fine-tuned models were evaluated for their cross-lingual transfer performance on \texttt{Belebele} (reading comprehension task). Columns indicate the target language of the evaluation, while the rows indicate the source language the models were fine-tuned with (e.g. ``zu$\rightarrow$" indicates models fine-tuned with data in Zulu). All numbers are the mean 0-shot performance accuracy of three evaluations followed by the standard deviation (±). Model fine-tuning that yielded mono-lingual improvements (relative to the baseline) more than two standard deviations are in \colorbox{myblue}{blue}, while cross-lingual improvements (relative to the baseline) more than two standard deviations are in \colorbox{mygreen}{green}.\\} \label{table:perf-crosslingual-llama3-tr-wino-ts-bele}
\end{table*}

\begin{table*}[t!]
\centering
% \small

\setlength{\tabcolsep}{5.5pt}
\begin{tabular}{r|llllllllllll}
                    \hline \hline
                     & \multicolumn{12}{c}{\texttt{Belebele}} \\
                     % & \multicolumn{4}{c|}{\begin{tabular}[c]{@{}c@{}}\texttt{MMLU-Clinical-ZA} \\ "clinical knowledge"\end{tabular}} 
                     % & \multicolumn{4}{c}{\texttt{Belebele}} \\
                     & \textbf{en} & \textbf{af} & \textbf{zu} & \textbf{xh} 
                     & \textbf{am} & \textbf{bm} & \textbf{ig} & \textbf{nso} 
                     & \textbf{sn} & \textbf{st} & \textbf{tn} & \textbf{ts} \\ \hline 

\begin{tabular}[c]{@{}c@{}}Base \\ \tiny{Llama 3} \vspace{-0.1cm}\\ \tiny{70B IT} \end{tabular} 
& 94.6 & 84.8 & 36.7 & 36.8 & 35.1 & 35.6 & 35.9 & 37.1 & 38.1 & 35.9 & 36.4 & 40.7
\\ \hline
\multicolumn{12}{l}{\begin{tabular}[c]{@{}c@{}}\texttt{MMLU College Medicine} \\ dev + test + val\end{tabular}} \\
{en$\rightarrow$}
& \begin{tabular}[c]{@{}c@{}} {94.6} \vspace{-0.1cm} \\ \scriptsize{±0.2}\end{tabular}
& \begin{tabular}[c]{@{}c@{}} \colorbox{mygreen}{90.1} \vspace{-0.1cm} \\ \scriptsize{±0.5}\end{tabular}
& \begin{tabular}[c]{@{}c@{}} {36.1} \vspace{-0.1cm} \\ \scriptsize{±1.3}\end{tabular}
& \begin{tabular}[c]{@{}c@{}} {36.5} \vspace{-0.1cm} \\ \scriptsize{±0.7}\end{tabular}
& \begin{tabular}[c]{@{}c@{}} {31.3} \vspace{-0.1cm} \\ \scriptsize{±1.2}\end{tabular}
& \begin{tabular}[c]{@{}c@{}} {33.6} \vspace{-0.1cm} \\ \scriptsize{±0.5}\end{tabular}
& \begin{tabular}[c]{@{}c@{}} \colorbox{mygreen}{38.6} \vspace{-0.1cm} \\ \scriptsize{±0.8}\end{tabular}
& \begin{tabular}[c]{@{}c@{}} {36.0} \vspace{-0.1cm} \\ \scriptsize{±0.6}\end{tabular}
& \begin{tabular}[c]{@{}c@{}} {38.5} \vspace{-0.1cm} \\ \scriptsize{±2.7}\end{tabular}
& \begin{tabular}[c]{@{}c@{}} {34.9} \vspace{-0.1cm} \\ \scriptsize{±1.7}\end{tabular}
& \begin{tabular}[c]{@{}c@{}} {37.6} \vspace{-0.1cm} \\ \scriptsize{±1.1}\end{tabular}
& \begin{tabular}[c]{@{}c@{}} {38.2} \vspace{-0.1cm} \\ \scriptsize{±0.5}\end{tabular}
\\
{af$\rightarrow$}
& \begin{tabular}[c]{@{}c@{}} {94.8} \vspace{-0.1cm} \\ \scriptsize{±0.1}\end{tabular}
& \begin{tabular}[c]{@{}c@{}} \colorbox{myblue}{90.9} \vspace{-0.1cm} \\ \scriptsize{±0.5}\end{tabular}
& \begin{tabular}[c]{@{}c@{}} \colorbox{mygreen}{42.6} \vspace{-0.1cm} \\ \scriptsize{±0.1}\end{tabular}
& \begin{tabular}[c]{@{}c@{}} {38.8} \vspace{-0.1cm} \\ \scriptsize{±1.5}\end{tabular}
& \begin{tabular}[c]{@{}c@{}} {35.3} \vspace{-0.1cm} \\ \scriptsize{±1.1}\end{tabular}
& \begin{tabular}[c]{@{}c@{}} {30.8} \vspace{-0.1cm} \\ \scriptsize{±1.2}\end{tabular}
& \begin{tabular}[c]{@{}c@{}} \colorbox{mygreen}{44.1} \vspace{-0.1cm} \\ \scriptsize{±1.4}\end{tabular}
& \begin{tabular}[c]{@{}c@{}} \colorbox{mygreen}{40.2} \vspace{-0.1cm} \\ \scriptsize{±1.2}\end{tabular}
& \begin{tabular}[c]{@{}c@{}} {40.5} \vspace{-0.1cm} \\ \scriptsize{±2.2}\end{tabular}
& \begin{tabular}[c]{@{}c@{}} \colorbox{mygreen}{40.2} \vspace{-0.1cm} \\ \scriptsize{±0.5}\end{tabular}
& \begin{tabular}[c]{@{}c@{}} \colorbox{mygreen}{42.3} \vspace{-0.1cm} \\ \scriptsize{±1.4}\end{tabular}
& \begin{tabular}[c]{@{}c@{}} {41.2} \vspace{-0.1cm} \\ \scriptsize{±0.5}\end{tabular}
\\
{zu$\rightarrow$}
& \begin{tabular}[c]{@{}c@{}} {94.7} \vspace{-0.1cm} \\ \scriptsize{±0.1}\end{tabular}
& \begin{tabular}[c]{@{}c@{}} \colorbox{mygreen}{90.3} \vspace{-0.1cm} \\ \scriptsize{±0.1}\end{tabular}
& \begin{tabular}[c]{@{}c@{}} \colorbox{myblue}{45.1} \vspace{-0.1cm} \\ \scriptsize{±0.5}\end{tabular}
& \begin{tabular}[c]{@{}c@{}} \colorbox{mygreen}{43.9} \vspace{-0.1cm} \\ \scriptsize{±1.1}\end{tabular}
& \begin{tabular}[c]{@{}c@{}} \colorbox{mygreen}{40.6} \vspace{-0.1cm} \\ \scriptsize{±0.7}\end{tabular}
& \begin{tabular}[c]{@{}c@{}} {31.5} \vspace{-0.1cm} \\ \scriptsize{±1.5}\end{tabular}
& \begin{tabular}[c]{@{}c@{}} \colorbox{mygreen}{45.0} \vspace{-0.1cm} \\ \scriptsize{±0.9}\end{tabular}
& \begin{tabular}[c]{@{}c@{}} \colorbox{mygreen}{41.9} \vspace{-0.1cm} \\ \scriptsize{±0.9}\end{tabular}
& \begin{tabular}[c]{@{}c@{}} \colorbox{mygreen}{44.3} \vspace{-0.1cm} \\ \scriptsize{±0.9}\end{tabular}
& \begin{tabular}[c]{@{}c@{}} \colorbox{mygreen}{41.2} \vspace{-0.1cm} \\ \scriptsize{±0.9}\end{tabular}
& \begin{tabular}[c]{@{}c@{}} \colorbox{mygreen}{42.5} \vspace{-0.1cm} \\ \scriptsize{±0.7}\end{tabular}
& \begin{tabular}[c]{@{}c@{}} {41.3} \vspace{-0.1cm} \\ \scriptsize{±1.1}\end{tabular}
\\
{xh$\rightarrow$}
& \begin{tabular}[c]{@{}c@{}} {94.7} \vspace{-0.1cm} \\ \scriptsize{±0.4}\end{tabular}
& \begin{tabular}[c]{@{}c@{}} \colorbox{mygreen}{91.0} \vspace{-0.1cm} \\ \scriptsize{±0.3}\end{tabular}
& \begin{tabular}[c]{@{}c@{}} \colorbox{mygreen}{46.4} \vspace{-0.1cm} \\ \scriptsize{±1.1}\end{tabular}
& \begin{tabular}[c]{@{}c@{}} \colorbox{myblue}{45.7} \vspace{-0.1cm} \\ \scriptsize{±1.4}\end{tabular}
& \begin{tabular}[c]{@{}c@{}} \colorbox{mygreen}{39.2} \vspace{-0.1cm} \\ \scriptsize{±0.5}\end{tabular}
& \begin{tabular}[c]{@{}c@{}} {31.3} \vspace{-0.1cm} \\ \scriptsize{±1.6}\end{tabular}
& \begin{tabular}[c]{@{}c@{}} \colorbox{mygreen}{46.3} \vspace{-0.1cm} \\ \scriptsize{±1.1}\end{tabular}
& \begin{tabular}[c]{@{}c@{}} \colorbox{mygreen}{42.6} \vspace{-0.1cm} \\ \scriptsize{±1.6}\end{tabular}
& \begin{tabular}[c]{@{}c@{}} \colorbox{mygreen}{44.0} \vspace{-0.1cm} \\ \scriptsize{±0.2}\end{tabular}
& \begin{tabular}[c]{@{}c@{}} \colorbox{mygreen}{42.6} \vspace{-0.1cm} \\ \scriptsize{±2.4}\end{tabular}
& \begin{tabular}[c]{@{}c@{}} \colorbox{mygreen}{45.8} \vspace{-0.1cm} \\ \scriptsize{±1.4}\end{tabular}
& \begin{tabular}[c]{@{}c@{}} \colorbox{mygreen}{42.7} \vspace{-0.1cm} \\ \scriptsize{±0.5}\end{tabular}
\\
{am$\rightarrow$}
& \begin{tabular}[c]{@{}c@{}} {94.5} \vspace{-0.1cm} \\ \scriptsize{±0.3}\end{tabular}
& \begin{tabular}[c]{@{}c@{}} \colorbox{mygreen}{90.1} \vspace{-0.1cm} \\ \scriptsize{±0.5}\end{tabular}
& \begin{tabular}[c]{@{}c@{}} \colorbox{mygreen}{42.9} \vspace{-0.1cm} \\ \scriptsize{±0.6}\end{tabular}
& \begin{tabular}[c]{@{}c@{}} \colorbox{mygreen}{42.5} \vspace{-0.1cm} \\ \scriptsize{±1.0}\end{tabular}
& \begin{tabular}[c]{@{}c@{}} \colorbox{myblue}{50.4} \vspace{-0.1cm} \\ \scriptsize{±0.3}\end{tabular}
& \begin{tabular}[c]{@{}c@{}} {33.1} \vspace{-0.1cm} \\ \scriptsize{±0.8}\end{tabular}
& \begin{tabular}[c]{@{}c@{}} \colorbox{mygreen}{44.1} \vspace{-0.1cm} \\ \scriptsize{±0.9}\end{tabular}
& \begin{tabular}[c]{@{}c@{}} \colorbox{mygreen}{39.0} \vspace{-0.1cm} \\ \scriptsize{±0.2}\end{tabular}
& \begin{tabular}[c]{@{}c@{}} \colorbox{mygreen}{42.2} \vspace{-0.1cm} \\ \scriptsize{±1.2}\end{tabular}
& \begin{tabular}[c]{@{}c@{}} \colorbox{mygreen}{40.5} \vspace{-0.1cm} \\ \scriptsize{±1.0}\end{tabular}
& \begin{tabular}[c]{@{}c@{}} \colorbox{mygreen}{41.3} \vspace{-0.1cm} \\ \scriptsize{±0.4}\end{tabular}
& \begin{tabular}[c]{@{}c@{}} {41.9} \vspace{-0.1cm} \\ \scriptsize{±0.7}\end{tabular}
\\
{bm$\rightarrow$}
& \begin{tabular}[c]{@{}c@{}} {94.3} \vspace{-0.1cm} \\ \scriptsize{±0.4}\end{tabular}
& \begin{tabular}[c]{@{}c@{}} \colorbox{mygreen}{89.2} \vspace{-0.1cm} \\ \scriptsize{±0.2}\end{tabular}
& \begin{tabular}[c]{@{}c@{}} {37.4} \vspace{-0.1cm} \\ \scriptsize{±0.7}\end{tabular}
& \begin{tabular}[c]{@{}c@{}} {37.7} \vspace{-0.1cm} \\ \scriptsize{±1.0}\end{tabular}
& \begin{tabular}[c]{@{}c@{}} {35.3} \vspace{-0.1cm} \\ \scriptsize{±0.8}\end{tabular}
& \begin{tabular}[c]{@{}c@{}} {32.8} \vspace{-0.1cm} \\ \scriptsize{±0.6}\end{tabular}
& \begin{tabular}[c]{@{}c@{}} \colorbox{mygreen}{40.9} \vspace{-0.1cm} \\ \scriptsize{±1.0}\end{tabular}
& \begin{tabular}[c]{@{}c@{}} {33.4} \vspace{-0.1cm} \\ \scriptsize{±2.2}\end{tabular}
& \begin{tabular}[c]{@{}c@{}} {37.2} \vspace{-0.1cm} \\ \scriptsize{±2.0}\end{tabular}
& \begin{tabular}[c]{@{}c@{}} {35.0} \vspace{-0.1cm} \\ \scriptsize{±0.8}\end{tabular}
& \begin{tabular}[c]{@{}c@{}} {37.8} \vspace{-0.1cm} \\ \scriptsize{±3.0}\end{tabular}
& \begin{tabular}[c]{@{}c@{}} {38.1} \vspace{-0.1cm} \\ \scriptsize{±0.5}\end{tabular}
\\
{ig$\rightarrow$}
& \begin{tabular}[c]{@{}c@{}} {94.6} \vspace{-0.1cm} \\ \scriptsize{±0.4}\end{tabular}
& \begin{tabular}[c]{@{}c@{}} \colorbox{mygreen}{91.5} \vspace{-0.1cm} \\ \scriptsize{±0.5}\end{tabular}
& \begin{tabular}[c]{@{}c@{}} \colorbox{mygreen}{44.3} \vspace{-0.1cm} \\ \scriptsize{±0.6}\end{tabular}
& \begin{tabular}[c]{@{}c@{}} \colorbox{mygreen}{44.0} \vspace{-0.1cm} \\ \scriptsize{±0.9}\end{tabular}
& \begin{tabular}[c]{@{}c@{}} \colorbox{mygreen}{42.7} \vspace{-0.1cm} \\ \scriptsize{±0.4}\end{tabular}
& \begin{tabular}[c]{@{}c@{}} {32.5} \vspace{-0.1cm} \\ \scriptsize{±1.0}\end{tabular}
& \begin{tabular}[c]{@{}c@{}} \colorbox{myblue}{48.3} \vspace{-0.1cm} \\ \scriptsize{±1.0}\end{tabular}
& \begin{tabular}[c]{@{}c@{}} \colorbox{mygreen}{43.0} \vspace{-0.1cm} \\ \scriptsize{±1.1}\end{tabular}
& \begin{tabular}[c]{@{}c@{}} \colorbox{mygreen}{45.5} \vspace{-0.1cm} \\ \scriptsize{±1.5}\end{tabular}
& \begin{tabular}[c]{@{}c@{}} \colorbox{mygreen}{44.1} \vspace{-0.1cm} \\ \scriptsize{±0.5}\end{tabular}
& \begin{tabular}[c]{@{}c@{}} \colorbox{mygreen}{45.9} \vspace{-0.1cm} \\ \scriptsize{±1.0}\end{tabular}
& \begin{tabular}[c]{@{}c@{}} \colorbox{mygreen}{43.1} \vspace{-0.1cm} \\ \scriptsize{±0.3}\end{tabular}
\\
{nso$\rightarrow$}
& \begin{tabular}[c]{@{}c@{}} {94.4} \vspace{-0.1cm} \\ \scriptsize{±0.6}\end{tabular}
& \begin{tabular}[c]{@{}c@{}} \colorbox{mygreen}{89.9} \vspace{-0.1cm} \\ \scriptsize{±0.7}\end{tabular}
& \begin{tabular}[c]{@{}c@{}} \colorbox{mygreen}{43.3} \vspace{-0.1cm} \\ \scriptsize{±1.5}\end{tabular}
& \begin{tabular}[c]{@{}c@{}} \colorbox{mygreen}{41.6} \vspace{-0.1cm} \\ \scriptsize{±1.2}\end{tabular}
& \begin{tabular}[c]{@{}c@{}} \colorbox{mygreen}{41.2} \vspace{-0.1cm} \\ \scriptsize{±0.9}\end{tabular}
& \begin{tabular}[c]{@{}c@{}} {31.4} \vspace{-0.1cm} \\ \scriptsize{±1.5}\end{tabular}
& \begin{tabular}[c]{@{}c@{}} \colorbox{mygreen}{43.2} \vspace{-0.1cm} \\ \scriptsize{±0.6}\end{tabular}
& \begin{tabular}[c]{@{}c@{}} \colorbox{myblue}{47.2} \vspace{-0.1cm} \\ \scriptsize{±1.4}\end{tabular}
& \begin{tabular}[c]{@{}c@{}} \colorbox{mygreen}{43.1} \vspace{-0.1cm} \\ \scriptsize{±1.4}\end{tabular}
& \begin{tabular}[c]{@{}c@{}} \colorbox{mygreen}{45.5} \vspace{-0.1cm} \\ \scriptsize{±0.4}\end{tabular}
& \begin{tabular}[c]{@{}c@{}} \colorbox{mygreen}{47.6} \vspace{-0.1cm} \\ \scriptsize{±1.8}\end{tabular}
& \begin{tabular}[c]{@{}c@{}} \colorbox{mygreen}{43.7} \vspace{-0.1cm} \\ \scriptsize{±0.8}\end{tabular}
\\
{sn$\rightarrow$}
& \begin{tabular}[c]{@{}c@{}} {94.8} \vspace{-0.1cm} \\ \scriptsize{±0.4}\end{tabular}
& \begin{tabular}[c]{@{}c@{}} \colorbox{mygreen}{90.2} \vspace{-0.1cm} \\ \scriptsize{±0.5}\end{tabular}
& \begin{tabular}[c]{@{}c@{}} \colorbox{mygreen}{45.9} \vspace{-0.1cm} \\ \scriptsize{±1.1}\end{tabular}
& \begin{tabular}[c]{@{}c@{}} \colorbox{mygreen}{43.8} \vspace{-0.1cm} \\ \scriptsize{±0.5}\end{tabular}
& \begin{tabular}[c]{@{}c@{}} \colorbox{mygreen}{41.1} \vspace{-0.1cm} \\ \scriptsize{±0.3}\end{tabular}
& \begin{tabular}[c]{@{}c@{}} {34.5} \vspace{-0.1cm} \\ \scriptsize{±2.4}\end{tabular}
& \begin{tabular}[c]{@{}c@{}} \colorbox{mygreen}{46.6} \vspace{-0.1cm} \\ \scriptsize{±0.4}\end{tabular}
& \begin{tabular}[c]{@{}c@{}} \colorbox{mygreen}{45.0} \vspace{-0.1cm} \\ \scriptsize{±1.4}\end{tabular}
& \begin{tabular}[c]{@{}c@{}} \colorbox{myblue}{50.9} \vspace{-0.1cm} \\ \scriptsize{±0.7}\end{tabular}
& \begin{tabular}[c]{@{}c@{}} \colorbox{mygreen}{44.4} \vspace{-0.1cm} \\ \scriptsize{±0.9}\end{tabular}
& \begin{tabular}[c]{@{}c@{}} \colorbox{mygreen}{46.4} \vspace{-0.1cm} \\ \scriptsize{±1.0}\end{tabular}
& \begin{tabular}[c]{@{}c@{}} \colorbox{mygreen}{45.3} \vspace{-0.1cm} \\ \scriptsize{±1.0}\end{tabular}
\\
{st$\rightarrow$}
& \begin{tabular}[c]{@{}c@{}} {94.5} \vspace{-0.1cm} \\ \scriptsize{±0.3}\end{tabular}
& \begin{tabular}[c]{@{}c@{}} \colorbox{mygreen}{91.0} \vspace{-0.1cm} \\ \scriptsize{±0.3}\end{tabular}
& \begin{tabular}[c]{@{}c@{}} \colorbox{mygreen}{45.3} \vspace{-0.1cm} \\ \scriptsize{±0.1}\end{tabular}
& \begin{tabular}[c]{@{}c@{}} \colorbox{mygreen}{44.4} \vspace{-0.1cm} \\ \scriptsize{±2.3}\end{tabular}
& \begin{tabular}[c]{@{}c@{}} \colorbox{mygreen}{42.2} \vspace{-0.1cm} \\ \scriptsize{±1.1}\end{tabular}
& \begin{tabular}[c]{@{}c@{}} {32.9} \vspace{-0.1cm} \\ \scriptsize{±0.4}\end{tabular}
& \begin{tabular}[c]{@{}c@{}} \colorbox{mygreen}{45.4} \vspace{-0.1cm} \\ \scriptsize{±1.2}\end{tabular}
& \begin{tabular}[c]{@{}c@{}} \colorbox{mygreen}{46.9} \vspace{-0.1cm} \\ \scriptsize{±1.1}\end{tabular}
& \begin{tabular}[c]{@{}c@{}} \colorbox{mygreen}{45.3} \vspace{-0.1cm} \\ \scriptsize{±1.4}\end{tabular}
& \begin{tabular}[c]{@{}c@{}} \colorbox{myblue}{49.6} \vspace{-0.1cm} \\ \scriptsize{±1.2}\end{tabular}
& \begin{tabular}[c]{@{}c@{}} \colorbox{mygreen}{49.4} \vspace{-0.1cm} \\ \scriptsize{±1.5}\end{tabular}
& \begin{tabular}[c]{@{}c@{}} \colorbox{mygreen}{44.4} \vspace{-0.1cm} \\ \scriptsize{±0.6}\end{tabular}
\\
{tn$\rightarrow$}
& \begin{tabular}[c]{@{}c@{}} {94.3} \vspace{-0.1cm} \\ \scriptsize{±0.4}\end{tabular}
& \begin{tabular}[c]{@{}c@{}} \colorbox{mygreen}{90.5} \vspace{-0.1cm} \\ \scriptsize{±0.5}\end{tabular}
& \begin{tabular}[c]{@{}c@{}} \colorbox{mygreen}{42.3} \vspace{-0.1cm} \\ \scriptsize{±0.8}\end{tabular}
& \begin{tabular}[c]{@{}c@{}} \colorbox{mygreen}{41.4} \vspace{-0.1cm} \\ \scriptsize{±1.5}\end{tabular}
& \begin{tabular}[c]{@{}c@{}} \colorbox{mygreen}{39.4} \vspace{-0.1cm} \\ \scriptsize{±2.1}\end{tabular}
& \begin{tabular}[c]{@{}c@{}} {31.0} \vspace{-0.1cm} \\ \scriptsize{±1.3}\end{tabular}
& \begin{tabular}[c]{@{}c@{}} \colorbox{mygreen}{45.6} \vspace{-0.1cm} \\ \scriptsize{±1.0}\end{tabular}
& \begin{tabular}[c]{@{}c@{}} \colorbox{mygreen}{44.1} \vspace{-0.1cm} \\ \scriptsize{±0.8}\end{tabular}
& \begin{tabular}[c]{@{}c@{}} \colorbox{mygreen}{43.0} \vspace{-0.1cm} \\ \scriptsize{±0.8}\end{tabular}
& \begin{tabular}[c]{@{}c@{}} \colorbox{mygreen}{44.4} \vspace{-0.1cm} \\ \scriptsize{±1.1}\end{tabular}
& \begin{tabular}[c]{@{}c@{}} \colorbox{myblue}{47.3} \vspace{-0.1cm} \\ \scriptsize{±0.4}\end{tabular}
& \begin{tabular}[c]{@{}c@{}} \colorbox{mygreen}{41.7} \vspace{-0.1cm} \\ \scriptsize{±0.4}\end{tabular}
\\
{ts$\rightarrow$}
& \begin{tabular}[c]{@{}c@{}} {94.5} \vspace{-0.1cm} \\ \scriptsize{±0.3}\end{tabular}
& \begin{tabular}[c]{@{}c@{}} \colorbox{mygreen}{91.1} \vspace{-0.1cm} \\ \scriptsize{±0.3}\end{tabular}
& \begin{tabular}[c]{@{}c@{}} \colorbox{mygreen}{44.6} \vspace{-0.1cm} \\ \scriptsize{±0.8}\end{tabular}
& \begin{tabular}[c]{@{}c@{}} \colorbox{mygreen}{43.3} \vspace{-0.1cm} \\ \scriptsize{±1.0}\end{tabular}
& \begin{tabular}[c]{@{}c@{}} \colorbox{mygreen}{40.6} \vspace{-0.1cm} \\ \scriptsize{±1.0}\end{tabular}
& \begin{tabular}[c]{@{}c@{}} {31.1} \vspace{-0.1cm} \\ \scriptsize{±1.1}\end{tabular}
& \begin{tabular}[c]{@{}c@{}} \colorbox{mygreen}{46.2} \vspace{-0.1cm} \\ \scriptsize{±0.9}\end{tabular}
& \begin{tabular}[c]{@{}c@{}} \colorbox{mygreen}{45.0} \vspace{-0.1cm} \\ \scriptsize{±1.2}\end{tabular}
& \begin{tabular}[c]{@{}c@{}} \colorbox{mygreen}{45.5} \vspace{-0.1cm} \\ \scriptsize{±0.3}\end{tabular}
& \begin{tabular}[c]{@{}c@{}} \colorbox{mygreen}{45.5} \vspace{-0.1cm} \\ \scriptsize{±1.3}\end{tabular}
& \begin{tabular}[c]{@{}c@{}} \colorbox{mygreen}{46.2} \vspace{-0.1cm} \\ \scriptsize{±0.6}\end{tabular}
& \begin{tabular}[c]{@{}c@{}} \colorbox{myblue}{48.2} \vspace{-0.1cm} \\ \scriptsize{±1.3}\end{tabular}
\\   
    \hline \hline

\end{tabular}
\caption{\textbf{Results of Cross-lingual Transfer using Llama 3 70B Instruct when Tuning on MMLU College Medicine and Testing on Belebele}. Llama 3 70B Instruct was independently fine-tuned on the 11 African languages of focus:  Afrikaans (af), Zulu (zu),
Xhosa (xh) (datasets for these three languages were sourced from (BMGF 2024)), Amharic (am), Bambara (bm), Igbo (ig),
Sepedi (nso), Shona (sn), Sesotho (st), Setswana (tn), and Tsonga (ts). Llama 3 70B Instruct was also fine-tuned on English (en) for reference. The translated \texttt{MMLU College Medicine} section (dev + test + val sets) was used for fine-tuning. The fine-tuned models were evaluated for their cross-lingual transfer performance on \texttt{Belebele} (reading comprehension task). Columns indicate the target language of the evaluation, while the rows indicate the source language the models were fine-tuned with (e.g. ``zu$\rightarrow$" indicates models fine-tuned with data in Zulu). All numbers are the mean 0-shot performance accuracy of three evaluations followed by the standard deviation (±). Model fine-tuning that yielded mono-lingual improvements (relative to the baseline) more than two standard deviations are in \colorbox{myblue}{blue}, while cross-lingual improvements (relative to the baseline) more than two standard deviations are in \colorbox{mygreen}{green}.\\} \label{table:perf-crosslingual-llama3-tr-mmlu-ts-bele}
\end{table*}

% Table: Data Quality x Quantity - Winogrande is Eval
\begin{table*}[]
\centering
\begin{tabular}{|l|c|ccccc|}
                            \hline
                           \multicolumn{1}{|c|}{\textbf{Fine-Tuning Dataset (language)}} & \multicolumn{1}{c|}{\textbf{Data Quality}} & \multicolumn{5}{c|}{\textbf{Data Quantity}} \\
                           & & 0\% & 25\%   & 50\%   & 75\%   & 100\%  \\ 
                           % & & (N/A) & (53; 17) & (107; 33) & (160; 50) & (213; 66) \\
                           \hline
\multicolumn{7}{|c|}{Evaluation set: \texttt{Winogrande}} \\ \hline
 \multicolumn{2}{|c|}{\textit{Samples}}& N/A & \textit{53} & \textit{107} & \textit{160} & \textit{213} \\ \hline
\multirow{2}{*}{Winogrande (en)} & \multicolumn{1}{c|}{Low}         
& \cellcolor{fgreen!4}61.2 \scriptsize{±{0.0}} & \cellcolor{fgreen!0}60.3 \scriptsize{±{0.9}} & \cellcolor{fgreen!90}82.4 \scriptsize{±{0.3}} & \cellcolor{fgreen!94}83.4 \scriptsize{±{0.5}} & \cellcolor{fgreen!100}84.9 \scriptsize{±{0.8}} \\
& High & \cellcolor{fgreen!4}61.2 \scriptsize{±{0.0}} & \cellcolor{fgreen!79}79.7 \scriptsize{±{0.1}} & \cellcolor{fgreen!91}82.6 \scriptsize{±{0.3}} & \cellcolor{fgreen!97}84.2 \scriptsize{±{0.4}} & \cellcolor{fgreen!100}84.8 \scriptsize{±{0.1}} \\
\hline
\multirow{2}{*}{Winogrande (af)} & \multicolumn{1}{c|}{Low}         
& \cellcolor{fgreen!0}51.0 \scriptsize{±{0.0}} & \cellcolor{fgreen!11}53.7 \scriptsize{±{1.1}} & \cellcolor{fgreen!56}64.3 \scriptsize{±{0.6}} & \cellcolor{fgreen!65}66.5 \scriptsize{±{0.7}} & \cellcolor{fgreen!69}67.5 \scriptsize{±{0.4}} \\
& High & \cellcolor{fgreen!0}51.0 \scriptsize{±{0.0}} & \cellcolor{fgreen!72}68.3 \scriptsize{±{0.5}} & \cellcolor{fgreen!88}72.0 \scriptsize{±{0.7}} & \cellcolor{fgreen!87}71.8 \scriptsize{±{0.4}} & \cellcolor{fgreen!100}74.9 \scriptsize{±{0.3}} \\
\hline
\multicolumn{2}{|c|}{\textit{Samples}}& N/A & \textit{17} & \textit{33} & \textit{50} & \textit{66} \\ \hline
\multirow{2}{*}{MMLU College Medicine (en)} & \multicolumn{1}{c|}{Low}         
& \cellcolor{fgreen!0}61.2 \scriptsize{±{0.0}} & \cellcolor{fgreen!13}62.3 \scriptsize{±{0.4}} & \cellcolor{fgreen!70}66.9 \scriptsize{±{0.5}} & \cellcolor{fgreen!12}62.2 \scriptsize{±{0.1}} & \cellcolor{fgreen!78}67.6 \scriptsize{±{0.4}} \\
& High & \cellcolor{fgreen!0}61.2 \scriptsize{±{0.0}} & \cellcolor{fgreen!44}64.8 \scriptsize{±{0.2}} & \cellcolor{fgreen!46}65.0 \scriptsize{±{0.5}} & \cellcolor{fgreen!66}66.6 \scriptsize{±{1.1}} & \cellcolor{fgreen!100}69.4 \scriptsize{±{0.4}} \\
\hline
\multirow{2}{*}{MMLU College Medicine (af)} & \multicolumn{1}{c|}{Low}         
& \cellcolor{fgreen!4}51.0 \scriptsize{±{0.0}} & \cellcolor{fgreen!26}52.9 \scriptsize{±{0.2}} & \cellcolor{fgreen!76}57.4 \scriptsize{±{0.3}} & \cellcolor{fgreen!66}56.5 \scriptsize{±{0.5}} & \cellcolor{fgreen!83}58.0 \scriptsize{±{0.5}} \\
& High & \cellcolor{fgreen!4}51.0 \scriptsize{±{0.0}} & \cellcolor{fgreen!0}50.6 \scriptsize{±{0.0}} & \cellcolor{fgreen!38}54.0 \scriptsize{±{0.3}} & \cellcolor{fgreen!65}56.4 \scriptsize{±{0.1}} & \cellcolor{fgreen!100}59.5 \scriptsize{±{0.6}} \\
\hline
\end{tabular}
\caption{\textbf{Results of Data Quality vs. Quantity using Llama 3 70B Instruct and Evaluating on Winogrande}. Llama 3 70B Instruct was fine-tuned on various GPT-4o LLM-as-an-Annotator quality and random-sampling quantity combination subsets of Winogrande train (small) and MMLU College Medicine where there was a mono-lingual lift of at least 5\% observed when using the full fine-tuning dataset (see Tables \ref{table:perf-crosslingual-llama3-tr-wino-ts-wino} and \ref{table:perf-crosslingual-llama3-tr-mmlu-ts-wino}). The fine-tuned models were then evaluated on the Winogrande test split in the same language as the fine-tuning dataset. Note that the exact fine-tuning dataset sample sizes when using Winogrande as the fine-tuning dataset (Wino) and MMLU College Medicine as the fine-tuning dataset (MMLU) are given above the performance results. Language codes are as follows: English (en), Afrikaans (af).}
\label{table:quality-x-quantity-ts-wino}
\end{table*}

% Table: Data Quality x Quantity - MMLU Clinical Knowledge is Eval
\begin{table*}[]
\centering
\begin{tabular}{|l|c|ccccc|}
                            \hline
                           \multicolumn{1}{|c|}{\textbf{Fine-Tuning Dataset (language)}} & \multicolumn{1}{c|}{\textbf{Data Quality}} & \multicolumn{5}{c|}{\textbf{Data Quantity}} \\
                           & & 0\% & 25\%   & 50\%   & 75\%   & 100\%  \\ 
                           % & & (N/A) & (53; 17) & (107; 33) & (160; 50) & (213; 66) \\
                           \hline
\multicolumn{7}{|c|}{Evaluation set: \texttt{MMLU Clinical Knowledge}} \\ \hline
 \multicolumn{2}{|c|}{\textit{Samples}}& N/A & \textit{53} & \textit{107} & \textit{160} & \textit{213} \\ \hline
\multirow{2}{*}{Winogrande (am)} & \multicolumn{1}{c|}{Low}         
& \cellcolor{fgreen!0}33.6 \scriptsize{±{0.0}} & \cellcolor{fgreen!72}44.3 \scriptsize{±{1.0}} & \cellcolor{fgreen!82}45.7 \scriptsize{±{2.0}} & \cellcolor{fgreen!76}44.8 \scriptsize{±{0.5}} & \cellcolor{fgreen!92}47.2 \scriptsize{±{1.6}} \\
& High & \cellcolor{fgreen!0}33.6 \scriptsize{±{0.0}} & \cellcolor{fgreen!73}44.4 \scriptsize{±{0.2}} & \cellcolor{fgreen!76}44.9 \scriptsize{±{0.4}} & \cellcolor{fgreen!100}48.4 \scriptsize{±{0.8}} & \cellcolor{fgreen!93}47.3 \scriptsize{±{0.2}} \\
\hline
\multirow{2}{*}{Winogrande (ig)} & \multicolumn{1}{c|}{Low}         
& \cellcolor{fgreen!0}33.6 \scriptsize{±{0.0}} & \cellcolor{fgreen!26}36.5 \scriptsize{±{4.5}} & \cellcolor{fgreen!100}44.8 \scriptsize{±{0.9}} & \cellcolor{fgreen!74}41.9 \scriptsize{±{3.5}} & \cellcolor{fgreen!63}40.6 \scriptsize{±{3.7}} \\
& High & \cellcolor{fgreen!0}33.6 \scriptsize{±{0.0}} & \cellcolor{fgreen!69}41.3 \scriptsize{±{2.4}} & \cellcolor{fgreen!83}42.9 \scriptsize{±{0.8}} & \cellcolor{fgreen!96}44.3 \scriptsize{±{1.8}} & \cellcolor{fgreen!82}42.8 \scriptsize{±{2.2}} \\
\hline
\multirow{2}{*}{Winogrande (nso)} & \multicolumn{1}{c|}{Low}         
& \cellcolor{fgreen!0}37.0 \scriptsize{±{0.0}} & \cellcolor{fgreen!24}38.9 \scriptsize{±{2.7}} & \cellcolor{fgreen!40}40.2 \scriptsize{±{2.2}} & \cellcolor{fgreen!65}42.2 \scriptsize{±{0.2}} & \cellcolor{fgreen!75}43.0 \scriptsize{±{1.7}} \\
& High & \cellcolor{fgreen!0}37.0 \scriptsize{±{0.0}} & \cellcolor{fgreen!81}43.5 \scriptsize{±{2.8}} & \cellcolor{fgreen!91}44.3 \scriptsize{±{1.2}} & \cellcolor{fgreen!62}42.0 \scriptsize{±{3.5}} & \cellcolor{fgreen!100}45.0 \scriptsize{±{0.9}} \\
\hline
\multicolumn{2}{|c|}{\textit{Samples}}& N/A & \textit{17} & \textit{33} & \textit{50} & \textit{66} \\ \hline
\multirow{2}{*}{MMLU College Medicine (af)} & \multicolumn{1}{c|}{Low}         
& \cellcolor{fgreen!60}71.3 \scriptsize{±{0.0}} & \cellcolor{fgreen!5}64.4 \scriptsize{±{3.1}} & \cellcolor{fgreen!0}63.8 \scriptsize{±{0.4}} & \cellcolor{fgreen!28}67.3 \scriptsize{±{2.7}} & \cellcolor{fgreen!58}71.1 \scriptsize{±{1.2}} \\
& High & \cellcolor{fgreen!60}71.3 \scriptsize{±{0.0}} & \cellcolor{fgreen!70}72.5 \scriptsize{±{2.5}} & \cellcolor{fgreen!93}75.4 \scriptsize{±{1.3}} & \cellcolor{fgreen!100}76.3 \scriptsize{±{1.7}} & \cellcolor{fgreen!94}75.6 \scriptsize{±{0.4}} \\
\hline
\multirow{2}{*}{MMLU College Medicine (zu)} & \multicolumn{1}{c|}{Low}         
& \cellcolor{fgreen!0}39.2 \scriptsize{±{0.0}} & \cellcolor{fgreen!49}44.6 \scriptsize{±{1.7}} & \cellcolor{fgreen!3}39.5 \scriptsize{±{1.7}} & \cellcolor{fgreen!30}42.5 \scriptsize{±{1.5}} & \cellcolor{fgreen!21}41.5 \scriptsize{±{1.6}} \\
& High & \cellcolor{fgreen!0}39.2 \scriptsize{±{0.0}} & \cellcolor{fgreen!12}40.5 \scriptsize{±{0.9}} & \cellcolor{fgreen!44}44.0 \scriptsize{±{2.4}} & \cellcolor{fgreen!96}49.8 \scriptsize{±{2.6}} & \cellcolor{fgreen!100}50.2 \scriptsize{±{2.0}} \\
\hline
\multirow{2}{*}{MMLU College Medicine (xh)} & \multicolumn{1}{c|}{Low}         
& \cellcolor{fgreen!29}38.5 \scriptsize{±{0.0}} & \cellcolor{fgreen!54}42.2 \scriptsize{±{3.1}} & \cellcolor{fgreen!0}34.3 \scriptsize{±{1.3}} & \cellcolor{fgreen!44}40.7 \scriptsize{±{2.3}} & \cellcolor{fgreen!55}42.3 \scriptsize{±{1.7}} \\
& High & \cellcolor{fgreen!29}38.5 \scriptsize{±{0.0}} & \cellcolor{fgreen!38}39.9 \scriptsize{±{0.2}} & \cellcolor{fgreen!99}48.8 \scriptsize{±{1.9}} & \cellcolor{fgreen!91}47.6 \scriptsize{±{1.4}} & \cellcolor{fgreen!100}48.9 \scriptsize{±{2.1}} \\
\hline
\multirow{2}{*}{MMLU College Medicine (am)} & \multicolumn{1}{c|}{Low}         
& \cellcolor{fgreen!0}33.6 \scriptsize{±{0.0}} & \cellcolor{fgreen!62}44.8 \scriptsize{±{1.6}} & \cellcolor{fgreen!51}42.9 \scriptsize{±{2.1}} & \cellcolor{fgreen!68}45.9 \scriptsize{±{2.9}} & \cellcolor{fgreen!82}48.4 \scriptsize{±{1.3}} \\
& High & \cellcolor{fgreen!0}33.6 \scriptsize{±{0.0}} & \cellcolor{fgreen!81}48.2 \scriptsize{±{0.8}} & \cellcolor{fgreen!81}48.2 \scriptsize{±{1.7}} & \cellcolor{fgreen!98}51.3 \scriptsize{±{1.9}} & \cellcolor{fgreen!100}51.7 \scriptsize{±{2.7}} \\
\hline
\multirow{2}{*}{MMLU College Medicine (bm)} & \multicolumn{1}{c|}{Low}         
& \cellcolor{fgreen!6}32.5 \scriptsize{±{0.0}} & \cellcolor{fgreen!11}32.8 \scriptsize{±{0.8}} & \cellcolor{fgreen!42}34.9 \scriptsize{±{2.5}} & \cellcolor{fgreen!9}32.7 \scriptsize{±{0.8}} & \cellcolor{fgreen!27}33.9 \scriptsize{±{2.1}} \\
& High & \cellcolor{fgreen!6}32.5 \scriptsize{±{0.0}} & \cellcolor{fgreen!0}32.1 \scriptsize{±{3.3}} & \cellcolor{fgreen!23}33.6 \scriptsize{±{3.1}} & \cellcolor{fgreen!100}38.7 \scriptsize{±{1.2}} & \cellcolor{fgreen!61}36.1 \scriptsize{±{2.3}} \\
\hline
\multirow{2}{*}{MMLU College Medicine (ig)} & \multicolumn{1}{c|}{Low}         
& \cellcolor{fgreen!0}33.6 \scriptsize{±{0.0}} & \cellcolor{fgreen!48}41.9 \scriptsize{±{3.7}} & \cellcolor{fgreen!57}43.4 \scriptsize{±{2.1}} & \cellcolor{fgreen!42}40.9 \scriptsize{±{2.4}} & \cellcolor{fgreen!81}47.5 \scriptsize{±{2.3}} \\
& High & \cellcolor{fgreen!0}33.6 \scriptsize{±{0.0}} & \cellcolor{fgreen!88}48.7 \scriptsize{±{2.1}} & \cellcolor{fgreen!83}47.9 \scriptsize{±{3.0}} & \cellcolor{fgreen!100}50.8 \scriptsize{±{2.9}} & \cellcolor{fgreen!97}50.3 \scriptsize{±{2.8}} \\
\hline
\multirow{2}{*}{MMLU College Medicine (nso)} & \multicolumn{1}{c|}{Low}         
& \cellcolor{fgreen!0}37.0 \scriptsize{±{0.0}} & \cellcolor{fgreen!15}39.5 \scriptsize{±{0.8}} & \cellcolor{fgreen!55}46.3 \scriptsize{±{1.5}} & \cellcolor{fgreen!29}41.9 \scriptsize{±{1.3}} & \cellcolor{fgreen!55}46.4 \scriptsize{±{2.0}} \\
& High & \cellcolor{fgreen!0}37.0 \scriptsize{±{0.0}} & \cellcolor{fgreen!58}46.9 \scriptsize{±{2.5}} & \cellcolor{fgreen!61}47.4 \scriptsize{±{1.7}} & \cellcolor{fgreen!100}54.0 \scriptsize{±{1.3}} & \cellcolor{fgreen!98}53.6 \scriptsize{±{0.6}} \\
\hline
\multirow{2}{*}{MMLU College Medicine (sn)} & \multicolumn{1}{c|}{Low}         
& \cellcolor{fgreen!0}43.4 \scriptsize{±{0.0}} & \cellcolor{fgreen!3}43.7 \scriptsize{±{1.3}} & \cellcolor{fgreen!37}46.7 \scriptsize{±{1.2}} & \cellcolor{fgreen!4}43.8 \scriptsize{±{1.4}} & \cellcolor{fgreen!12}44.5 \scriptsize{±{2.0}} \\
& High & \cellcolor{fgreen!0}43.4 \scriptsize{±{0.0}} & \cellcolor{fgreen!38}46.8 \scriptsize{±{3.6}} & \cellcolor{fgreen!61}48.9 \scriptsize{±{1.4}} & \cellcolor{fgreen!74}50.1 \scriptsize{±{3.0}} & \cellcolor{fgreen!100}52.4 \scriptsize{±{2.1}} \\
\hline
\multirow{2}{*}{MMLU College Medicine (st)} & \multicolumn{1}{c|}{Low}         
& \cellcolor{fgreen!28}43.8 \scriptsize{±{0.0}} & \cellcolor{fgreen!61}48.8 \scriptsize{±{2.8}} & \cellcolor{fgreen!32}44.4 \scriptsize{±{2.5}} & \cellcolor{fgreen!0}39.5 \scriptsize{±{2.3}} & \cellcolor{fgreen!30}44.0 \scriptsize{±{3.1}} \\
& High & \cellcolor{fgreen!28}43.8 \scriptsize{±{0.0}} & \cellcolor{fgreen!55}47.9 \scriptsize{±{0.4}} & \cellcolor{fgreen!74}50.7 \scriptsize{±{1.0}} & \cellcolor{fgreen!91}53.3 \scriptsize{±{2.7}} & \cellcolor{fgreen!100}54.7 \scriptsize{±{1.0}} \\
\hline
\multirow{2}{*}{MMLU College Medicine (tn)} & \multicolumn{1}{c|}{Low}         
& \cellcolor{fgreen!30}39.2 \scriptsize{±{0.0}} & \cellcolor{fgreen!74}44.0 \scriptsize{±{2.4}} & \cellcolor{fgreen!76}44.3 \scriptsize{±{1.4}} & \cellcolor{fgreen!98}46.7 \scriptsize{±{1.1}} & \cellcolor{fgreen!0}35.9 \scriptsize{±{4.1}} \\
& High & \cellcolor{fgreen!30}39.2 \scriptsize{±{0.0}} & \cellcolor{fgreen!100}46.9 \scriptsize{±{2.0}} & \cellcolor{fgreen!100}46.9 \scriptsize{±{4.2}} & \cellcolor{fgreen!89}45.7 \scriptsize{±{2.3}} & \cellcolor{fgreen!85}45.3 \scriptsize{±{1.6}} \\
\hline
\multirow{2}{*}{MMLU College Medicine (ts)} & \multicolumn{1}{c|}{Low}         
& \cellcolor{fgreen!0}34.0 \scriptsize{±{0.0}} & \cellcolor{fgreen!18}36.4 \scriptsize{±{1.0}} & \cellcolor{fgreen!21}36.8 \scriptsize{±{1.0}} & \cellcolor{fgreen!2}34.3 \scriptsize{±{1.3}} & \cellcolor{fgreen!49}40.5 \scriptsize{±{2.1}} \\
& High & \cellcolor{fgreen!0}34.0 \scriptsize{±{0.0}} & \cellcolor{fgreen!29}37.9 \scriptsize{±{1.0}} & \cellcolor{fgreen!72}43.6 \scriptsize{±{0.8}} & \cellcolor{fgreen!69}43.3 \scriptsize{±{3.6}} & \cellcolor{fgreen!100}47.4 \scriptsize{±{3.4}} \\
\hline
\end{tabular}
\caption{\textbf{Results of Data Quality vs. Quantity using Llama 3 70B Instruct and Evaluating on MMLU Clinical Knowledge}. Llama 3 70B Instruct was fine-tuned on various GPT-4o LLM-as-an-Annotator quality and random-sampling quantity combination subsets of Winogrande train (small) and MMLU College Medicine where there was a mono-lingual lift of at least 5\% observed when using the full fine-tuning dataset (see Tables \ref{table:perf-crosslingual-llama3-tr-wino-ts-mmlu} and \ref{table:perf-crosslingual-llama3-tr-mmlu-ts-mmlu-ck}). The fine-tuned models were then evaluated on the MMLU Clinical Knowledge test split in the same language as the fine-tuning dataset. Note that the exact fine-tuning dataset sample sizes when using Winogrande as the fine-tuning dataset (Wino) and MMLU College Medicine as the fine-tuning dataset (MMLU) are given above the performance results. Language codes are as follows: Amharic (am), Igbo (ig), Sepedi (nso), Afrikaans (af), Zulu (zu), Xhosa (xh), Bambara (bm), Shona (sn), Sesotho (st), Setswana (tn), Tsonga (ts).}
\label{table:quality-x-quantity-ts-mmlu-ck}
\end{table*}

% Table: Data Quality x Quantity - MMLU Virology is Eval
\begin{table*}[]
\centering
\begin{tabular}{|l|c|ccccc|}
                            \hline
                           \multicolumn{1}{|c|}{\textbf{Fine-Tuning Dataset (language)}} & \multicolumn{1}{c|}{\textbf{Data Quality}} & \multicolumn{5}{c|}{\textbf{Data Quantity}} \\
                           & & 0\% & 25\%   & 50\%   & 75\%   & 100\%  \\ 
                           % & & (N/A) & (53; 17) & (107; 33) & (160; 50) & (213; 66) \\
                           \hline
\multicolumn{7}{|c|}{Evaluation set: \texttt{MMLU Virology}} \\ \hline
 \multicolumn{2}{|c|}{\textit{Samples}}& N/A & \textit{53} & \textit{107} & \textit{160} & \textit{213} \\ \hline
\multirow{2}{*}{Winogrande (zu)} & \multicolumn{1}{c|}{Low}         
& \cellcolor{fgreen!0}28.9 \scriptsize{±{0.0}} & \cellcolor{fgreen!38}32.1 \scriptsize{±{3.5}} & \cellcolor{fgreen!52}33.3 \scriptsize{±{4.3}} & \cellcolor{fgreen!88}36.3 \scriptsize{±{2.5}} & \cellcolor{fgreen!62}34.1 \scriptsize{±{1.2}} \\
& High & \cellcolor{fgreen!0}28.9 \scriptsize{±{0.0}} & \cellcolor{fgreen!64}34.3 \scriptsize{±{1.0}} & \cellcolor{fgreen!95}36.9 \scriptsize{±{0.7}} & \cellcolor{fgreen!57}33.7 \scriptsize{±{4.4}} & \cellcolor{fgreen!100}37.3 \scriptsize{±{2.2}} \\
\hline
\multirow{2}{*}{Winogrande (xh)} & \multicolumn{1}{c|}{Low}         
& \cellcolor{fgreen!0}27.1 \scriptsize{±{0.0}} & \cellcolor{fgreen!29}29.7 \scriptsize{±{2.4}} & \cellcolor{fgreen!100}36.1 \scriptsize{±{1.0}} & \cellcolor{fgreen!47}31.3 \scriptsize{±{2.2}} & \cellcolor{fgreen!76}33.9 \scriptsize{±{1.2}} \\
& High & \cellcolor{fgreen!0}27.1 \scriptsize{±{0.0}} & \cellcolor{fgreen!62}32.7 \scriptsize{±{2.3}} & \cellcolor{fgreen!47}31.3 \scriptsize{±{2.6}} & \cellcolor{fgreen!69}33.3 \scriptsize{±{1.7}} & \cellcolor{fgreen!71}33.5 \scriptsize{±{3.0}} \\
\hline
\multirow{2}{*}{Winogrande (nso)} & \multicolumn{1}{c|}{Low}         
& \cellcolor{fgreen!0}26.5 \scriptsize{±{0.0}} & \cellcolor{fgreen!58}33.3 \scriptsize{±{2.8}} & \cellcolor{fgreen!43}31.5 \scriptsize{±{1.2}} & \cellcolor{fgreen!74}35.1 \scriptsize{±{2.6}} & \cellcolor{fgreen!100}38.2 \scriptsize{±{1.3}} \\
& High & \cellcolor{fgreen!0}26.5 \scriptsize{±{0.0}} & \cellcolor{fgreen!32}30.3 \scriptsize{±{0.3}} & \cellcolor{fgreen!46}31.9 \scriptsize{±{3.1}} & \cellcolor{fgreen!70}34.7 \scriptsize{±{1.2}} & \cellcolor{fgreen!94}37.5 \scriptsize{±{2.6}} \\
\hline
\multirow{2}{*}{Winogrande (st)} & \multicolumn{1}{c|}{Low}         
& \cellcolor{fgreen!22}29.5 \scriptsize{±{0.0}} & \cellcolor{fgreen!0}28.3 \scriptsize{±{1.8}} & \cellcolor{fgreen!100}33.7 \scriptsize{±{1.0}} & \cellcolor{fgreen!93}33.3 \scriptsize{±{0.7}} & \cellcolor{fgreen!81}32.7 \scriptsize{±{2.4}} \\
& High & \cellcolor{fgreen!22}29.5 \scriptsize{±{0.0}} & \cellcolor{fgreen!89}33.1 \scriptsize{±{3.2}} & \cellcolor{fgreen!41}30.5 \scriptsize{±{2.1}} & \cellcolor{fgreen!81}32.7 \scriptsize{±{1.2}} & \cellcolor{fgreen!85}32.9 \scriptsize{±{0.9}} \\
\hline
\multicolumn{2}{|c|}{\textit{Samples}}& N/A & \textit{17} & \textit{33} & \textit{50} & \textit{66} \\ \hline
\multirow{2}{*}{MMLU College Medicine (zu)} & \multicolumn{1}{c|}{Low}         
& \cellcolor{fgreen!0}28.9 \scriptsize{±{0.0}} & \cellcolor{fgreen!68}35.3 \scriptsize{±{2.5}} & \cellcolor{fgreen!51}33.7 \scriptsize{±{3.1}} & \cellcolor{fgreen!2}29.1 \scriptsize{±{3.0}} & \cellcolor{fgreen!36}32.3 \scriptsize{±{2.8}} \\
& High & \cellcolor{fgreen!0}28.9 \scriptsize{±{0.0}} & \cellcolor{fgreen!57}34.3 \scriptsize{±{3.8}} & \cellcolor{fgreen!100}38.3 \scriptsize{±{4.1}} & \cellcolor{fgreen!77}36.1 \scriptsize{±{1.6}} & \cellcolor{fgreen!74}35.9 \scriptsize{±{0.7}} \\
\hline
\multirow{2}{*}{MMLU College Medicine (xh)} & \multicolumn{1}{c|}{Low}         
& \cellcolor{fgreen!0}27.1 \scriptsize{±{0.0}} & \cellcolor{fgreen!86}33.5 \scriptsize{±{3.9}} & \cellcolor{fgreen!78}32.9 \scriptsize{±{2.7}} & \cellcolor{fgreen!89}33.7 \scriptsize{±{2.2}} & \cellcolor{fgreen!16}28.3 \scriptsize{±{3.0}} \\
& High & \cellcolor{fgreen!0}27.1 \scriptsize{±{0.0}} & \cellcolor{fgreen!97}34.3 \scriptsize{±{2.6}} & \cellcolor{fgreen!86}33.5 \scriptsize{±{0.9}} & \cellcolor{fgreen!78}32.9 \scriptsize{±{2.8}} & \cellcolor{fgreen!100}34.5 \scriptsize{±{1.5}} \\
\hline
\multirow{2}{*}{MMLU College Medicine (nso)} & \multicolumn{1}{c|}{Low}         
& \cellcolor{fgreen!0}26.5 \scriptsize{±{0.0}} & \cellcolor{fgreen!37}29.3 \scriptsize{±{2.4}} & \cellcolor{fgreen!50}30.3 \scriptsize{±{1.4}} & \cellcolor{fgreen!16}27.7 \scriptsize{±{1.2}} & \cellcolor{fgreen!76}32.3 \scriptsize{±{0.3}} \\
& High & \cellcolor{fgreen!0}26.5 \scriptsize{±{0.0}} & \cellcolor{fgreen!82}32.7 \scriptsize{±{0.9}} & \cellcolor{fgreen!100}34.1 \scriptsize{±{1.2}} & \cellcolor{fgreen!76}32.3 \scriptsize{±{0.3}} & \cellcolor{fgreen!95}33.7 \scriptsize{±{0.0}} \\
\hline
\multirow{2}{*}{MMLU College Medicine (sn)} & \multicolumn{1}{c|}{Low}         
& \cellcolor{fgreen!27}31.3 \scriptsize{±{0.0}} & \cellcolor{fgreen!0}28.5 \scriptsize{±{1.8}} & \cellcolor{fgreen!31}31.7 \scriptsize{±{3.4}} & \cellcolor{fgreen!51}33.7 \scriptsize{±{2.2}} & \cellcolor{fgreen!73}35.9 \scriptsize{±{2.5}} \\
& High & \cellcolor{fgreen!27}31.3 \scriptsize{±{0.0}} & \cellcolor{fgreen!45}33.1 \scriptsize{±{1.2}} & \cellcolor{fgreen!55}34.1 \scriptsize{±{1.2}} & \cellcolor{fgreen!100}38.7 \scriptsize{±{3.0}} & \cellcolor{fgreen!63}34.9 \scriptsize{±{0.6}} \\
\hline
\multirow{2}{*}{MMLU College Medicine (tn)} & \multicolumn{1}{c|}{Low}         
& \cellcolor{fgreen!0}28.3 \scriptsize{±{0.0}} & \cellcolor{fgreen!19}29.5 \scriptsize{±{3.6}} & \cellcolor{fgreen!69}32.7 \scriptsize{±{4.4}} & \cellcolor{fgreen!41}30.9 \scriptsize{±{3.5}} & \cellcolor{fgreen!78}33.3 \scriptsize{±{3.5}} \\
& High & \cellcolor{fgreen!0}28.3 \scriptsize{±{0.0}} & \cellcolor{fgreen!100}34.7 \scriptsize{±{1.9}} & \cellcolor{fgreen!59}32.1 \scriptsize{±{2.5}} & \cellcolor{fgreen!47}31.3 \scriptsize{±{2.6}} & \cellcolor{fgreen!94}34.3 \scriptsize{±{2.2}} \\
\hline
\multirow{2}{*}{MMLU College Medicine (ts)} & \multicolumn{1}{c|}{Low}         
& \cellcolor{fgreen!0}31.3 \scriptsize{±{0.0}} & \cellcolor{fgreen!71}33.7 \scriptsize{±{2.2}} & \cellcolor{fgreen!0}31.3 \scriptsize{±{1.6}} & \cellcolor{fgreen!18}31.9 \scriptsize{±{1.8}} & \cellcolor{fgreen!82}34.1 \scriptsize{±{0.3}} \\
& High & \cellcolor{fgreen!0}31.3 \scriptsize{±{0.0}} & \cellcolor{fgreen!12}31.7 \scriptsize{±{3.7}} & \cellcolor{fgreen!76}33.9 \scriptsize{±{1.7}} & \cellcolor{fgreen!59}33.3 \scriptsize{±{0.9}} & \cellcolor{fgreen!100}34.7 \scriptsize{±{2.9}} \\
\hline
\end{tabular}
\caption{\textbf{Results of Data Quality vs. Quantity using Llama 3 70B Instruct and Evaluating on MMLU Virology}. Llama 3 70B Instruct was fine-tuned on various GPT-4o LLM-as-an-Annotator quality and random-sampling quantity combination subsets of Winogrande train (small) and MMLU College Medicine where there was a mono-lingual lift of at least 5\% observed when using the full fine-tuning dataset (see Tables \ref{table:perf-crosslingual-llama3-tr-wino-ts-mmlu-vir} and \ref{table:perf-crosslingual-llama3-tr-mmlu-ts-mmlu-vir}). The fine-tuned models were then evaluated on the MMLU Virology test split in the same language as the fine-tuning dataset. Note that the exact fine-tuning dataset sample sizes when using Winogrande as the fine-tuning dataset (Wino) and MMLU College Medicine as the fine-tuning dataset (MMLU) are given above the performance results. Language codes are as follows: Zulu (zu), Xhosa (xh), Sepedi (nso), Sesotho (st), Shona (sn), Setswana (tn), Tsonga (ts).}
\label{table:quality-x-quantity-ts-mmlu-vir}
\end{table*}

% Table: Data Quality x Quantity - Belebele is Eval
\begin{table*}[]
\centering
\begin{tabular}{|l|c|ccccc|}
                            \hline
                           \multicolumn{1}{|c|}{\textbf{Fine-Tuning Dataset (language)}} & \multicolumn{1}{c|}{\textbf{Data Quality}} & \multicolumn{5}{c|}{\textbf{Data Quantity}} \\
                           & & 0\% & 25\%   & 50\%   & 75\%   & 100\%  \\ 
                           % & & (N/A) & (53; 17) & (107; 33) & (160; 50) & (213; 66) \\
                           \hline
\multicolumn{7}{|c|}{Evaluation set: \texttt{Belebele}} \\ \hline
 \multicolumn{2}{|c|}{\textit{Samples}}& N/A & \textit{53} & \textit{107} & \textit{160} & \textit{213} \\ \hline
\multirow{2}{*}{Winogrande (af)} & \multicolumn{1}{c|}{Low}         
& \cellcolor{fgreen!0}84.8 \scriptsize{±{0.0}} & \cellcolor{fgreen!75}90.1 \scriptsize{±{0.2}} & \cellcolor{fgreen!92}91.3 \scriptsize{±{0.4}} & \cellcolor{fgreen!94}91.5 \scriptsize{±{0.8}} & \cellcolor{fgreen!80}90.5 \scriptsize{±{0.5}} \\
& High & \cellcolor{fgreen!0}84.8 \scriptsize{±{0.0}} & \cellcolor{fgreen!93}91.4 \scriptsize{±{0.2}} & \cellcolor{fgreen!89}91.1 \scriptsize{±{0.1}} & \cellcolor{fgreen!100}91.9 \scriptsize{±{0.2}} & \cellcolor{fgreen!96}91.6 \scriptsize{±{0.2}} \\
\hline
\multirow{2}{*}{Winogrande (zu)} & \multicolumn{1}{c|}{Low}         
& \cellcolor{fgreen!0}36.7 \scriptsize{±{0.0}} & \cellcolor{fgreen!8}37.6 \scriptsize{±{1.2}} & \cellcolor{fgreen!46}41.9 \scriptsize{±{1.4}} & \cellcolor{fgreen!60}43.5 \scriptsize{±{0.6}} & \cellcolor{fgreen!66}44.2 \scriptsize{±{1.8}} \\
& High & \cellcolor{fgreen!0}36.7 \scriptsize{±{0.0}} & \cellcolor{fgreen!11}37.9 \scriptsize{±{1.1}} & \cellcolor{fgreen!58}43.3 \scriptsize{±{1.2}} & \cellcolor{fgreen!100}48.1 \scriptsize{±{1.4}} & \cellcolor{fgreen!18}38.8 \scriptsize{±{0.5}} \\
\hline
\multirow{2}{*}{Winogrande (xh)} & \multicolumn{1}{c|}{Low}         
& \cellcolor{fgreen!0}36.8 \scriptsize{±{0.0}} & \cellcolor{fgreen!38}40.5 \scriptsize{±{1.6}} & \cellcolor{fgreen!79}44.5 \scriptsize{±{1.0}} & \cellcolor{fgreen!3}37.1 \scriptsize{±{1.2}} & \cellcolor{fgreen!43}41.0 \scriptsize{±{1.5}} \\
& High & \cellcolor{fgreen!0}36.8 \scriptsize{±{0.0}} & \cellcolor{fgreen!64}43.0 \scriptsize{±{0.8}} & \cellcolor{fgreen!100}46.5 \scriptsize{±{0.5}} & \cellcolor{fgreen!69}43.5 \scriptsize{±{0.6}} & \cellcolor{fgreen!91}45.6 \scriptsize{±{0.9}} \\
\hline
\multirow{2}{*}{Winogrande (am)} & \multicolumn{1}{c|}{Low}         
& \cellcolor{fgreen!0}35.1 \scriptsize{±{0.0}} & \cellcolor{fgreen!65}48.6 \scriptsize{±{1.0}} & \cellcolor{fgreen!63}48.3 \scriptsize{±{0.6}} & \cellcolor{fgreen!75}50.7 \scriptsize{±{1.0}} & \cellcolor{fgreen!68}49.3 \scriptsize{±{0.9}} \\
& High & \cellcolor{fgreen!0}35.1 \scriptsize{±{0.0}} & \cellcolor{fgreen!75}50.6 \scriptsize{±{1.0}} & \cellcolor{fgreen!100}55.9 \scriptsize{±{1.1}} & \cellcolor{fgreen!91}54.0 \scriptsize{±{1.2}} & \cellcolor{fgreen!89}53.7 \scriptsize{±{0.4}} \\
\hline
\multirow{2}{*}{Winogrande (ig)} & \multicolumn{1}{c|}{Low}         
& \cellcolor{fgreen!0}35.9 \scriptsize{±{0.0}} & \cellcolor{fgreen!60}44.9 \scriptsize{±{1.1}} & \cellcolor{fgreen!54}44.0 \scriptsize{±{1.5}} & \cellcolor{fgreen!95}50.3 \scriptsize{±{1.6}} & \cellcolor{fgreen!100}51.0 \scriptsize{±{0.9}} \\
& High & \cellcolor{fgreen!0}35.9 \scriptsize{±{0.0}} & \cellcolor{fgreen!70}46.5 \scriptsize{±{0.4}} & \cellcolor{fgreen!85}48.8 \scriptsize{±{1.5}} & \cellcolor{fgreen!74}47.1 \scriptsize{±{0.8}} & \cellcolor{fgreen!91}49.7 \scriptsize{±{0.8}} \\
\hline
\multirow{2}{*}{Winogrande (nso)} & \multicolumn{1}{c|}{Low}         
& \cellcolor{fgreen!0}37.1 \scriptsize{±{0.0}} & \cellcolor{fgreen!38}41.5 \scriptsize{±{1.0}} & \cellcolor{fgreen!19}39.3 \scriptsize{±{0.4}} & \cellcolor{fgreen!47}42.5 \scriptsize{±{1.4}} & \cellcolor{fgreen!21}39.5 \scriptsize{±{1.0}} \\
& High & \cellcolor{fgreen!0}37.1 \scriptsize{±{0.0}} & \cellcolor{fgreen!41}41.9 \scriptsize{±{2.2}} & \cellcolor{fgreen!100}48.7 \scriptsize{±{1.2}} & \cellcolor{fgreen!69}45.1 \scriptsize{±{1.7}} & \cellcolor{fgreen!79}46.3 \scriptsize{±{1.7}} \\
\hline
\multirow{2}{*}{Winogrande (sn)} & \multicolumn{1}{c|}{Low}         
& \cellcolor{fgreen!24}38.1 \scriptsize{±{0.0}} & \cellcolor{fgreen!0}35.3 \scriptsize{±{1.4}} & \cellcolor{fgreen!63}42.7 \scriptsize{±{0.9}} & \cellcolor{fgreen!41}40.1 \scriptsize{±{1.0}} & \cellcolor{fgreen!57}42.0 \scriptsize{±{1.4}} \\
& High & \cellcolor{fgreen!24}38.1 \scriptsize{±{0.0}} & \cellcolor{fgreen!53}41.5 \scriptsize{±{1.2}} & \cellcolor{fgreen!86}45.4 \scriptsize{±{1.1}} & \cellcolor{fgreen!100}47.1 \scriptsize{±{0.2}} & \cellcolor{fgreen!84}45.2 \scriptsize{±{0.8}} \\
\hline
\multirow{2}{*}{Winogrande (st)} & \multicolumn{1}{c|}{Low}         
& \cellcolor{fgreen!0}35.9 \scriptsize{±{0.0}} & \cellcolor{fgreen!55}42.7 \scriptsize{±{2.2}} & \cellcolor{fgreen!67}44.2 \scriptsize{±{1.1}} & \cellcolor{fgreen!50}42.1 \scriptsize{±{1.9}} & \cellcolor{fgreen!100}48.3 \scriptsize{±{0.6}} \\
& High & \cellcolor{fgreen!0}35.9 \scriptsize{±{0.0}} & \cellcolor{fgreen!39}40.7 \scriptsize{±{0.8}} & \cellcolor{fgreen!61}43.5 \scriptsize{±{0.7}} & \cellcolor{fgreen!65}43.9 \scriptsize{±{1.1}} & \cellcolor{fgreen!77}45.5 \scriptsize{±{2.5}} \\
\hline
\multirow{2}{*}{Winogrande (tn)} & \multicolumn{1}{c|}{Low}         
& \cellcolor{fgreen!0}36.4 \scriptsize{±{0.0}} & \cellcolor{fgreen!49}43.3 \scriptsize{±{0.5}} & \cellcolor{fgreen!57}44.4 \scriptsize{±{0.5}} & \cellcolor{fgreen!77}47.2 \scriptsize{±{0.9}} & \cellcolor{fgreen!85}48.3 \scriptsize{±{0.3}} \\
& High & \cellcolor{fgreen!0}36.4 \scriptsize{±{0.0}} & \cellcolor{fgreen!39}41.8 \scriptsize{±{0.6}} & \cellcolor{fgreen!57}44.4 \scriptsize{±{0.3}} & \cellcolor{fgreen!66}45.7 \scriptsize{±{0.9}} & \cellcolor{fgreen!100}50.4 \scriptsize{±{0.6}} \\
\hline
\multicolumn{2}{|c|}{\textit{Samples}}& N/A & \textit{17} & \textit{33} & \textit{50} & \textit{66} \\ \hline
\multirow{2}{*}{MMLU College Medicine (af)} & \multicolumn{1}{c|}{Low}         
& \cellcolor{fgreen!0}84.8 \scriptsize{±{0.0}} & \cellcolor{fgreen!71}88.5 \scriptsize{±{0.7}} & \cellcolor{fgreen!56}87.7 \scriptsize{±{0.8}} & \cellcolor{fgreen!25}86.1 \scriptsize{±{0.6}} & \cellcolor{fgreen!17}85.7 \scriptsize{±{1.1}} \\
& High & \cellcolor{fgreen!0}84.8 \scriptsize{±{0.0}} & \cellcolor{fgreen!62}88.0 \scriptsize{±{0.5}} & \cellcolor{fgreen!77}88.8 \scriptsize{±{0.4}} & \cellcolor{fgreen!79}88.9 \scriptsize{±{0.2}} & \cellcolor{fgreen!100}90.0 \scriptsize{±{0.5}} \\
\hline
\multirow{2}{*}{MMLU College Medicine (zu)} & \multicolumn{1}{c|}{Low}         
& \cellcolor{fgreen!13}36.7 \scriptsize{±{0.0}} & \cellcolor{fgreen!0}35.5 \scriptsize{±{1.1}} & \cellcolor{fgreen!79}42.5 \scriptsize{±{0.6}} & \cellcolor{fgreen!31}38.3 \scriptsize{±{2.7}} & \cellcolor{fgreen!63}41.1 \scriptsize{±{1.3}} \\
& High & \cellcolor{fgreen!13}36.7 \scriptsize{±{0.0}} & \cellcolor{fgreen!40}39.1 \scriptsize{±{1.1}} & \cellcolor{fgreen!84}43.0 \scriptsize{±{0.4}} & \cellcolor{fgreen!100}44.4 \scriptsize{±{1.1}} & \cellcolor{fgreen!98}44.2 \scriptsize{±{0.8}} \\
\hline
\multirow{2}{*}{MMLU College Medicine (xh)} & \multicolumn{1}{c|}{Low}         
& \cellcolor{fgreen!0}36.8 \scriptsize{±{0.0}} & \cellcolor{fgreen!35}40.5 \scriptsize{±{1.7}} & \cellcolor{fgreen!24}39.3 \scriptsize{±{0.9}} & \cellcolor{fgreen!52}42.3 \scriptsize{±{1.7}} & \cellcolor{fgreen!89}46.1 \scriptsize{±{0.9}} \\
& High & \cellcolor{fgreen!0}36.8 \scriptsize{±{0.0}} & \cellcolor{fgreen!41}41.1 \scriptsize{±{1.0}} & \cellcolor{fgreen!70}44.2 \scriptsize{±{1.4}} & \cellcolor{fgreen!84}45.6 \scriptsize{±{0.9}} & \cellcolor{fgreen!100}47.3 \scriptsize{±{1.7}} \\
\hline
\multirow{2}{*}{MMLU College Medicine (am)} & \multicolumn{1}{c|}{Low}         
& \cellcolor{fgreen!0}35.1 \scriptsize{±{0.0}} & \cellcolor{fgreen!92}48.3 \scriptsize{±{1.3}} & \cellcolor{fgreen!100}49.4 \scriptsize{±{1.5}} & \cellcolor{fgreen!97}48.9 \scriptsize{±{1.0}} & \cellcolor{fgreen!83}47.0 \scriptsize{±{0.3}} \\
& High & \cellcolor{fgreen!0}35.1 \scriptsize{±{0.0}} & \cellcolor{fgreen!84}47.1 \scriptsize{±{1.6}} & \cellcolor{fgreen!75}45.8 \scriptsize{±{0.6}} & \cellcolor{fgreen!78}46.2 \scriptsize{±{0.1}} & \cellcolor{fgreen!84}47.1 \scriptsize{±{0.6}} \\
\hline
\multirow{2}{*}{MMLU College Medicine (ig)} & \multicolumn{1}{c|}{Low}         
& \cellcolor{fgreen!0}35.9 \scriptsize{±{0.0}} & \cellcolor{fgreen!51}42.1 \scriptsize{±{0.9}} & \cellcolor{fgreen!44}41.2 \scriptsize{±{1.0}} & \cellcolor{fgreen!60}43.1 \scriptsize{±{0.9}} & \cellcolor{fgreen!88}46.6 \scriptsize{±{2.0}} \\
& High & \cellcolor{fgreen!0}35.9 \scriptsize{±{0.0}} & \cellcolor{fgreen!47}41.6 \scriptsize{±{1.3}} & \cellcolor{fgreen!68}44.1 \scriptsize{±{1.1}} & \cellcolor{fgreen!100}48.0 \scriptsize{±{1.1}} & \cellcolor{fgreen!93}47.1 \scriptsize{±{1.4}} \\
\hline
\multirow{2}{*}{MMLU College Medicine (nso)} & \multicolumn{1}{c|}{Low}         
& \cellcolor{fgreen!19}37.1 \scriptsize{±{0.0}} & \cellcolor{fgreen!50}40.7 \scriptsize{±{0.8}} & \cellcolor{fgreen!32}38.6 \scriptsize{±{0.4}} & \cellcolor{fgreen!83}44.5 \scriptsize{±{0.5}} & \cellcolor{fgreen!70}43.0 \scriptsize{±{1.9}} \\
& High & \cellcolor{fgreen!19}37.1 \scriptsize{±{0.0}} & \cellcolor{fgreen!0}34.9 \scriptsize{±{0.4}} & \cellcolor{fgreen!44}40.0 \scriptsize{±{0.4}} & \cellcolor{fgreen!86}44.8 \scriptsize{±{1.9}} & \cellcolor{fgreen!100}46.4 \scriptsize{±{1.3}} \\
\hline
\multirow{2}{*}{MMLU College Medicine (sn)} & \multicolumn{1}{c|}{Low}         
& \cellcolor{fgreen!18}38.1 \scriptsize{±{0.0}} & \cellcolor{fgreen!45}40.1 \scriptsize{±{2.0}} & \cellcolor{fgreen!51}40.6 \scriptsize{±{1.0}} & \cellcolor{fgreen!74}42.3 \scriptsize{±{1.3}} & \cellcolor{fgreen!100}44.2 \scriptsize{±{0.5}} \\
& High & \cellcolor{fgreen!18}38.1 \scriptsize{±{0.0}} & \cellcolor{fgreen!15}37.9 \scriptsize{±{0.4}} & \cellcolor{fgreen!0}36.8 \scriptsize{±{0.4}} & \cellcolor{fgreen!81}42.8 \scriptsize{±{0.6}} & \cellcolor{fgreen!66}41.7 \scriptsize{±{0.5}} \\
\hline
\multirow{2}{*}{MMLU College Medicine (st)} & \multicolumn{1}{c|}{Low}         
& \cellcolor{fgreen!0}35.9 \scriptsize{±{0.0}} & \cellcolor{fgreen!84}45.1 \scriptsize{±{0.2}} & \cellcolor{fgreen!51}41.5 \scriptsize{±{1.1}} & \cellcolor{fgreen!61}42.6 \scriptsize{±{1.6}} & \cellcolor{fgreen!75}44.1 \scriptsize{±{1.6}} \\
& High & \cellcolor{fgreen!0}35.9 \scriptsize{±{0.0}} & \cellcolor{fgreen!57}42.2 \scriptsize{±{0.8}} & \cellcolor{fgreen!55}42.0 \scriptsize{±{1.1}} & \cellcolor{fgreen!74}44.0 \scriptsize{±{0.7}} & \cellcolor{fgreen!100}46.9 \scriptsize{±{0.8}} \\
\hline
\multirow{2}{*}{MMLU College Medicine (tn)} & \multicolumn{1}{c|}{Low}         
& \cellcolor{fgreen!0}36.4 \scriptsize{±{0.0}} & \cellcolor{fgreen!82}44.3 \scriptsize{±{1.0}} & \cellcolor{fgreen!100}46.0 \scriptsize{±{0.7}} & \cellcolor{fgreen!94}45.4 \scriptsize{±{0.8}} & \cellcolor{fgreen!79}44.0 \scriptsize{±{0.8}} \\
& High & \cellcolor{fgreen!0}36.4 \scriptsize{±{0.0}} & \cellcolor{fgreen!35}39.8 \scriptsize{±{0.3}} & \cellcolor{fgreen!56}41.8 \scriptsize{±{1.3}} & \cellcolor{fgreen!76}43.7 \scriptsize{±{0.6}} & \cellcolor{fgreen!99}45.9 \scriptsize{±{1.0}} \\
\hline
\multirow{2}{*}{MMLU College Medicine (ts)} & \multicolumn{1}{c|}{Low}         
& \cellcolor{fgreen!32}40.7 \scriptsize{±{0.0}} & \cellcolor{fgreen!32}40.7 \scriptsize{±{0.1}} & \cellcolor{fgreen!11}38.7 \scriptsize{±{0.9}} & \cellcolor{fgreen!0}37.6 \scriptsize{±{0.4}} & \cellcolor{fgreen!40}41.4 \scriptsize{±{0.6}} \\
& High & \cellcolor{fgreen!32}40.7 \scriptsize{±{0.0}} & \cellcolor{fgreen!41}41.5 \scriptsize{±{0.7}} & \cellcolor{fgreen!51}42.5 \scriptsize{±{0.2}} & \cellcolor{fgreen!48}42.2 \scriptsize{±{1.6}} & \cellcolor{fgreen!100}47.2 \scriptsize{±{0.9}} \\
\hline
\end{tabular}
\caption{\textbf{Results of Data Quality vs. Quantity using Llama 3 70B Instruct and Evaluating on Belebele}. Llama 3 70B Instruct was fine-tuned on various GPT-4o LLM-as-an-Annotator quality and random-sampling quantity combination subsets of Winogrande train (small) and MMLU College Medicine where there was a mono-lingual lift of at least 5\% observed when using the full fine-tuning dataset (see Tables \ref{table:perf-crosslingual-llama3-tr-wino-ts-bele} and \ref{table:perf-crosslingual-llama3-tr-mmlu-ts-bele}). The fine-tuned models were then evaluated on Belebele in the same language as the fine-tuning dataset. Note that the exact fine-tuning dataset sample sizes when using Winogrande as the fine-tuning dataset (Wino) and MMLU College Medicine as the fine-tuning dataset (MMLU) are given above the performance results. Language codes are as follows: Afrikaans (af), Zulu (zu), Xhosa (xh), Amharic (am), Igbo (ig), Sepedi (nso), Shona (sn), Sesotho (st), Setswana (tn), Tsonga (ts).}
\label{table:quality-x-quantity-ts-bele}
\end{table*}

\begin{table*}
\centering
\footnotesize
\setlength\extrarowheight{2pt}
\begin{tabular}{|p{0.7cm}p{0.6cm}p{0.4cm}p{0.4cm}p{0.7cm}p{1cm}p{0.6cm}p{1.4cm}p{2.8cm}p{3.5cm}|}
\hline
\textbf{ID} & \textbf{Badge} & \textbf{Loc.} & \textbf{Jobs} & \textbf{Hours} & \textbf{Earned} & \textbf{JSS} & \textbf{Eng. Prof.} & \textbf{Assigned Tasks} & \textbf{Target Prof.} \\
\hline
A.B. & Top & ZA & 10 & 381 & \$2,000 & 100\% & F & E (af) & af (N) \\
A.D. & - & ZA & 65 & 1,265 & \$10,000 & 100\% & N & E/V (af) & af (N) \\
A.H. & - & ZA & 65 & 283 & \$10,000 & 100\% & F & V (af) & af (N) \\
A.M. & - & ZA & 4 & 0 & \$200 & 76\% & N & E (xh) & xh (N) \\
A.S.1 & - & ZA & 16 & 136 & \$2,000 & 90\% & N & T (nso); E (zu) & zu (N); nso (F) \\
A.S.2 & - & ZW & 6 & 10 & \$500 & 100\% & N & V (sn) & sn (N) \\
A.V. & Top+ & ZA & 490 & 1,703 & \$90,000 & 99\% & N & V/T (af) & af (N) \\
A.Z.1 & - & ZA & 6 & 2 & \$100 & 100\% & N & E (zu,ts) & zu (N); ts(N) \\
A.Z.2 & - & ZA & 23 & 1,480 & \$10,000 & 94\% & F & T (zu) & zu (N) \\
B.A. & Top & ET & 259 & 93 & \$20,000 & 100\% & N & E (am) & am (N) \\
B.M. & - & ZA & 0 & 0 & \$0 & - & N & T (zu) & zu (N) \\
C.B. & Top & NG & 137 & 893 & \$30,000 & 97\% & N & E (ig) & ig (N) \\
C.P. & - & ZA & 4 & 21 & \$300 & 63\% & N & E (xh) & xh (N) \\
D.C. & - & ML & 26 & 11 & \$1,000 & 100\% & N & E (bm) & bm (N) \\
D.D. & Top+ & ZA & 26 & 1,004 & \$20,000 & 100\% & N & V (af) & af (N) \\
D.E. & Top & NG & 15 & 749 & \$10,000 & 100\% & N & E/V/T (ig) & ig (N) \\
D.H. & Top+ & ET & 255 & 626 & - & 100\% & F & T (am) & am (N) \\
D.M. & Top+ & ML & 131 & 13,676 & \$100,000 & 100\% & N & E/T (bm) & bm (N) \\
F.K. & - & ZW & 5 & 6 & \$600 & 100\% & N & E/V (sn) & sn (N) \\
H.B. & - & ZA & 5 & 65 & \$1,000 & 100\% & N & E (af) & af (N) \\
H.P. & - & ZA & 0 & 0 & \$0 & - & N & T (af) & af (N) \\
I.M. & Top & ZA & 22 & 7 & \$9,000 & 96\% & N & T (st, tn, af, zu) & af (F); zu (C); st (N); tn (N) \\
L.A. & - & ZA & 24 & 822 & \$10,000 & 100\% & N & E (af) & af (N) \\
L.M. & - & ZA & 3 & 0 & \$0 & 100\% & N & V (zu) & zu (N) \\
L.R. & - & ZA & 2 & 0 & \$70 & - & N & E (st) & st (F) \\
L.S. & - & ZA & 15 & 0 & \$5,000 & 100\% & N & E (af) & af (N) \\
M.B. & Top & ZA & 70 & 1,258 & \$10,000 & 100\% & F & E (af) & af (F) \\
M.H. & - & ZA & 3 & 0 & \$200 & 100\% & N & E/V/T (zu) & zu (N) \\
M.L. & Top & ET & 67 & 250 & \$10,000 & 100\% & F & V (am) & am (N) \\
M.M.1 & - & ZA & 76 & 386 & \$10,000 & 100\% & N & V (zu) & zu (N) \\
M.M.2 & Top & ZA & 14 & 974 & \$6,000 & 100\% & N & E/T/V (tn) & tn (N) \\
M.P. & - & ZA & 2 & 0 & \$200 & - & N & E (nso) & nso (N) \\
M.S. & - & SN & 25 & 108 & \$3,000 & 100\% & N & E/V (bm) & bm (F) \\
M.T.1 & Top & ET & 38 & 998 & \$10,000 & 100\% & N & E (am) & am (N) \\
M.T.2 & Top & ZA & 90 & 3,354 & \$90,000 & 90\% & F & E/V (xh); V (zu) & xh (N); zu (N) \\
M.W. & - & ZA & 1 & 1 & \$15 & - & N & E (af) & af (N) \\
N.B. & - & ZA & 6 & 0 & \$200 & 100\% & N & V (xh) & xh (N) \\
N.C. & - & ZA & 71 & 231 & \$10,000 & 100\% & N & E/V (af) & af (N) \\
N.H. & - & ZA & 8 & 0 & \$200 & 100\% & N & E/T/V (ts) & ts(F) \\
N.K. & - & ZW & 4 & 2 & \$1,000 & - & N & V (sn) & sn (N) \\
N.M. & - & ZA & 13 & 1,367 & \$10,000 & 81\% & N & E/T/V (ts) & ts(F) \\
N.N. & - & ZW & 11 & 120 & \$10,000 & 100\% & N & T (sn) & sn (N) \\
N.R. & - & ZA & 18 & 35 & \$1,000 & 100\% & N & E (st) & st (N) \\
P.H. & - & ZA & 98 & 167 & \$5,000 & 95\% & N & E/T/V (af) & af (N) \\
P.N. & Top+ & NG & 24 & 1,583 & \$20,000 & 96\% & N & V/T (ig) & ig (N) \\
P.R. & - & ZA & 9 & 42 & \$500 & 88\% & N & E (zu, xh); E/V (nso) & xh (N); zu (N); nso (N) \\
P.T. & - & ZA & 13 & 14 & \$400 & 100\% & N & E (xh) & xh (N) \\
R.M. & Top & ZA & 85 & 1,166 & \$40,000 & 100\% & N & T (zu, nso); V (st, nso) & zu (N); nso (N); st (N) \\
\hline
\end{tabular}
\end{table*}

\begin{table*}[t]
\centering
\footnotesize
\setlength \extrarowheight{2pt}
\begin{tabular}{|p{0.7cm}p{0.6cm}p{0.4cm}p{0.4cm}p{0.7cm}p{1cm}p{0.6cm}p{1.4cm}p{2.8cm}p{3.5cm}|}
\hline
\textbf{ID} & \textbf{Badge} & \textbf{Loc.} & \textbf{Jobs} & \textbf{Hours} & \textbf{Earned} & \textbf{JSS} & \textbf{Eng. Prof.} & \textbf{Assigned Tasks} & \textbf{Target Prof.} \\
\hline
S.B. & - & ZA & 5 & 7 & \$300 & 100\% & N & E (af) & af (N) \\
S.D. & - & ZA & 0 & 0 & \$50 & - & N & E/T (xh) & xh (N) \\
S.M.1 & - & ZA & 0 & 0 & \$50 & - & N & E/T (xh) & xh (N) \\
S.M.2 & Top & ZA & 15 & 257 & \$5,000 & 100\% & N & E (zu) & zu (N) \\
S.N.1 & - & ZA & 0 & 0 & \$0 & - & F & T (xh) & xh (N) \\
S.N.2 & - & ZA & 14 & 467 & \$6,000 & 100\% & N & V (zu) & zu (N) \\
S.N.3 & Top & ZA & 13 & 29 & \$5,000 & 100\% & N & E/V (zu) & zu (N) \\
S.S. & - & ZA & 15 & 2 & \$1,000 & 100\% & N & E/V (xh) & xh (N) \\
S.V. & - & ZA & 1 & 0 & \$8 & - & F & E (zu) & zu (N) \\
T.B. & - & ZA & 7 & 152 & \$5,000 & 100\% & N & T (af) & af (C) \\
T.M.1 & - & BW & 9 & 19 & \$600 & 100\% & N & E/V (tn) & tn (N) \\
T.M.2 & - & ZA & 3 & 0 & \$50 & 100\% & N & T (xh) & xh (N) \\
T.M.3 & - & ZA & 10 & 80 & \$2,000 & 100\% & N & E (zu) & zu (N) \\
V.K. & Top & MW & 44 & 139 & \$4,000 & 100\% & N & E (zu) & zu (F) \\
X.P. & - & ZA & 1 & 0 & \$50 & - & N & T (xh) & xh (N) \\
Y.M. & - & ZW & 7 & 1 & \$200 & 100\% & N & E/V (sn) & sn (N) \\
Z.P. & Top & ZA & 291 & 424 & \$60,000 & 100\% & F & E (af) & af (N) \\
\hline
\end{tabular}
    \caption{\textbf{Profile of Translators / Annotators Sourced from Upwork.com}. This table lists the profiles of individuals recruited to perform the translation, review, and cultural appropriateness tasks. ``Badge" is the profile badge present on Upwork to indicate performance, if applicable (``Top" is for the ``Top Rated" badge, and ``Top+" is for the ``Top Rated Plus" badge). ``Loc." is the worker's country of residence (given as an ISO 3166-1 alpha-2 country code). ``Jobs" is the number of prior jobs completed on Upwork. ``Hrs." is the number of prior working hours reported on Upwork (does not count fixed-price jobs). ``Earned" is the total prior earned amount reported on Upwork (in USD). ``JSS" is a metric used by Upwork to represent ``Job Success Score", expressed as a percentage (if not present, the worker likely did not have enough jobs at the time for it to be calculated). ``Eng. Prof." is the self-reported English language proficiency. ``Assigned Task" is a semicolon-separated list of tasks assigned to the worker, with the task displayed as a single letter followed by the target language for that task. Tasks are represented as follows: ``E":  cultural appropriateness \textit{evaluator}, ``T": \textit{translator}, ``V": translation \textit{validator}. ``Target Prof." is the self-reported proficiency in the target languages for the assigned tasks. Language proficiency is represented as follows: ``N": native or bilingual, ``F": fluent, ``C": conversational.\\\\\\\\\\\\\\\\\\\\\\\\\\\\\\\\\\\\\\\\\\\\\\\\\\\\\\\\}
\label{table:upwork-profiles}
\end{table*}

%
%
%
%
%
%
% ALL APPENDIX FIGURES
%
%
%
%
%
%
%
%

\FloatBarrier

\FloatBarrier

\begin{figure*}[h]
\centering
    \begin{mdframed}
    \noindent\textbf{Prompt:}\\
        \noindent\texttt{
        The following are sentences that are missing a word or a few words (denoted with an underscore), each followed by two options to fill in the missing word or words. The correct option is given for each sentence: \\ \\
        Sentence 1: \{Dev Set Sentence \#1\} \\
        Option1: \{Dev Set Option 1 \#1\} \\
        Option2: \{Dev Set Option 2 \#1\} \\
        Correct Option: \{Dev Set Correct Option \#1\} \\ \\
        ... \\ \\ Sentence 5: \{Dev Set Sentence \#5\} \\ Option1: \{Dev Set Option 1 \#5\} \\ Option2: \{Dev Set Option 2 \#5\} \\ Correct Option: \{Dev Set Correct Option \#5\} \\ \\ Now, given the following sentence and options, output only the number corresponding to the correct option. Do not add any explanation. \\ \\ Sentence: \{Test Set Sentence\} \\ Option1: \{Test Set Option 1\} \\ Option2: \{Test Set Option 2\} \\ Correct Option: } \\

    \noindent\textbf{Example Output:}\\
    \noindent\texttt{1}
    \end{mdframed}
    \caption{\textbf{Winogrande Evaluation Prompt.} The 5-shot, hard prompt template given to LLMs in order to evaluate them on Winogrande. The 5-shot examples were all sourced randomly from the translated Winogrande dev set such that the same examples were used for each language (i.e. the examples were versions of the same English examples but translated to different African languages). Moreover, within each language, the same 5-shot examples were provided with each question in the translated Winogrande test set. LLMs were evaluated on all questions in the test set (1,767 questions per language). If the LLM gave the same number (1 or 2) as the correct answer but not both, the LLM's response was considered correct. Otherwise, the LLM's response was considered incorrect. This prompt was adapted from the one used for MMLU (see Figure \ref{fig:mmlu-prompt} below), since the GitHub repository for Winogrande lacks LLM evaluation prompts.}
    \label{fig:winogrande-prompt}
\end{figure*}

\FloatBarrier

\begin{figure*}[h]
\centering
    \begin{mdframed}
    \noindent\textbf{Prompt:}\\
        \noindent\texttt{
        The following are multiple choice questions (with answers) about \{MMLU subject name\}. \\ \\
        Question 1: \{Dev Set Question \#1\} \\
        A. \{Dev Set Option A \#1\} \\
        B. \{Dev Set Option B \#1\} \\ C. \{Dev Set Option C \#1\} \\
        D. \{Dev Set Option D \#1\} \\
        Answer: \{Dev Set Answer \#1\} \\ \\
        ... \\ \\ Question 5: \{Dev Set Question \#5\} \\
        A. \{Dev Set Option A \#5\} \\
        B. \{Dev Set Option B \#5\} \\ C. \{Dev Set Option C \#5\} \\
        D. \{Dev Set Option D \#5\} \\
        Answer: \{Dev Set Answer \#5\} \\ \\ Now, given the following question and answer choices, output only the letter corresponding to the correct answer. Do not add any explanation. \\ \\ Question: \{Test Set Question\} \\
        A. \{Test Set Option A\} \\
        B. \{Test Set Option B\} \\ C. \{Test Set Option C\} \\
        D. \{Test Set Option D\} \\
        Answer:  } \\

    \noindent\textbf{Example Output:}\\
    \noindent\texttt{A}
    \end{mdframed}
    \caption{\textbf{MMLU Evaluation Prompt.} The 5-shot, hard prompt template given to LLMs in order to evaluate them on the three clinical sections of MMLU. The 5-shot examples were all sourced from the translated MMLU dev set for the given subject (each subject had exactly 5 examples in its dev set). LLMs were evaluated on all questions in the test set (265/173/166 questions per language for clinical knowledge/college medicine/virology). If the LLM gave the correct letter answer after extra characters like parentheses or periods were removed, the LLM's response was considered correct; otherwise, the LLM's response was considered incorrect. This prompt was inferred from the code provided by the authors of MMLU in their GitHub repository \cite{hendrycks2021ethics}. Find the repository here: https://github.com/hendrycks/test}
    \label{fig:mmlu-prompt}
\end{figure*}

\FloatBarrier

\begin{figure*}[h]
\centering
    \begin{mdframed}
    \noindent\textbf{Prompt:}\\
        \noindent\texttt{
        Given the following passage, query, and answer choices, output only the letter corresponding to the correct answer. Do not add any explanation. \\ \#\#\# \\ Passage: \\ \{Test Set Passage\} \\\#\#\# \\ Query: \\ \{Test Set Question\} \\\#\#\# \\ Choices: \\(A) \{Test Set Option A\} \\(B) \{Test Set Option B\} \\(C) \{Test Set Option C\} \\(D) \{Test Set Option D\} \\\#\#\# \\ Answer:} \\

    \noindent\textbf{Example Output:}\\
    \noindent\texttt{A}
    \end{mdframed}
    \caption{\textbf{Belebele Evaluation Prompt.} The 0-shot, hard prompt template given to LLMs in order to evaluate them on Belebele. LLMs were evaluated on all questions in Belebele (900 questions per language; Belebele does not have a dev set). If the LLM gave the correct letter answer after extra characters like parentheses or periods were removed, the LLM's response was considered correct; otherwise, the LLM's response was considered incorrect. This prompt follows the one provided by the authors of Belebele in their GitHub repository \cite{bandarkar-etal-2024-belebele}. Find the repository here: https://github.com/facebookresearch/belebele}
    \label{fig:belebele-prompt}
\end{figure*}

\FloatBarrier

\begin{figure*}[h]
\centering
    \begin{mdframed}
    \noindent\textbf{Prompt:}\\
        \noindent\texttt{
        Given the following sentence and options, output only the number corresponding to the correct option. Do not add any explanation. \\ \\ Sentence: \{Train Set Sentence\} \\ Option1: \{Train Set Option 1\} \\ Option2: \{Train Set Option 2\} \\ Correct Option: } \\

    \noindent\textbf{Expected Output:}\\
    \noindent\texttt{(1 or 2, depending on which is correct)}
    \end{mdframed}
    \caption{\textbf{Winogrande Supervised Fine-Tuning Prompt.} The prompt template used for all experiments that involved fine-tuning on all of Winogrande or a subset of Winogrande (e.g. for quality and quantity combinations). As is the case for supervised fine-tuning, the LLMs were given examples of Winogrande questions with correct answers following the template above (similar to few-shot learning) and were expected to adjust their parameters to perform better on Winogrande as well as tasks like Winogrande (e.g. other question-answering tasks like MMLU or Belebele), ideally with lifts in African languages (especially the language used for fine-tuning). The fine-tuning/training examples were sourced from the Winogrande train (small) set, which has 640 examples per language. The prompt was designed to be as similar as possible to the Winogrande evaluation prompt we used, which can be seen in Figure \ref{fig:winogrande-prompt}.}
    \label{fig:winogrande-ft-prompt}
\end{figure*}

\FloatBarrier

\begin{figure*}[h]
\centering
    \begin{mdframed}
    \noindent\textbf{Prompt:}\\
        \noindent\texttt{
        Given the following question and answer choices, output only the letter corresponding to the correct answer. Do not add any explanation. \\ \\ Question: \{Train Set Question\} \\
        A. \{Train Set Option A\} \\
        B. \{Train Set Option B\} \\ C. \{Train Set Option C\} \\
        D. \{Train Set Option D\} \\
        Answer: } \\

    \noindent\textbf{Expected Output:}\\
    \noindent\texttt{(A, B, C, or D, depending on which is correct)}
    \end{mdframed}
    \caption{\textbf{MMLU College Medicine Supervised Fine-Tuning Prompt.} The prompt template used for all experiments that involved fine-tuning on all of MMLU College Medicine or a subset of MMLU College Medicine (e.g. for quality and quantity combinations). As is the case for supervised fine-tuning, the LLMs were given examples of MMLU College Medicine questions with correct answers following the template above (similar to few-shot learning) and were expected to adjust their parameters to perform better on other clinical MMLU sections as well as other, similar question-answering tasks like Winogrande or Belebele, ideally with lifts in African languages (especially the language used for fine-tuning). The fine-tuning/training examples were sourced from the entire MMLU College Medicine section, which has 200 examples per language (MMLU subjects do not have non-test sets large enough for ample fine-tuning, so all of MMLU College Medicine was used for fine-tuning and excluded from fine-tuned model evaluations). The prompt was designed to be as similar as possible to the MMLU evaluation prompt we used, which can be seen in Figure \ref{fig:mmlu-prompt}.}
    \label{fig:mmlu-ft-prompt}
\end{figure*}

\FloatBarrier

\begin{figure*}[h]
\centering
    \begin{mdframed}
    \noindent\textbf{Prompt:}\\
        \noindent\texttt{I am trying to curate a supervised fine-tuning dataset that will be used to improve the performance of my LLM on \{benchmark\} in \{target language\} (I have translated \{benchmark\} from English into many other languages). Please rate the following fine-tuning dataset sample on a scale of 1-10. Your rating should reflect the anticipated usefulness of the sample if included in the fine-tuning dataset for improving my LLM's performance on \{benchmark\} (with 1 implying least useful and 10 implying most useful). Do not include any additional explanation; just provide the number rating: \{fine-tuning dataset sample\}} \\

    \noindent\textbf{Example Output:}\\
    \noindent\texttt{9}
    \end{mdframed}
    \caption{\textbf{GPT-4o LLM-as-an-Annotator Prompt.} The prompt template given to GPT-4o in order to assign quality scores to rows in our fine-tuning datasets. \texttt{\{benchmark\}} was substituted for the name of the benchmark (e.g. MMLU, Belebele) as well as the MMLU section (if the benchmark was MMLU) and a short description of the benchmark (e.g. ``a diversity-focused reading comprehension evaluation benchmark" for Belebele). \texttt{\{target language\}} was substituted for the full name of the language that the model would be both fine-tuned in and evaluated on (e.g. ``Xhosa"). Finally, \texttt{\{fine-tuning dataset sample\}} was substituted with a JSONL row representing a supervised fine-tuning dataset sample that would be used to fine-tune a GPT model using OpenAI's fine-tuning API (see https://platform.openai.com/docs/guides/fine-tuning/example-format for the format of such JSONL dataset fine-tuning rows).}
    \label{fig:gpt-4o-quality-prompt}
\end{figure*}

\FloatBarrier

\FloatBarrier

\begin{figure*}[h] % 'h' for "here", other options: t (top), b (bottom), p (page of floats), etc.
    \centering  % Center the figure
    \includegraphics[width=0.99\textwidth]{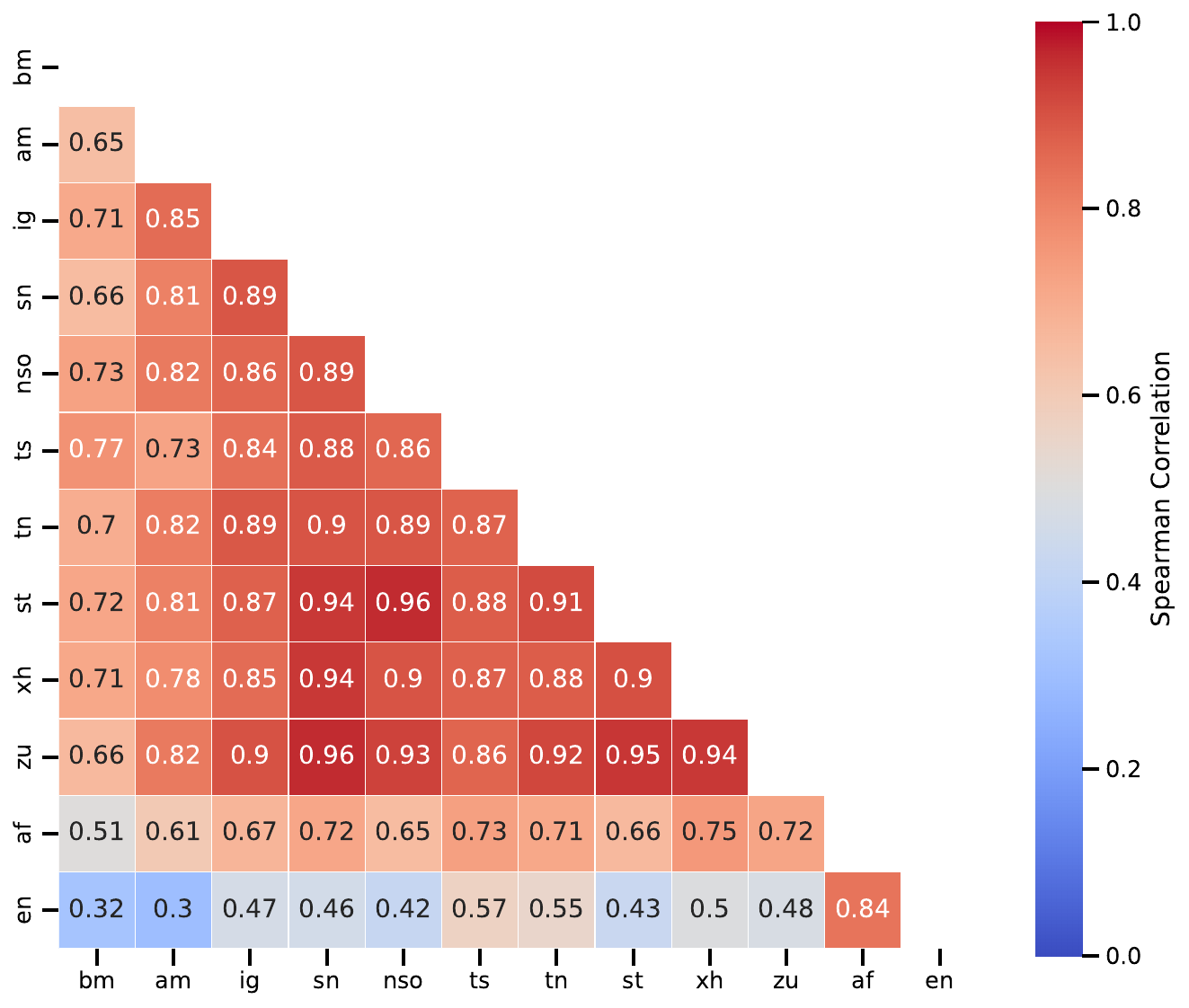}
    \caption{\textbf{Correlation of State-of-the-Art Models on Translated Benchmarks.} The correlation heat map shows similarities between out-of-the-box LLM performances from Table \ref{table:perf-results} by language. The color bar shows the Spearman rank correlation, where values closer to 1 are more correlated. English is the least correlated with the other languages, while the seven Bantu languages (Zulu, Xhosa, Sesotho, Setswana, Tsonga, Sepedi, and Shona) and Igbo (Volta-Niger) had the highest correlation values (see Figure \ref{fig:languagemap} for language families). Bambara and Amharic are the least correlated with the other languages. Language codes are as follows: Bambara (bm), Amharic (am), Igbo (ig), Shona (sn), Sepedi (nso), Tsonga (ts), Setswana (tn), Sesotho (st), Xhosa (xh), Zulu (zu), Afrikaans (af), English (en).}
    \label{fig:out-of-box-correlation}
\end{figure*}

\FloatBarrier

\FloatBarrier

\begin{figure*}[h] % 'h' for "here", other options: t (top), b (bottom), p (page of floats), etc.
    \centering  % Center the figure
    \includegraphics[width=0.99\textwidth]{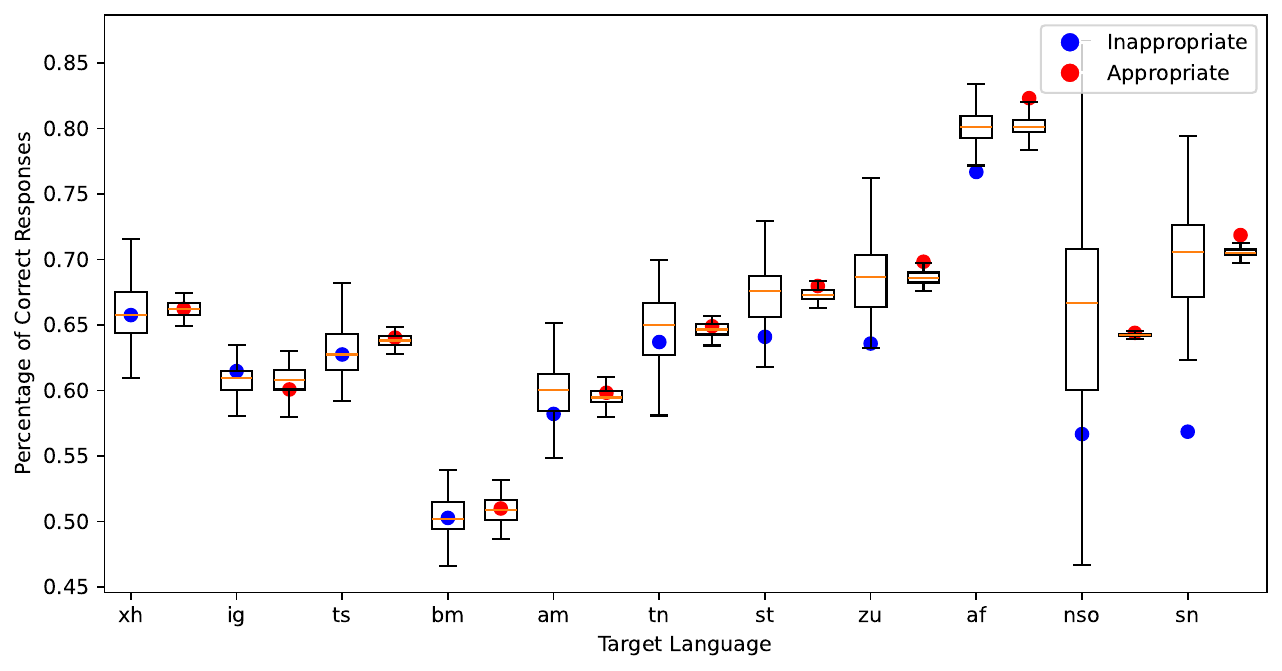}
    \caption{\textbf{Comparison of GPT-4o Performance on ``appropriate" Versus ``inappropriate" Winogrande Questions.} The percentage of correct responses for the GPT-4o model when the translation was appropriate (red/dots to the right of language code x-axis labels) and when the translation was inappropriate (blue/dots directly above language code x-axis labels) are shown for each target language. The boxplots show the distribution of the percentage of correct responses when repeated random samples of the same size as the appropriate and inappropriate counts for each target language were drawn; the figure implies that the differences in performance on the appropriate vs. inappropriate subsets for many languages (e.g. Shona/sn) are unlikely to be due to chance. Language codes are as follows: Xhosa (xh), Igbo (ig), Tsonga (ts), Bambara (bm), Amharic (am), Setswana (tn), Sesotho (st), Zulu (zu), Afrikaans (af), Sepedi (nso), Shona (sn).}
    \label{fig:appropriateness_boxplots}
\end{figure*}

\FloatBarrier

\FloatBarrier

\begin{figure*}[h]
    \centering
    \includegraphics[width=0.99\textwidth, trim={0 0 0 1.2cm},clip]{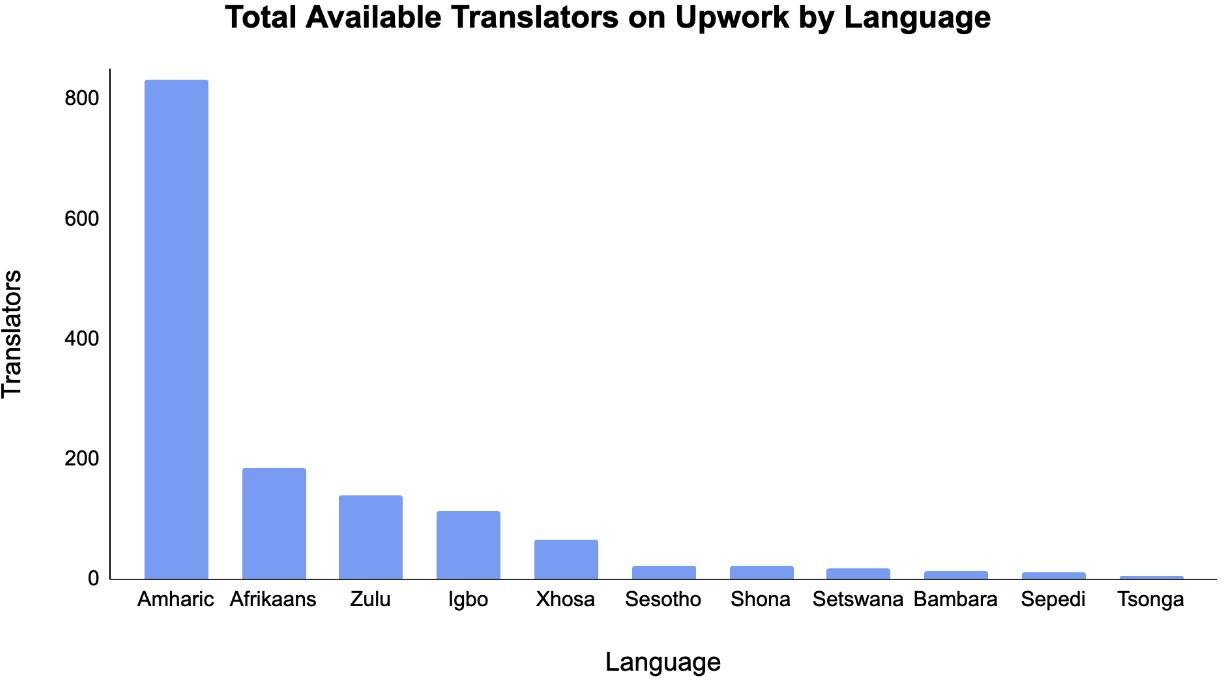}
    \caption{\textbf{Availability of Translators on Upwork.com}. We conducted a simple search on Upwork.com for ``\{Language\} Translator" (where \{Language\} was substituted with the actual language name) and counted the number of results, indicating roughly how many translators for each language were available on Upwork.com. The total number of available translators on Upwork.com was severely limited (less than 100) for several target languages: Xhosa, Sesotho, Shona, Setswana, Bambara, Sepedi, and Tsonga.}
    \label{fig:upworker_pool}
\end{figure*}

\FloatBarrier

\FloatBarrier

\FloatBarrier

\FloatBarrier

 % that most increased accuracy for obtaining the probability of a ``Good translation" rating from both evaluators using a random forest model were the Zulu, Igbo, and Bambara languages as the target language of translation. 
    
     % The ROUGE-1 (vs MT) threshold for being ``Very Dissimilar" to the corresponding machine translation was \emph{less than 0.5}, while the threshold for being ``Very Similar" to the corresponding machine translation was \emph{greater than 0.95}. Naturally, to be listed as ``Matches MT", the ROUGE-1 score had to be \emph{exactly 1}. ``Initial Translation" denotes the Winogrande row after the first round of translations, while the ``Corrected Translation" denotes the Winogrande row after it was reviewed, and possibly corrected, during the second round of translations. A Winogrande row translation ``Needed Correction" if and only if the second translator made changes to the initial translation.

\FloatBarrier

\FloatBarrier

\FloatBarrier

\end{document}